%% file: thesis-umich-style.tex
\documentclass[thesis]{thesis-umich}
 \pdfoutput=1

\usepackage{blindtext} 
\usepackage[export]{adjustbox}
\usepackage{tabularx}
\usepackage{float}
\usepackage{wrapfig}
\usepackage{lscape}

\usepackage{geometry}
\usepackage{pdfpages}
\usepackage{graphicx}
\usepackage{fullpage}
\usepackage{amsfonts}
\usepackage{amsmath}
\usepackage{algpseudocode}
\usepackage{algorithm}
\usepackage{hyperref}
\usepackage{comment}

\usepackage{natbib}

\usepackage{subcaption}
\usepackage{amssymb}
\usepackage{appendix}
\usepackage{bbm}
\usepackage{multirow}
\usepackage{booktabs}
\usepackage{algorithm}
\usepackage{algpseudocode}
\usepackage{hhline}

\usepackage{setspace}
\usepackage{scalerel}
\usepackage{makecell}
\usepackage{bbding}

\usepackage{pifont}

\def\v#1{\mathbf{#1}}

\DeclareMathOperator*{\softmax}{softmax}

\DeclareMathOperator*{\argmax}{\arg\!\max}

\newcommand\SingleLineDown[1]{%
\Statex\hspace*{-\algorithmicindent}\textbf{#1}%
\vspace*{-.7\baselineskip}\Statex\hspace*{\dimexpr-\algorithmicindent-2pt\relax}\rule{\linewidth}{0.4pt}%
}

\newcommand\DoubleLine[1]{%
\vspace*{-0.7\baselineskip}\Statex\hspace*{\dimexpr-\algorithmicindent-2pt\relax}\rule{\linewidth}{0.4pt}%
\Statex\hspace*{-\algorithmicindent}\textbf{#1}%
\vspace*{-.7\baselineskip}\Statex\hspace*{\dimexpr-\algorithmicindent-2pt\relax}\rule{\linewidth}{0.4pt}%
}

\newcommand{\bs}[1]{\textcolor{magenta}{#1 --Bowen}}
\newcommand{\checkwriting}[1]{\textcolor{magenta}{This part is revised or rewritten from original paper. Check writing. --Bowen}}
\newcommand{\red}[1]{\textcolor{red}{#1}}
\newcommand{\brown}[1]{\textcolor{brown}{#1}}

\newcommand{\kl}[1]{}
\newcommand{\kledit}[1]{#1}
\newcommand{\bsedit}[1]{#1}

\renewcommand{\bs}[1]{}
\renewcommand{\checkwriting}[1]{}

\newcommand{\cmark}{\ding{51}}%
\newcommand{\xmark}{\ding{55}}%

\title{Toward American Sign Language Processing in the Real World:\\ Data, Tasks, and Methods}
\author{Bowen Shi}
\date{Toyota Technological Institute at Chicago}

\email{bshi@ttic.edu}
\orcid{0000-0003-0065-9692}

\frontispiece{\includegraphics[width=4in]{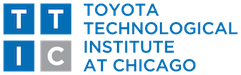}}

\frontpagestyle{7} 

\abstract{
\input{abstract}
}

\acknowledgments[7]{\input{acknowledgments}}

\committee{ %
Professor Karen Livescu (Thesis Advisor) \\
Professor Gregory Shakhnarovich \\
Professor Diane Brentari \\
Professor Chris Dyer}

\chair{Professor First Name}

\showlistoftables
\showlistofacronyms

\input{abbr}

\begin{document}




\input{intro}
\input{background}
\input{data}
\input{pipeline}
\input{end2end}
\input{closing_gap}

\input{fs_detection}
\input{search}
\input{translation}
\input{conclusion}

\bibliographystyle{plain}

\bibliography{ref}

\end{document}

%% file: abstract.tex
Sign language, which conveys meaning through gestures, is the chief means of communication among deaf people. Developing sign language processing techniques would bridge the communication barrier between deaf and hearing individuals and make artificial intelligence technologies more accessible to the Deaf community.
Although prior research on sign language processing has made use of studio datasets collected in a controlled environment, these datasets may not accurately reflect the complexity of real-world signing. Recognizing sign language in natural settings presents significant challenges due to factors such as lighting, background clutter, and variations in signer characteristics.

In this thesis, I study automatic sign language processing ``in the wild'', using signing videos collected from the Internet. 
This thesis contributes new datasets, tasks, and methods to advance the state of the art in this area.
Most chapters of this thesis address tasks related to fingerspelling, which is an important component of sign language and yet has not been studied widely by prior work.
In American Sign Language (ASL), fingerspelling accounts for 12-35\% of ASL discourse and is used frequently for content words in conversations involving current events or technical topics. 
In Deaf online media,
transcribing even only the fingerspelled portions could add a great deal of value since these portions are often dense in content words.
In Chapter~\ref{ch:data},
I present three new large-scale ASL datasets ``in the wild'': ChicagoFSWild, ChicagoFSWild+, and OpenASL. ChicagoFSWild and ChicagoFSWild+ are two
datasets of fingerspelling sequences trimmed from raw sign language videos. OpenASL is a large-scale open-domain real-world ASL-English translation dataset based on online subtitled sign language videos. Using ChicagoFSWild and ChicagoFSWild+, I address fingerspelling recognition (Chapter~\ref{ch:pipeline}-\ref{ch:closing-gap}), which consists of transcribing fingerspelling sequences into text. 
To tackle the visual challenges in real-world data, I propose a recognition pipeline composed of a special-purpose signing hand detector and
a fingerspelling recognizer (Chapter~\ref{ch:pipeline}), and an end-to-end approach based on the iterative attention mechanism that allows recognizing fingerspelling from a raw video without
explicit hand detection (Chapter~\ref{ch:e2efsr}). I further show that using a Conformer-based network jointly modeling handshape and mouthing can bring significant gains to fingerspelling recognition (Chapter~\ref{ch:closing-gap}), bringing performance close to that of humans. Next, I
propose two important tasks for building real-world 
fingerspelling-based applications:
fingerspelling detection (Chapter~\ref{ch:fsdet}) and fingerspelling search (Chapter~\ref{ch:fs-search}). For fingerspelling detection, I introduce a suite of evaluation metrics and a new model
that learns to detect fingerspelling via multi-task training. 
To address the problem of searching for fingerspelled keywords or key phrases in raw sign language videos, we propose a novel method that jointly localizes and matches fingerspelling segments to text based on fingerspelling detection. Finally, I
will describe a benchmark for large-vocabulary open-domain sign language translation (Chapter~\ref{ch:slt-wild}). To address the challenges of sign language translation in realistic settings
and without glosses, we propose a set of techniques including sign search as a pretext task
for pre-training and fusion of mouthing and handshape features. 

\newpage

%% file: acknowledgments.tex
First and foremost, I would like to express my deepest gratitude to my advisor, Karen Livescu. In the initial stages of my PhD, when my research experience was little, she mentored me into the field of sign language processing. From grasping the basic concepts of machine learning to delving into the latest advancements in language technologies, I have continuously learned under her mentorship. Karen is always patient, letting me make mistake after mistake, so that I can learn from them.  Whenever I encountered challenging research questions, she never hesitated to provide careful assistance in finding solutions. Beyond that, I’m also grateful to Karen for allowing me the space to explore my interests and work on them, helping me grow as an independent researcher. Her meticulousness and research ethics are values I aim to uphold throughout my research career.

Next, I would like to thank my committee members: Greg Shakhnarovich, Diane Brentari, and Chris Dyer. Greg's comments and questions consistently encouraged me to consider the broader scope of computer vision, providing the inspiration to develop essential visual processing tools. His sense of humor always turned our discussions into enjoyable interactions. Diane's expertise in sign language linguistics has been a guiding light, ensuring that my research remains closely aligned with real-life applications and the interest of Deaf communities. Chris's interdisciplinary viewpoint, bridging linguistics, NLP, and machine learning, has motivated me to explore the core linguistic aspects of sign language processing.

I am grateful to the Sign language Linguistics lab of the University of Chicago, whose assistance has been crucial for my research. The data collection wouldn't have been possible without their help. In particular, I would like to thank Aurora Martinez Del Rio and Jonathan Keane. Aurora has always been patient and helpful with my data inquiries. Jon's thesis has been my go-to for linguistics questions about fingerspelling.

During my PhD, I was fortunate to do internships at Amazon Alexa and Meta AI. These experiences provided me with opportunities to engage in exciting projects within the field of audio and speech research, thereby enriching my industrial experience. I benefit from interactions with Ming Sun, Chieh-Chi Kao, Chao Wang, Spyros Matsoukas, Krishna Puvvada, Viktor Rozgic from Amazon Alexa and Wei-Ning Hsu, Abdelrahman Mohamed, Kushal Lakhotia from Meta AI. In particular, I would like to thank Ming Sun and Wei-Ning Hsu for hosting my internships at Amazon and Meta, respectively.  I am grateful to Ming for overseeing my various projects and organizing the pleasant social events. Wei-Ning's meticulous guidance and constructive advice pushed the boundaries of my intern project beyond my initial expectation. The fellow interns, Gustavo Aguilar,  Srinivas Parthasarathy and Yang Chen also made the internship experience fun and rewarding.

I would like to express my gratitude to professors at TTIC: Kevin Gimpel, David McAllester, Julia Chuzhoy, Madhur Tulsiani, Matthew Walter, Nati Srebro, Avrim Blum. I enjoy interacting with them during coffee breaks and have learned a lot from their fantastic lessons and research talks. I am also thankful to the administrative staff at TTIC: Adam Bohlander, Erica Cocom, Chrissy Coleman, Jessica Jacobson, Mary Marre, and Amy Minick, who made my PhD life easy and smooth in so many ways. I would also like to thank Matthew Turk, the president of TTIC, for his interest in sign language research. It has always been a great pleasure chatting with him during the student workshops.

I want to thank all the fellow students, post-docs and research assistant professors of the speech group at TTIC: Ankita Pasad, Shane Settle, Freda Shi, Ju-Chieh Chou, Chung-Ming Chien, Hongyuan Mei, Kartik Goyal,  David Yunis, Shubham Toshniwal, Qingming Tang, Weiran Wang, Hao Tang and Herman Kamper, for making each group meeting fun and rewarding. Collaborating with them is both enjoyable and highly rewarding to me. The hackathons organized by Ankita and Shubham are among the most unforgettable experiences I had during the PhD years.  Herman and Shane’s presentations are always enlightening. 

I am also thankful to the students at TTIC, including Sudarshan Babu, Nick Kolkin, Takuma Yoneda, Tomoki Tsujimura, Yota Toyama, Shuning Jin, Lifu Tu, Lingyu Gao, Ruotian Luo, Mingda Chen, Jose Marcelo Sandoval-Castañeda, Igor Vasiljevic, Charles Schaff and many more.  The fun social activities such as game nights and holiday parties have made my PhD journey truly wonderful.

Finally, thanks to my family. I am profoundly grateful to my parents for their unconditional support over the years, a debt of thanks that words cannot possibly express. 

%% file: abbr.tex
\acronyms{
\acro{AI}{Artificial Intelligence}
\acro{AP}{Average Precision}
\acro{ASL}{American Sign Language}
\acro{AUSLAN}{Australian Sign Language}
\acro{Acc}{Accuracy}
\acro{Attn-KWS}{Attention-based Keyword Search}
\acro{BLEU}{BiLingual Evaluation Understudy}
\acro{BLEURT}{Bilingual Evaluation Understudy with Representations from Transformers}
\acro{BMN}{Boundary Matching Network}
\acro{BSL}{British Sign Language}
\acro{Bi-LSTM}{Bidirectional Long short-term memory}
\acro{CNN}{Convolutional Neural Network}
\acro{CSL}{Chinese Sign Language}
\acro{CTC}{Connectionist Temporal Classification}
\acro{ChicagoFSWild}{Chicago Fingerspelling in the Wild}
\acro{DGS}{German Sign Language}
\acro{DTW}{Dynamic Time Warping}
\acro{Ext-Det}{External detector}
\acro{FFN}{Feedforward Network}
\acro{FS}{Fingerspelling}
\acro{FSR}{Fingerspelling Recognition}
\acro{FSS-Net}{Fingerspelling Search Network}
\acro{FVS}{Fingerspelling-based Video Search}
\acro{FWS}{Fingerspelled Word Search}
\acro{GAN}{Generative Adversarial Network}
\acro{GCN}{Graph Convolutional Neural Network}
\acro{GMM}{Gaussian Mixture Model}
\acro{GRU}{Gated Recurrent Unit}
\acro{HMM}{Hidden Markov Model}
\acro{HoG}{Histogram of Gradient}
\acro{I3D}{Inflated 3D ConvNet}
\acro{ISLR}{Isolated Sign Language Recognition}
\acro{KSL}{Korean Sign Language}
\acro{LER}{Letter Error Rate}
\acro{LSTM}{Long short-term memory}
\acro{MS-TCN}{Multi-Stage Temporal Convolutional Network}
\acro{Mo-cap}{Motion Capture}
\acro{NAD}{National Association of the Deaf}
\acro{NMS}{Non-Maximum Suppression}
\acro{RC3D}{Region Convolutional 3D Network}
\acro{RCNN}{Region-based Convolutional Network}
\acro{RGB}{Red Green Blue}
\acro{RGB-D}{Red Green Blue-Depth}
\acro{RNN}{Recurrent Neural Network}
\acro{ROI}{Region-Of-Interest}
\acro{ROUGE}{Recall-Oriented Understudy for Gisting Evaluation}
\acro{RPN}{Region Proposal Network}
\acro{ResNet}{Residual Network}
\acro{SCRF}{Segmental Conditional Random Field}
\acro{SGD}{Stochastic Gradient Descent}
\acro{SIFT}{Scale-Invariant Feature Transform}
\acro{SLD}{Sign Language Detection}
\acro{SLP}{Sign Language Production}
\acro{SLR}{Sign Language Recognition}
\acro{SLT}{Sign Language Translation}
\acro{SOTA}{State-of-the-art}
\acro{ST-GCN}{Spatial-temporal Graph Convolutional Neural Network}
\acro{Seq2seq}{Sequence-to-Sequence model}
\acro{TSL}{Turkish Sign Language}
\acro{WER}{Word Error Rate}
\acro{WLASL}{Word-Level American Sign Language Dataset}
\acro{mAP}{mean Average Precision}
\acro{mF1}{mean F1 score}
}

%% file: intro.tex
\chapter{Introduction}
\label{ch:intro}

Artificial Intelligence (AI) is entering our daily life. In particular, speech and language technologies enable AI agents to communicate and converse fluently with humans, leading to intelligent virtual assistants (e.g., Alexa, Siri) that greatly enhance our lives. Primarily based on spoken languages, these emerging technologies do not benefit people who cannot speak or hear. According to the World Health Organization, over 430 million people worldwide have ``disabling'' hearing loss,\footnote{``Disabling'' hearing loss refers to hearing loss greater than 35 decibels (dB) in the better hearing ear~\cite{who-deaf-stats}.} and the number is expected to increase to 500 million by 2050~\cite{who-deaf-stats}.
Sign language, the primary means of communication among deaf people, is an area we must address to make AI technologies widely accessible to Deaf communities.\footnote{The lowercase ``deaf'' refers to the audiological condition of not hearing, while the uppercase "Deaf" refers to a
cultural identity.}
 What are the challenges of building automatic sign language processing technologies in real world scenarios?
 How do we use modern machine learning approaches to tackle the challenges? 
This thesis 
explores my progress in answering these questions in the past few years.

\section{Background}
\label{sec:intro-sign-language}

\subsection{Sign Language}
Sign language is a type of full-fledged natural language that conveys meaning through gestures. 
According to the World Federation of the Deaf~\cite{worldfeddeaf}, there are more than 200 sign languages worldwide.   
In the US, more than 400,000 people~\citep{ethnologue} use American Sign Language (ASL) to communicate.
Sign language is also used by hearing individuals with speech disorders. For example, sign language can serve as an augmentative and alternative tool by speech-language therapists to treat aphasia~\citep{marks2017stitch}.

It is worth noting that each sign language is a distinct language with its own grammar and lexicon, which is not a signing form of the spoken language used in the same geographical area. Though people in the US and UK both use English, American Sign Language (ASL) and British Sign Language (BSL) are two different languages and are mutually unintelligible. 
Sign language is expressed through the hands combined with non-manual elements.
Compared to simple hand gestures occasionally used when people speak, sign languages display more internal complexity~\citep{Mcneill1994hand}. 
The basic units of a sign language are handshape, movement, orientation, location, and non-manual features. Changing one of the above units may make a subtle visual difference but lead to different meanings. For example, the signs GOOD and BAD in ASL are highly similar. Both signs have the same handshape, movement, and location, while their only difference lies in the hand orientation at the end of the sign.
Additionally, non-manual features such as facial expression, mouthing, and body movement or posture, are critical components of sign language.
Non-manual features resemble the role of prosody (e.g., pitch, intonation) in spoken language~\citep{sandler2012visual}.
Moreover, in contrast to spoken languages, where only one sound can be made at a time, sign languages frequently use multiple body parts simultaneously to transmit a message. For example, two hands encoding distinct signs can be used simultaneously~\citep{Vermeerbergen2007Simultaneity}.

The development of sign language technologies can benefit both Deaf and non-Deaf communities by bridging the communication barrier between deaf and hearing individuals.
For instance, an automatic interpretation system between sign languages and spoken languages can enable deaf individuals to express their opinions on online social platforms in their native language, thus allowing them to better engage as equal members of cultural and political discourse.  
With the vastly growing amount of online deaf social and news media,
sign language search can also enable efficient indexing and searching of sign language videos, which are typically associated with little text. Compared to a text-based searching interface,
tools allowing people to express sign by demonstration via video input are more straightforward and friendly to Deaf users. 
Such a video-based interface can also improve search performance for online sign language dictionaries, thus helping students to learn sign language~\citep{alonzo2019effect,bragg2015user,Athitsos2010LargeLP}.
Furthermore, with the advancement of speech technologies, electronic devices and appliances (e.g., phones, smart home devices, self-driving cars) are ubiquitously controlled through voice,  which poses challenges to Deaf users.
Though some Deaf people are able to use spoken language, their speech usually sounds different from hearing speech and is not generally recognizable with existing automatic speech recognition engines~\citep{Glasser2019AutomaticSR}. 
Furthermore, Deaf users generally prefer an interface based on sign language over text input~\citep{Glasser2020Accessibility,Glasser2019AutomaticSR,Rodolitz2019accessibility}. 
For all the reasons mentioned above, developing sign language processing techniques can significantly help to improve the accessibility of digital devices and personal assistants to Deaf communities.

From a scientific perspective, research on sign language processing can contribute to a more comprehensive understanding of language processing. While much of the current research on language modeling focuses on spoken and written languages, the unique characteristics of sign language are often neglected. Studying sign language processing could shed light on how the brain integrates visual information to understand language, which could have broader implications for our understanding of human perception. 
Overall, conducting research on sign language processing is essential to enhance our understanding and modeling of languages.

\subsection{Fingerspelling}
\label{sec:intro-fingerspelling}

\begin{figure}[btp]
    \centering
    \includegraphics[width=\linewidth]{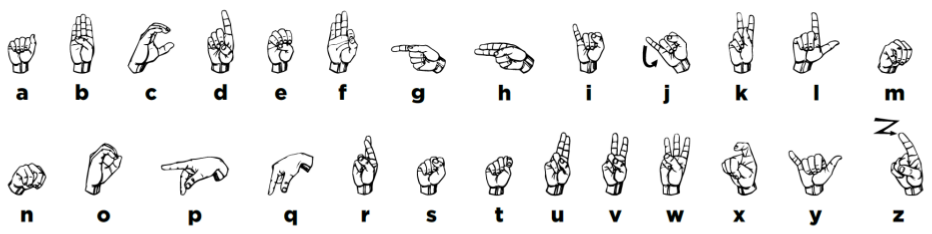}
    \caption{The ASL fingerspelling alphabet, from~\cite{jkean2014thesis}}
    \label{fig:intro-fs-alphabet}
\end{figure}

Most of this thesis will focus on fingerspelling, a component of sign language in which words are spelled out with a sequence of handshapes or hand trajectories corresponding to single letters using the alphabet in a writing system (see Figure~\ref{fig:intro-fs-alphabet}). 
As the manual alphabet commonly involves a limited set of handshapes, it is comparatively easier to learn than regular signs~\cite{padden2003alphabet}. In deaf education, teachers use various fingerspelling-based tools to develop literacy skills for deaf students~\cite{padden2000american,Alawad2018examining}.
In principle, any word in a spoken language can be represented in its manual form with a manual alphabet~\cite{padden2003alphabet}. Thus, fingerspelling is a bridge between sign language and spoken language and is commonly used to borrow spoken language vocabulary into ASL~\cite{padden2003alphabet,jkean2014thesis}.
Fingerspelling is also used for various other purposes, including signing names, emphasis and clarification in bilingual settings and scientific discourses~\cite{jkean2014thesis,padden2003alphabet}.

From a practical perspective, fingerspelling is essential in sign language processing due to its frequent usage in some sign languages.
According to a linguistic study on ASL~\cite{padden2003alphabet}, fingerspelled items account for 12-35\% of the total number of signs in continuous sign segments.
In Australian Sign Language (AUSLAN), $\sim$10\% of lexical items are fingerspelled~\cite{Schembri2007socio}.
In ASL, fingerspelled words are important content words frequently appearing in technical language, colloquial conversations involving names, conversations involving current events, emphatic forms, and the context of codeswitching between ASL and English~\cite{brentari2001native,padden1998asl,Montemurro2018emphatic}.
Thus, transcribing only the fingerspelling portion of online sign language videos can add great value in building a search system for sign language videos. 
Within a sign language translation system, a separate fingerspelling recognition module may remain important and play a role similar to that of transliteration in written language translation~\cite{Durrani2014IntegratingAU}. 

\begin{figure}[htp]
\centering
\begin{tabular}{c}
\includegraphics[width=\linewidth]{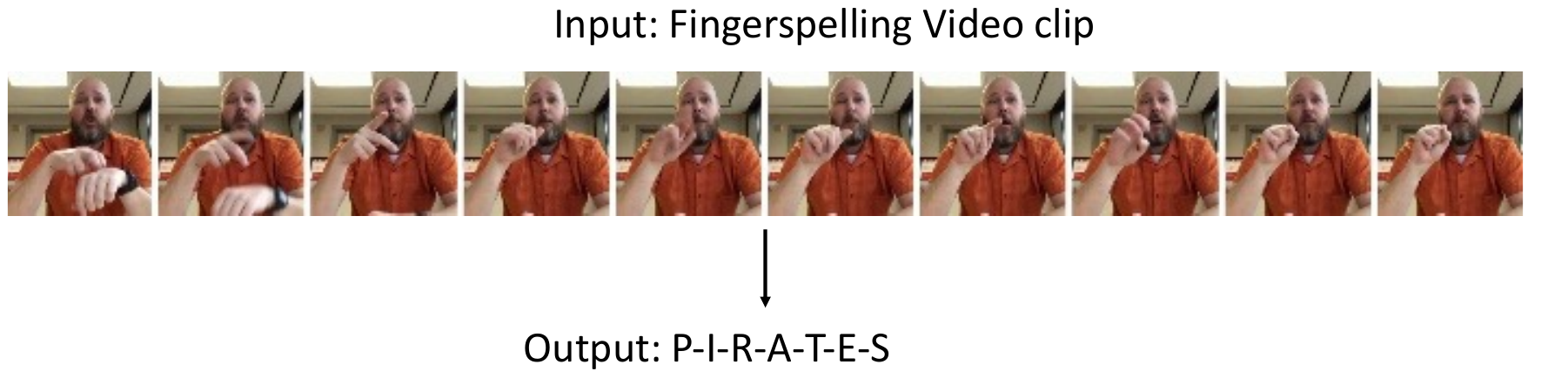}\\ 
\end{tabular}
\caption{\label{fig:fsr-thesis-intro} Fingerspelling recognition task}
\end{figure}

Accurately recognizing fingerspelling can enhance the usability of fingerspelling as an efficient input method for digital devices among Deaf users. In terms of methodology, fingerspelling recognition (see Figure~\ref{fig:fsr-thesis-intro}) has much in common with sign language recognition and translation, as all those tasks can be viewed as sequence transduction. 
Many handshapes used for fingerspelling are also used in other signs.
In addition, the problem of dealing with visual variations (e.g., lighting, appearance) is also largely shared in these three tasks.
The smaller
set of handshapes makes fingerspelling recognition relatively simpler. Nonetheless, due to the high similarity of the three tasks, techniques developed for fingerspelling recognition can be a stepping stone toward sign language recognition or translation.

\section{Overview of Contributions}
\label{sec:intro-contribution}

While a significant amount of prior work is dedicated to tackling sign language processing, most studies have been conducted under very controlled settings (see Chapter~\ref{ch:background} for more details), limiting their applicability in real-world scenarios.
The primary goal of the this thesis is collecting sign language data and developing methods to handle sign language processing tasks in more realistic settings.
In particular, we make contributions in three directions: data collection, task formulation and modeling. The contributions in dataset collection are as follows.

\begin{enumerate}
\item \textbf{In-house annotated fingerspelling dataset: Chicago Fingerspelling in the Wild (ChicagoFSWild)} [Chapter~\ref{ch:data}] We collect the
 first fingerspelling dataset in the wild, ChicagoFSWild. Compared to previous fingerspelling datasets, ChicagoFSWild is much larger (3 times the number of sequences in the previous largest dataset) and {poses several new} challenges including much larger visual variability and an increased number of signers.
  
\item\textbf{Crowdsourced fingerspelling dataset: ChicagoFSWild+} [Chapter~\ref{ch:data}] To further increase
 the scale of fingerspelling data available for researchers, we crowdsource fingerspelling annotations on Amazon Mechanical Turk and collect the largest existing fingerspelling dataset in the wild, ChicagoFSWild+, which is 10 times larger than ChicagoFSWild. Experiments demonstrate that, despite its noisier labels, utilizing ChicagoFSWild+ as supervision can improve recognition performance by a large margin.

\item\textbf{Open-domain sign language translation dataset: OpenASL} [Chapter~\ref{ch:data}]
We collect Open-ASL,  a large-scale ASL-English dataset collected from online video sites. {OpenASL} contains 288 hours of ASL videos in {multiple domains}
from over 200 signers and is the largest publicly available ASL translation {dataset} to date.

\end{enumerate}

\begin{figure}[htp]
\centering
\begin{tabular}{c}
\includegraphics[width=\linewidth]{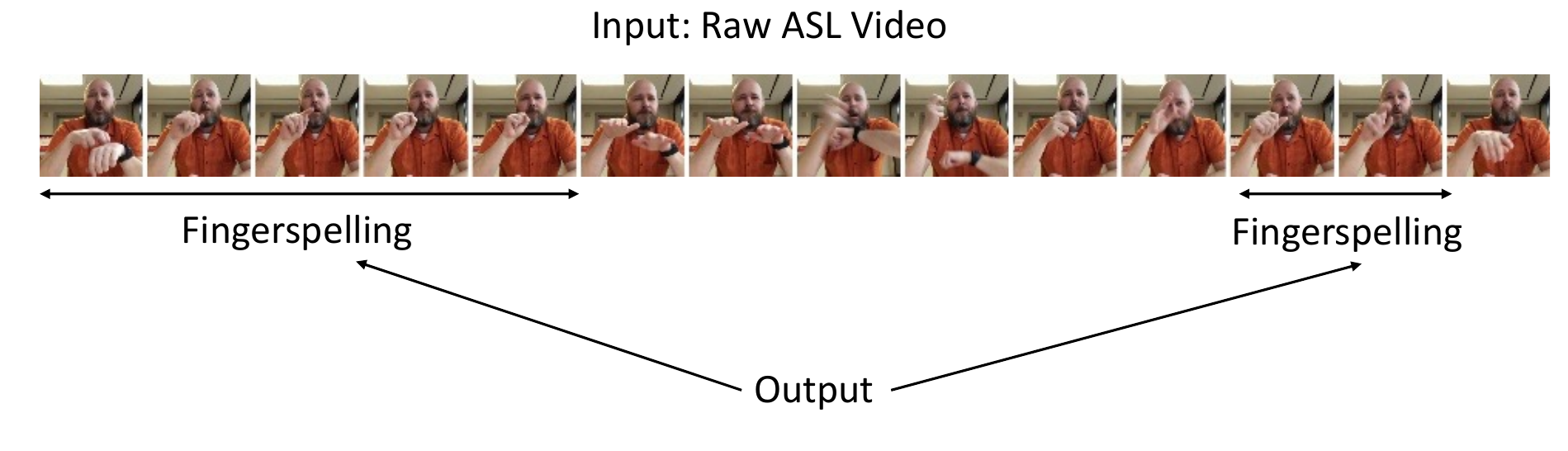}\\ 
\end{tabular}
\caption{\label{fig:fsd-thesis-intro} Fingerspelling detection task}
\end{figure}

\begin{figure}[htp]
\centering
\begin{tabular}{c}
\includegraphics[width=\linewidth]{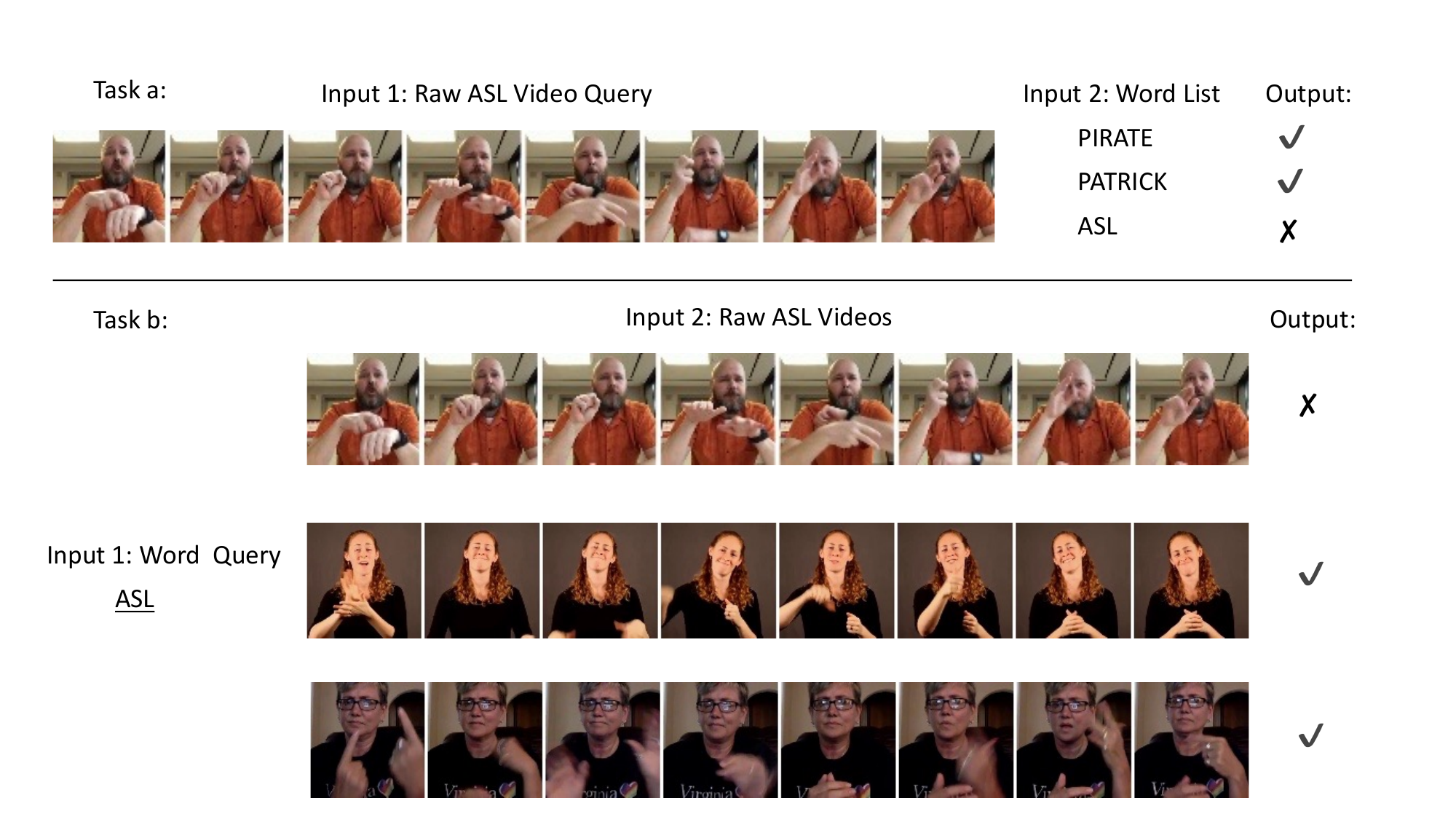}\\ 
\end{tabular}
\caption{\label{fig:fss-thesis-intro} Fingerspelling search and retrieval tasks (task a: Fingerspelled word search, task b: fingerspelling video search). Task a involves identifying all the words that are fingerspelled in a given video, while task b involves searching for videos that include a specific fingerspelled word.}
\end{figure}

The contributions in task formulation are as follows:

\begin{enumerate}
    \item \textbf{Fingerspelling detection} [Chapter~\ref{ch:fsdet}]  Prior fingerspelling recognizers have all been based on pre-segmented fingerspelling video clips. In order to recognize fingerspelling from raw untrimmed ASL videos, we must first detect where fingerspelling starts and ends. We establish the task of fingerspelling detection (see Figure~\ref{fig:fsd-thesis-intro}) and propose several evaluation metrics that take the performance of a downstream recognition model into account.
    \item \textbf{Fingerspelling search and retrieval} [Chapter~\ref{ch:fs-search}] 
    No prior work on fingerspelling-based search and retrieval exists. We define two tasks for open-vocabulary fingerspelling search from untrimmed sign language videos (see Figure~\ref{fig:fss-thesis-intro}): fingerspelled word search (video$\rightarrow$text) and fingerspelling-based video search (text$\rightarrow$video).
\end{enumerate}

\begin{figure}[htp]
\centering
\begin{tabular}{c}
\includegraphics[width=\linewidth]{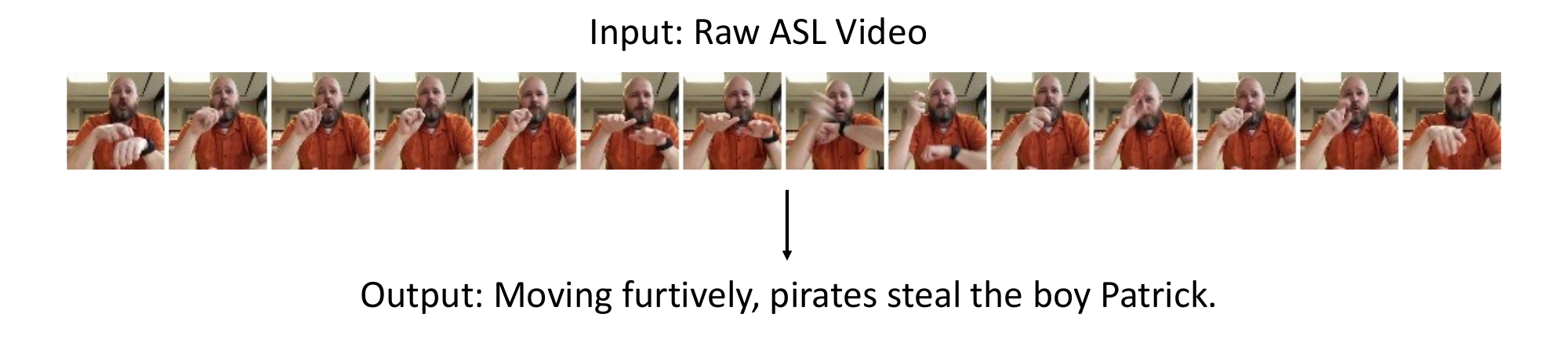}\\ 
\end{tabular}
\caption{\label{fig:slt-thesis-intro} Sign language translation task}
\end{figure}

The modeling contributions are as follows:

\begin{enumerate}
    \item \textbf{Fingerspelling recognition with a signing hand detector} [Chapter~\ref{ch:pipeline}]
    To tackle visual challenges, we propose a recognition pipeline composed of a special-purpose signing hand detector and a sequence model for the fingerspelling recognition for real-life data. We find that letter error rates are higher than in prior work on studio data, which is as expected. As the first attempt at fingerspelling recognition in the wild, this work is intended to serve as a baseline for future work on fingerspelling recognition in realistic conditions.
    \item \textbf{End-to-end fingerspelling recognition based on iterative attention} [Chapter~\ref{ch:e2efsr}]
    All prior work on fingerspelling recognition has relied on hand detection. To reduce the human effort in annotating hands, we propose a new approach, \textit{iterative attention}, which gradually zooms into the hand region of interest (ROI) in the original image. The model is trained end-to-end from raw image frames and does not rely on an external hand detector. Experiments on ChicagoFSWild/ChicagoFSWild+ show that the model outperforms
    prior work by a large margin.
    \item \textbf{Improving fingerspelling recognition with multi-stream Conformer} [Chapter~\ref{ch:closing-gap}] 
    To further boost recognition performance, we propose to incorporate mouthing in fingerspelling recognition.
    Based on the Conformer architecture from prior work in speech recognition, we propose a new multi-stream Conformer model that encodes both the handshape sequence and mouth movement.
    The proposed recognizer achieves a new state-of-the-art for real-world fingerspelling recognition, matching a proficient human student of ASL.
    \item \textbf{Fingerspelling detection via multi-task training} [Chapter~\ref{ch:fsdet}]  To address fingerspelling detection, we propose a new multi-task model which jointly estimates pose, detects fingerspelling, and recognizes it. Experiments on ChicagoFSWild/ChicagoFSWild+ show that the model outperforms baselines derived from related work on similar tasks by a large margin.
    \item \textbf{Fingerspelling search network} [Chapter~\ref{ch:fs-search}] 
    We propose a novel model to tackle fingerspelling search, Fingerspelling Search Network (FSS-Net), which generates temporal proposals via an internal fingerspelling detector and jointly learns the visual and textual embeddings. The model outperforms baselines adapted from related work in video search and retrieval by a large margin. 
    \item \textbf{Real-world sign language translation with multi-stream features and sign search pre-training} [Chapter~\ref{ch:slt-wild}] Prior work on sign language translation (see Figure~\ref{fig:slt-thesis-intro}) relies on glosses.
    To alleviate the need for gloss annotation, which is expensive and time-consuming, we propose a {set} 
of novel techniques including pre-training with spotted signs and fusion of multiple local visual features. Our model improves over existing sign language
translation baselines by a large margin. 

\end{enumerate}

It is important to acknowledge that the sign language translation approach presented in this thesis is still in its early stages and far from being production-ready. To advance the development of this technology, there is an urgent need to collect larger-scale real-world datasets and refine the modeling (see Chapter~\ref{ch:conclusion}). These efforts will also facilitate the study of variation and change in signing, further advancing our understanding of sign languages. In the long run, sustained efforts in this direction will lead to the creation of more sophisticated sign language processing techniques that can facilitate seamless communication between Deaf and hearing communities.

\newpage

%% file: background.tex
\chapter{Background: Sign Language Processing}
\label{ch:background}

This chapter gives an overview of sign language processing. We will discuss existing approaches and their limitations on several sign language processing tasks commonly addressed in the literature.
Sign language processing may broadly refer to a series of topics on using computational methods to study sign languages.
Here we mainly review tasks involving perception and understanding from sign language videos, primarily studied in the field of computer vision.

\section{How to Represent Sign Language}

Before diving into the techniques for processing sign language, it is worth considering how sign language is represented. We discuss this topic from two aspects: recording and notation.

\subsection{Sign language recording}
Video recording of people signing is the most natural medium for representing sign language. Videos are convenient to acquire and capture almost all necessary information for sign language processing. In addition to standard RGB cameras, researchers have also used depth cameras~\citep{oszust2013polish,Zafrulla2011american,ozge2020autsl,ozge2021chaplearn,huang2019attention} to capture RGB-D videos, which incorporate 3D information in signing.
Sign language videos are of low cost to collect and are widely accessible. It is also the only choice under certain circumstances (e.g., online archives). 
However, processing sign language videos is challenging and requires sophisticated computer vision algorithms (e.g., background removal).
To overcome the challenges in visual modeling, prior work has adopted some "intrusive" approach ~\citep{fels1993glove,Lu2010CollectingAM,liang1998realtime,Oz2011AmericanSL,wang2009realtime} to record sign language, where sensors are attached to the signer during data collection. 
As sign language is expressed through the movement of multiple body parts, various sensors are typically needed to record signing information fully. For instance, in~\citep{Lu2010CollectingAM} signers had to wear a full suite of appliances, including gloves, eye-tracker, and motion-capture (mo-cap) bodysuit.
Despite their high precision in capturing body movement, sign language glove-data or mo-cap data are more expensive to collect compared to videos. The intrusiveness also makes glove-based systems generally inconvenient to use for deaf users.

\subsection{Sign language notation}
 Multiple notation systems (e.g., SignWriting~\citep{signwriting}, HamNoSys~\citep{hanke2004hamnosys}) have been invented to document sign language. 
These handwriting systems are commonly based on a combination of iconic symbols for basic units of sign language, such as handshape, location, and movement.
These abstract notation systems are used in deaf education~\citep{floos2002how} and sign language linguistic research~\citep{galea2014signwriting,Eccarius2008handshape}. In sign language processing, prior work (e.g., \citep{Vogler1999towards,koller2013may}) has also explored their usage in sign language recognition.

In addition to handwriting systems, glossing is another commonly used way to represent sign language in writing. Sign language glosses are word(s) from another written language (e.g., English) used to name the signs. Glossing is a way of transcribing sign language instead of translation in the target language. Glossing does not precisely represent sign language as some signs can have multiple glosses. For example, three different glosses: "IMPORTANT", "WORTH" and "VALUE" all map to the same sign in ASL.
Compared to sign language handwriting systems, glosses represent a sign's meaning rather than its visual shape. In sign language processing, glosses have been used widely as labels in datasets~\citep{Dreuw2007speech,Athitsos2008TheAS,forster-etal-2014-extensions,Martnez2002PurdueRA,Camgoz2018neural}. 

Each system of representation has unique advantages and can be useful to different users. The focus of this thesis is on the processing of sign language from videos and is not intended to replace any existing systems of representation.

\section{Sign Language Processing Tasks}

\subsection{Sign Language Recognition}
\begin{figure}[btp]
    \centering
    \includegraphics[width=\linewidth]{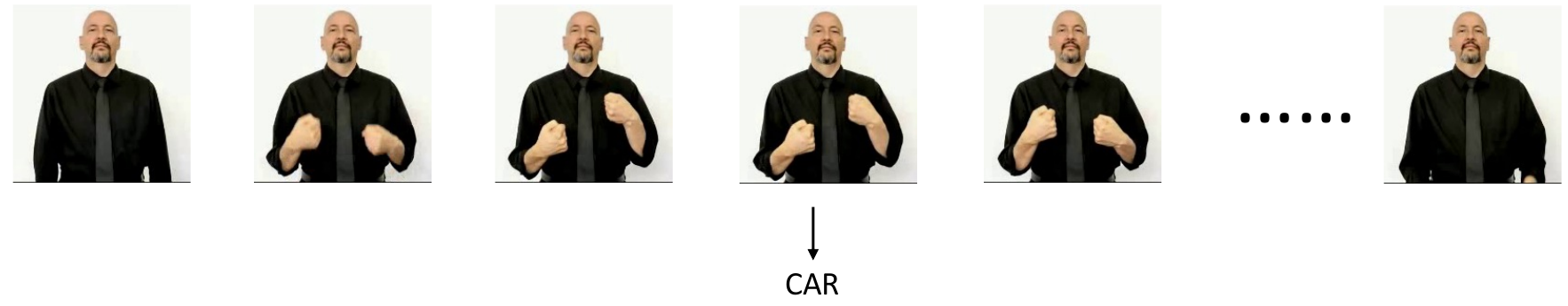}
    \caption{\label{fig:islr-task}Illustration of isolated sign recognition (ISLR). Images are from WLASL dataset~\cite{li2020word}}
\end{figure}

\begin{figure}[btp]
    \centering
    \includegraphics[width=1\linewidth]{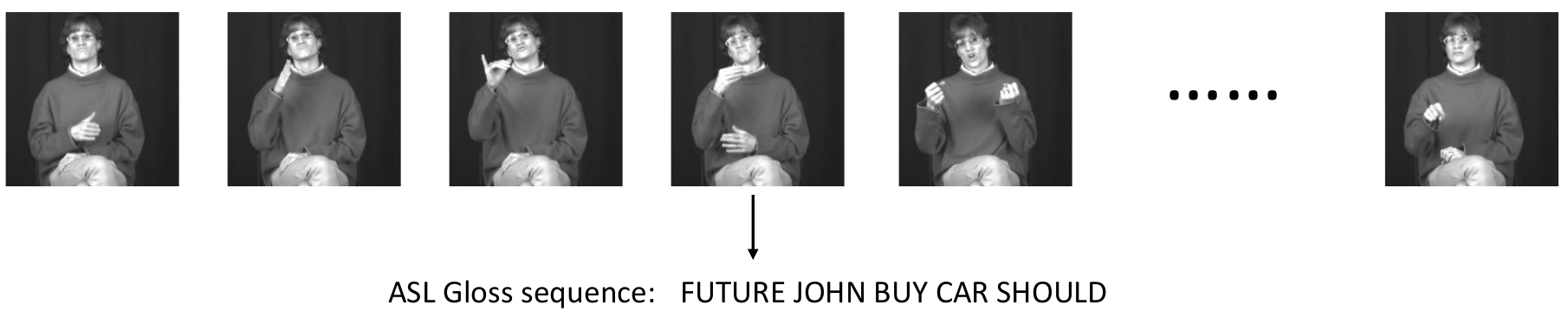}
    \caption{\label{fig:cslr-task}Illustration of continuous sign language recognition (CSLR). Images are from Boston-104 dataset~\cite{Dreuw2007speech}.}
\end{figure}

Sign language recognition (SLR) aims at transcribing a continuous signing stream (e.g., video, glove data) into glosses. It can be used in various scenarios ranging from constructing sign language dictionaries~\citep{Wang2010ASF} to serving as an intermediate step in sign language translation~\citep{Camgoz2018neural,camgoz2020sign} and has been widely studied in the literature. Depending on whether the input stream corresponds to a single sign or a sequence of signs, it can be further categorized into \emph{isolated sign recognition} (a single sign, Figure~\ref{fig:islr-task}) and \emph{continuous sign language recognition} (a sequence of signs, Figure~\ref{fig:cslr-task}). The two tasks are reviewed below, respectively.

\textbf{Isolated Sign Language Recognition}
Research on isolated sign language recognition (ISLR) dates back to the 1980s. In 1988, Tamura et al.~\citep{Tamura1988RecognitionOS} presented an RGB-based system to recognize ten signs from Japanese Sign Language (JSL) with motion images. Most of the early work for ISLR is based on glove or mo-cap data~\citep{Kadous95grasp,majula1995isolated,ouhyoung1996sign,Wang2002AnAB,fang2007large,wang2006expanding}. Another notable characteristic of the early studies is their restricted vocabulary size, which typically varies between 10 and 300. Among the first studies 
on large-vocabulary isolated sign recognition, Wang et al.~\citep{wang2000fast} realized a 5000-sign recognizer for Chinese Sign Language (CSL) with Hidden Markov Model (HMM) based on motion data captured by gloves and a 3D tracker. The system was further improved in~\citep{Wang2002AnAB} by modeling sub-sign units and incorporating a language model.

The vocabulary size in studies using RGB or RGB-D videos~\citep{infantino2007framework,cooper2007sign} was restricted ($<$300 signs), possibly due to inadequate computer vision tools. 
Since 2010, video-based recognition has become the mainstream~\citep{Koller2020QuantitativeSO}. 
In contrast to glove or mo-cap data where spatial coordinates of different body parts of the signer can be precisely estimated, videos must undergo pre-processing steps to localize regions of interest (e.g., hand). A common pre-processing step is thresholding based on skin colors
and has been adopted by much prior work~\citep{Pitsikalis2011advances,cooper2012sign,infantino2007framework,Wang2010ASF,Athitsos2009ADF} for hand segmentation. The video is commonly represented by some human-engineered features, which were fed subsequently to a downstream classifier. Specifically, Wang et al.~\citep{wang2009similarity} used hand trajectories to describe signing and classified 921 distinct ASL signs based on Dynamic Time Warping (DTW). Pitsikalis et al.~\citep{Pitsikalis2011advances} proposed an HMM model based on motion features to model phonetic sub-units. Cooper et al.~\citep{cooper2012sign} also incorporated handshape features using Histogram-of-Gradient (HoG)~\citep{freeman1994orientation} and constructed a classifier predicting linguistically inspired units. The classifiers were combined with a Markov chain to recognize signs. 
Instead of using the whole video stream, Wang et al.~\citep{wang2016sparse} also considered extracting specific posture fragments. They used algorithms in stable marriage problem~\cite{Gale1962College} for sequence alignment to efficiently classify signs under the large-vocabulary setting.

With the advance of deep learning, more work attempt to learn the isolated sign recognizer in an end-to-end fashion while abandoning the feature engineering pipeline. 
In recent approaches for ISLR, the signing videos are directly fed into a visual encoder for classification, and the whole model is trained with cross-entropy loss.
The visual encoder, commonly comprised of stacks of convolutional layers, is usually pre-trained on large-scale external image or video datasets (e.g., ImageNet~\cite{deng2009imagenet}).
For instance, Liao et al.~\cite{liao2019dynamic} proposed a residual network combined with a bidirectional long-short term memory network (ResNet-BiLSTM) that jointly localizes hands, extracts spatial-temporal visual feature and classifies the feature sequence. 
Li et al.~\cite{li2020word} and Joze et al.~\cite{Joze2019MSASLAL} used 3D convolutional neural networks (CNN)~\cite{Carreira2017QuoVA} pre-trained on a large action classification data~\cite{Carreira2017QuoVA} and showed its superiority over hybrid convolutional recurrent models. 
Prior work~\cite{li2020word,Joze2019MSASLAL,Hosain2021HandPG,Hu2021HandModelAwareSL,DeCoster2021IsolatedSR} has also represented video frames as signer pose, which is a skeleton connecting pre-defined human body keypoints estimated from some off-the-shelf pose estimation models (e.g., Openpose~\cite{cao2019openpose}). However, due to errors in pose estimation, an ISLR model based on pose alone usually achieves lower performance than RGB-based models~\cite{li2020word,Joze2019MSASLAL}. To address this issue, Jiang et al.~\cite{Jiang2021SkeletonAM} proposed to fuse multi-modal representation, which incorporates both pose and RGB-based visual feature. 
Hosain et al.~\cite{Hosain2021HandPG} used body pose to guide spatial-temporal feature pooling.
Hu et al.~\cite{Hu2021HandModelAwareSL} learned an intermediate sign representation based on hand pose through regularization from a hand prior model~\cite{romerio2017embodied}.
Despite being scaled up to 1,000-2,000 signs and $>$100 signers, existing ISLR research mainly operates on data from sign language dictionaries~\cite{li2020word,Joze2019MSASLAL} or lexicon projects~\cite{Athitsos2008TheAS,li2020word,Joze2019MSASLAL}. From a practical point of view, ISLR systems may be necessary for certain applications, such as building a sign language dictionary. However, natural signing in daily life usually involves complete sentences or longer utterances instead of isolated individual signs. 

\textbf{Continuous Sign Language Recognition}
Instead of classifying a signing stream corresponding to a single sign, the goal of continuous sign language recognition (CSLR) is to transcribe a continuous signing stream to a sequence of glosses.
A large body of work on continuous sign recognition~\cite{starner1995visual,nam1997recognition,liang1998realtime,Assan1997video,Dreuw2007speech,Koller2015ContinuousSL,Koller2016DeepSH,koller2017resign} is based on HMM, adapted from speech recognition. Compared to ISLR, CSLR is complicated by multiple sources
of variation, such as coarticulation and social factors (e.g., dialect, age, gender).
Furthermore,
there is often extra movement inserted between adjacent signs, which is also known as \emph{movement epenthesis}~\citep{liddel1989american} (e.g., between "FATHER" and "STUDY" in the sentence of "FATHER READ" in ASL).
To cope with the above phenomenon, Vogler and Metaxas~\citep{volger1997adapting} adopted context-dependent HMM to explicitly model coarticulated signs and the transitional movement. 
Movement epenthesis has also been exploited in~\citep{gao2004transition}
to segment continuous signing data into individual signs to train a visual model. 
In addition to transitional movement, prior work has also used other motion features such as time-varying parameter~\citep{liang1998realtime} and the hand configuration change~\citep{sagawa2000method} to detect signing boundaries in the input, which are utilized to enhance the learning of a sign model.
The visual feature in early work on CSLR is mostly human-engineered, such as 3D-HoG~\citep{Forster2013ImprovingCS} or geometric feature~\citep{Zahedi2006GeometricFF}, similar to ISLR. 

Neural networks are currently the dominant approach for CSLR. Commonly used models for visual feature extraction 
include hybrid  Conv-RNN~\cite{Cui2017RecurrentCN,koller2017resign},
3D-CNN~\cite{huang2018video,pu2018dilated}.
Despite the architectural difference, the overall iterative training paradigm arising from HMM-based models in early work is still commonly used. In general, such a training process consists of labeling frames based on a trained recognizer and training a new recognizer based on frame labels. For example, Koller et al.~\cite{koller2017resign} trained a hybrid Conv-RNN-HMM recognizer with Expectation-Maximization (EM)~\cite{Dempster1977MaximumLF}. Based on a recognizer trained with Connectionist Temporal Classification (CTC)~\cite{Graves2006ConnectionistTC}, Cui et al.~\cite{Cui2017RecurrentCN} further fine-tuned the convolutional and recurrent layers  separately with pseudo-labels generated by the original model. Pu et al.~\cite{pu2019iterative} trained a recognizer based on sequence-to-sequence model~\citep{Cho2014LearningPR} and applied DTW to produce frame labels to fine-tune the convolutional layers in the encoder. The iterative training approach has shown to be effective in training the visual module of a recognizer.  
Meanwhile, end-to-end training is gaining increasing attention~\citep{Cheng2020FullyCN,Min2021VisualAC,niu2020Stochastic,zuo2022c2slr} due to its simple training pipeline. Specifically, Min et al.~\cite{Min2021VisualAC} proposed a CTC-based recognizer
jointly trained with an alignment loss between features from the visual subnetwork and sequential model.
Inspired by HMM, Niu and Mak~\cite{niu2020Stochastic} modeled each gloss with multiple states following a learned probability distribution and proposed a set of data augmentation techniques to enhance visual feature learning. Zuo and Mak~\cite{zuo2022c2slr} incorporated pose distillation and visual-sequential feature alignment as auxiliary tasks for training a recognizer, which outperformed many methods based on iterative training.
Most recently-proposed CSLR models are evaluated on a few public benchmarks (e.g., Phoenix-2014~\cite{Koller2015ContinuousSL}, Phoenix-2014T~\cite{Camgoz2018neural}, CSL-Video~\cite{huang2018video}), which are constrained in various aspects, including domain (e.g., weather forecast), vocabulary size ($\sim$ 1K) and the number of signers ($<$10).
Typically, the current state-of-the-art method~\cite{zuo2022c2slr} achieves $\sim20\%$ word error rate (WER) under such a setting.
A continuous sign language recognizer commonly serves as an upstream component for sign language translation~\cite{Camgoz2018neural,camgoz2020sign}. Though most existing work treats CSLR as a standalone task, it is worthwhile to study this task within a whole translation system.

\subsection{Sign Language Translation}
\begin{figure}[btp]
    \centering
    \includegraphics[width=1\linewidth]{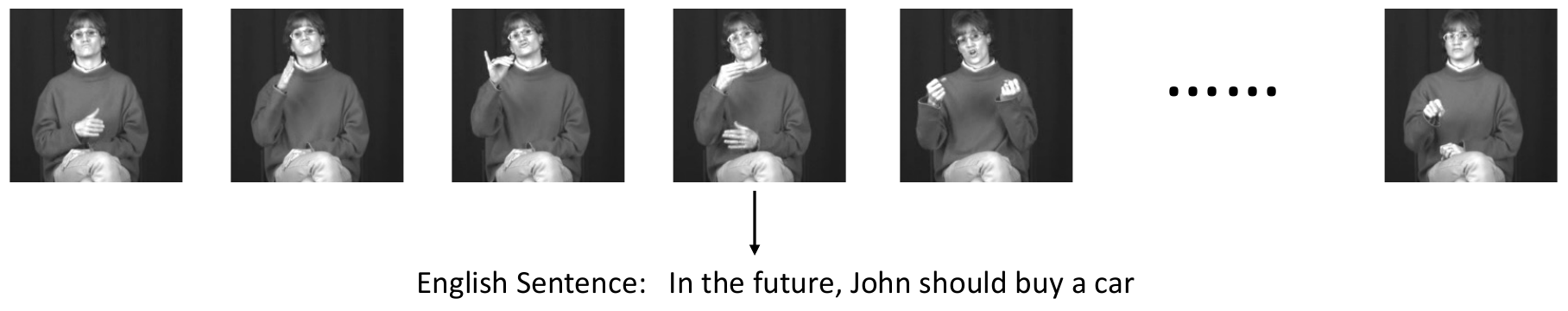}
    \caption{\label{fig:slt-task}Illustration of sign language translation (SLT). Images are from Boston-104 dataset~\cite{Dreuw2007speech}.}
\end{figure}

We use sign language translation (SLT) to refer to translating sign language into spoken language, following the convention used in prior work~\citep{Yin2021IncludingSL,Camgoz2018neural}. Translating from spoken language into sign language is called sign language production, which will be reviewed in Section~\ref{sec:background-other-tasks}.
 
In contrast to CSLR, the signing stream and text in spoken language are no longer monotonically aligned. Thus an SLT model must learn the mapping between the signs and text in spoken language in addition to a proper visual representation.
Computational linguists studied SLT based on the written form of sign language~\citep{Morrissey2008DataDrivenMT,Morrissey2005example,Morrissey2010BuildingAS,othman2012english}. For example, Othman et al.~\cite{othman2012english} built a parallel corpus between ASL glosses and English written text using a rule-based approach, which was used to study gloss-to-text translation in~\citep{Yin2020BetterSL}.
Translating sign language videos into text has many practical applications (e.g., building conversational agents for Deaf people).
The task of translating a sign language video stream into text in spoken language was introduced in~\citep{SystemBritta1999TowardsAA}. Specifically, Britta et al.~\cite{SystemBritta1999TowardsAA} proposed a system consisting of a CSLR module transcribing DGS video into glosses (i.e., sign$\rightarrow$gloss) and a statistical machine translation module translating gloss sequence into German (i.e., gloss$\rightarrow$text). Cascading a sign recognizer and a gloss-to-text machine translation model is widely adopted by much follow-up work. Based on the same framework, Dreuw et al.~\cite{dreuw2009enhancing} enhanced the translation module by using visual features. Schmidt et al.~\cite{Schmidt2013UsingVR} further incorporated a lip-reading model into the whole system. In addition to RGB-based video, depth camera has also been explored in~\citep{Chai2013SignLR} in SLT for hand tracking. 

Early work in video-based SLT is restricted in vocabulary size ($\sim$100 signs). Camgoz et al.~\cite{Camgoz2018neural} proposed a large-vocabulary DGS-German translation video corpora RWTH-PHOENIX-Weather 2014T (Phoenix14T), based on sign language interpretations of German weather forecasts.
Phoenix14T, covering 1066 DGS signs and 2887 German words, has been used as a community benchmark for SLT. Under Phoenix14T, Camgoz et al.~\cite{Camgoz2018neural} studied several setups including the two-stage setting (sign$\rightarrow$gloss + gloss$\rightarrow$text) and direct sign-to-text translation using a RNN-based sequence-to-sequence model.
The cascaded system benefited from the direct supervision provided by glosses~\cite{camgoz2020sign} and hence outperformed the sign-to-text model by a large margin. 
Therefore, most follow-up work~\citep{camgoz2020sign,Yin2020BetterSL,Zhou2021ImprovingSL,Kan2022SignLT,Chen2022ASM} opted for gloss during training to achieve high performance. For instance, Camgoz et al.~\citep{camgoz2020sign} trained a Transformer-based encoder-decoder architecture~\citep{Vaswani17transformer} jointly with CTC~\citep{Graves2006ConnectionistTC} and cross-entropy loss. Local visual cues (e.g., face, hand) have also been incorporated into the system to improve translation quality~\citep{camgoz2020multi,Yin2020BetterSL}. Chen et al.~\cite{Chen2022ASM} studied transfer learning for SLT and proposed an end-to-end translation model outperforming the cascaded system, though gloss is still required for training.
Despite its popularity, Phoenix14T is not a realistic dataset 
due to its limited vocabulary size and domain-specific nature.
The real-life scenario is highly challenging for existing SLT methods. In a wide-domain BSL-English translation task~\citep{Albanie2021bobsl}, a standard SLT approach based on Transformers only achieved a 1.0 BLEU-4 score~\citep{Papineni2002BleuAM} and struggled to capture the meaning of the whole sentence. 
Overall, SLT research is still in its infancy and a video-based translation system satisfying real-world needs is not readily available.

\subsection{Fingerspelling Recognition}
\label{sec:sign-language-fingerspelling}

%
%
%
%
%
%
%

\begin{figure}[btp]
    \centering
    \includegraphics[width=1\linewidth]{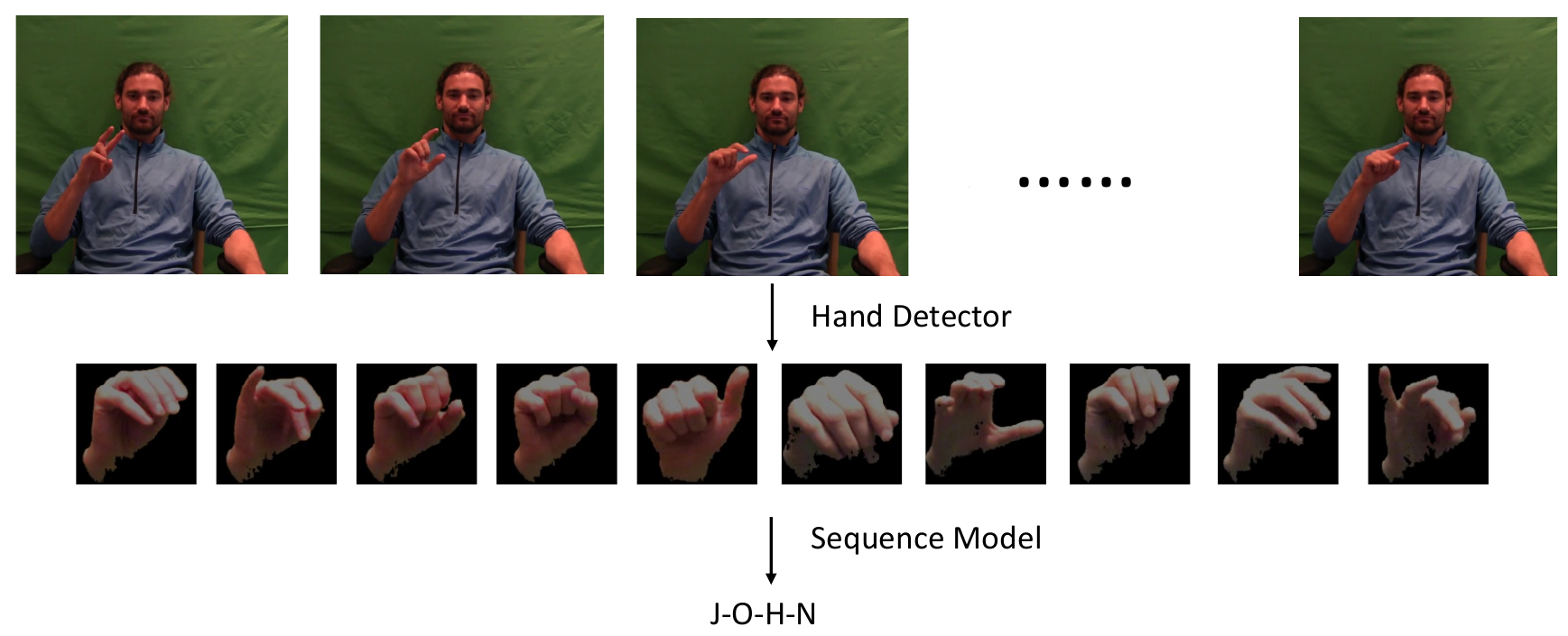}
    \caption{\label{fig:slt-task}Illustration of fingerspelling recognition (FSR). Images are from ChicagoFSVid dataset~\cite{Kim2017LexiconfreeFR}.}
\end{figure}
Fingerspelling Recognition (FSR) aims at transcribing a fingerspelling video clip into text. 
FSR shares many similarities with SLR as both tasks can be viewed as sequence transduction. %
In some prior work (e.g., \cite{liang1999realtime,Wang2010ASF,Athitsos2009ADF,Dreuw2007speech}), the employed data included a small number of fingerspelling instances, and the fingerspelled signs were usually treated in the same way as regular signs. 
Compared to SLR in general, FSR is a relatively simpler task given its limited set of handshapes. 
However, FSR presents its own challenges as it involves quick and small motions that can be highly coarticulated. Some letters have similar handshapes that can only be distinguished from fine-grained visual cues. For example, the fist-like handshape is used across the 3-letter group of (\emph{A}, \emph{S}, \emph{T}) and 2-letter pair of (\emph{M}, \emph{N}), while the only difference among those letters primarily lies in the position of the thumb. 
The recognition is further complicated by other factors in daily signing such as variance in hand appearance or camera perspective (see Chapter~\ref{ch:data} for qualitative examples), making FSR extremely challenging.

Given the significance of fine-grained details in the hand region to FSR, much prior work is dedicated to using special-purpose sensors (e.g., gloves~\cite{Lamari1999HandAR,Oz2005recognition}) to capture hand data for FSR. Oz and Leu~\cite{Oz2005recognition} developed a system for classifying static fingerspelling handshapes based on Cyberglove and motion tracker. The sensor data, including finger joint angles, hand position, and motion, are directly fed into a feedforward neural network for classification. Color gloves, which facilitate hand tracking, have also been used to acquire fingerspelling data~\cite{Lamari1999HandAR,wang2009realtime} and can boost FSR by enabling 3D hand pose estimation~\cite{wang2009realtime}.
To circumvent the inconvenience caused by the intrusiveness in glove data, researchers have also resorted to depth sensors~\cite{Keskin2012HandPE,pugeault,zhang2015histogram,Feris2004ExploitingDD} to capture fingerspelling data. These tools enable the extraction of depth information for hands, which increases the robustness of hand tracking to illumination or color variance. 

Compared to the fingerspelling data from special equipment, RGB-based videos are more widely accessible. However, video processing requires more sophisticated computer vision tools.
To ease pre-processing (e.g., hand detection), prior work on video-based FSR mainly uses data collected under a controlled environment (e.g., with uniform background). 
Hand detection is usually based on skin color via a simple heuristic procedure~\cite{Lockton2002RealtimeGR,Goh2006DynamicFR,Ricco2009accv,Kim2017LexiconfreeFR} or Markov Random Field (MRF)~\cite{Park2005vision,Liwicki2009automatic}. 
The detected hand is commonly represented with human-engineered features, which are fed to downstream recognition. 
Specifically,
Lamar et al.~\cite{Lamari1999HandAR} segmented the hand by color thresholding and described the hand posture by the position of its regions obtained from clustering.
Park et al.~\cite{Park2005vision} represented the hand with the direction of hand boundaries and  constructed a sign classifier with DTW for Korean Sign Language (KSL) fingerspelling.  Similarly,  Lockton and Fitzgibbon~\cite{Lockton2002RealtimeGR} classified 46 ``hand'' gestures, including the ASL fingerspelling handshapes, with hand masking and template matching. Pugeault and Bowden~\cite{pugeault} proposed a fingerspelling handshape classifier using Garbor filters and random forests. 

Apart from the classification of static handshapes, FSR from a continuous signing stream is more practical and has also been widely studied in the literature~\cite{Goh2006DynamicFR,Ricco2009accv,Liwicki2009automatic,Kim2012AmericanSL,Kim2013SCRF,Kim2016Adaptation,Kim2017LexiconfreeFR,Shi2017MultitaskTW}. While the recognition models for FSR share many similarities to CSLR, FSR focuses more on handshape modeling. The sequence models for FSR primarily rely on HMM~\cite{Goh2006DynamicFR,Liwicki2009automatic,Ricco2009accv,Kim2012AmericanSL}, adapted from speech recognition. Commonly used hand features include HoG~\cite{Liwicki2009automatic,Kim2013SCRF,Kim2017LexiconfreeFR}, the scale-invariant feature transform
 (SIFT) feature~\cite{Lowe1999ObjectRF}, or hand silouhettes~\cite{Ricco2009accv}. One of the notable restrictions of early work in FSR is its limited vocabulary size (20-100). In addition, the allowed fingerspelling sequences are usually constrained by a given small lexicon. Under such a setting, several prior works have achieved high recognition accuracy. With a 100-word vocabulary, Liwicki and Everingham~\cite{Liwicki2009automatic} proposed an HMM-based recognizer and reached 98.6\% word accuracy on 1000 BSL fingerspelling sequences from a single signer. A similar model in~\cite{Ricco2009accv} achieved word accuracy of 92.6\% for ASL fingerspelling when the vocabulary size is 86,
Kim et al.~\cite{Kim2012AmericanSL} first studied lexicon-free FSR systematically and found that unconstraining words to a lexicon leads to a significant drop in performance. Specifically, a tandem-HMM model trained with phonological features can achieve $<$3\% letter error rate (LER) when the allowed fingerspelling sequence is constrained to a 300-word vocabulary, while the error rate increases to over 10\% in the lexicon-free setting. 
The follow-up studies~\cite{Kim2013SCRF,Kim2017LexiconfreeFR} further improved lexicon-free FSR by using segmental conditional random fields (SCRF) and lowered the LER to $\sim 8\%$  when the model is trained and tested on the same signer.
In addition to the constraints on vocabulary, signer dependence is another factor affecting FSR performance.
Most prior studies follow a signer-dependent setup, where the training and testing data came from the same signer.
Signer variation, including speed, hand appearance, and hand motion variation have a significant impact on FSR performance. 
In~\cite{Kim2017LexiconfreeFR}, the gap in LER between an SCRF-based recognizer in the signer-dependent and the signer-independent setting is higher than $40\%$. To bridge the gap between the two scenarios, Kim et al.~\cite{Kim2016Adaptation} suggested using a small amount of labeled data from the test signer for adaptation.
Shi et al.~\cite{Shi2017MultitaskTW} also explored sequence-to-sequence models for FSR and demonstrated their superiority over tandem HMM or SCRF for signer-independent FSR on the same benchmark.  

Overall, existing work on FSR is conducted under controlled settings, limited in terms of visual conditions, vocabulary size, and the number of signers. Many fingerspelling datasets used in prior work (e.g., \cite{Park2005vision,Feris2004ExploitingDD,Liwicki2009automatic}) are private, hindering direct comparison between different FSR models. Despite being publicly available, the dataset introduced in~\cite{Kim2017LexiconfreeFR} is collected in lab conditions and is very different from real-world fingerspelling. Furthermore, prior FSR approaches are all built upon pure fingerspelling sequences, which are not directly applicable to fingerspelling transcription from raw sign language videos.

\subsection{Other Tasks}
\label{sec:background-other-tasks}

\textbf{Sign Language Production}
Sign Language Production (SLP) consists of translating a spoken language utterance into a sign language video. As it is only loosely connected to the tasks addressed by this thesis, we only briefly review existing SLP techniques. See~\cite{Rastgoo2021SignLP,Bragg2019SignLR} for a comprehensive survey.
A traditional approach for SLP is sign language Avatar~\cite{Bangham2000Virtual,Cox2002TessaAS,Kipp2011Sign,McDonald2015AnAT}. Avatars are animated computer graphics models generated from motion plans inferred from the spoken language~\cite{Bragg2019SignLR}. Avatar-based SLP approaches~\cite{Karpouzis2007educational,Elliott2008Linguistic} usually require a front-end module translating the spoken language into sign language notated by some handwritten forms (e.g., HamNoSys~\cite{hanke2004hamnosys}). Building sign language Avatars typically involves a series of complex steps and much human intervention~\cite{Bragg2019SignLR}. Furthermore, sign language Avatars tend to have unnatural movement and under-articulation, leading to loss of important meaning~\cite{Stoll2020text2sign}. Hence, it has not received wide popularity among Deaf community~\cite{Stoll2020text2sign}. 
There has been much recent work that tackles SLP with deep-learning approaches. For example, Saunders et al.~\cite{saunders2020progressive} used pose skeleton as the intermediate representation of a sign language video and built a two-step German-DGS translation system. The model employs a sequence-to-sequence model to translate German into DGS glosses, mapped to DGS skeletons via a look-up table~\cite{Stoll2018SignLP} or a second sequence-to-sequence model~\cite{saunders2020progressive}. Based on the pose, Saunders et al.~\cite{Saunders2020EverybodySN,Saunders2022signing} extended the framework to generate raw RGB sign language videos with a Generative Adversarial Network (GAN)~\cite{Goodfellow2014GenerativeAN}.  
Despite being able to produce more realistic sign language videos, existing methods are restricted by the domain and number of signers. 
As many SLP models~\cite{saunders2020progressive,Stoll2018SignLP,Stoll2020text2sign,Saunders2020EverybodySN,Saunders2022signing} are built upon the same datasets as SLT, collecting large-scale general real-world SLT datasets is also expected to bring gains in SLP.

\textbf{Sign Language Detection} Sign language detection (SLD) aims at identifying whether a video clip is signing or not. It has various applications in real-world sign language applications, including video tagging for online video sites (e.g., YouTube) or empowering deaf users on videoconferencing platforms~\cite{Moryossef2020sld}. 
Moreover, SLD can serve as one module in a captioning system for general videos involving sign language to trigger the downstream SLT/SLR models automatically.
Existing methods for SLD are based on human body pose~\cite{Moryossef2020sld} or raw image frames~\cite{Monteiro2012DesignAE,shipman2015towards,Shipman2017SpeedAccuracyTF,karappa2014detection,Borg2019SignLD}. For pose-based SLD, the detector is built upon human body pose estimated by a separate pose estimation module. Specifically, Moryossef et al.~\cite{Moryossef2020sld} used Openpose~\cite{cao2019openpose} to extract keypoints and conducted frame classification with an LSTM~\cite{Hochreiter1997long} network on a video clip represented by keypoint motion. The method is evaluated on sign language videos collected in a studio environment, which lack interference and distractions from non-signing instances. SLD on more realistic sign language videos has also been studied in prior work~\cite{Borg2019SignLD}.
Monteiro et al.~\cite{Monteiro2012DesignAE} mined 192 online videos containing confounding non-signing videos, including gesturing presenters or persons with moving arms and heads. Through an SVM classifier based on several types of movement features, they observed that motion relative to the face is effective for determining whether a video corresponds to signing or not. This detection method was further improved in~\cite{shipman2015towards,Shipman2017SpeedAccuracyTF,karappa2014detection} by incorporating more fine-grained motion information (e.g., the distance between body parts).
Borg and Camilleri~\cite{Borg2019SignLD} collected a large-scale SLD dataset from 1,120 YouTube videos and proposed a two-steam fusion approach based on the convolutional gated recurrent units (GRU) for classification.
In general, SLD is relatively under-studied. Existing approaches for SLD are often developed and evaluated as a standalone model. How to integrate SLD in an end-to-end SLR or SLT system has not been studied yet.

\textbf{Sign Language Segmentation} Sign language segmentation (SLS) generally refers to detecting the time boundaries of some meaningful units in a continuous signing stream. The definition of units varies in different studies. In the context of CSLR, many recognizers (e.g., HMM-based models~\cite{Koller2015ContinuousSL,Koller2016DeepSH}) can produce frame-level gloss labels through forced alignment. In those prior works, the gloss segmentation commonly serves as a tool to obtain intermediate pseudo-labels for training a sequence-level recognizer and is not evaluated on its own.
Much prior work~\cite{Farhadi2006AligningAF,Alon2009AUF,Santemiz2009automatic,Buehler2009learning,albanie2020bsl1k} in SLS involves discovering boundaries of individual signs based on sentence translation of a continuous signing stream. 
For example, Farhadi and Forsyth~\cite{Farhadi2006AligningAF} aligned 31 English words to its ASL signing from a $\sim$1-hour ASL movie accompanied by an English translation based on HMM and a discriminative word model. Similarly, Alon et al.~\cite{Alon2009AUF} proposed a spatial-temporal gesture model and temporally spotted 3 signs from an ASL video clip of $\sim$15 minutes.
Along this line of research, prior work has also explored using sign language videos with weakly aligned subtitles from television programs. Based on a template matching method, Santemiz et al.~\cite{Santemiz2009automatic} proposed to fuse DTW with multi-modal HMMs sequentially and spotted 40 signs from $\sim$2 hours of Turkish Sign Language (TSL) news broadcast videos.
Buehler et al.~\cite{Buehler2009learning} used multiple-instance learning to spot 210 BSL signs from 10.5 hours of BBC sign language videos. 
SLS has also been tackled under a category-agnostic setting which aims at border discovery and does not require assigning semantic labels for each sign. 
Specifically, Farag and Bull~\cite{Farag2019learning} identified whether a frame is an active signing frame or a transitional frame in a continuous signing video based on random forests and geometric features from the pose skeleton. Renz et al.~\cite{Renz2021signsegmentation} studied the same task for BSL and DGS signing streams with multi-stage temporal convolutional neural networks. To address the data scarcity in SLS, Renz et al.~\cite{Renz2021signsegmentation_b} further leveraged unlabeled data through a pseudo-labeling framework to boost segmentation performance.  
Bull et al.~\cite{Bull2020AutomaticSO} addressed a similar task of segmenting online subtitled signing video into sentence-like units based on spatial-temporal graph convolutional networks (ST-GCN).
Currently, there lacks a unified definition of units in SLS, some of which are not inherent to sign language (e.g., English word as the unit in~\cite{Farhadi2006AligningAF}). Whether a well-performing segmentation model can benefit other related tasks such as CSLR or SLT requires more investigation.   

\section{Summary}

In this chapter, we gave an overview of sign language processing.  
A large amount of work tackled the problem from various perspectives, including recognition, translation, segmentation. 
However, existing approaches are mainly studied under a controlled setting and are restricted in multiple aspects such as vocabulary size, signer diversity, and visual conditions. 
Though models have been developed and improved for specific tasks, 
each task is commonly treated independently instead of as one component in an end-to-end system. How a task-specific model affects its downstream task and hence the performance of a whole sign language processing system is largely understudied.
The above limitation hinders building sign language technologies that directly serve real-world use cases.

%% file: data.tex
\chapter{Data}
\label{ch:data}

This chapter describes three large-scale real-life sign language video corpora collected from video sites: ChicagoFSWild~\citep{shi2018american}, ChicagoFSWild+~\citep{shi2019fingerspelling} and OpenASL~\citep{shi2022open}. ChicagoFSWild and ChicagoFSWild+, annotated respectively by in-house annotators and crowdsourced workers, consist of fingerspelling only. OpenASL, which is based on online subtitled ASL videos, contains full signing streams and is proposed for sign language translation. 

In this chapter, our focus is primarily on comparing our proposed datasets to prior data that were collected in a studio-like environment.
We highlight the significance of the variations present in sign language videos.
It's worth noting that carefully recorded sign language videos are highly valuable assets for both linguistic research and sign language processing tasks, such as SLP. To achieve success in recognition problems, it's essential to include real-life data with greater variability. 

\section{Existing Datasets}
\label{sec:existing-data}

\input{table-tex/data/continuous-data}

\input{figure-tex/data/prior-data}

Table~\ref{tab:continuous-datasets} summarizes the statistics of typical public datasets of continuous signing.\footnote{Typically, many datasets of continuous signing include written language translations. Therefore, we only review the translation datasets here.} The limitation of those datasets can be categorized into the following aspects.
(\romannumeral 1) The collection environment is heavily controlled. Majority of the datasets (e.g., Boston-104~\cite{Dreuw2007speech}, RVL-SLLL~\cite{Wilbur2006purdue}, NCSLSGR~\cite{Neidle2012ChallengesID}) are collected in a studio-like environment, where signers are required to sign the content of a desired corpus. 
Figure~\ref{fig:data-existing-datasets} shows typical image frames from such datasets. Compared to natural day-to-day signing, the data involves much less visual variability including more uniform camera perspective, simpler background, less motion blur.
More realistic datasets (e.g., Phoenix14T~\cite{Camgoz2018neural}), typically collected from sign language interpreted sources in television programs, is still fairly limited in visual variability (see Phoenix14T in Figure~\ref{fig:data-existing-datasets}). 
(\romannumeral 2) Many datasets are specific to one domain, which one can tell from their restrictive vocabulary size. For example, the popular SLT benchmark Phoenix14T~\cite{Camgoz2018neural} is from DGS-interpreted weather forecasts.
The KSL dataset KETI~\cite{ko2019neural} only consists of 105 KSL-interpreted sentences used in emergency situations (e.g., "Help me", "The house is burning").  Note the domain restriction has a huge impact on the model performance. In~\cite{Yin2021IncludingSL}, the same translation model is performing $\sim$6 times worse (in BLEU score) in an open-domain benchmark (public DGS~\cite{Hanke2020ExtendingTP}) compared to a special-domain benchmark (Phoenix14T~\cite{Camgoz2018neural}).
(\romannumeral 3) Most datasets are limited in terms of signers. 
Target signs are typically performed by experts or native signers, hence the different levels of fluency in real-life signing are not reflected in those datasets. Furthermore, due to the generally small number of signers involved, a model developed using those data tend to have difficulty in generalization to unseen signers. In~\cite{koller2017resign}, the same CSLR recognizer is 20\% (absolute) worse in the signer-independent setting compared to the closed-signer scenario.
(\romannumeral 4) Overall, the sizes of existing datasets are small. The largest continuous signing data are of 1.4K hours, while most datasets are fewer than 100 hours. This poses a potential challenge to developing sophisticated deep learning models, which typically require a large amount of training data.
(\romannumeral 5) Almost all datasets listed in Table~\ref{tab:continuous-datasets} are from interpreted sources. Those data are either from television channels that provide sign language interpretation (e.g., BOBSL~\cite{albanie2020bsl1k}, Phoenix14T~\cite{Camgoz2018neural}), or formed of signers requested to sign the desired content (e.g., Boston-104~\cite{Dreuw2007speech}, How2Sign~\cite{Duarte2021how}). This differs from signing in daily life (e.g., conversations), that are self-generated instead of prompted.

\input{table-tex/data/fs-data}

In Table~\ref{tab:fs-datasets}, we show a summary of fingerspelling-specific data. 
Many existing fingerspelling datasets (e.g., JSL-fingerspelling~\cite{Kwolek2016recognition}) only consist of static hand gestures. 
Although such data are useful in specific applications like building gesture-based human-computer interaction systems, real-world sign language media almost always involve continuous signing. Furthermore, some letters are fingerspelled (e.g., "Z" in ASL) with a hand trajectory instead of a static handshape. Thus, a continuous fingerspelling dataset is more valuable to real-world sign language processing. Thus hereafter, we use "fingerspelling dataset" to refer to the continuous fingerspelling dataset specifically.
Existing fingerspelling datasets are either from a general ASL corpus (ASLLVD~\cite{Neidle2012ChallengesID}, NCSLGR~\cite{Vogler2012ANW}) or collected on its own (ChicagoFSVid~\cite{Kim2017LexiconfreeFR}). 
ASLLVD~\cite{Neidle2012ChallengesID} is an ASL lexicon dataset of about 4 hours, consisting of 47 fingerspelling sequences. NCSLGR~\cite{Vogler2012ANW} is for continuous ASL signing and includes 1,000 fingerspelling sequences.
The largest fingerspelling dataset is ChicagoFSVid~\cite{Kim2017LexiconfreeFR}, which consists of 641 words fingerspelled by 4 ASL signers.
Current fingerspelling data exhibits the same aspects of limitations (e.g., lack of visual variability, a restricted set of signers, the small scale) as continuous signing data described before. Figure~\ref{fig:data-existing-datasets} shows image examples from ASLLVD~\cite{Neidle2012ChallengesID}, NCSLGR~\cite{Vogler2012ANW} and ChicagoFSVid~\cite{Kim2017LexiconfreeFR}. Though ChicagoFSVid is several times larger than the other datasets, its fingerspelling sequences are isolated, in contrast to real-life signing where fingerspelling is embedded in continuous signing.

\section{ChicagoFSWild}
\label{sec:data-fswild}

\begin{figure}[htp]
    \centering
    \includegraphics[width=\linewidth]{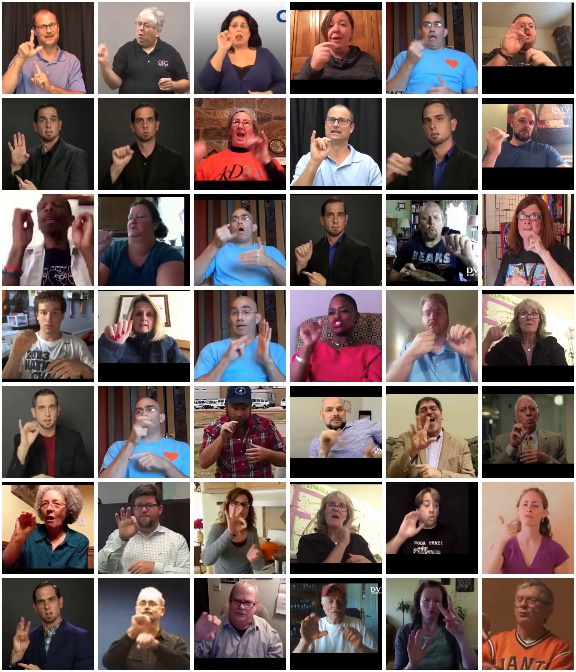}
    \caption{Example image frames in ChicagoFSWild.}
    \label{fig:data-fswild}
\end{figure}

The fingerspelling video clips of ChicagoFSWild are collected from YouTube, \texttt{aslized.org} and \texttt{deafvideo.tv}.
ASLized is an organization that creates educational videos that pertain to the use, study, and structure of ASL. DeafVideo.tv is a social media website for deaf vloggers, where users post videos on a wide range of topics.  The videos include a variety of viewpoints and styles, such as webcam videos and lectures.
214 raw ASL videos were collected, and all fingerspelling clips
within these videos were manually located and annotated. Figure~\ref{fig:data-fswild} shows typical image frames from ChicagoFSWild.

The videos were annotated by in-house annotators at TTIC and the U.~Chicago Sign Language Linguistics lab, using the ELAN video annotation tool~\cite{elan} Annotators viewed the videos, identified instances of fingerspelling within these videos, marked the beginning and end of each fingerspelling sequence, and labeled each sequence with the letter sequence being fingerspelled.  No frame-level labeling has been done; we use only sequence-level labels.
Annotators also marked apparent misspellings 
and instances of fingerspelling articulated with two hands.
The fingerspelled segments include proper nouns, other words, and abbreviations (e.g., N-A-D for National Association of the Deaf).  
The handshape vocabulary contains the 26 English letters and the 5
special characters \{$<$sp$>$, \&, \emph{'},\emph{.},\emph{@}\}
that occur very rarely.

We estimate the inter-annotator agreement on the label sequences to be about 94\%, as measured for two annotators who both labeled a small subset of the videos; this is the letter accuracy of one annotator, viewing the other as reference.

\begin{figure}[t]
\centering
\includegraphics[width=0.9\linewidth]{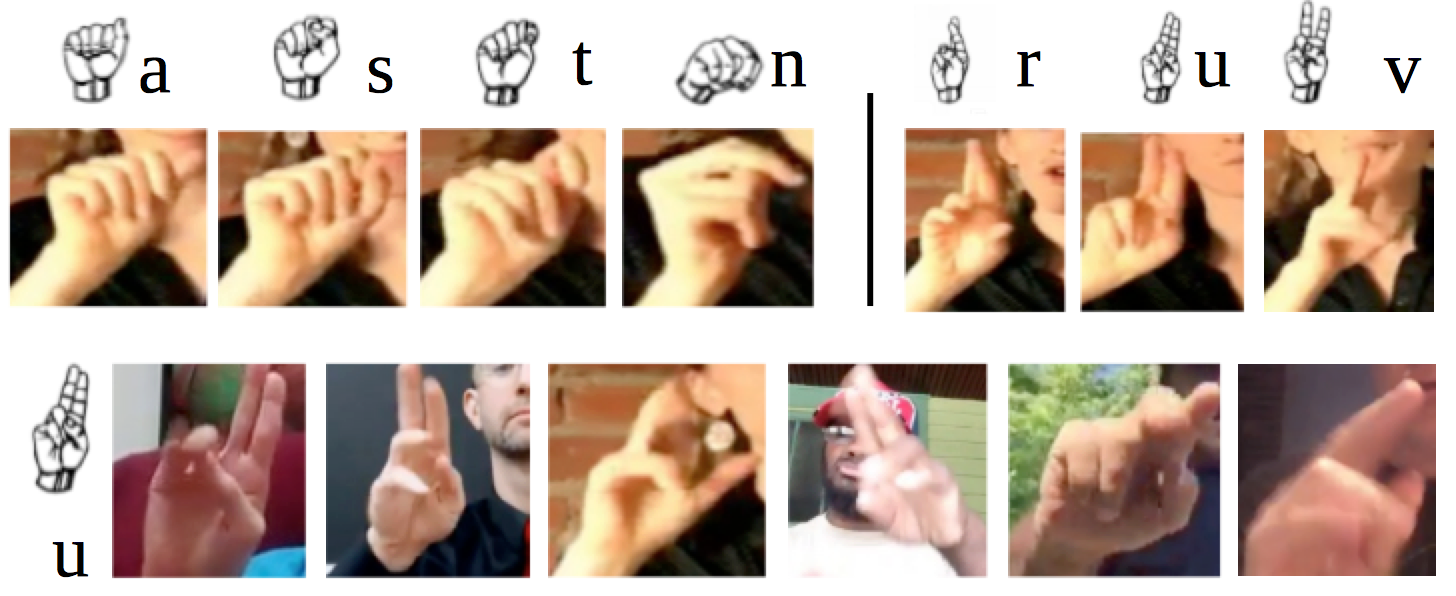}
\caption{Illustrations of ambiguity in fingerspelled handshapes.  Upper row: different letters with similar handshapes, all produced by the same signer.  Lower row: the same letter (u) signed by different signers.}
\label{fig:conf}
\end{figure}

\begin{figure}[btp]
\centering
\includegraphics[width=\linewidth]{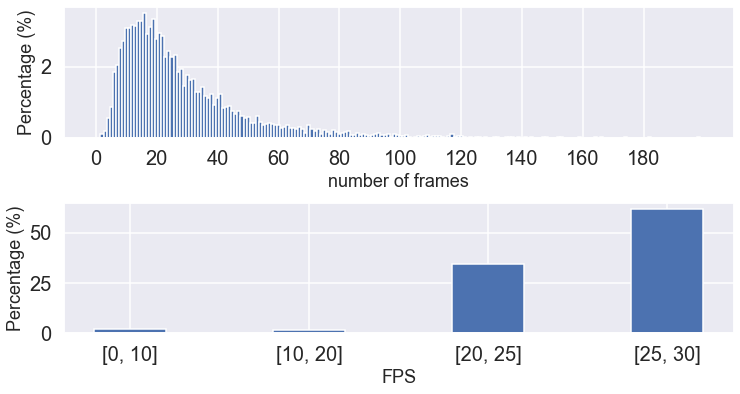}
\caption{\label{fig:statistics}Histograms of the number of frames per fingerspelled sequence and frames per second (FPS) for fingerspelled sequences in the data set.}
\end{figure}

As a pre-processing step,
we removed all fingerspelling video sequences containing fewer frames
than the number of labels.  We split the remaining data (7304
sequences) into 5455 training sequences, 981 development (dev) sequences, and 868 test sequences.  
Using frames per second (FPS) as a proxy for video quality, we ensured that the distribution of FPS was roughly the same across the three data partitions.  The dataset includes about 168 unique signers (91 male, 77 female).\footnote{These numbers are estimated by visual inspection of the videos, as most do not include meta-data about the signer.}  192 of the raw videos contain a single signer, while 22 videos contain multiple people.
Each unique signer is assigned to only one of the data
partitions.  The majority of the fingerspelling sequences are
right-handed (6782 sequences), with many fewer being left-handed (522
sequences) and even fewer two-handed (121 sequences).  Roughly half of
the sequences come from spontaneous sources such as blogs and
interviews; the remainder comes from scripted sources such as news,
commercials, and academic presentations. The frame resolution has a
mean and standard deviation of $640\times 360\,\pm\,290\times 156$. Additional statistics are
given in Figure \ref{fig:statistics}.

This data set collected ``in the wild'' poses serious challenges, such as great visual variability (due to lighting, background, camera angle, recording quality) and signing variability
(due to speed and hand appearance).
To illustrate some of these challenges, Figure~\ref{fig:conf} shows a number of representative frames from our data set.

There can be a great deal of variability in fingerspelling the same letter, as illustrated in the bottom row of Figure~\ref{fig:conf}.  In addition, many fingerpselled letters have similar handshapes.  For example, the letters \emph{a}, \emph{s}, \emph{t} and \emph{n} are only distinguished by the position of the thumb, and the letters \emph{r}, \emph{u} and \emph{v} are all signed with the index and middle fingers extended upward.  The small differences among these letters can be even harder to detect in typical lower-quality online video with highly coarticulated fingerspelling, as seen in the top row of Figure~\ref{fig:conf}.

\section{ChicagoFSWild+}
\label{sec:fswild+}

Similar to ChicagoFSWild, ChicagoFSWild+ is excised from sign language video ``in the wild'', collected from same online sources such as YouTueb and \texttt{deafvideo.tv}. However, the fingerspelling labels of ChicagoFSWild+ is crowdsourced instead of being carefully annotated by experts.

We developed a fingerspelling
video annotation interface
derived from VATIC~\citep{vatic} and have used it to collect our new data set,
ChicagoFSWild+, by crowdsourcing the annotation process via Amazon
Mechanical Turk.  Annotators are presented with one-minute ASL video clips
and are asked to mark the start and
end frames of fingerspelling within the clips (if any is present), to provide a transcription
(a sequence of English letters) for each fingerspelling sequence, but
not to align the transcribed letters to video frames.  Two annotators
are used for each clip.

\begin{figure}[htp]
  \centering
  \includegraphics[width=\linewidth]{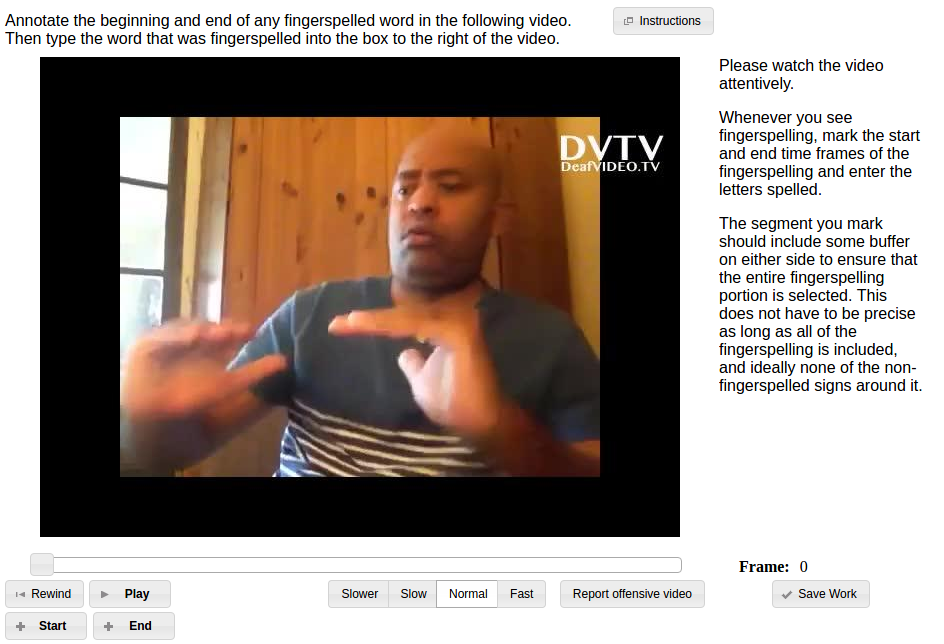}
  \caption{\label{fig:vatic} The crowdsourced annotation interface.}
\end{figure}

\begin{table}[htp]
  \centering
  \begin{tabular}{l|l|l|l}\hline
    { Perceived Gender (\%)} & { female \; 32.7} & { male \; 63.2} &
                                                                       { other \; 4.1} \\ \hline
    { Handedness (\%)} & { left \;\;\;\;\;\; 10.6} & { right
                                                   \; 86.9} & { other \; 2.5} \\ \hline
  \end{tabular}
  \caption{\label{tab:fswildplus-stats} Statistics of ChicagoFSWild+. ``Other'' includes multiple signers, unknown. Both labels are as perceived by an annotator. 
  }
\end{table}

ChicagoFSWild+ includes 50,402 training sequences from 216 signers (Table~\ref{tab:fswildplus-stats}), 3115
development sequences from 22 signers, and 1715 test sequences from 22
signers.  This data split has been
done in such a way as to approximately evenly distribute certain
attributes (such as signer gender and handedness) between the three
sets.  In addition, in order to enable clean comparisons between results
on ChicagoFSWild and ChicagoFSWild+, we used the signer labels in the two
data sets to ensure that there are no overlaps in signers between the
ChicagoFSWild training set and the ChicagoFSWild+ test set.  Finally, the annotations in the development and test sets were proofread by human annotators and a single ``clean'' annotation kept for each sequence in these sets.  For the training set, no proofreading has been done and both annotations of each sequence are used.
Compared to ChicagoFSWild, the crowdsourcing setup allows us to collect
dramatically more training data in ChicagoFSWild+, with significantly
less expert/researcher effort.

\section{OpenASL}
\label{sec:openasl}

\begin{figure}[htp]
    \centering
    \includegraphics[width=\linewidth]{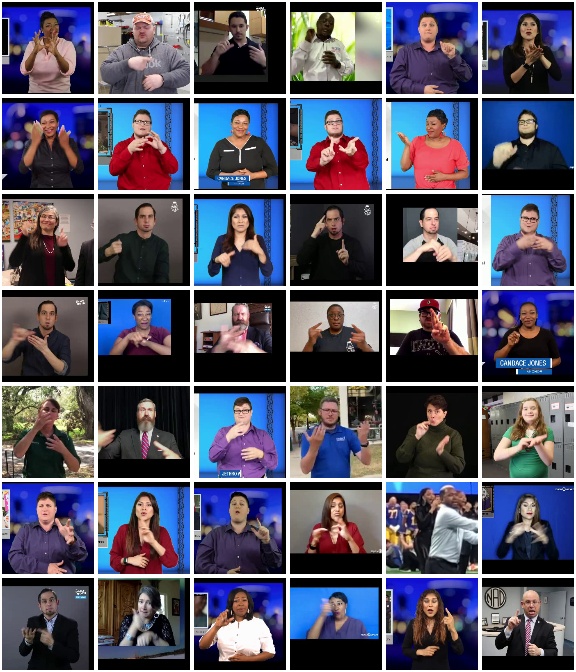}
    \caption{Example image frames in OpenASL.}
    \label{fig:data-openasl}
\end{figure}

OpenASL is a large-scale ASL-English translation dataset (see Figure~\ref{fig:data-openasl}).
The videos are collected from 
\kledit{video} websites, mainly YouTube. A large portion of our data consists of ASL news, which come primarily from the %
\kledit{YouTube channels} \texttt{TheDailyMoth} and \texttt{Sign1News}.
We download all videos with English captions in \kledit{these} two channels \kledit{though} June 2021. 
The rest \bsedit{of }
\kledit{the dataset is} collected from short \kledit{YouTube videos}
uploaded 
by the National Association of the Deaf (\texttt{NAD}).
Those videos are mostly in the form of sign VLOGs 
\kledit{of various} types including announcements, daily tips, and short conversations. 

\begin{figure}[htp]
\centering
\begin{tabular}{c}
\includegraphics[width=0.8\linewidth]{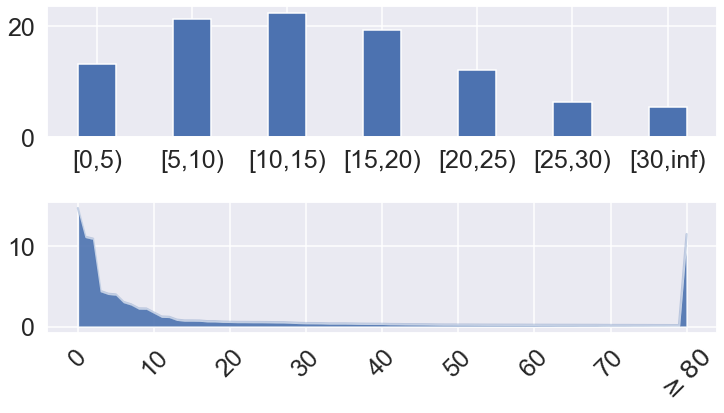} \\
\end{tabular}
\caption{\label{fig:distribution-signer-length} \kledit{Empirical distribution} of 
sentence length (upper) and percentage of sequences per signer (lower). The sentence length is in \# words per sentence. \kledit{The signer identity is obtained from metadata of the original video. The sequences where the identity is unknown are not counted.}}
\end{figure}

The raw video is divided into \kledit{roughly sentence-sized} clips based on the associated subtitles. Specifically, we split the 
\kledit{transcript} into sentences with the NLTK\footnote{\url{https://www.nltk.org/}}
sentence segmentation tool and retrieve \kledit{the corresponding} (video clip, sentence) pairs. \kledit{This procedure produces}
98,417
\kledit{translation} pairs in \kledit{total, with} 
\bsedit{33,549 unique words.}
Figure~\ref{fig:distribution-signer-length} shows the distribution of sentence length in our data. 
\kledit{We randomly select 966 and 975 translation pairs from our data} as validation and test sets respectively.

\kledit{The annotation of the validation and test sets is manually verified.  Specifically, the English translation and time boundaries of each video clip are proofread and corrected as needed by professional ASL captioners.}
Each annotator views the video clip and is given the original English sentence from the subtitle for reference. The annotator 
\kledit{marks the corrected} beginning and end of \kledit{the} sentence, and provides \kledit{a corrected} English translation \kledit{if needed} as well as the corresponding gloss sequence. During translation \kledit{of each sentence}, the annotator has access to the whole video in case \kledit{the} context is needed for \kledit{accurate translation}. 

\begin{figure}[btp]
    \centering
    \includegraphics[width=0.8\linewidth]{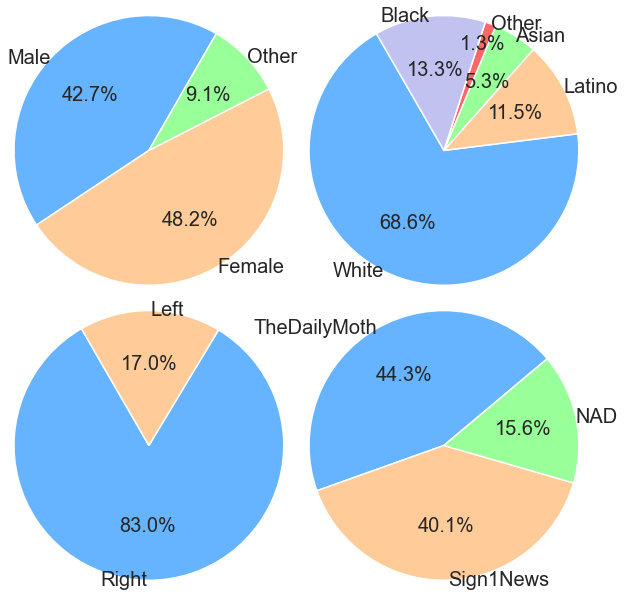}
    \caption{\label{fig:data-distribution}Distribution of \kledit{several}
    properties in \kledit{video clips in a subset of} OpenASL (top left: perceived gender, top right: race, bottom left: handedness, bottom right: \bsedit{sources}). The gender, race and handedness labels are as perceived by an annotator.}
\end{figure}

Figure~\ref{fig:data-distribution} shows the distribution of several properties in our dataset. \kledit{Note that these are not ground-truth labels, but rather approximate labels as perceived by an annotator.  The goal is to give an idea of the degree of diversity in the data, not to provide ground-truth metadata for the dataset.  The label "other" covers a variety of categories, including examples where the annotator is unsure of the label and examples that contain multiple signers.}

\begin{figure}[btp]
    \centering
    \includegraphics[width=0.8\linewidth]{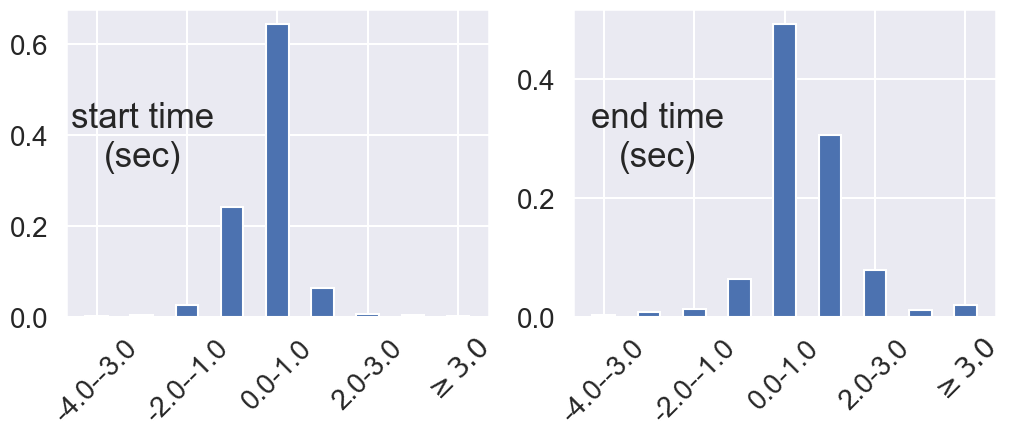}
    \caption{\label{fig:alignment-error}\kledit{Empirical distribution} of alignment \kledit{errors} (in seconds) \kledit{in a manually checked subset of our data} \bsedit{(500 sequences)} Left: start time, right: end time.}
\end{figure}

One feature of our data is \kledit{the use} of subtitles associated with the video as the English translation, thus saving effort on human annotation. Subtitled videos have also been employed in prior \kledit{work}~\cite{camgoz2021content,Albanie2021bobsl} for constructing sign language datasets. As prior \kledit{work has mostly focused} on interpreted signing videos where 
content originally in the spoken language is interpreted into sign language, the subtitles {used there} are naturally aligned to the audio instead of the signing stream. As is shown in~\cite{bull2021aligning}, there exists a large time boundary shift between the two. The videos used in OpenASL are "self-generated" 
\kledit{rather than} interpreted, \kledit{so} the English subtitles \kledit{are already} 
aligned to the video accurately. As can be seen from \kledit{Figure}~\ref{fig:alignment-error}, the sentence alignment in the subtitles \kledit{is} of overall high quality (\kledit{usually less} than 2 second \kledit{time shifts), although} 
a small percentage ($<5\%$) \kledit{are larger}. 

\bsedit{We measure \kledit{the} degree of agreement between the original and corrected \kledit{translations} using BLEU score~\citep{Papineni2002BleuAM}. The original translation achieves 81.0 BLEU-4 score when it is compared against the corrected one. The high agreement in translation, as well as the small alignment error from Figure~\ref{fig:alignment-error}, shows the overall high quality of the subtitles. Thus to save annotation effort, we do not proofread the training data.}
\section{Summary}
\label{sec:summary}
This chapter introduced three real-world ASL video corpora: ChicagoFSWild, ChicagoFSWild+ and OpenASL. The first two datasets are fingerspelling-based, while OpenASL is an ASL-English translation dataset. We described in detail the source of collected videos, annotation of the data, and their statistics. In the following chapters, we will use those three datasets to evaluate the sign language processing techniques.

%% file: table-tex/data/continuous-data.tex
\begin{table}[htp]
    \centering
    \resizebox{\textwidth}{!}{
    \begin{tabular}{lcccccccccc}
    \toprule
         Dataset & src & tgt & src vocab & tgt vocab & \#signers & \#hours & domain & interpreted & studio & w/ gloss \\
         \midrule
         Boston-104~\cite{Dreuw2007speech} & ASL & EN & 104 & 167 & 3 & 0.2 & - & \cmark & \cmark & \cmark \\
         RVL-SLLL~\cite{Wilbur2006purdue} & ASL & EN & 104 & 130 & 14 & - & - & \cmark & \cmark & \cmark \\
         NCSLGR~\cite{Vogler2012ANW} & ASL & EN & 2.4k & 1.8k & 4 & 5.3 & - & \cmark & \cmark & \cmark\\         How2Sign~\cite{Duarte2021how} & ASL & EN & - & 16k & 11 & 79 & diverse & \cmark & \cmark & \xmark \\
         CSL-Daily~\cite{Zhou2021ImprovingSL} & CSL & CN & 2k & 2k & 10 & 23 & - & \cmark & \cmark & \xmark \\      SIGNUM~\cite{Agris2007TowardsAV} & DGS & DE & 450 & 1k & 25 & 55 & - & \cmark & \cmark & \cmark \\  
         Public-DGS~\cite{Hanke2020ExtendingTP} & DGS & DE & - & 31k & 330 & 50 & diverse & \xmark & \cmark & \cmark \\ 
        Phoenix14T~\cite{Camgoz2018neural} & DGS & DE & 1k & 3k & 9 & 11 & weather & \cmark & \xmark & \cmark \\  
         KETI~\cite{ko2019neural} & KSL & KO & 524 & 419 & 14 & 28 & emergency & \cmark & \cmark & \cmark\\
        GSL SI~\cite{adaloglou2021comprehensive} & GSL & EL & 310 & - & 7 & 9.6 & - & \cmark & \xmark & \cmark \\ 
         Swisstxt-News~\cite{camgoz2021content} & DSGS & DE & - & 10k & - & 9.5 & news & \cmark & \xmark & \xmark\\ 
         VRT-News~\cite{camgoz2021content} & VGT & NL & - & 7k & - & 9 & news & \cmark & \xmark & \xmark\\ 
         BOBSL~\cite{Albanie2021bobsl} & BSL & EN & 2.3k & 78k & 39 & 1.4k & diverse & \cmark & \xmark & \xmark\\ 
         \bottomrule
    \end{tabular}
    }
    \caption{\label{tab:continuous-datasets}Summary of public datasets for sign language translation}
\end{table}

%% file: figure-tex/data/prior-data.tex
\begin{figure*}[htp]
\centering
\begin{tabular}{c}
\toprule
How2Sign~\cite{Duarte2021how} \\ 
 \includegraphics[width=\linewidth]{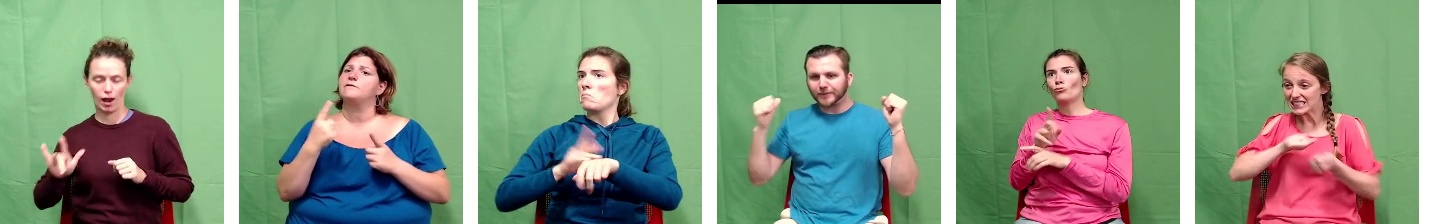}\\ \midrule
 Boston-104~\cite{Dreuw2007speech} \\ 
 \includegraphics[width=\linewidth]{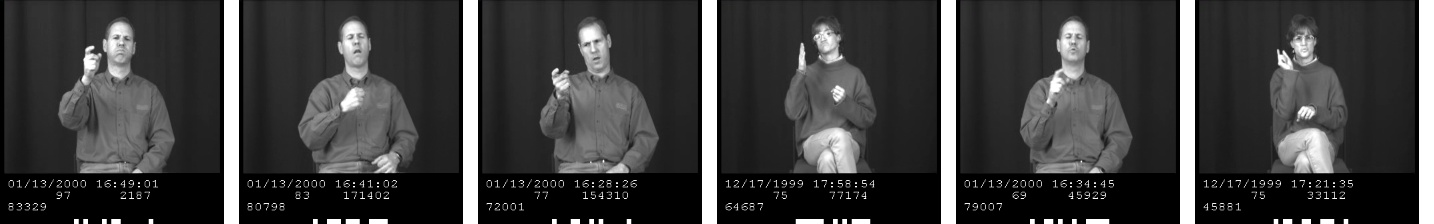}\\ \midrule
 Phoenix-14T~\cite{Camgoz2018neural} \\ 
 \includegraphics[width=\linewidth]{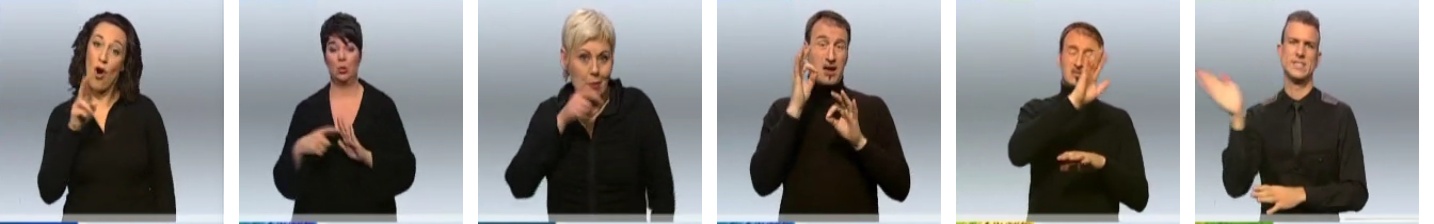}\\ \midrule
 ASLLVD~\cite{Neidle2012ChallengesID} \\ 
 \includegraphics[width=\linewidth]{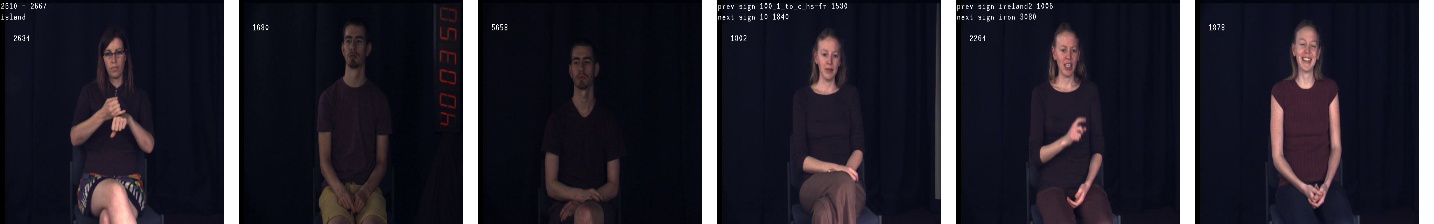}\\ \midrule
 NCSLGR~\cite{Vogler2012ANW} \\ 
 \includegraphics[width=\linewidth]{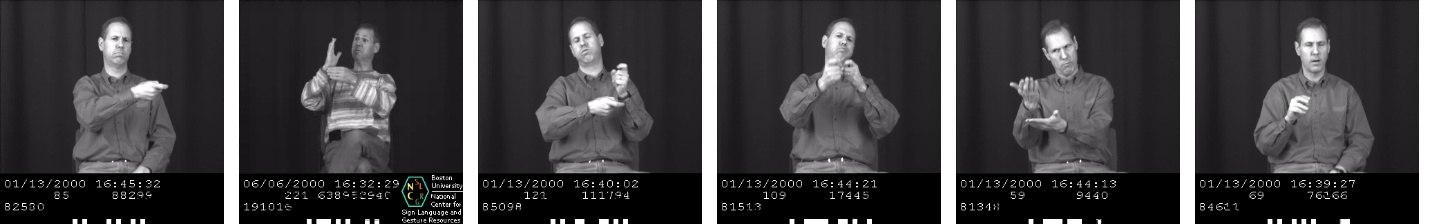}\\ \midrule
 ChicagoFSVid~\cite{Kim2017LexiconfreeFR} \\ 
 \includegraphics[width=\linewidth]{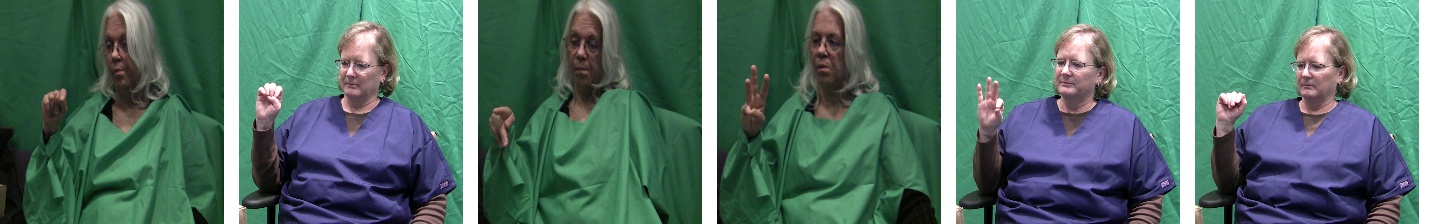}\\ \bottomrule
\end{tabular}
\caption{\label{fig:data-existing-datasets}Example image frames from existing sign language datasets}
\end{figure*}

%% file: table-tex/data/fs-data.tex
\begin{table}[htp]
    \centering
    \resizebox{\textwidth}{!}{
    \begin{tabular}{lcccccccc}
    \toprule
         Dataset & lang & vocab & \#images & \#sequences & \#signer & static & isolated & studio \\
         \midrule
         JSL-fingerspelling~\cite{Kwolek2016recognition}& DGS & - & 5k & - & 10 & \cmark &\cmark & \cmark \\
         LSA16~\cite{Ronchetti2016HandshapeRF} & LSA & - & 800 & - & 10 & \cmark &\cmark & \cmark \\
         ISL-HS~\cite{Oliveira2017dataset} & ISL & - & 52k & - & 6 & \cmark & \cmark & \cmark \\
         ArSL2018~\cite{Latif2019arasl} & ArSL & - & 54k & - & 40 & \cmark &\cmark & \cmark \\
         ASL-gesture~\cite{pugeault} & ASL & - & 131k & - & 5 & \cmark &\cmark& \cmark \\
         ASLLVD~\cite{Neidle2012ChallengesID}& ASL & 32 & 2k & 47 & 5 & \xmark &\cmark& \cmark \\
         NCSLGR~\cite{Vogler2012ANW}& ASL & 216 & $\sim$11k & 1k & 4 & \xmark&\xmark & \cmark \\
         ChicagoFSVid~\cite{Kim2017LexiconfreeFR} & ASL & 641 & 347k & 2.2k & 4 & \xmark &\cmark& \cmark \\
         \bottomrule
    \end{tabular}
    }
    \caption{\label{tab:fs-datasets}Summary of public fingerspelling datasets}
\end{table}

%% file: pipeline.tex
\chapter{Pipeline Approach for Fingerspelling Recognition}
\label{ch:pipeline}
In this chapter, we study fingerspelling recognition for real-world videos, also known as fingerspelling recognition in the wild.
The task consists of transcribing a fingerspelling video segment into words or phrases.
Compared to recognition with studio data, fingerspelling recognition in the wild is much harder due to the lower frame rates and greater visual variability of input videos.
Under this challenging scenario, we find training a special-purpose signing hand detector and only feeding the signing hand regions into a downstream recognizer provides an effective way to tackle the vision problems. Such an approach is referred to as {the pipeline approach} for fingerspelling recognition. In the following sections, we describe the pipeline approach in detail and show its performance on ChicagoFSWild. This chapter is based on~\citep{shi2018american}.

\section{Related Work}
\label{sec:pipeline-intro}
As most information in fingerspelling is carried by hand, localizing the signing hand is commonly used as a preprocessing step in fingerspelling recognition for studio data~. Signing hand localization is realized by a hand segmentation module, which is often based on skin-color models~\cite{Ricco2009accv,Goh2006DynamicFR,Kim2017LexiconfreeFR}.

Below we use the approach proposed in~\cite{Kim2017LexiconfreeFR,Kim2016Adaptation,Kim2013SCRF,Kim2012AmericanSL} as an example to illustrate hand segmentation. Specifically, given an RGB image frame, a Gaussian mixture model and a single Gaussian are employed respectively to model the distribution of hand color pixels $P_{GMM}(x;\theta_{GMM})$ and background pixels $P_{G}(x;\theta_{G})$. 
Any pixel $x$ satisfying condition~\ref{eq:prem_studio_hand_model} is considered as hand pixel, where $\pi_{hand}$ is the prior of hand pixel. Model parameters $\theta_{GMM}$,$\theta_{G}$, $\pi_{hand}$ are estimated with training data.

\begin{equation}
  \label{eq:prem_studio_hand_model}
  \centering
  P_{GMM}(x; \theta_{GMM})\pi_{hand} > P_{G}(x; \theta_{G})(1-\pi_{hand})
\end{equation}

Several post-processing steps were further applied to clean up the result: pixels that fall into face regions given by an external face detector, or whose hand probability $p_{GMM}(x,\theta_{GMM})$ is below a threshold, or which is located outside a pre-defined signing region, are suppressed. The detected hand corresponds to the largest connected component in the resulting binary image.

As the visual content of studio data is simple, the above-mentioned hand segmentation does not require much training data. For example, Kim et al.~\citep{Kim2017LexiconfreeFR} only use 30 images per signer (120 images in total) to conduct segmentation. However, this thresholding approach with color models is hard to generalize into fingerspelling data in the wild, where image frames involve a more complex background with diverse colors.
Moreover, the segmentation model is signer-specific and does not fit into a signer-independent recognition setting.

\section{Method}
\label{sec:pipeline-method}
\begin{figure}[htp]
\centering
\includegraphics[width=\linewidth]{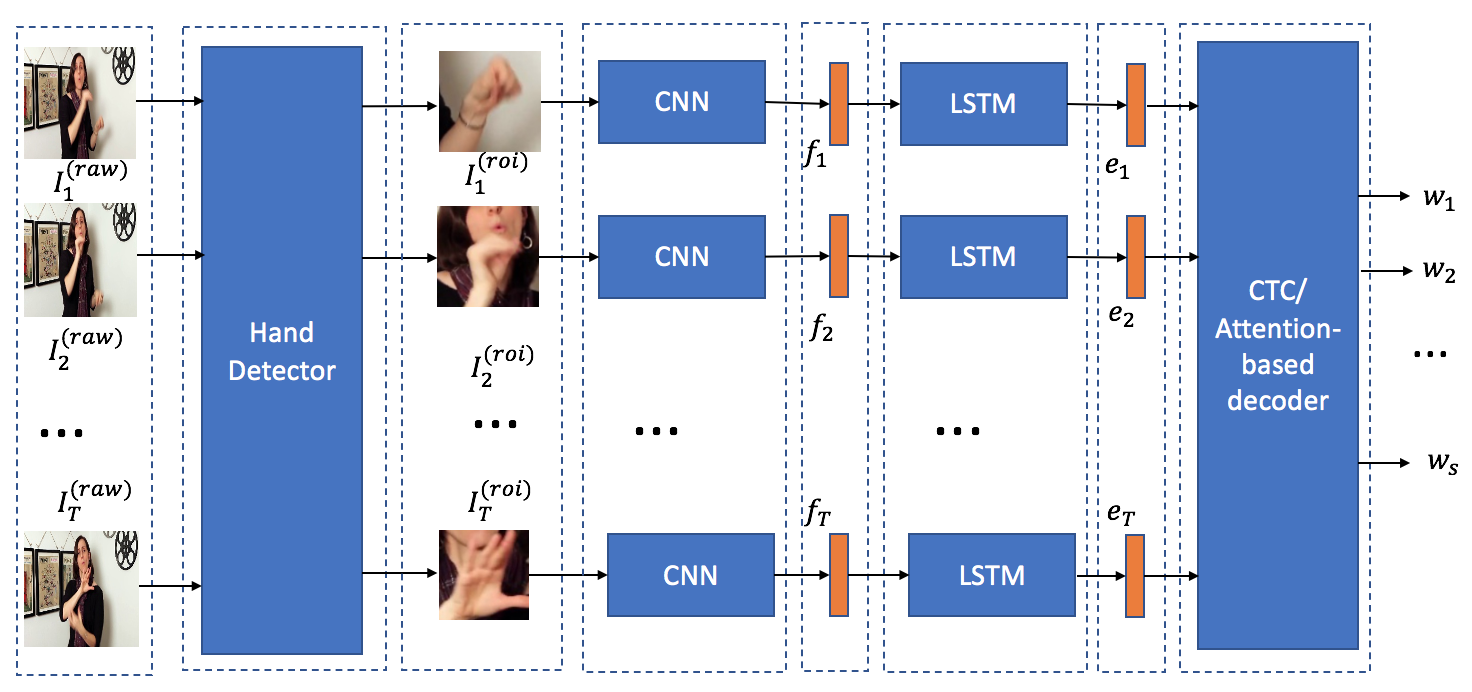}
\caption{\label{fig:pipeline-model}Sketch of the pipeline approach for fingerspelling recognition in the wild.  After the hand detector component, the rest of the model is trained end-to-end.}
\end{figure}

An overview of our method for fingerspelling recognition in the wild is illustrated in Figure~\ref{fig:pipeline-model}. Similar to approaches for studio videos, the pipeline consists of a signing hand detector followed by a sequence recognizer.

\subsection{Signing Hand Detection}
\label{sec:tube_gen}
The hand detection problem here is somewhat different from typical
hand detection.
A large proportion of the video frames contain more than one hand,
but since ASL fingerspelling generally involves a single hand, the
objective here is to detect the {\it signing hand}.  This can be
viewed as a problem of action localization~\cite{Weinzaepfel2015Learning}.
As in prior work on action localization \cite{Weinzaepfel2015Learning}, we
train a detection network that takes as input both the image
appearance and optical flow, represented as a motion vector for every
pixel computed from two neighboring frames
\cite{Horn1980determining}.\footnote{For optical flow we use the OpenCV
  implementation of~\cite{Farnebck2003TwoFrameME}.}  For the detection
network, we adapt the design of the Faster R-CNN object detector
\cite{ren2015faster}. As in \cite{ren2015faster}, the detector is based on
an ImageNet-pretrained VGG-16 network \cite{deng2009imagenet,Simonyan2015very}.
Unlike the general object detector, we only preserve the first 9
layers of VGG-16 and the stride
of the network is reduced to 4. Lower layers able to capture more fine
details~\cite{Weinzaepfel2015Learning} combined with finer stride/localization
are beneficial for detecting signing hands, which tend to be small
relative to the frame size.

Unlike much work in action localization \cite{Weinzaepfel2015Learning},
which processes optical flow and appearance images in two distinct
streams, we concatenate the optical flow and RGB image as the input to
a single CNN. In our video data, motion involves many
background objects like faces and non-signing hands, so a separate optical flow stream may be misleading.

Given bounding boxes predicted framewise by the Faster R-CNN, we first
filter them by spatial non-maxima
suppresion (NMS)~\cite{girshick2015fast}, greedily removing any box
with high overlap with a higher-scoring box in the same frame. Next, we
link the surviving boxes across time to form a video region likely to be associated with a fingerspelling sequence,
which we call a ``signing tube'' (analogously to action tubes in
action recognition~\cite{Gkioxari2015finding}). Even after NMS, there
may be multiple boxes in a single frame (e.g., when the signer is
signing with both hands). Our temporal linking process helps prevent
switching between hands in such cases. It also has a smoothing effect, which can
reduce errors in prediction compared to that based on a single frame.

More formally, the input to the signing tube prediction is a sequence of sets of bounding box coordinate and score pairs: $\{(b_t^1, s_t^1), (b_t^2, s_t^2), ..., (b_t^n, s_t^n)\}$, $1\leq t\leq T$, produced by the frame-level signing hand detector. The score $s_t^i$ is the probability of a signing hand output by the Faster R-CNN. We define the {\it linking score} of two boxes $b_t^i$ and $b_{t+1}^j$ in two consecutive frames as:
\begin{equation}
e(b_t^i, b_{t+1}^j)=s_t^i+s_{t+1}^j+\lambda * IoU(b_t^i, b_{t+1}^j)
\end{equation}
\noindent where $IoU(b_t^i, b_{t+1}^j)$ is the intersection over union of $b_t^i$ and $b_{t+1}^j$ and $\lambda$ is a hyperparameter that is tuned on held-out data.
Generation of the optimal signing tube is the problem of finding a sequence of boxes $\{b_1^{l_1}, ..., b_T^{l_T}\}$ that maximizes the sequence score, defined as
\begin{equation}
E(l)=\frac{1}{T}\sum_{t=1}^{T-1}e(b_t^{l_t}, b_{t+1}^{l_{t+1}})
\end{equation}
\noindent This optimization problem is solved via a Viterbi-like dynamic programming algorithm \cite{Viterbi1967error}. 

\subsection{Fingerspelling Sequence Model}
We next take the signing tube, represented as a sequence of image patches $\{\mathbf{I}_1, \mathbf{I}_2,..., \mathbf{I}_T\}$, as input to a sequence model that outputs the fingerspelled word(s) $w$.  We work in a lexicon-free setting, in which the word vocabulary size is unlimited, and represent the output $w$ as a sequence of letters $w_1, w_2, ..., w_s$.  The model begins by applying several convolutional layers to individual image frames to extract feature maps.  
The convolutional layers transform the frame sequence $\{\mathbf{I}_1, \mathbf{I}_2,..., \mathbf{I}_T\}$ into a sequence of features $\{\mathbf{f}_1, \mathbf{f}_2,..., \mathbf{f}_T\}$.

The sequence of image features $\{\mathbf{f}_1, \mathbf{f}_2,..., \mathbf{f}_T\}$ is then fed as input to a long short-term memory recurrent neural network (LSTM)~\cite{Hochreiter1997long} that models the temporal structure, producing a sequence of hidden state vectors (higher-level features) $\{\mathbf{e}_1, \mathbf{e}_2,..., \mathbf{e}_T\}$.{rephrased}  Given the
features produced by the LSTM, the next step is to compute the probabilities of the letter sequences $w_1, w_2, ..., w_s$. 
We consider two approaches for decoding, neither of which requires
frame-level labels at training time: an attention-based LSTM decoder,
and connectionist temporal classification (CTC)~\cite{Graves2006ConnectionistTC}.  In the former case, the whole sequence model becomes a recurrent encoder-decoder with attention~\cite{vinyals2015grammar}.

In the \textbf{attention-based model}, temporal attention weights are applied to $(\mathbf{e}_1, \mathbf{e}_2,..., \mathbf{e}_T)$ during decoding, which allows the decoder to focus on certain chunks of visual features when producing each output letter.  If the hidden state of the decoder LSTM at timestep $t$ is $\mathbf{d}_t$, the probability of the output letter sequence is given by 
\begin{equation}
\begin{split}
& \alpha_{it} = \softmax(\mathbf{v}_d^T \tanh(\mathbf{W}_e\mathbf{e}_i + \mathbf{W}_d \mathbf{d}_t)) \\
& \mathbf{d}_t^\prime = \displaystyle\sum_{i=1}^T \alpha_{it}\mathbf{e}_i \\
& p(w_t|w_{1:t-1}, \mathbf{e}_{1:T})=\softmax (\mathbf{W}_o[\mathbf{d}_t;\mathbf{d}_t^\prime]+\mathbf{b}_o) \\
& p(w_1, w_2, \ldots, w_s|\mathbf{e}_{1:T})=\displaystyle\prod_{t=1}^s{p(w_t|w_{1:t-1}, \mathbf{e}_{1:T})} \\
\end{split}
\end{equation}
\noindent where $\mathbf{d}_t$ is given by the standard LSTM update equations~\citep{Hochreiter1997long}.  The model is trained to minimize log loss.

In the \textbf{CTC-based model},  for an input sequence of $m$-dimensional visual feature vectors $\mathbf{e}_{1:T}$, we define a continuous map $\mathcal{N}_w: (\mathcal{R}^m)^T \mapsto (L^\prime)^T$ representing the transformation from $m$-dimensional
features $\mathbf{e}_{1:T}$ to frame-level label probabilities and a many-to-one map $\mathcal{B}: {{L^\prime}^T} \mapsto L^{\leq T}$ where $L^{\leq T}$ is the set of all possible labelings. Letting $L$ be the output label vocabulary,
$L^\prime = L\cup \{blank\}$, and $y_k^t$ the probability of
label $k$ at time $t$, the posterior probability of any labeling $\pi\in {L^\prime}^T$ is

\begin{equation}
p(\pi|\mathbf{e}_{1:T}) = \displaystyle\prod_{t=1}^T{y_{\pi_t}^t}=\displaystyle\prod_{t=1}^T {\textrm{softmax}_{\pi_t}(\mathbf{A}^e\mathbf{e}^t
+\mathbf{b}^e)}
\end{equation}

At training time, the probability of a given labeling $w = w_1, w_2, ..., w_s$
is obtained by summing over all the possible frame-level labelings $\pi$,
which can be computed by a forward/backward algorithm:
\begin{equation}\label{eq:ctc_s2}
p(w|\mathbf{e}_{1:T})=\displaystyle\sum_{\pi\in\mathcal{B}^{-1}(w)}p(\pi|\mathbf{e}_{1:T})
\end{equation}
\noindent The CTC model is trained to optimize this probability for the ground-truth label sequences.

Finally, in decoding we combine these basic sequence models with an RNN language model.  To decode with a language model, we use beam search to produce several candidate words at each time step and then rescore the hypotheses in the beam using the summed score of the sequence model, weighted language model, and an insertion penalty to balance the insertion and deletion errors. The language model weight and insertion penalty are tuned.

\section{Experiments}
\label{sec:pipeline-experiments}
\subsection{Setup}

All of the experiments are done in a signer-independent, lexicon-free (open-vocabulary) setting using ChicagoFSWild and partitions described in Section~\ref{ch:data}.

\textbf{Evaluation} We measure the letter accuracy of predicted sequences, as is commonly used in sign language recognition and speech recognition:
$Acc = 1 - \frac{S+I+D}{N}$, 
where S, I and D are the numbers of substitutions, insertions, and deletions (with respect to the ground truth) respectively, and N is the number of letters in the ground-truth transcription. 

\textbf{Hand detection details}
We manually annotated every frame in 180 video clips from our training set  with signing and non-signing hand bounding boxes.\footnote{The non-signing hand category is annotated in
  training to help the detector learn the distinction between signing
  hands, other hands, and background; once the detector is trained we
  ignore the non-signing hand category, and only use the signing hand detections.}
Of these, 123 clips (1667 frames) are used for training and 19 clips (233 frames) for validation.  All images are resized to $640\times 368\times 3$.  We use stochastic gradient descent (SGD) for optimization, with initial learning rate 0.001
and decreased by a factor of $2$ every 5 epochs.
We apply greedy per-frame NMS with intersection-over-union (IoU)\footnote{IoU is
the ratio of the area of overlap over area of union of two regions.}
threshold of 0.9, until 50 boxes/frame remain. The bounding boxes are then smoothed as described in Section~\ref{sec:tube_gen}. $\lambda$ is tuned to 0.3, which maximizes the proportion of validation set bounding boxes with IoU $> 0.5$.  
Using our bounding box smoothing approach, the proportion of bounding
boxes with IoU $> 0.5$ is increased from 70.0\% to 77.5\%.

\textbf{Letter sequence recognition details} The input to the recognizer is a bounding box of the predicted signing hand region. All bounding boxes are resized to $224\times 224$ before being fed to the sequence model.
For the convolutional layers of the sequence model, we use AlexNet \citep{Krizhevsky2012imagenet} pre-trained on ImageNet as the base architecture.
For the recurrent network, we use a single-layer long short-term memory (LSTM) network with 600 hidden units. (A model with more recurrent layers does not consistently improve performance.)  The network weights are learned using mini-batch stochastic gradient descent (SGD) with weight decay. The initial learning rate is 0.01 and is decayed by a factor of 10 every 15 epochs. Dropout with a rate of 0.5 is used between fully connected layers of AlexNet.  The batch size is 1 sequence in all experiments. The hyperparameters were tuned to maximize the dev set letter accuracy.
The language model is a single-layer LSTM with 600 hidden units, trained on the letter sequences in our training set.

\subsection{Main Results}
\label{sec:main_baseline}

Table~\ref{tab:pipeline-res} shows the performance of our models (``Hand'') using the cropped hand region as input to the sequence model, as well as of a baseline model (``Global'') with the same sequence model architecture but without hand detection (i.e., taking the whole image as input).  This baseline model is based on commonly used approaches for video description~\cite{Donahue2015long}.
For the Global baseline, image frames are resized to $224\times 224$ due to memory constraints.  We also report the result of a ``guessing'' baseline (``LM'') that predicts words directly from our language model.
This baseline
only uses statistics of fingerspelled letter sequences, uses no visual input for prediction, and always predicts the
output of a greedy decoding of the language model.

The Global baseline outperforms the language model baseline by a small margin, suggesting that the full-image model is unable to use much of the visual information.
Compared to the baseline, our approach with hand detection performs much better.  The hand detection step both filters out irrelevant information (e.g. background, non-signing hand) and allows us to use higher resolution image regions.
CTC-based models consistently outperform the encoder-decoder models on this task.
Since fingerspelling sequences are expected to have largely monotonic alignment with the video, this may benefit the simpler CTC model.

\begin{table}[htp]
\centering
\begin{tabular}{lccccc}\\ \toprule
 & LM & Global enc-dec & Global CTC & Hand enc-dec & Hand CTC \\ \midrule
Test Acc (\%)& 9.4 & 12.7 & 10.0 & 35.0 & 41.9 \\ \bottomrule
\end{tabular}
\caption{\label{tab:pipeline-res}Letter accuracies (\%) for the language model and global baseline models and our hand detector-based models.}
\end{table}

\textbf{Additional model variants}
Besides the proposed model, we also considered a number of variants that we ultimately rejected.  The bounding boxes output by the hand detector fail to contain the whole hands in many cases.  We considered enlarging the predicted bounding box by a factor of $s$ in width and height before feeding it to the recognizer. 
In addition, we also considered using optical flow as an additional input channel to the sequence model (in addition to the hand detector), since motion information is important in our task.

We find that neither of these variants consistently and/or significantly improves
performance (on the dev set) compared to the baseline with $s=1$
and no optical flow input. 
Thus we do not pursue these model variants further.

\subsection{Error Analysis}

The most common types of errors are deletions, followed by insertions.  The encoder-decoder model makes more insertion errors and fewer deletion errors than the CTC model, that is its error types are more balanced, but its overall performance is worse.  The most common substitution pairs for the CTC model are (u $\rightarrow$ r), (o $\rightarrow$ e), (y $\rightarrow$ i), (w $\rightarrow$ u), (y $\rightarrow$ i) and (j $\rightarrow$ i) (see Table~\ref{tab:sub_error}).
 (u $\rightarrow$ r), (y $\rightarrow$ i) and (w $\rightarrow$ u) involve errors with infrequent letters, which may be due to the relative dearth of training data for these letters.  The pair (j $\rightarrow$ i) is interesting in that the most important difference between them is whether the gesture is dynamic or static.  Compared to studio data, the frame rates in our data are much lower, which may make it more difficult to distinguish between static and dynamic letters with otherwise similar handshapes.

Since deletions are the most frequent error type, applying an insertion/deletion penalty is one possible way to improve performance.  Using such a penalty produces a negligibly small improvement, as seen in Table~\ref{tab:lm}.

\begin{table}
\centering
\begin{tabular}{c|ccccc}\toprule
 & (u $\rightarrow$ r) & (o $\rightarrow$ e) & (y $\rightarrow$ i) & (w $\rightarrow$ u) & (j $\rightarrow$ i) \\ \midrule
 \%& 17.0 & 11.7 & 7.9 & 7.6 & 6.7 \\ \bottomrule
\end{tabular}
\caption{\label{tab:sub_error}
  Percentage of several important substitution error pairs on the development set.  For a given label pair ($x_1 \rightarrow x_2$), this is the percentage of occurrences of the ground-truth label $x_1$ that are recognized as $x_2$.}
\end{table}

\begin{figure}[btp]
\centering
\includegraphics[width=0.8\linewidth]{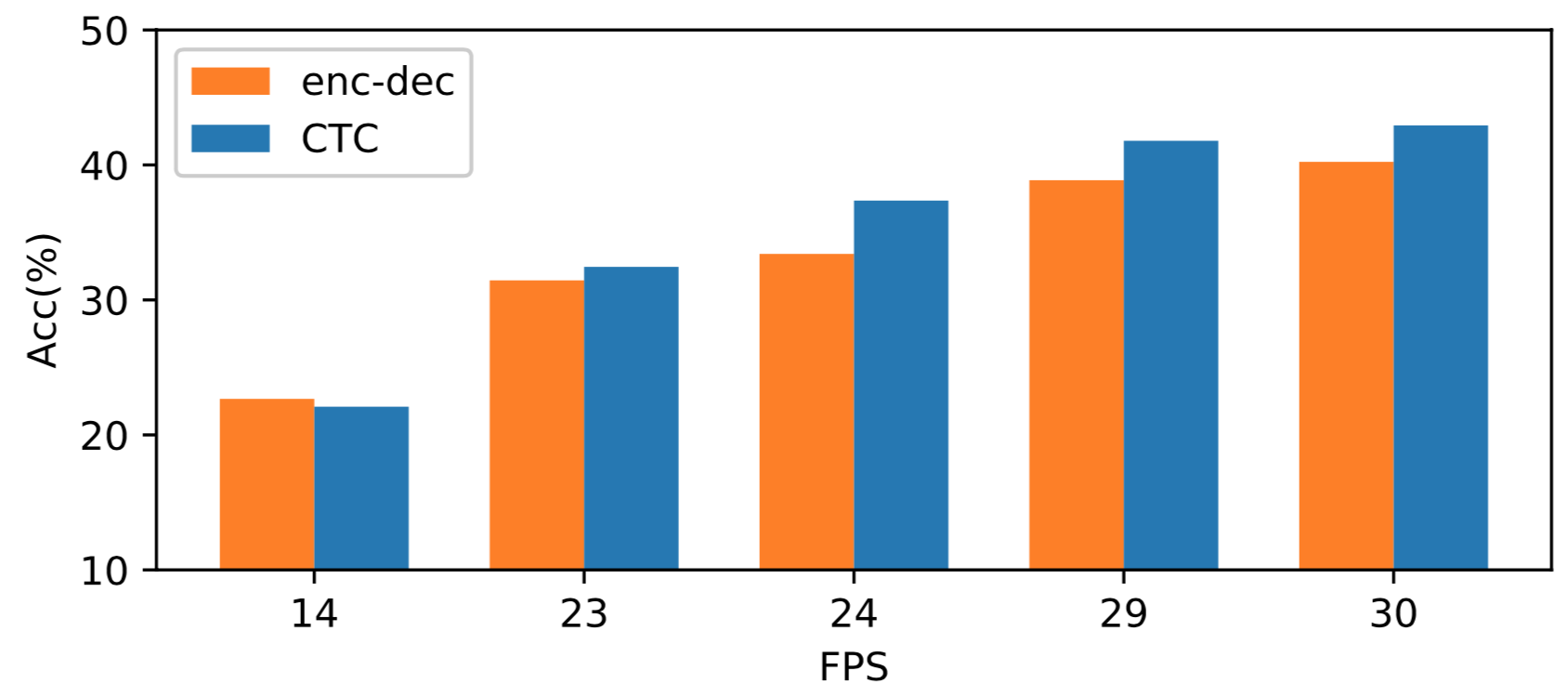}
\caption{\label{fig:fps}Development set accuracy for sequences with different frame rates (FPS) for our CTC and encoder-decoder models.}
\end{figure}

To measure the impact of video quality on performance, we divided the dev set into subsets according to the frame rate (FPS) and report the average error in each subset (see Figure~\ref{fig:fps}). In general, higher frame rate corresponds to higher
accuracy.

\subsection{Model Analysis}
\label{sec:pipeline-model-analysis}

\textbf{Effect of the language model}
Next we consider to what extent the language model 
improves performance.
It is not clear how much the language model can help, or what training material is best, 
since fingerspelling does not follow the same distribution as English words and there is not a great deal of transcribed fingerspelling data available.  
In addition to training on the letter sequences in our own training set, we also consider training on all words in the CMUdict (version 0.7a) dictionary~\cite{cmu_dict}, which contains English words and common names, and no improvement was found.  The development set perplexity of our LM trained 
with in-house data is 17.3.  Since the maximum perplexity is 32 (31 characters plus end-of-sequence), this perplexity reflects the difficulty of learning the statistics of fingerpselled letter sequences.
We also experimentally check the effect of the insertion penalty and beam search.
The beam size, language model weight, and insertion penalty are tuned and the best development set results are given in Table \ref{tab:lm}.    Using a language model, the accuracy is improved by a small margin ($\sim$1\%).

\begin{table}
\centering
\begin{tabular}{ccccc}\hline
 & no LM  &  + beam & + beam + ins & + beam+ ins +lm \\ \hline
 \small{CTC} & 41.1 & 41.1 & 41.4 & 42.8 \\ \hline
 \small{Enc-dec} & 35.7 & 35.8 & 35.9 & 36.7 \\ \hline
\end{tabular}
\caption{\label{tab:lm} Development set letter accuracies (\%) when decoding with a language model (lm: LM trained with words from our training set, beam: beam search, ins: insertion penalty, no LM: greedy decoding).}
\end{table}

\subsection{Human Performance}

We estimate human performance on the ChicagoFSWild test set. Specifically, the human is presented with the fingerspelling sequences only and does not have access to the surrounding video.
We measured letter accuracy on the ChicagoFSWild test set of a native signer and two additional proficient signers.\footnote{The two signers, who are research assistants in the Sign Language Linguistics Lab at the University of Chicago, have a minimum of two years of prior experience studying ASL before joining the lab.}  The native signer has an accuracy of 86.1\%; the non-native signers have somewhat lower accuracies (74.3\%, 83.1\%).  
These results indicate that the task is not trivial even for humans.

%
%

%
%

\section{Summary}
\label{sec:pipeline-summary}
In this chapter, we addressed the recognition of ASL fingerspelling in naturally occurring online videos with a pipeline composed of a hand detector and an end-to-end neural sequence model. The experiments on ChicagoFSWild demonstrates the importance of hand localization for obtaining high-resolution regions of interest, and that CTC-based models outperform encoder-decoder models on our task. We analyze the sources of error and the effect of a number of design choices. The best test set letter accuracies we obtain, using a CTC-based recognizer, are around 42\%, indicating that there is much room for improvement. 

%% file: end2end.tex
\chapter{End-to-End Fingerspelling Recognition}
\label{ch:e2efsr}

As we have seen previously in Chapter~\ref{ch:pipeline}, localizing the signing hand is an important step to obtain high-resolution ROIs, which is essential to achieve high accuracy. Training a special-purpose hand detector requires hand annotation, which becomes increasingly challenging in a larger fingerspelling dataset (e.g., ChicagoFSWild+).  Moreover, errors in hand detection may propagate to the downstream recognizer, leading to a decline in recognition performance.

This chapter presents an alternative approach to fingerspelling recognition that eliminates the need for hand detection. Our model is based on a spatial attention mechanism and is trained end-to-end from raw image frames.

Notably, recent advances in general-domain hand detectors, such as those based on human pose estimation techniques (e.g., MediaPipe~\cite{Lugaresi2019mediapipe}), may alleviate some of the limitations of the pipeline approach we previously established. The purpose of this work is to explore the possibilities of recognizing fingerspelling in an end-to-end manner, rather than replacing the use of hand detectors in fingerspelling recognition. As we will demonstrate in this chapter, our approach can be combined with hand detection, resulting in further improvements in recognition accuracy.

This chapter is based on~\citep{shi2019fingerspelling}.

\section{Related Work}
\label{sec:end2end-related-work}

Spatial attention has been applied in 
vision tasks including image captioning \citep{show_attend_tell} and
fine-grained image recognition \citep{multi_grain, Xiao2014TheAO,Zhao2016DiversifiedVA, Liu2016FullyCA}. Our use of attention to
iteratively zoom in on regions of interest is most similar to the work of
Fu~{\it et al.}~\citep{look_closer} using a similar ``zoom-in''
attention for image classification. Their model is trained directly from the full image, and iterative
localization provides small
gains; their approach is also
limited to a single image. In contrast, our model is applied to a
frame sequence, producing an ``attention tube'', and 
is iteratively trained with frame sequences of increasing resolution,
yielding sizable benefits.

\section{Method}
\label{sec:end2end-method}

\subsection{Attention-Based Recurrent Neural Network}\label{sec:arnn}

The attention-based recurrent neural network transcribes the input image sequence $\v I_1, \v I_2, ..., \v I_T$ into a letter sequence $w = w_1, w_2, ..., w_K$.
One option is to extract visual features with a 2D-CNN on individual frames and feed those features to a
recurrent neural network
to incorporate temporal
structure.  Alternatively, one can obtain a spatio-temporal representation
of the frame sequence by applying a 3D-CNN to the stacked frames. Both
approaches lack \emph{attention} -- a mechanism to focus on the
informative part of an image.  
In our case, most information is
conveyed by the hand, which often occupies only a small portion of
each frame.
This suggests using a spatial attention mechanism.

\begin{figure}[btp]
\centering
  \includegraphics[width=\linewidth]{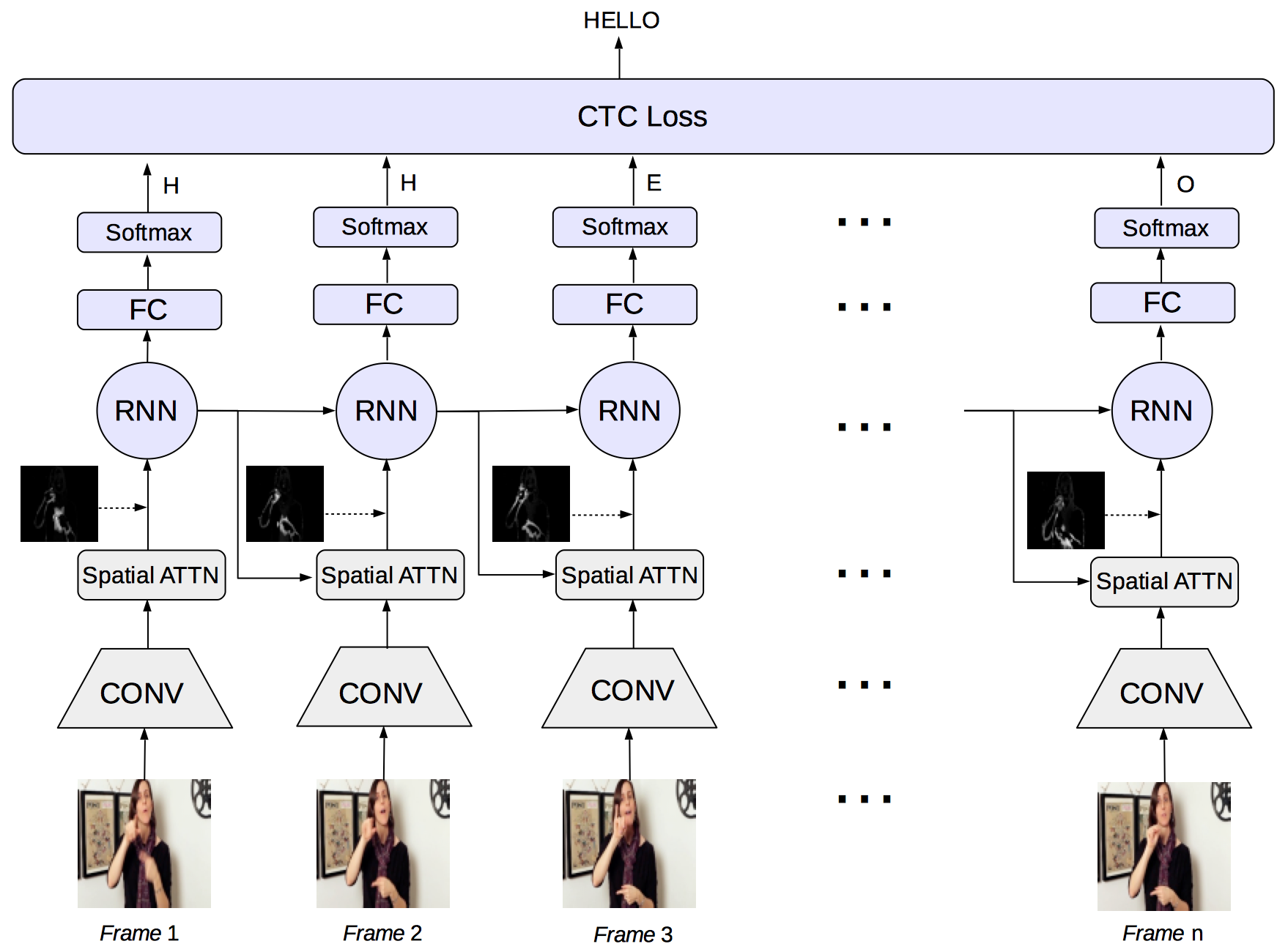}
  \caption{\label{fig:model_architecture} Recurrent CNN with attention.}
\end{figure}

Our attention model is based on a convolutional recurrent architecture
(see Figure \ref{fig:model_architecture}). At frame $t$, a fully
convolutional neural network is applied on the image frame $\v I_t$ to
extract a feature map $\v f_t$. Suppose the hidden state of the recurrent unit at
timestep $t-1$ is $\v e_{t-1}$.  We compute the attention map $\pmb\beta_t$
based on $\v f_t$ and $\v e_{t-1}$ (where $i,j$ index spatial locations):

\begin{equation*}\label{eq:beta}
v_{tij} = \v u_f^T \tanh\left(\v W_d \v e_{t-1}+\v W_f \v f_{tij}\right)\quad
\beta_{tij} = \frac{\exp\left(v_{tij}\right)}{\displaystyle\sum_{i, j}\exp\left(v_{tij}\right)}
\end{equation*}

The attention map $\pmb\beta_t$ reflects the
importance of features at different spatial locations to the
letter sequence. Optionally, we include a \emph{prior-based
  attention} term $\mathbf{M}$, which represents prior knowledge we
have about the importance of different spatial locations for our task.  For instance, $\mathbf{M}$
may be based on optical flow, as regions in motion are more likely than static regions to
correspond to a signing hand. The visual feature vector 
at time step $t$ is a weighted average of $\v f_{tij}$, $1\leq i\leq h,
1\leq j\leq w$, where $w$ and $h$ are the width and height of the feature
map respectively: 
\begin{align}
  \label{eq:attn_map}
       \v A_t = \frac{\pmb\beta_t\odot \v M_t^\alpha}{\displaystyle\sum_{p,
        q}\beta_{tpq}M_{tpq}^\alpha},\qquad
    \v h_t=\displaystyle\sum_{i, j}\v f_{tij}A_{tij}
\end{align}
\noindent where $\mathbf{A}$ represents the (posterior) attention map and $\alpha$ controls the relative weight of the prior and attention weights learned by the model.
The state of the recurrent unit at time step $t$ is updated via
$\v e_{t} = LSTM(\v e_{t-1}, \v h^t)$
where $LSTM$ refers to a long short-term memory
unit~\citep{Hochreiter1997long} (though other types of RNNs could be used here as
well).

The sequence $\mathbf{E} = (\v e_1, \v e_2, \ldots, \v e_T)$ can be viewed as high-level features for the image frames.  Once we have this sequence of features, the
next step is to decode it into a letter sequence: $(\v e_1, \v e_2,..., \v e_T)\rightarrow w = w_1, w_2, ..., w_K$. 
Our model is based on connectionist temporal classification (CTC)~\citep{Graves2006ConnectionistTC}, which requires no frame-to-letter alignment for training.
For a sequence of visual features $\mathbf{e}$ of length $T$, we generate frame-level label posteriors via a fully-connected layer followed by a softmax, as shown in Figure \ref{fig:model_architecture}.  In CTC, the frame-level labels are drawn from $L\cup \{blank\}$, where $L$ is the true label set and $blank$ is a special label that can be interpreted as ``none of the above''.
The probability of a complete frame-level labeling $\pi = (\pi_1, \pi_2, \ldots, \pi_T)$ is then
\begin{equation}
  p(\pi|\mathbf{e}_{1:T}) = \displaystyle\prod_{t=1}^T {\softmax_{\pi_t}(\mathbf{W}^e\mathbf{e}_t+\mathbf{b}^e)}
\end{equation}
At test time, we can produce a final frame-level label sequence by taking the highest-probability label at each frame (greedy search).  Finally, the label sequence $w = w_1, w_2, ..., w_K$ is produced from the frame-level sequence $\pi$ via the CTC ``label collapsing function'' $\mathcal{B}$, which removes duplicate frame labels and then $blank$s.

At training time CTC maximizes log probability of the final label
sequence, by summing over all compatible frame-level labelings using a
forward-backward algorithm.

\textbf{Language model} In addition to this basic model, at test time
we can also incorporate a language model
providing a probability for each possible next letter given the
previous ones.
We use a beam search to find the best
letter sequence, in a similar way to decoding approaches used for
speech recognition:  The score for hypotheses in the beam is composed
of the CTC model's score (softmax output) combined with a weighted
language model probability, and an additional bias term for balancing insertions and deletions.

\subsection{Iterative Visual Attention via Zooming in}
The signing hand(s) typically constitute only a small
portion of each frame.  In order to recognize fingerspelling sequences, the model needs to be able to reason about fine-grained motions and minor differences in handshape.  The attention mechanism enables the model 
to focus on informative regions, but high resolution is needed in order to retain sufficient information in the attended region.  One straightforward approach is to use very high-resolution input images.  However, since the
convolutional recurrent encoder covers the full image sequence, 
using large images can lead to prohibitively large memory
footprints. Using the entire frame in real-world videos also increases
vulnerability to distractors/noise.

\begin{figure}[btp]
\centering
  \includegraphics[width=\linewidth]{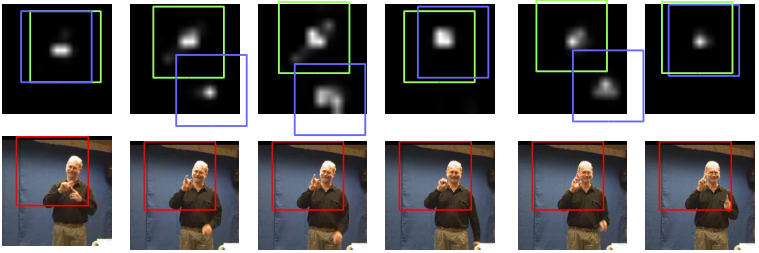}
  \caption{\label{fig:vtb} Illustration of one iteration of iterative
    attention, consisting of finding a zoomed-in ROI sequence based on
    the sequence of visual attention maps. 1st row: sequence of
    attention maps overlaid by candidate boxes of every frame. Green
    boxes are selected by dynamic programming. 2nd row: final sequence
    of bounding boxes after averaging. }
\end{figure}

To get the benefit of high resolution without the extra computational
burden, we propose to iteratively focus on regions within the input image frames, by
refining the attention map. Given a trained attention model
$\mathcal{H}$, we run inference with $\mathcal{H}$ on the target
image sequence $\v I_1,\v I_2,\ldots,\v I_T$ to generate the
associated sequence of posterior attention maps: $\mathbf{A}_1, \mathbf{A}_2,\ldots,\mathbf{A}_T$. We use the sequence of attention maps to obtain a new
sequence of images $\v I_1^\prime, \v I_2^\prime,\ldots,\v I_T^\prime$
consisting of smaller bounding boxes within the original images. The
fact that we extract the new frames by \emph{zooming in on the
  original images} is key, since this allows us to retain the highest
resolution available in the original video, while restricting our
attention and without paying the price for using large high
resolution frames.

We then train a new model $\mathcal{H^\prime}$ that takes $\v I_1^\prime,\v I_2^\prime,\ldots,\v I_T^\prime$ as input.  We can iterate this process, finding increasingly smaller regions of interest (ROIs).
This iterative process runs for
$S$ steps (producing $S$ trained models) until ROI images of sufficiently high resolution are obtained.  In practice, the stopping criterion for iterative attention is based on fingerspelling accuracy on held-out data.

Given a series of zooming ratios (ratios between the size of the
bounding box and the full frame) $R_1, R_2..., R_S$, the
zooming process  sequentially finds a series of bounding box sequences
$\{b_t^1\}_{1\leq t\leq T}, ..., \{b_t^S\}_{1\leq t\leq T}$. We
describe in the experiments section how we choose $R$s.

This iterative process generates $S$ models.  At test time, for each input image sequence $\v I_1, \v I_2, ..., \v I_T$, the models 
$\mathcal{H}_{1:S-1}$ are run in sequence to get a sub-region sequence
$\v I^{S-1}_1, \v I^{S-1}_2, ..., \v I^{S-1}_T$.
For simplicity we just use the last model $\mathcal{H}_S$ for
word decoding based on input $\v I^{S-1}_1, \v I^{S-1}_2, ..., \v
I^{S-1}_T$
.
The iterative attention process is illustrated in
Algorithm \ref{alg:iter} and Figures~\ref{fig:vtb}.


\begin{algorithm}[htp]
\centering
  \caption{\label{alg:iter} Iterative attention via zooming in. }
  \begin{algorithmic}[1]
    \SingleLineDown{Training, Input: $\{(\v I^{n, 0}_{1:T_n}, w^n)\}_{1\leq n\leq N}$}\;
    \For{$s \in \{1, 2,..., S\}$}
    \State Train model $\mathcal{H}_s$ with $(\v I^{n, s-1}_{1:T_n}, w^n)_{1\leq n\leq N}$\;
    \For{$n=1,...N$}
    \State Run inference on $\v I^n_{1:T_n}$ with $\mathcal{H}_s$ to
    obtain\\
    \hskip4em attention map $\v A^n_{1:T_n}$\;
    \State Solve Equation \ref{eq:vtb_max} to obtain sequence of\\
    \hskip4em bounding boxes $b^n_{1:T_n}$ \;
    \State Crop and resize $\v I^{n, 0}_{1:T_n}$ with $b^n_{1:T_n}$ to get $\v I^{n, s}_{1:T_n}$\;
    \EndFor
    \EndFor
    \State Return $\{\mathcal{H}_s\}$, $1\leq s\leq S$\;
  \end{algorithmic}
    \begin{algorithmic}[1]
    \DoubleLine{Test, Input: $\v I^0_{1:T}$, $\{\mathcal{H}_s\}$,
      $1\leq s\leq S$}
    \For{$s \in \{1, 2,..., S\}$}
    \State Run inference on $\v I^{s-1}_{1:T}$ with $\mathcal{H}_s$ to
    obtain attention\\
    \hskip4em map $\v A_{1:T}$ and predicted words $w^s$\;
    \State Solve Equation \ref{eq:vtb_max} to obtain sequence of\\
    \hskip4em bounding boxes $b_{1:T}$ \;
    \State Crop and resize $\v I^0_{1:T}$ with $b_{1:T}$ to get $\v I^{s}_{1:T}$\;
    \EndFor
    \State Return $\v w^S$
  \end{algorithmic}
  \end{algorithm}

 We next detail how bounding boxes are obtained in each
iteration of iterative attention (illustrated in Figure~\ref{fig:vtb}).  In each iteration $s$, the goal is
to find a sequence of bounding boxes $\{b_1, b_2, ..., b_T\}$ based on
the posterior attention map sequence $\{\v A_1, \v A_2, ..., \v A_T\}$ and
the zoom factor $R$, which determines the size of $b_i$ relative to the size of
$\v I^s$.  In each frame $\v I^s_t$, we put a
box of size $R_s|\v I^s|$ centered at each of the top $k$ peaks in the
attention map $\v A_t$. Each box $b_t^i$, $t\in[T]$, $i\in[k]$, is
assigned a score $a_t^i$ equal to the attention
value at its center.
We define a linking score between two bounding boxes $b_t^i$ in consecutive frames as follows:
\begin{equation}\label{eq:vtb_link}
  sc(b_t^i, b_{t+1}^j)=a_t^i+a_{t+1}^j+\lambda * IoU(b_t^i, b_{t+1}^j),
\end{equation}
\noindent where $IoU(b_t^i, b_{t+1}^j)$ is the Jaccard index (intersection over
union) of $b_t^i$ and $b_{t+1}^j$ and $\lambda$ is a hyperparameter
that trades off between the box score and
smoothness. Using $IoU$ has a smoothing effect and
ensures that the framewise bounding box does not switch between 
hands. This formulation is analogous to finding an ``action tube'' in
action recognition~\citep{Gkioxari2015finding}. Finding the sequence of
bounding boxes with highest average $s$ can be written as the optimization problem
\begin{equation}\label{eq:vtb_max}
  \argmax_{i_1, \ldots, i_T}\frac{1}{T}\displaystyle\sum_{t=1}^{T-1} sc(b_t^{i_t}, b_{t+1}^{i_{t+1}})
\end{equation}
which can be efficiently solved by dynamic programming.
Once the zooming boxes are found, we
take the average of all boxes within a sequence for further
smoothing, and finally crop the zoom-in region from the original (full
resolution) image frames to avoid any repetitive interpolation
artifacts from unnecessary resizing.  
We describe our process for
determining the zoom ratios $R_{1:S}$ in Section~\ref{sec:end2end-experiment}.

\newcommand{\set}[1]{{\it{#1}}}

\section{Experimental Setup}
\label{sec:end2end-experiment}

We report results on three evaluation sets: \set{ChicagoFSWild/dev} is used to initially assess various methods and select the
most promising ones; \set{ChicagoFSWild/test} results are directly
comparable to the pipeline approach presented in chapter~\ref{ch:pipeline}; and finally, results on \set{ChicagoFSWild+/test} set
provide an additional measure of accuracy in the wild on a set more
than twice the size of \set{ChicagoFSWild/test}. 
These are the only existing data sets for sign language fingerspelling recognition ``in the wild'' to our knowledge.  
Performance is measured in terms of letter accuracy (in percent), computed by finding the minimum edit (Hamming) distance alignment between the hypothesized and ground-truth letter sequences. 

\subsection{Initial Frame Processing}
We consider the following scenarios for initial processing of the input frames:

\noindent\textbf{Whole frame} Use the full video frame, with no cropping.\\
\noindent\textbf{Face ROI} Crop a region centered on the face detection box, but 3 times larger.\\
\noindent\textbf{Face scale} Use the face detector, but instead of cropping, resize the entire frame to bring the face box to a canonical size (36 pixels).\\
\noindent\textbf{Hand ROI} Crop a region centered on the box resulting from the signing hand detector, either the same size as the bounding box or twice larger (this choice is a tuning parameter).

\subsection{Face Detection}
Here we provide detail on how we
apply the face detector in the {\bf face ROI} and {\bf face scale} setups, in
particular on how the ROI is extraced and scaled.

To save
computation, the face detector is run on one in every five frames per
sequence, interpolating the detections for the remaining 80\% of the frames. If only one face is detected, we take the average of all bounding boxes for the whole
sequence. In cases where multiple faces are
detected,  we first find a smooth ``face tube'' by successively taking
the bounding box in the next frame that has the highest IoU with the face
bounding box in the current frame. For every tube, a motionness score is
defined as the average value of optical flow within a surrounding
region ($3\times$ size of bounding box). Finally the tube with the highest score is selected and again the box is averaged over the whole sequence. In cases where face detection fails, we use the mean of all face bounding boxes detected in all images of the same size in the training set. We empirically observe that the failure case where no face is detected is rare ($\sim 0.5\%$ of the training set).

In the \textbf{face ROI} setting, a large region centered on the detected face bounding box is cropped and resized to serve as input.
This is because the signing hand(s) are spatially close to the face during fingerspelling. 
Specifically we crop a region centered on the bounding box which is 3
times larger. The ROI is resized with a ratio of
$\frac{224}{max(w_{roi}, h_{roi})}$ and then padded on the short side to make a squared target image of size $224\times 224$.

 In the \textbf{face scale} setting, we only scale the original frame based on the size of the face bounding box to avoid artifacts arising from cropping. The purpose of scaling is to make the scale of hands in different input sequences roughly uniform.  As our data are from videos with a large variety of viewpoints and resolutions, the scale of the hands varies over a wide range.
For instance the proportion of the hand in an image from a webcam video can be several times larger than that in an image from a third-person view.
Specifically we pre-set a base size $b$ (36 in our experiments) for the face bounding box. Input images of original size $W_{I}\times H_I$ with a bounding box of size $w_I\times h_I$ are rescaled with a ratio of $\frac{b}{max(w_I, h_I)}$. If the image area is larger than $224\times 224$ after rescaling, we further rescale by a ratio of $\alpha$ to ensure the resulting image has at most $224\times 224$ resolution due to memory constraints. $\alpha$ is multiplied in the iterative zooming-in for that input sequence.

\subsection{Model Variants}
Given any initial frame processing \textbf{X} from the list above, we compare several types of models: 

\noindent\textbf{X} Use the sequence of frames/regions in a recurrent convolutional CTC model directly (as in Figure~\ref{fig:model_architecture}, but without visual attention).  For \textbf{X = Hand ROI}, this is the approach used in chapter~\ref{ch:pipeline}, the pipeline approach of open-vocabulary fingerspelling recognition in the wild.\\
\noindent\textbf{X+attention} Use the model of Figure~\ref{fig:model_architecture}.\\
\noindent\textbf{Ours+X} Apply our iterative attention approach starting with the input produced by \textbf{X}.\\

The architecture of the recognition model described in Sec.~\ref{sec:arnn} is the same in all of
the approaches, except for the choice of visual attention model.
All input frames (cropped or
whole) are resized to a max size of 224 pixels, except for \textbf{Face scale} which yields arbitrary sized frames.
Images of higher resolution are not used due to memory constraints.

\newcommand{\nores}{{\textcolor{red}{\textbf{?}}}}

\begin{table}[t!]
  \centering{
  \begin{tabular}{l|c}
    \toprule
    {\bf Method}& {\bf Letter accuracy (\%)}\\
    \hhline{|==|}
    Whole frame & 11.0\\
    Whole frame+attention & 23.0\\
    Ours+whole frame & 42.8\\
    Ours+whole frame\textcolor{blue}{+LM}  & 43.6\\
    \hhline{|==|}
    Face scale & 10.9\\
    Face scale+attention & 14.2\\
    Ours+face scale& 42.9\\
    Ours+face scale\textcolor{blue}{+LM}& 44.0\\
    \hhline{|==|}
    Face ROI & 27.8\\
    Face ROI+attention & 33.4\\
    Face ROI+attention\textcolor{blue}{+LM} &35.2 \\
    Ours+face ROI & 45.6\\
    Ours+face ROI\textcolor{blue}{+LM} & \textbf{46.8}\\
    \hhline{|=|=|}
    Hand ROI~\citep{shi2018american} & 41.1\\
    Hand ROI\textcolor{blue}{+LM} ~\citep{shi2018american}& 42.8\\
    Hand ROI+attention & 41.4\\
    Hand ROI+attention\textcolor{blue}{+LM} & 43.1\\
    Ours+hand ROI& 45.0\\
    Ours+hand ROI\textcolor{blue}{+LM}  & 45.9\\

    \bottomrule
  \end{tabular}}
  \caption{Results on \set{ChicagoFSWild/dev}; training on
    \set{ChicagoFSWild/train}. Ours+X: iterative attention (proposed method)
    applied to input obtained with X. \textcolor{blue}{+LM}: add language model trained on
    \set{ChicagoFSWild/train}.}
  \label{tab:dev}
\end{table}

We use the signing \textbf{hand detector} described in Chapter~\ref{ch:pipeline}, and the two-frame motion (\textbf{optical flow}) estimation algorithm of
Farneback~\citep{Farnebck2003TwoFrameME}.

We use the DLIB face detector~\citep{King2009DlibmlAM},
       trained on the WIDER data set~\citep{yang2016wider}.  To save computation we run the face detector
       on one in every five frames in each sequence and interpolate to get bounding boxes for the remaining frames.       
In cases where multiple faces are detected, we form ``face tubes'' by connecting boxes in subsequent frames
with high overlap. Tubes are scored by average optical flow within an 
(expanded) box along the tube, and the highest scoring tube is
selected. Bounding box positions along the tube are averaged,
producing the final set of face detection boxes for the sequence. See
supplementary material for additional details.

\textbf{Model training} The convolutional layers of our model are based on
AlexNet~\citep{Krizhevsky2012imagenet} pre-trained on ImageNet~\citep{deng2009imagenet}. The last
max-pooling layer of AlexNet is removed so that we have a
sufficiently large feature map. When the input images are of size $224\times 224$,
the extracted feature map is of size $13\times 13$; larger inputs
yield larger feature maps.
 We include 2D-dropout layers between the last three convolutional layers with drop rate 0.2. For the RNN, we use a one-layer 
 LSTM network with 512 hidden units.  
 The model is trained with SGD, with an initial learning rate of 0.01 for 20 epochs and 0.001 for an additional 10 epochs. We use development set accuracy for early stopping. We average optical flow images at timestep $t-1$, $t$ and $t+1$, and use the magnitude as the prior map $\v M_t$ (Equation~\ref{eq:attn_map}) for time step $t$. The language model is an LSTM with 256 hidden units, trained on the training set annotations.\footnote{Training on external English text is not appropriate here, since the distribution of fingerspelled words is quite different from that of English.} 
Experiments are run on an NVIDIA Tesla K40c GPU.

\textbf{Zoom ratios} For each iteration of the iterative visual
attention, we consider zoom ratios $R\in\{0.9, 0.9^2, 0.9^3, 0.9^4\}$,
and find the optimal sequence of ratios by beam search, with beam size
2, using accuracy on \set{ChicagoFSWild+/dev} as the evaluation
criterion. The parameter $\lambda$ in Equation~\eqref{eq:vtb_link} is tuned to 0.1.

\section{Results and Analysis}

\subsection{Main Results}
\noindent\textbf{Results on dev} Table~\ref{tab:dev} shows results on \set{ChicagoFSWild/dev} for models
trained on \set{ChicagoFSWild/train}.

First, for all types of initial frame processing, performance is improved by
introducing a standard visual attention mechanism. The whole-frame approach (whether scaled
by considering face detections, or not) is improved the
most
, since without attention too much of model capacity is wasted on
irrelevant regions; however, attention applied to whole frames remains inferior to ROI-based
methods. Using the pre-trained hand or face detector to guide the ROI
extraction produces a large boost in accuracy, confirming that
focusing the model on a high-resolution, task-relevant ROI is
important.
These ROI-based methods still benefit from adding standard
attention, but the improvements are smaller (face ROI: $5.6\%$, hand ROI: $0.3\%$). 

In contrast, our iterative attention approach, which does not
rely on any pretrained detectors, gets better performance than
detector-based methods, including the pipeline approach (\textbf{Hand ROI}), even when attention is
added to the latter (42.8\% for \textbf{Ours+whole frame} vs.~41.4\% for \textbf{Hand ROI+attention}). Our approach of
(gradually) zooming in on an ROI therefore outperforms a
signing hand detector. Specifically in \textbf{Hand ROI}, the improvement suggests signing hands can get more precisely located with our approach after initialization from a hand detector.

Finally, adding a language model yields modest accuracy improvements across the board.
The language model has a development set perplexity of 17.3, which is quite high but still much lower than the maximum possible perplexity (the number of output labels). 
Both the high perplexity and small improvement from the language model are unsurprising, since fingerspelling is often used for rare words.

\begin{table}[th!]
  \centering{
  \begin{tabular}{l|c|c}
    \toprule
    {\bf Method}& {\bf ChicagoFSWild} & {\bf ChicagoFSWild+}\\
        \hline
    Hand ROI\textcolor{blue}{+LM}~\citep{shi2018american} & 41.9 & 41.2\\
    ~~~\textcolor{green!50!black}{+new data} & \textcolor{green!50!black}{57.5} & \textcolor{green!50!black}{58.3}\\
    \hline
    Ours+whole frame\textcolor{blue}{+LM} & 42.4& 43.8\\

    ~~~\textcolor{green!50!black}{+new data}      &\textcolor{green!50!black}{57.6} &
                                                                                               \textcolor{green!50!black}{61.0}\\
    \hline
    Ours+hand ROI\textcolor{blue}{+LM} & 42.3 & 45.9\\
    ~~~\textcolor{green!50!black}{+new data}      &\textcolor{green!50!black}{60.2} &
                                                                                               \textcolor{green!50!black}{61.1}\\

    \hline
    Ours+face ROI\textcolor{blue}{+LM} & \textbf{45.1} &
                                                         \textbf{46.7}\\
    ~~~\textcolor{green!50!black}{+new data} & \textbf{\textcolor{green!50!black}{61.2}} &\textbf{\textcolor{green!50!black}{62.3}}\\
    \bottomrule
  \end{tabular}}
  \caption{Results on \set{ChicagoFSWild/test}
    and
    \set{ChicagoFSWild+/test}.
    Black: 
    trained
    on \set{ChicagoFSWild/train}; \textcolor{green!50!black}{Green}:
    trained on
    \set{ChicagoFSWild/train + ChicagoFSWild+/train}. 
  }
  \label{tab:test}
\end{table}

\noindent\textbf{Results on test} We report results on
\set{ChicagoFSWild/test} for the methods that are most
competitive on dev (Table~\ref{tab:test}). All of these use some
form of attention (standard or our iterative approach) and a language
model.

The combination of face-based initial ROI with our iterative attention
zooming produces the best results overall.
This is likely due to the complexity of
our image data. In cases of multiple moving objects in the same image,
the zooming-in process may fail especially in initial iterations of whole frame-based processing, when the resolution of the hand is very low because of downsampling given memory constraints.  On the other hand, the initial face-based
ROI cropping is likely to remove clutter and distractors without loss
of task-relevant information. However, even without cropping to the
face-based ROI, our approach ({\bf Ours+whole frame+LM})  still improves over
the hand detection-based pipeline approach.

\noindent\textbf{Training on additional data} Finally, we report the
effect of extending the training data with the \set{ChicagoFSWild+/train}
set, 
increasing the number of training
sequences from 5,455 to 55,856.  
The crowdsourced annotations in 
ChicagoFSWild+ may be noisier, but they are much more plentiful. In
addition, 
the crowdsourced training data includes two annotations of each sequence, which can be seen as a form of natural data augmentation. 
As Table~\ref{tab:test} shows (in \textcolor{green!50!black}{green}),
all tested models benefit significantly from the new data. But the large
gap between our iterative attention approach and the hand
detector-based approach of~\cite{shi2018american} remains.
The improvement of end-to-end approach over \cite{shi2018american} applied to whole frames is \emph{larger} on the ChicagoFSWild+ test set. 
The hand detector could become less accurate due to possible domain discrepancy
between ChicagoFSWild (on which it was trained) and ChicagoFSWild+. 
In contrast, our model replacing the off-the-shelf hand detector
with an iterative attention-based ``detector'' is not influenced
by such a discrepancy.

\begin{figure}[btp]
\centering
\begin{tabular}{@{}c@{}}
\includegraphics[width=0.6\linewidth]{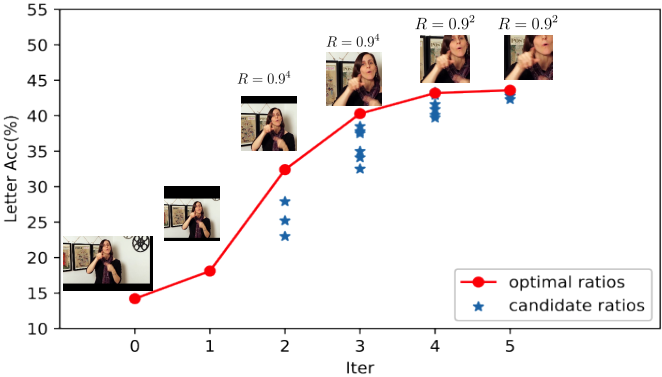}    \\
\includegraphics[width=0.6\linewidth]{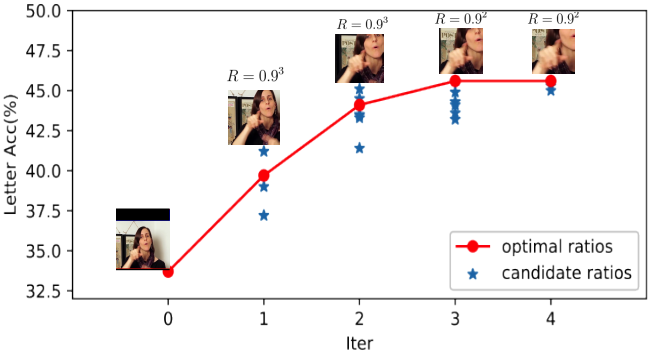} 
\end{tabular}
  \caption{\label{fig:acc_iter}Letter accuracy vs.~iteration in
    the {\bf Ours+face ROI} setting, showing an example ROI zooming ratio sequence found by
    beam search (shown,
    red curve).  Blue stars: accuracy with other zooming ratios considered.}
\end{figure}

\subsection{Effect of Iterative Zooming}

 The results of Table \ref{tab:dev} indicate that iterative zooming gives a large performance boost over the basic model.  In both face ROI and whole frame setups,
the hand corresponds to only a small portion of the input image.  
Figure \ref{fig:acc_iter} shows how the accuracy and the input image evolve in successive zooming iterations. Though no supervision regarding the hand is used for training, the location of the signing hand is implicitly learned through the attention mechanism. Higher recognition accuracy suggests that  the learned attention locates the hand more precisely than a separately trained detector.  To test this hypothesis, we measure hand detection performance on the dev set from the hand annotation data in ChicagoFSWild.  At the same miss rate of 0.158, the average IoU's of the attention-based detector and of the separately trained hand detector are 0.413 and 0.223, respectively. Qualitatively, we also compare the sequences of signing hands output by the two detectors. 
See supplementary material for more details.

As we zoom in, two things happen:  The resolution of
the hand increases, and more of the potentially distracting background
is removed. One could achieve the former without the latter by \emph{enlarging}
the initial input by $1/R$. We compared this approach to iterative
attention.

In particular, we compare the accuracy of zooming at ratio $R$ and enlarging the input images by $\frac{1}{R}$ in the {\bf face ROI} setting. For this experiment, $R$ is set to $0.9^3$, corresponding to the zooming ratio we use in the first iteration.  Comparison on smaller ratios is not feasible due to GPU memory constraints (12GB in our case).  For both zooming and enlarging, the resolution of the signing hand is the same. As can be seen from Table \ref{tab:zoom_vs_enlarge}, zooming outperforms enlarging.  When the prior map is used, the gap between the two approaches is small. This is mainly because distracting portions can be filtered via the motion-based prior in our model. The gain of zooming becomes much larger when we do not use optical flow as a complementary prior, demonstrating the benefit of distraction removal in our approach. Additionally, the motion-based prior has a negligible effect on the accuracy of our approach in this setting. 

\begin{table}[htb]
  \centering
  \begin{tabular}{lcc}\toprule
    $R=0.9^3$ & Zooming & Enlarging \\ \midrule
    with prior & \textbf{39.6} & 39.3 \\ 
    without prior & \textbf{39.8} & 38.1 \\ \bottomrule
  \end{tabular}
  \caption{\label{tab:zoom_vs_enlarge}Accuracy comparison between zooming and enlarging in the {\bf face ROI} setting.}
\end{table}

To summarize, the {\it enlarging} approach has lower performance than iterative attention and was prohibitively memory-intensive (we could
not proceed past one zooming iteration). Moreover, the prior on attention became more important. Therefore, iterative attention allows us to operate at much
higher resolution than would have been possible without it, and in
addition helps by removing distracting portions of the input frames.

\subsection{Other Analysis}

\textbf{Robustness to Face Detection}  Here we investigate the sensitivity of 
the results to the accuracy of face detection.

A face detector is used in two experimental setups: {\bf face ROI} and {\bf face scale}. To see how robust the model is to face detection errors,
we add noise to the bounding box output by the face detector. Specifically, two types of noise were separately added: size noise and position noise.
For size noise, we perturb the actual face detection boxes by multiplying the width and height of the box by
factors each drawn from $\mathcal{N}(1,\sigma_{s}^2)$.
For position noise, we add values drawn from $\mathcal{N}(0,\sigma_{p}^2)$ to the center coordinates of the face detection boxes. Note that position noise only affects the {\bf face ROI} experiments. We vary $\sigma_s$, $\sigma_p$ and show results in Table \ref{tab:size_noise}, \ref{tab:position_noise}. 

\begin{table}[h]
  \centering
  \begin{tabular}{lccc}\toprule
    $\sigma_s$ & IoU & Face ROI & Face Scale \\ \midrule
    0.0 & 1.000 & \textbf{45.6} & 42.9 \\ 
    0.1 & 0.858 & 45.2 & 42.7 \\ 
    0.2 & 0.741 & 44.7 & 43.3 \\ 
    0.3 & 0.641 & 44.3 & \textbf{44.0} \\ 
    0.4 & 0.556 & 42.6 & 43.3 \\ \bottomrule
  \end{tabular}
  \caption{\label{tab:size_noise} Impact of size noise on letter accuracy for {\bf face ROI} and {\bf face scale} setups. IoU is measured between the perturbed and original bounding boxes.}
\end{table}

\begin{table}[h]
  \centering
  \begin{tabular}{lcc}\toprule
    $\sigma_p$ & IoU & Face ROI \\ \midrule
    0.0 & 1.000 & \textbf{45.6} \\
    0.5 & 0.780 & 45.2 \\ 
    1.0 & 0.621 & 45.0 \\ 
    1.5 & 0.499 & 44.6 \\ 
    2.0 & 0.402 & 44.2 \\ \bottomrule
  \end{tabular}
  \caption{\label{tab:position_noise} Impact of position noise on letter accuracy for the {\bf face ROI} setup. Note the {\bf face scale} is not affected by position noise. IoU is measured between the perturbed and original bounding boxes.}
\end{table}

Overall we find that position noise has a smaller impact on accuracy compared to size noise. The {\bf face scale} setup, where no cropping is done in pre-processing, is more robust to size noise than the {\bf face ROI} setup is. Adding size noise brings a small improvement in this setting, which provides evidence that the face detector we use is not perfect. To summarize, the recognition performance degrades gracefully with face detector errors.
%

\begin{figure}[htp]
  \begin{minipage}{1.0\textwidth}
    \vspace{0.1in}
    (2).
    \includegraphics[width=\textwidth]{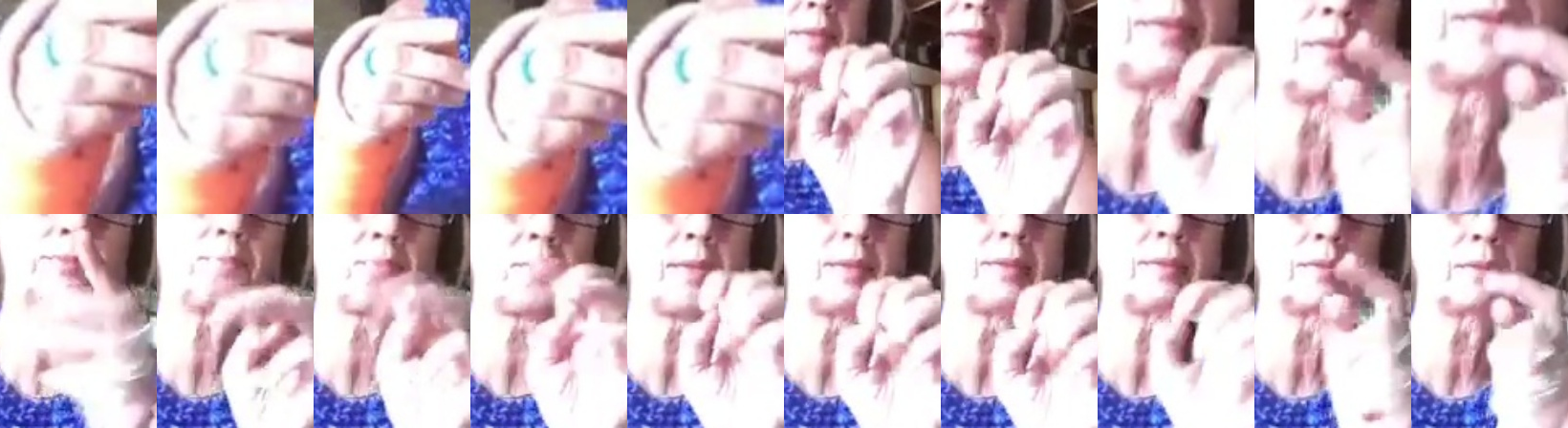}
  \end{minipage}
  \begin{minipage}{1.0\textwidth}
    \vspace{0.1in}
    (3).
    \includegraphics[width=\textwidth]{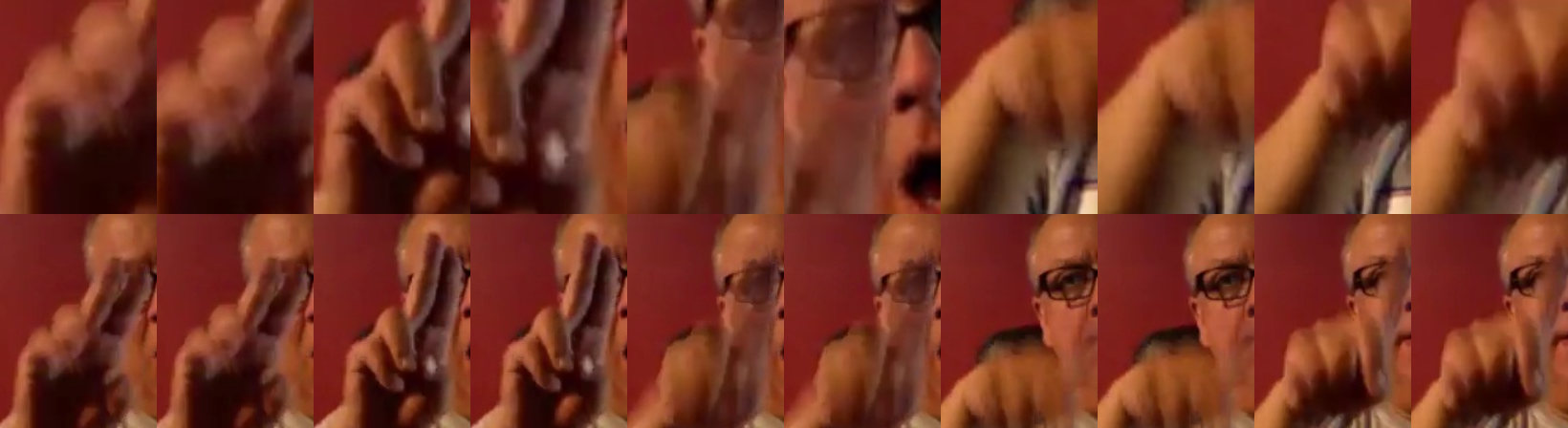}
  \end{minipage}
  \begin{minipage}{1.0\textwidth}
    \vspace{0.1in}
    (4). 
    \includegraphics[width=\textwidth]{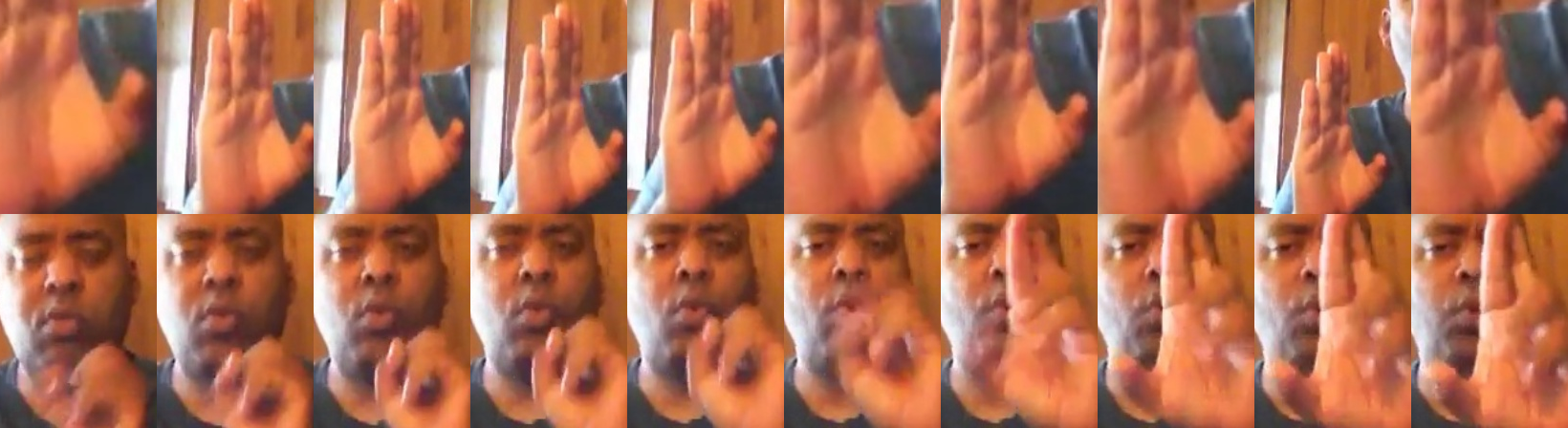}
  \end{minipage}
  \caption{\label{fig:comp_detector}Signing hands detected by the iterative attention detector vs.~the off-the-shelf signing hand detector~\cite{shi2018american}, taken from the ChicagoFSWild dev set. In each example, the upper row is from off-the-shelf detector and the lower row is from iterative attention. Signing hands are successfully detected by iterative attention in all cases. \\ Errors made by the off-the-shelf detector: In (1) and (2), bounding boxes are switched between signing and non-signing hand; in (3), the detected signing hand is incomplete; in (4), the non-signing hand is mis-detected as the signing hand. Note that sequence-level smoothing has already been incorporated in the off-the-shelf detector.}
\end{figure}

\textbf{Performance on signing hand detection}
Iterative attention serves as an implicitly learned ``detector'' of signing hands. We compare the performance of this detector
with a separately trained signing hand detector here. The signing hand detector is the one used in~\cite{shi2018american} and has been described in the chapter~\ref{ch:pipeline}.
We convert the iterative attention ROI to an explicit detector through the following steps: take the input image of the last iteration, backtrack to the original image frame to get its coordinates, and use these coordinates as the bounding box. We take a model trained in the {\bf face ROI} setting and compare it with an off-the-shelf detector. Figure \ref{fig:comp_detector}  shows example sequences from the ChicagoFSWild dev set, where our approach successfully finds signing sequences while the off-the-shelf detector fails.

\begin{table}[hbt]
  \centering
  \begin{tabular}{lcc}\toprule
     & Off-the-shelf~\citep{shi2018american} & Iterative-Attn \\ \midrule
    Avg IoU & 0.213 & \textbf{0.443} \\ 
    Avg Miss Rate & 0.158 & 0.158 \\ \bottomrule
  \end{tabular}
  \caption{\label{tab:comp_detector} Comparison of IoU between an off-the-shelf signing hand detector and a detector produced by iterative attention.}
\end{table}

For quantitative evaluation, we take the dev set of hand annotation data in ChicagoFSWild, which includes 233 image frames from 19 sequences, and remove all frames with two signing hands. That amounts to 200 image frames in total. We compute average IoU and miss rate between the target bounding box and ground truth. The miss rate is defined as 1-intersection/ground-truth area. 
As the two detectors have different IoU's and miss rates, for ease of comparison we resize the bounding box of the off-the-shelf detector to keep its miss rate consistent with that of the iterative-attention detector. As is shown in Table \ref{tab:comp_detector}, our detector almost doubles the average IoU of the off-the-shelf detector at the same miss rate. Though numerical differences between IoU's may be exaggerated due to the small amount of evaluation data, the effectiveness of our approach for localization of signing hands can also be inferred from improvements in recognition accuracy.

\section{Summary}
\label{sec:end2end-summary}

This chapter describes a new end-to-end model for ASL fingerspelling
recognition in the wild using an iterative attention mechanism. Our model gradually reduces its area of attention
while simultaneously increasing the resolution of its ROI
within the input frames, yielding a sequence of models of
increasing accuracy. In contrast to the pipeline approach, our end-to-end model
does not rely on hand detection, segmentation, or pose
estimation modules.  The
results of our method on both ChicagoFSWild and ChicagoFSWild+ are better than the results of previous methods by a large margin.

%% file: closing_gap.tex
\chapter{Improving Fingerspelling Recognition with Multi-Stream Conformers}
\label{ch:closing-gap}

In the previous chapters, we described the pipeline and end-to-end approaches for fingerspelling recognition. However, there is still a significant gap between the recognizer (61.2\%) and human performance (74.3\%-86.1\%). This chapter investigates further ways to improve fingerspelling recognition in the wild. To this end, we propose a multi-stream Conformer-based recognizer that jointly models hand and mouth sequences. The proposed model achieves 77.5\% letter accuracy on the ChicagoFSWild test set, which is on par with the performance of proficient signers.

\section{Conformers}

\subsection{Transformer}
Conformer~\cite{Gulati2020ConformerCT} is a variant of Transformer~\cite{Vaswani17transformer}, a popular sequence encoder applied in various areas~\cite{Vaswani17transformer,DosovitskiyB0WZ2021image,Dong2018SpeechTransformerAN}. 
Unlike LSTM, Transformer employs an attention mechanism to encode a sequence, which injects global information into the feature at every time step.

More formally, a Transformer encoder maps an input sequence $\v E\in\mathbb{R}^{T\times D}$ into an output sequence $\v O\in\mathbb{R}^{T\times D}$, where $T$ is the sequence length and $D$ is the feature dimension. An $N$-layer transformer stacks $N$ layers of identical structure, each composed mainly of a multi-head self-attention (MSA) and feedforward network (FNN).

%

\textbf{Multi-head Self Attention} At each timestep, the MSA layer conducts a weighted sum of input features.
Given three matrices of query $\v Q$, key $\v K$ and value $\v V$ where $\v Q,\v K,\v V\in\mathbb{R}^{N\times d}$, the Attention operation is defined as:

\begin{equation}
    \text{Attention}(\v Q,\v K,\v V)=\text{softmax}(\frac{\v Q\v K^T}{\sqrt{d}})\v V
\end{equation}
where $d$ is the feature dimension. 
To allow the model to attend to features representing different information, often multiple attention heads are used:

\begin{equation}
    \begin{split}
        & \text{MultiHead}(\v Q, \v K, \v V) = \text{Concat}(\text{head}_1,\text{head}_2,...,\text{head}_h)\v W^o \\
        &\text{head}_i = \text{Attention}(\v Q \mathbf{W}_i^Q, \v K\mathbf{W}_i^K, \v V\mathbf{W}_i^V), 1\leq i\leq h \\
    \end{split}
\end{equation}
where $\mathbf{W}_i^{Q}, \mathbf{W}_i^{K}, \mathbf{W}_i^{V}, \mathbf{W}^O$ are learnable parameters. 

Based on multi-head attention, the MSA operation is then defined as Equation~\ref{eq:close-gap-msa}. Intuitively, it computes attention where key, query and values are linear projections of the same feature sequence $\v X$. 

\begin{equation}
\label{eq:close-gap-msa}
    \text{MSA}(\v X)=\text{MultiHead}(\v X, \v X, \v X)
\end{equation}

\textbf{Feed-Forward Network} A Feed-Forward Network (FFN) is a multi-layer perceptron that is shared across timestamps: 

\begin{equation}
    \text{FFN}(\v X)=\text{ReLU}( \mathbf{X}\mathbf{W}_1+\mathbf{b}_1)\mathbf{W}_2+\mathbf{b}_2
\end{equation}
where $\v W_{\{1,2\}}, \v b_{\{1,2\}}$ are learnable parameters.

Suppose the input sequence fed into the $l$-th layer is $\v Z^l$,  the output $\v Y_l$ is computed as:

\begin{equation}
    \begin{split}
        &\v Y_{l} =\text{MSA}(\text{LN}(\v Z_l)) +\v Z_l \\
        &\v Z_{l+1} =\text{FFN}(\v Y_l) + \v Y_l \\
    \end{split}
\end{equation}
where LN denotes layer normalization~\cite{Ba2016LayerN}.

\subsection{Conformers}
Conformer~\citep{Gulati2020ConformerCT}, which stands for \textit{Convolution augmented Transformer}, is a popular encoder architecture in speech recognition~\citep{Guo2021RecentDO,Chung2021w2vBERTCC}.  
Conformer complements Transformer by inserting a convolutional module within each block. 
Moreover, it replaces the original FFN layer in the Transformer with two half-step FFN layers located before and after the self-attention module. 

\begin{equation}
    \begin{split}
        &\tilde{\v Y}_{l}=\v Z_l + \frac{1}{2}\text{FFN}(\v Z_l) \\
        & \mathbf{Y}_l^\prime = \tilde{\v Y}_{l} + \text{MSA}(\tilde{\v Y}_{l})\\
        %
        & \v Z_{l+1}=\text{LN}(\mathbf{Y}_l^\prime + \text{Conv}(\mathbf{Y}_l^\prime)+\frac{1}{2}\text{FFN}(\mathbf{Y}_l^\prime + \text{Conv}(\mathbf{Y}_l^\prime)))\\
    \end{split}
\end{equation}

Conv is a module consisting of gated convolution and normalization layers~\cite{Gulati2020ConformerCT}.
The convolutional layers in Conformer complement the Transformer by capturing local patterns in encoding. In FSR, the image sequence is monotonically aligned to letters. The local feature plays a vital role in correctly recognizing fingerspelling.

\section{Including Mouthing in FSR}
Mouth movement is one non-manual feature of sign language and has been used for various purposes, including modifying verbs or forming questions~\cite{liddel1989american,Liddell1978nonmanual}. 
Linguists use the term "mouthing" to refer to the mouth patterns in a sign language derived from its surrounding spoken language~\cite{Braem2001functions}. In ASL, prior linguistic studies~\cite{Davis1989distinguishing} described mouthing as a type of lexical borrowing, which is also one typical function of fingerspelling.  Specifically, in~\cite{Nadolske2008Occurence}, Nadolske and Rosenstock observed that mouthing frequently occurs in fingerspelled words. From a practical perspective, this suggests mouth movement may benefit fingerspelling recognition as a complementary signal to hand movement.
In sign language processing, including mouth movement has been shown to be effective in sign language recognition~\cite{koller2020weakly} or translation~\cite{camgoz2020multi}. However, most prior work~\cite{Kim2017LexiconfreeFR,Ricco2009accv,Goh2006DynamicFR,Liwicki2009automatic} in FSR is focused on modeling hands only, while the role of mouthing has not been studied to our knowledge. 

%

%
%
%
%

Here, we propose a framework that transcribes fingerspelling from two streams of input: hand and mouth, represented as two sequences of regions of interest (ROIs). The previous chapters described how to localize hands in a whole image frame. Specifically, one can either utilize a hand detector or an attention-based fingerspelling recognizer to detect the signing hand sequence from the whole image frames. To extract mouth ROIs, we employ a facial landmark detector and crop the mouth region based on the mouth keypoints (see Section~\ref{sec:better-exp} for implementation details).

In a single-stream fingerspelling recognizer, hand ROIs are encoded into visual features with a convolutional neural network (CNN), which are further fed into a sequence model for transcription. Two critical questions for a multi-stream recognizer receiving both hand and mouth sequences as input are: (1) how to encode mouth ROIs into the mouthing feature, and (2) how to fuse mouthing and hand features in the recognizer. We address those two questions as follows.

\input{figure-tex/better-fsr/mouthing-examples}

\textbf{Mouth Encoding}
Although mouth ROIs can be encoded with a randomly initialized CNN similar to hand encoding, feature learning for mouthing may benefit from other models for related tasks. In our datasets, we noticed many fingerspelled words are mouthed similarly as people speaking English (see Figure~\ref{fig:better-fsr-mouthing-examples}). 
A similar phenomenon was also observed in a prior linguistic study~\cite{Nadolske2008Occurence}, where Nadolske and Rosenstock hypothesized that elements considered part of English signing have a high occurrence of mouthing. In~\cite{Nadolske2008Occurence}, it was observed that the mouth movement in ASL signs "IMPORTANT", "SEE" and "TOO" match those of the complete English words "important", "see," and "too". Other example signs such as "FIRST", "DURING" and "BORROW" are mouthed like "ff", "doo" and "buh", which resembles the beginning movements of their English correlates.
Hence mouth-based fingerspelling recognition can potentially benefit from an English lip reading model, which is usually trained with more abundant training data. 
Motivated by this hypothesis, we use
the external lip-reading model AV-HuBERT~\cite{shi2022avhubert} to
extract 
features 
from the mouth ROIs.
We assume there is sufficient shared lip motion between English speaking and ASL mouthing though mouthing in ASL is not used to directly "say" words.

\begin{figure}[hbp]
    \centering
    \includegraphics[width=0.9\linewidth]{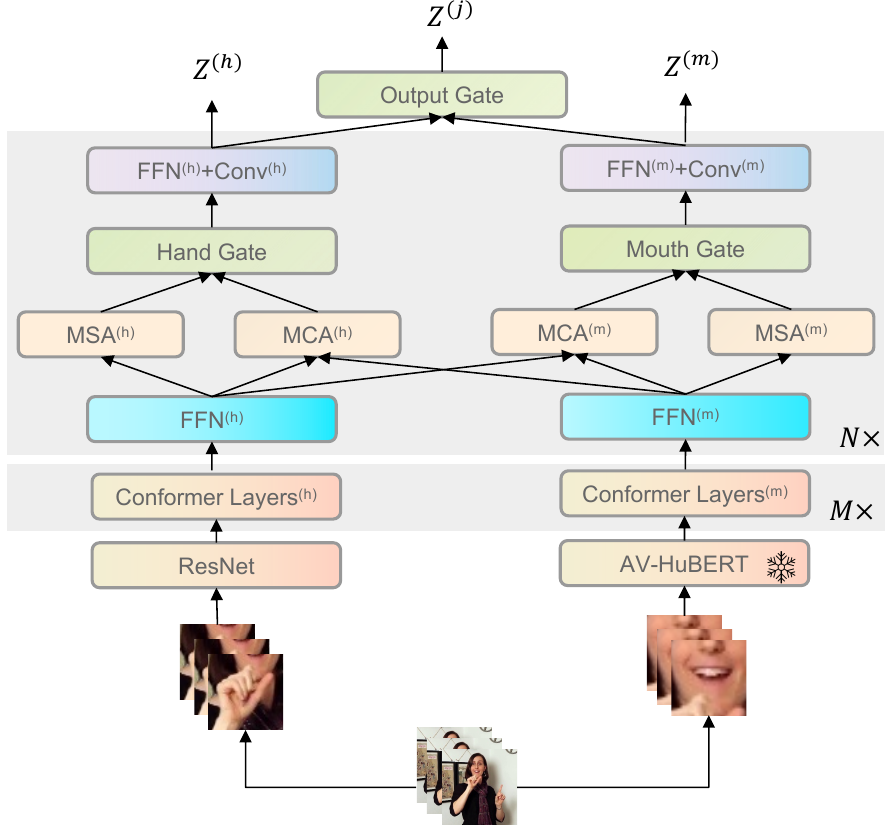}
    \caption{\label{fig:closing-gap-multimodal-conformer}Multi-stream Conformer for fingerspelling recognition. For clarity, residual connections are ignored. \SnowflakeChevron: frozen.}
\end{figure}

\textbf{Hand and mouth fusion}
A na\"ive way to fuse mouthing and handshape features is framewise concatenation, where mouthing and hand features are concatenated at the frame level before feeding into a shared Conformer model for sequence encoding.
%
%
%
%
%
However, such a strategy does not encode modality-specific sequential structure in the hand and mouth sequences.
Furthermore, the two streams are not always aligned on a frame basis. For example, the fingerspelled word "SEA" is composed of 3 different handshapes while it only has one single mouth shape. 
Thus using one single Conformer to encode the concatenated feature sequence may not well capture temporal relations presented in the mouth and hand stream. 
To address this issue, we propose a multi-stream Conformer encoder for fingerspelling recognition (see Figure~\ref{fig:closing-gap-multimodal-conformer}).  It is composed of separate Conformer modules for two streams which also model the interaction between the two modalities through the attention mechanism in intermediate layers.

Our multi-stream encoder produces two sequence encodings - $\v Z^{(m)}$ and $\v Z^{(h)}$, respectively for mouth and hand streams.
The mouth and hand features, encoded respectively by AV-HuBERT~\citep{shi2022avhubert} and ResNet~\citep{he2016deep}, are first transformed by $M$ modality-specific Conformer layers before feeding into an $N$-layer multi-stream module.
In the $l$-th layer of the multi-stream module, the mouthing features $\v Z_l^{(m)}$ and hand features $\v Z_l^{(h)}$ are initially transformed through half-step FFN and self-attention layers:

\begin{equation}
\label{eq:close-gap-multi-conformer-ffn}
\begin{split}
    & \tilde{\v Y}_l^{(x)}=\v Z_l^{(x)}+\frac{1}{2}\text{FFN}^{(x)}(\v Z_l^{(x)}) \\
    & \v Y_{l,s}^{(x)} = \tilde{\v Y}^{(x)}_l+\text{MSA}^{(x)}(\tilde{\v Y}^{(x)}_l),x\in\{m,h\}\\
\end{split}
\end{equation}
To facilitate information exchange between the mouthing and hand encoders,
we impose a cross-attention layer between the two modalities. The multi-head cross attention (MCA) layer, applied between features of two modalities $\v X$ and $\v Y$, is defined as:
$\text{MCA}(\v X, \v Y)=\text{MultiHead}(\v X, \v Y, \v Y)$,
where query and key lie in different modalities. We compute the cross-modality attended features $\v Y_l^{(x),c}$ for mouth and hand, respectively:
\begin{equation}
    \v Y_{l,c}^{(x)} = \tilde{\v Y}^{(x)}_l+\text{MCA}^{(x)}(\tilde{\v Y}^{(x)}_l),x\in\{m,h\}
\end{equation}
where $\text{MCA}^{(m)}$ and $\text{MCA}^{(h)}$ are cross-attention modules specific to mouthing and hand. To balance the weight of two modalities in each encoder, we employ a gating mechanism to combine self-attended and cross-attended features: 

\begin{equation}
\begin{split}
    & \v R^{(x)}=\sigma(\v W_s^{(x)}\v Y_{l,s}^{(x)}+\v W_c^{(x)}\v Y_{l,c}^{(x)}+\v b^{(x)})\\
    & \text{Gate}(\v Y_{l,s}^{(x)},\v Y_{l,c}^{(x)})=\v Y_{l,s}^{(x)}\odot \v R^{(x)}+\v Y_{l,c}^{(x)}\odot(1-\v R^{(x)})\\
    & \v Y_l^{(x)\prime} = \text{Gate}(\v Y_{l,s}^{(x)},\v Y_{l,c}^{(x)})+\tilde{\v Y}^{(x)},x\in\{m,h\}\\
\end{split}
\end{equation}
Applying the gating mechanism allows the model to attribute weights to different modalities in making predictions. In FSR, hand movement is usually more critical to recognition than mouthing. The use of gating helps the model to learn such bias implicitly. The attended feature $\v Y_l^{(x)\prime}$ is subsequently encoded with convolutional and feedforward layers similar to a single-modality Conformer:
\begin{equation}
    \v Z^{(x)}_{l+1}=\text{LN}(\mathbf{Y}_l^{(x)\prime} + \text{Conv}(\mathbf{Y}_l^{(x)\prime})+\frac{1}{2}\text{FFN}(\mathbf{Y}_l^{(x)\prime} + \text{Conv}(\mathbf{Y}_l^{(x)\prime}))),x\in\{m,h\}
\end{equation}

Finally, we fuse mouthing and hand encoding into a joint embedding similarly:

\begin{equation}
    \v Z^{(j)} = \text{Gate}(\v Z^{(m)}, \v Z^{(h)})
\end{equation}

The final encoding 
$\v Z^{(j)}$ is mapped into frame-level label posteriors via a fully-connected
layer followed by a softmax, as in Chapter ~\ref{ch:pipeline} and~\ref{ch:e2efsr}. 
Based on the frame posteriors, at training time we compute CTC loss.
In addition to joint embedding, we also compute two auxiliary CTC losses based on mouth and hand encodings $\v Z^{(m)}$/$\v Z^{(h)}$. The final loss is defined as:

\begin{equation}
    L_{\text{total}}=L_{\text{ctc}}(\v Z^{(j)}, w) + \lambda_h L_{\text{ctc}}(\v Z^{(h)}, w)+\lambda_{m}L_{\text{ctc}}(\v Z^{(m)}, w)
\end{equation}
where $w$ is the ground-truth letter sequence, and $\lambda_h$/$\lambda_m$ are two hyperparameters controlling the weight of hand-based/mouth-based losses.

\section{Experiments}
\label{sec:better-exp}
We follow the protocol in chapter~\ref{ch:e2efsr} for data pre-processing.  ImageNet-pretrained ResNet-18~\citep{he2016deep} is used as the backbone CNN to extract visual features for each image frame. The number of modality-specific ($M$) and multi-stream layers ($N$) are 4 and 8, respectively. The hidden dimension and feedforward dimension are 512 and 2048, respectively. 
To extract the mouth ROI, we run a facial landmark detector~\citep{King2009DlibmlAM} on the whole image and crop an ROI based on the mouth keypoints. Each ROI is resized to $88\times 88$. To handle the different frame rates of raw videos, both hand and mouth ROI video clips are resampled to 25 fps.
We extract visual embeddings for mouth ROIs with AV-HuBERT Base~\cite{shi2022avhubert}, a state-of-the-art English lip reading model. The AV-HuBERT model is frozen during recognizer training.  
The model is trained with Adam~\citep{Kingma2015AdamAM} for $S_{t}$ steps with a learning rate linearly increased to $0.001$ for $S_w$ steps and decayed to 0 in the remaining steps. $S_w$/$S_t$ are 2K/48K and 4K/96K for ChicagoFSWild and ChicagoFSWild+ respectively. To accelerate training, we form batches such that the total number of frames per batch does not exceed 300 per GPU. The model is trained on 1 and 3 GPUs for ChicagoFSWild and ChicagoFSWild+ respectively.
$\lambda_{m}$ and $\lambda_{h}$ are tuned based on recognition accuracy on the combined validation sets from the two datasets, respectively. During decoding, we use a 5-gram character-level language model trained on the training set of ChicagoFSWild and ChicagoFSWild+, which has achieved a perplexity of 9.67 on the ChicagoFSWild dev set.

\subsection{Main Results}

The performance of our proposed multi-stream Conformer on the ChicagoFSWild and ChicagoFSWild+ test sets is shown in Table~\ref{tab:better-comparison-sota}, where we compare our results with prior approaches and human performance. AlexNet-RNN and Attn-RNN, which are described in Chapter~\ref{ch:pipeline} and~\ref{ch:e2efsr}, respectively, are also included in the comparison. Our proposed model outperforms all prior approaches and achieves human-level performance. When using only the ChicagoFSWild dataset for training (5,455 sequences), the multi-stream Conformer achieves a letter accuracy of 68.4\%, which is a significant improvement over the former state-of-the-art Pose TDC (50.0\%) by $\sim18$\% (absolute). Moreover, when both ChicagoFSWild and ChicagoFSWild+ are used for training, the letter accuracy further increases to 77.5\%, surpassing the accuracy of a proficient signer (74.3\%). However, our best model still falls behind the performance of a native signer (86.1\%) by a considerable margin, indicating that there is still ample room for future research in FSR.

%


\begin{table}[htp]
    \centering
    \resizebox{\linewidth}{!}{
    \begin{tabular}{ccccc}
    \toprule
        Model & Training Data & \# Training sequences & CFW Test & CFW$_{+}$ Test   \\ \midrule
        AlexNet-RNN~\citep{shi2018american} & CFW & 5,455 & 41.1 & 41.9 \\ 
        Attn-RNN~\citep{shi2019fingerspelling} &  &  & 45.1& 46.7   \\ 
        Pose-GRU~\citep{Parelli2020slpose}&  & & 47.9& -  \\
        Context-TF~\citep{Gajurel2021AFV} &  & & 48.3& -  \\
        Siamese-Embedding~\citep{Kumwilaisak2022American} &  & & 49.6 \\ 
        Pose-TDC~\citep{Papadimitriou2020MultimodalSL} &  & & 50.0& -  \\ 
        Multi-stream Conformer (ours) &  & & \textbf{68.4} & \textbf{60.2}  \\ \midrule
        Attn-RNN~\citep{shi2019fingerspelling} & CFW+$\text{CFW}_{+}$ & 55,857 & 61.2 & 62.3 \\ 
        Query-Attn~\citep{Kruthiventi2021fingerspelling} &  & & 69.0 & 66.2 \\
        Multi-stream Conformer (ours) &  & & \textbf{77.5} & \textbf{73.0} \\
         Human & - & - & [74.3, 86.1] & - \\ \bottomrule
    \end{tabular}
    }
    \caption{\label{tab:better-comparison-sota}Letter accuracy (\%) of our proposed transformer-based fingerspelling recognizer and prior work on the ChicagoFSWild test sets. CFW: ChicagoFSWild, CFW+: ChicagoFSWild+}
\end{table}

\subsection{Model Analysis}

In this section, we will conduct an ablation study on the multi-stream Conformer and compare it with several baselines for using mouthing in FSR.

\begin{table}[htp]
    \centering
    \begin{tabular}{ccc}
    \toprule
        Method & CFW training & CFW+CFW$_+$ training \\
        \midrule
        AlexNet-LSTM~\cite{shi2018american} & 42.8 & 58.0 \\
        ResNet-BiLSTM & 53.1 & 63.5 \\
        ResNet-Transformer & 58.0 & 68.1 \\
        ResNet-Conformer & 64.3 & 73.5 \\
        Multi-stream Conformer (ours) & \textbf{66.5} & \textbf{76.1} \\
        \bottomrule
    \end{tabular}
    \caption{\label{tab:closing-gap-ablation}Ablation Study on ChicagoFSWild dev}
\end{table}

\textbf{Ablation Study} Table~\ref{tab:closing-gap-ablation} compares the proposed multi-stream Conformer to a few baselines based on hand input. The first model we evaluate is ResNet-LSTM, which uses a single-layer LSTM for sequence modeling, similar to the recognizers described in Chapters~\ref{ch:pipeline} and~\ref{ch:e2efsr}. We observe that using ResNet for image encoding already outperforms previous models that use AlexNet. When we replace the LSTM with a Transformer for sequence modeling, we see a significant improvement in performance for both low and high-data regimes. Specifically, the Conformer-based recognizer outperforms a Transformer-based model, indicating the importance of local context modeling provided by the convolution module. Finally, our proposed multi-stream Conformer yields additional gains over the single-stream Conformer model that only uses hand input, demonstrating the benefits of incorporating mouth information for improved fingerspelling recognition.

\textbf{Mouth-only baseline} To investigate the effectiveness of using mouth-only input for fingerspelling recognition, we trained a model on ChicagoFSWild dataset using visual features encoded by AV-HuBERT. The resulting model achieved 14.6\% letter accuracy on the test set, surpassing the LM baseline (9.4\%, Chapter~\ref{ch:pipeline}) but falling significantly short of the hand-based model (64.3\%). This performance gap is expected as many fingerspelled words are only partially mouthed, making it challenging to distinguish them solely based on mouth movement. Additionally, predicting words based on mouthing is inherently ambiguous, with some groups of words (e.g., \emph{mark}, \emph{bark}, \emph{park}) being indistinguishable from mouth movement alone. In summary, our findings suggest that hand input is a more reliable modality than mouth input for fingerspelling recognition.

\textbf{Vanilla fusion} To combine the hand and mouth sequences into a multi-stream recognizer, we compared our proposed model to a vanilla fusion approach, where the mouth feature is concatenated with the hand feature (output from the ResNet) before feeding into a single Conformer. However, as shown in Table~\ref{tab:better-comparison-using-mouthing}, this fusion strategy performed worse than our proposed model and even worse than the hand-only baseline (see Table~\ref{tab:closing-gap-ablation}). We believe this is because the hand and mouth sequences are not aligned at the frame level, making it challenging for the model to learn meaningful correlations between them. In contrast, our proposed model encodes the two asynchronous sequences via attention, allowing features to be combined with arbitrary frames in the other modality. Moreover, the gating mechanism of multi-stream Conformer can implicitly learn the unequal importance of different modalities to fingerspelling recognition, which the vanilla fusion approach lacks. Overall, our proposed model outperforms vanilla fusion in leveraging both hand and mouth input for fingerspelling recognition.

\begin{table}[htp]
    \centering
    \begin{tabular}{ccc}
    \toprule
        Method & CFW training & CFW+CFW$_+$ training \\ \midrule
        Vanilla fusion  & 63.1 & 71.9 \\
        Face ROI  & 58.4 & 70.3 \\
        RGB-based mouthing  & 64.6 & 73.9 \\
        Multi-stream Conformer (ours)  & \textbf{66.5} & \textbf{76.1} \\
        \bottomrule
    \end{tabular}
    \caption{\label{tab:better-comparison-using-mouthing} Comparison between our model and the alternative ways of using mouthing for FSR.}
\end{table}

 \textbf{Face ROI} In principle, a recognizer based on the face region of interest (ROI) can utilize both handshape and mouth movement information for fingerspelling recognition, since this region usually includes both the mouth and hand areas. To test this, we compare our model to a ResNet-Conformer model based on the face ROI. The preprocessing steps for the face ROI can be found in Chapter~\ref{ch:e2efsr}, and we do not apply iterative attention for further zooming in. However, this baseline model falls far behind our proposed model (see Table~\ref{tab:better-comparison-using-mouthing}). This is partially due to the lower resolution of input images in the face ROI setting, which are downsampled before being fed into the model due to memory constraints. One additional benefit of explicitly using mouth and hand ROIs is that it facilitates background removal, which prevents the recognizer from being distracted by irrelevant regions in the whole image.

\textbf{AV-HuBERT vs. Raw Image}
Our default approach for representing mouth movement involves using a sequence of features extracted by an external lip reading model, AV-HuBERT. In this section, we compare this approach to a raw-image-based recognizer that uses a mouth-specific ResNet pre-trained on ImageNet to encode image frames. Our proposed model consistently outperforms the image-based recognizer in both CFW and CFW+CFW$_{+}$ training regimes, as shown in Table~\ref{tab:better-comparison-using-mouthing}.

Given that mouthing does not always occur during fingerspelling, learning image features for mouthing from fingerspelling sequences requires a large amount of data. However, a well-performing lip reading model like AV-HuBERT can provide visual features of mouth movement, learned from larger spoken language datasets and shared between speaking and mouthing. This relieves the data scarcity issue for mouthing and enhances the performance of our proposed model.

\textbf{Impact of handedness} Our dataset includes both left-handed and right-handed signers. To evaluate the impact of handedness on the recognizer, we conducted an analysis by flipping all the left-hand videos horizontally. We detected handedness using an off-the-shelf hand detector~\cite{shan20understanding} and trained a new fingerspelling recognizer with the same architecture as the original model, but specifically for right-hand videos. We conducted experiments using three different random seeds for both settings. The right-hand-specific recognizer achieved a letter accuracy of $76.0{\pm0.2}\%$ on the ChicagoFSWild dev set, comparable to $76.1{\pm0.3}\%$ achieved by the original model. This result indicates that our recognizer is robust to handedness.

\subsection{Error Analysis and Qualitative Examples}
\begin{figure}[htp]
    \centering
    \includegraphics[width=0.8\linewidth]{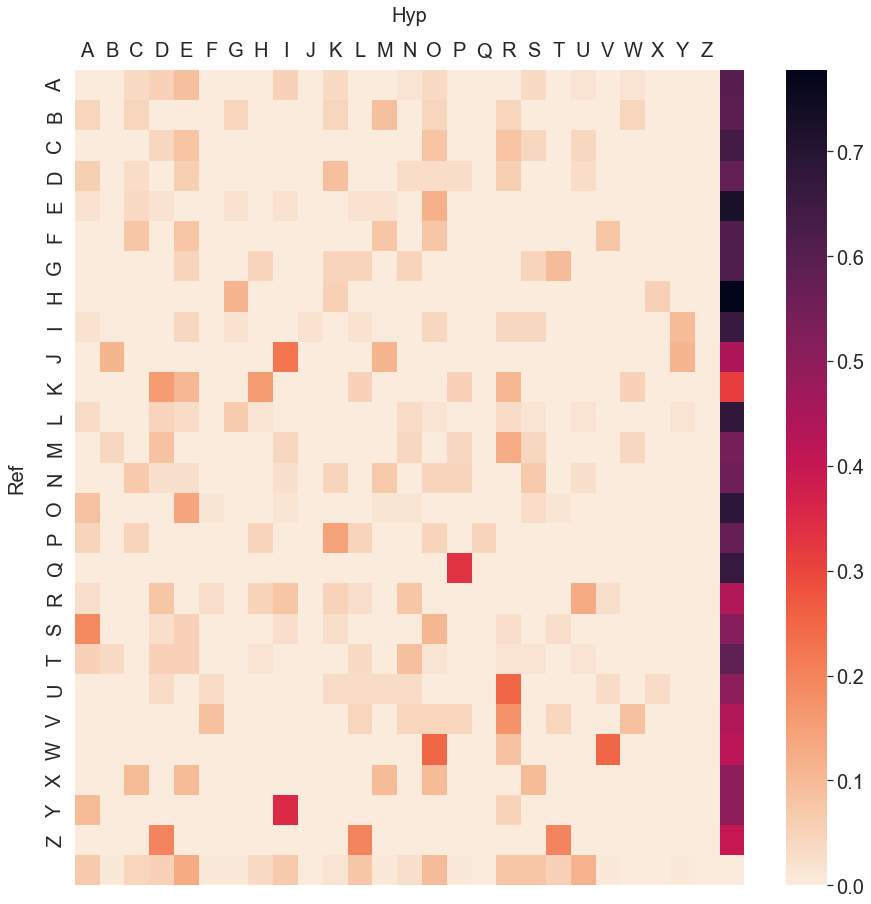}
    \caption{\label{fig:better-fsr-confmat}Confusion matrix of the proposed conformer-based fingerspelling recognizer. The diagonal has been removed for visual clarity. Row-wise normalization is applied to each cell after diagonal removal.}
\end{figure}

Figure~\ref{fig:better-fsr-confmat} shows the errors made by our fingerspelling recognizer. Deletions are the most common type of error, followed by substitutions and insertions. Our experiments show that tuning length penalty during beam search decoding can balance the three types of errors, but does not improve accuracy.
The distribution of errors is similar to that of the models presented in the previous chapters (chapter~\ref{ch:pipeline},~\ref{ch:e2efsr}), but the total number of errors made by our model is significantly lower. The most common substitution pairs are (O $\rightarrow$ E), (E $\rightarrow$ O), (Y $\rightarrow$ I), (S $\rightarrow$ A), (P $\rightarrow$ K), and (O $\rightarrow$ A). Handshapes for (E, A, S) and (Y, I) are similar, with the major difference being the position of the thumb. Similarly, the difference in hand orientation between P and K, as well as O and E, can be difficult for the model to recognize due to varying camera perspectives in real-world videos.
Figure~\ref{fig:better-fsr-correct-examples} and~\ref{fig:better-fsr-incorrect-examples} show qualitative examples of correct and errorful predictions made by the recognizer.

\input{figure-tex/better-fsr/correct-examples}

\input{figure-tex/better-fsr/incorrect-examples}

%

%

%
%

%
%
%

\section{Summary}

In this chapter, we proposed a multi-stream Conformer-based model for fingerspelling recognition, which uses both hand and mouth sequences. Our findings indicate that using Conformer as the sequence encoder significantly improves recognition accuracy, and including mouthing information brings further gains. Our proposed model achieved a recognition performance level comparable to that of a proficient signer. However, the accuracy of our best model is still considerably lower than that of a native signer, indicating that there is room for further research.

%% file: figure-tex/better-fsr/mouthing-examples.tex
\begin{figure*}[htp]
\centering
\begin{tabular}{c}
\toprule
GARDEN \\ 
 \includegraphics[width=\linewidth]{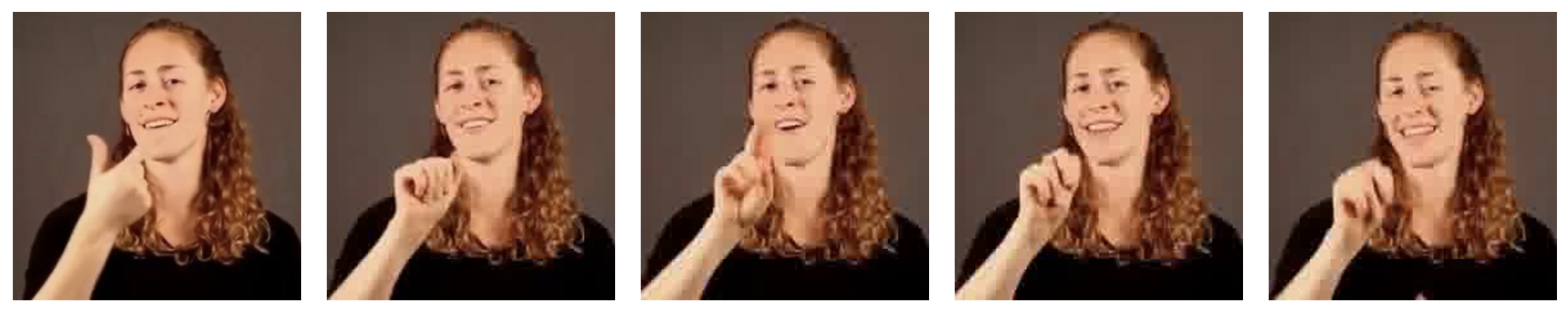}\\ \midrule
FUN \\ 
 \includegraphics[width=\linewidth]{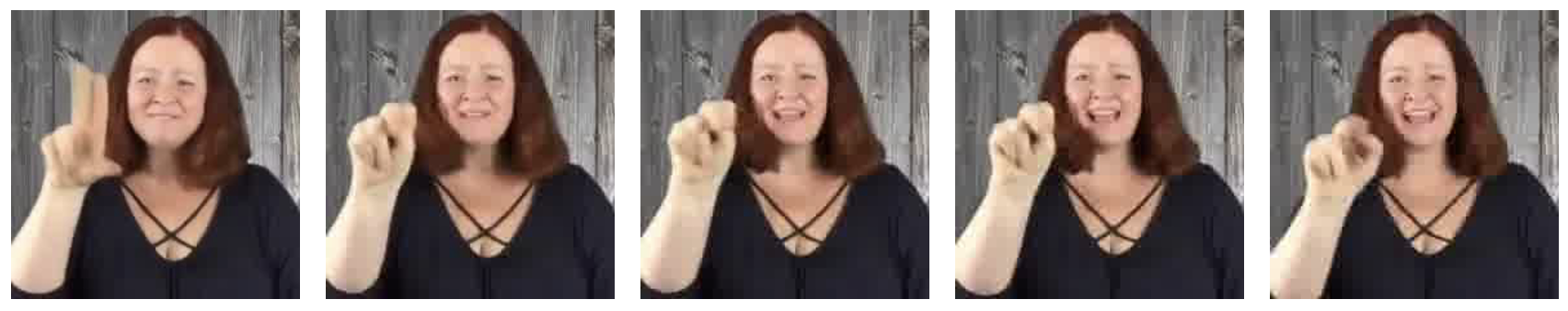}\\ 
\bottomrule
\end{tabular}
\caption{\label{fig:better-fsr-mouthing-examples}Mouthing examples in ChicagoFSWild}
\end{figure*}

%% file: figure-tex/better-fsr/correct-examples.tex
\begin{figure*}[htp]
\centering
\begin{tabular}{c}
\hline
IF \\ 
 \includegraphics[width=\linewidth]{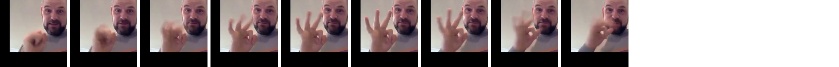}\\ \hline
\hline
OK \\ 
 \includegraphics[width=\linewidth]{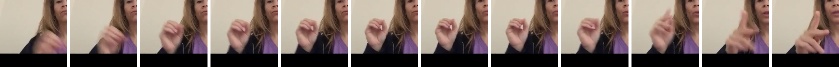}\\ \hline
\hline
ALL \\ 
 \includegraphics[width=\linewidth]{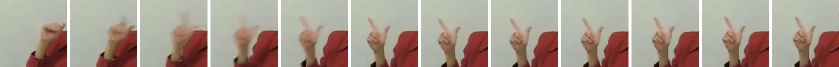}\\ \hline
\hline
ELITE \\ 
 \includegraphics[width=\linewidth]{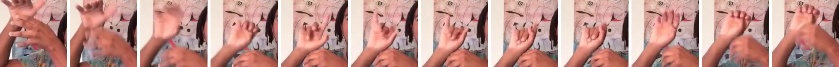}\\ \hline
\hline
OWN \\ 
 \includegraphics[width=\linewidth]{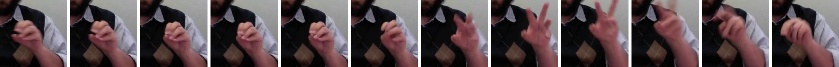}\\ \hline
\hline
LA \\ 
 \includegraphics[width=\linewidth]{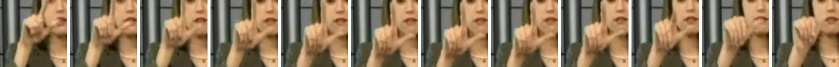}\\ \hline
\hline
MISN \\ 
 \includegraphics[width=\linewidth]{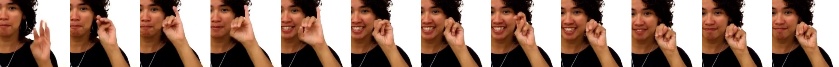}\\ \hline
\hline
ALL \\ 
 \includegraphics[width=\linewidth]{figure/better-fsr/correct-examples/ALL}\\ \hline
\hline
LITEACY \\ 
 \includegraphics[width=\linewidth]{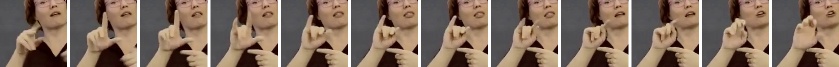}\\ \hline
\end{tabular}
\caption{\label{fig:better-fsr-correct-examples}Correct predictions from the recognizer}
\end{figure*}

%% file: figure-tex/better-fsr/incorrect-examples.tex
\begin{figure*}[htp]
\centering
\begin{tabular}{l}
\hline
REF: H\red{S} \\ 
 HYP: H\brown{A} \\ 
 \includegraphics[width=\linewidth]{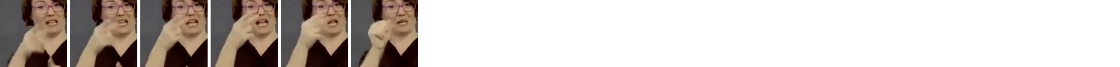}\\ \hline
\hline
REF: DRDONAM\red{M}ERT\red{E}NS \\ 
 HYP: DRDONAMER\brown{O}NS \\ 
 \includegraphics[width=\linewidth]{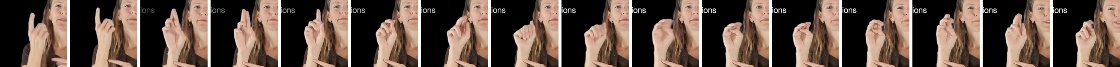}\\ \hline
\hline
REF: S\red{UP}ER \\ 
 HYP: S\brown{RK}ER \\ 
 \includegraphics[width=\linewidth]{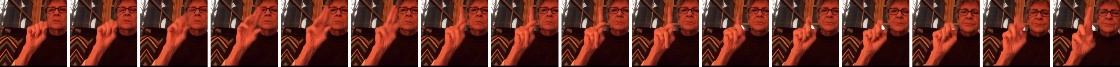}\\ \hline
\hline
REF: B\red{Y}E \\ 
 HYP: B\brown{I}E \\ 
 \includegraphics[width=\linewidth]{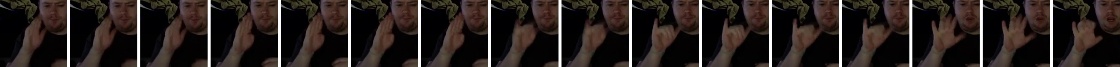}\\ \hline
\hline
REF: AV\red{VTAUY}DITORYVEBAL \\ 
 HYP: AV\brown{NARI}DITORYVEBALL \\ 
 \includegraphics[width=\linewidth]{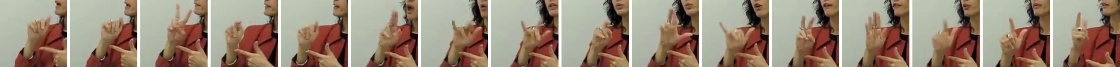}\\ \hline
\hline
REF: D\red{E}BT \\ 
 HYP: D\brown{O}BT \\ 
 \includegraphics[width=\linewidth]{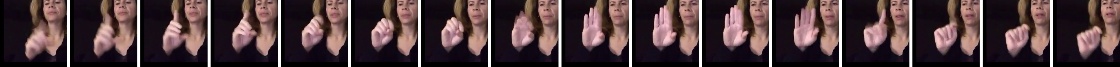}\\ \hline
\hline
REF: DAVIDORE\red{Y}NYOLDS \\ 
 HYP: DAVIDORE\brown{I}NYOLDS \\ 
 \includegraphics[width=\linewidth]{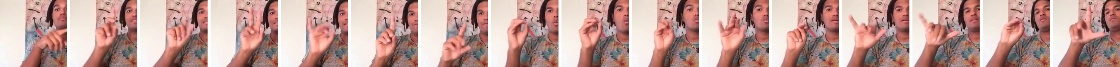}\\ \hline
\hline
REF: MYTHBUS\red{TE}RS \\ 
 HYP: MYTHBS\brown{DO}RS \\ 
 \includegraphics[width=\linewidth]{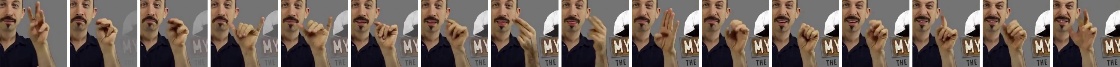}\\ \hline
\hline
REF: LITERAC\red{Y}ST\red{O} \\ 
 HYP: LITERAC\brown{I}ST \\ 
 \includegraphics[width=\linewidth]{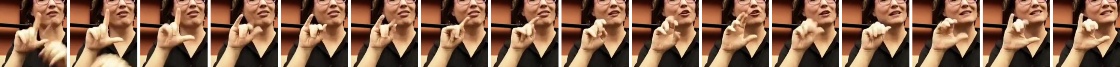}\\ \hline

\end{tabular}
\caption{\label{fig:better-fsr-incorrect-examples}Recognition errors. The deleted characters in the references are highlighted in \red{red}. The substituted and inserted characters in the hypotheses are highlighted in \brown{brown}.}
\end{figure*}

%
%

%% file: fs_detection.tex
\chapter{Fingerspelling Detection}
\label{ch:fsdet}

Chapter~\ref{ch:pipeline}-\ref{ch:closing-gap} study fingerspelling recognition and have assumed that the boundaries of fingerspelling regions in signing videos are known beforehand. In this chapter, we define the task of fingerspelling detection in raw,
untrimmed sign language videos. This is an important step
towards building real-world fingerspelling recognition systems. In the following sections, we will present a benchmark and a suite of evaluation
metrics for fingerspelling detection, some of which reflect the effect of detection on the
downstream fingerspelling recognition task. In addition, we
will propose a model that learns to detect fingerspelling
via multi-task training, 
incorporating pose estimation and
fingerspelling recognition (transcription) along with detection. The proposed
model outperforms 
alternative approaches adapted from related areas  and establishes a state of the art on the benchmark. This chapter is based on the paper~\citep{Shi2021FingerspellingDI}.

\section{Introduction}

Although fingerspelling recognition
has been widely
studied,
in most prior work~\citep{pugeault,Goh2006DynamicFR,Ricco2009accv,Liwicki2009automatic,Kim2012AmericanSL,Kim2013SCRF,Kim2016Adaptation,Kim2017LexiconfreeFR,Shi2017MultitaskTW,shi2018american,shi2019fingerspelling}, it is assumed that the
input sequence contains fingerspelling only, sometimes extracted from longer sequences of signing via human annotation.
Replacing human annotation with fully automatic
detection of fingerspelling -- identifying time spans in the
video containing fingerspelling -- is a hurdle that
must be cleared to enable truly practical fingerspelling
recognition ``in the wild''.

Fingerspelling detection has not been {widely} studied before. 
In principle, it can be treated as a special
case of action
detection~\citep{xu2017rc3d,chao2018rethinking,zhao2017temporal,Escorcia2016DAPsDA}. However,
in contrast to typical action detection scenarios, the actions in the
fingerspelling ``class'' are highly {heterogeneous} and many fingerspelling handshapes are also used in non-fingerspelled signs.  In addition, considering the
goal of using the detector as part of a complete sign language processing system, a fingerspelling detector should be evaluated based on its effect on a downstream recognition
model, a step not normally
included in evaluation of action recognition. This makes common
detection metrics, like average precision (AP) for action detection, less informative for fingerspelling detection.

Our design of a detection model is motivated by two
observations. The first is that articulated pose, in particular
handshape, plays a role in the distinctiveness of fingerspelling from other types of sign. At the same time, pose
estimation, while increasingly successful in some domains, may be
insufficiently accurate for directly informing fingerspelling
recognition, as shown in~\citep{shi2019fingerspelling}. Instead we
incorporate pose estimation as part of
 training 
our model, but do not rely on explicit pose estimates at test time.
The second observation concerns the goal of optimizing 
fingerspelling detection as a means to an end of improving downstream
recognition. We address this by including a fingerspelling recognizer
in model training. Our results show that this multi-task learning
approach produces a superior detector compared to baselines that omit
the pose and/or recognition losses.

\section{Related Work}
\label{sec:detection-related-work}
Prior work on fingerspelling detection~\citep{yanovich2016detection,tsechpenakis2006robust,tsechpenakis2006learning,yang2010simultaneous} employs visual features from optical
flow or pre-defined hand keypoints.  {For sign language data collected in the wild, the quality of pose estimates is usually low, making them a poor choice as input to a detector (as we show in our experiments).}
Tsechpenakis \emph{et al.}~\citep{tsechpenakis2006robust,tsechpenakis2006learning}
proposed a statistical procedure to detect changes in video for fingerspelling detection; however, these were tested anecdotally, in controlled environments and simplified scenarios.
Related work on detection of sign language segments (in video containing both signing and non-signing) uses recurrent neural networks (RNN) to classify individual frames into signing/non-signing categories~\citep{Moryossef2020sld}.  {We compare our approach to baselines that use ideas from this prior work.} 

Prior work~\citep{Moryossef2020sld,yanovich2016detection} has largely
treated the detection task as frame classification, evaluated via classification accuracy.  Though sequence prediction is considered in~\citep{tsechpenakis2006robust,tsechpenakis2006learning}, the model evaluation is qualitative.  In practice, a detection model is often intended to serve as one part of a larger system, producing candidate segments to a downstream recognizer.  Frame-based metrics ignore the quality of the segments and their effect on a downstream recognizer.

{Our modeling approach is based on the intuition that training
  with related tasks including recognition and pose estimation should
  help in detecting fingerspelling segments.
Multi-task approaches have been studied for other related tasks. For example, Li {\it et al.}~\citep{Li2017TowardsET} jointly train an object detector and a word recognizer to perform text spotting. In contrast to this approach of treating detection and recognition as two parallel tasks, our approach further allows the detector to account for the performance of the recognizer. In sign language recognition, Zhou {\it et al.}~\citep{Zhou2020SpatialTemporalMN} estimate human pose keypoints while training the recognizer.  However, the keypoints used in~\cite{Zhou2020SpatialTemporalMN} are manually annotated. Here we study whether we can distill knowledge from an external imperfect pose estimation model (OpenPose).}

\section{Task and Metrics}
\label{sec:metrics}

\newcommand{\letseq}{\mathbf{l}}

We are given a sign language video clip with $N$
frames $I_1,..., I_N$ containing $n$
fingerspelling segments {$\{x^{\ast}_i\}_{1\leq i\leq n}$} ($\ast$ denotes ground truth), where
$x^{\ast}_i=(s^{\ast}_i, t^{\ast}_i)$ 
  and $s^{\ast}_i, t^{\ast}_i$ are the start and end frame indices of the $i^\textrm{th}$ segment. {The corresponding ground-truth letter sequences are $\{\letseq_i^\ast\}_{1\leq i\leq n}$.}

 \textbf{Detection task} The task of fingerspelling
  detection is to find the fingerspelling segments within the clip. A detection model outputs $m$ predicted segment-score pairs $\{(\widehat{x}_i, {f}_i)\}_{1\leq i\leq m}$ where $\widehat{x}_i$ and $f_i$ are the $i^\text{th}$ predicted segment and its confidence score, respectively.

 \textbf{AP@IoU} A metric commonly used in object
detection~\citep{lin2014microsoft} and action
detection~\citep{idrees2016thumos,Heilbron2015ActivityNetAL} is AP@IoU. Predicted
segments, sorted by score $f_i$, are sequentially matched to the
ground-truth segment $x^{\ast}_j$ with the highest IoU (intersection
over union, a measure of overlap) above a
threshold $\delta_{IoU}$. Once $x^\ast_j$ is matched to a
$\widehat{x}_i$ it is removed from the candidate set.  Let
$k(i)$ be the index of the ground-truth segment $x^\ast_{k(i)}$ matched to $\widehat{x}_i$; then formally,
\begin{equation}
  \label{eq:ap_iou}
  k(i)=\displaystyle\argmax_{j: IoU(\widehat{x}_i,
    x^{\ast}_j)>\delta_{IoU},\,j\ne k(t)\forall t<i}IoU(\widehat{x}_i, x^{\ast}_j).
\end{equation}
Precision and recall are defined as the proportions of matched
examples in the {predicted and ground-truth segment sets},
respectively. Varying the number of predictions $m$ gives a {precision-recall curve $p(\tilde{r})$}.  The average precision (AP) is defined as the mean precision {over} $N_r+1$ equally spaced recall levels $[0, 1/N_r,...,1]$\footnote{$N_r$ is set to 100 as in~\cite{lin2014microsoft}.}, {where} the precision {at a given recall is defined as}
the maximum precision at a recall exceeding that level:
$AP=\frac{1}{N_r}\displaystyle\sum_{i=1}^{N_r}\max_{\tilde{r}:\tilde{r}\geq
  i/N_r}p(\tilde{r})$. AP@IoU can be reported for a range of values of
$\delta_{IoU}$.

\newcommand{\acc}{\textrm{Acc}}

\textbf{Recognition task} The task of recognition is to transcribe $\widehat{x}=(\widehat{s},\widehat{t})$ into a letter
sequence $\widehat{\letseq}$. The recognition accuracy of a predicted sequence
$\widehat{\letseq}$ w.r.t.~the ground truth $\letseq^\ast$ is defined as
$\acc(\letseq^\ast,\widehat{\letseq}) = 1 -
D(\letseq^\ast,\widehat{\letseq})/|\letseq^\ast|$, where $D$ is the
edit (Levenstein) distance
and $|\letseq|$ is
the length of a sequence. Note that $\acc$ can be negative. 

Prior work has considered recognition mainly applied to ground-truth segments $x^\ast$; in contrast, here we are concerned
with {detection for the purpose of recognition}.
%
We match a predicted $\widehat{x}_j$ to a ground-truth
$x^{\ast}_i$
and
then evaluate the accuracy of $\widehat{\letseq}_j$
w.r.t.~$\letseq_i^\ast$. Thus, in contrast to a typical action
detection scenario, here
%
IoU may not be
perfectly correlated with recognition accuracy. For example, a
detected segment that is too short can hurt recognition much more than
a segment that is too long.
We propose a new metric, AP@Acc, to measure the
performance of a fingerspelling detector in the context of a
given downstream recognizer. 

\textbf{AP@Acc} 
This metric uses the letter accuracy of a
recognizer to match between predictions and ground-truth. It also
requires {an IoU threshold} to prevent matches between non-overlapping
segments. As in AP@IoU, predicted segments are sorted by score and
sequentially matched:

\begin{equation}
  \label{eq:ap_acc}
    k(i)=\displaystyle\argmax_{\substack{j: IoU(\widehat{x}_i,
        x^{\ast}_j)>\delta_{IoU},\;j\ne k(t)\,\forall t<i\\
        Acc(\letseq^\ast_j,Rec(I_{\widehat{s}_i:\widehat{t}_i}))>\delta_{acc}}}
    \hspace{-2em}Acc(\letseq_j^\ast,Rec(I_{\widehat{s}_i:\widehat{t}_i})),
\end{equation}
where $Rec(I_{s:t})$ is the output (predicted letter sequence) from a
recognizer given the frames $I_{s:t}$.  We can
report AP@Acc for multiple values of $\delta_{acc}$.

%

\textbf{Maximum Sequence Accuracy (MSA)} Both AP@IoU and AP@Acc measure the precision of a set of detector predictions.
%
Our last metric directly measures just the performance of a
given downstream recognizer when given the detector's
predictions. We form the ground-truth letter sequence for the entire
video $I_{1:N}$ by concatenating the
letters of all ground-truth segments, with a special ``no-letter''
symbol $\emptyset$ separating consecutive letter sequences:

\begin{equation}
  \mathbf{L}^\ast=\emptyset,\letseq^\ast_1,\emptyset,\ldots,\emptyset,\letseq^\ast_n,\emptyset.
  \label{eq:lstar}
\end{equation}
{Note that $\emptyset$ is inserted only where non-fingerspelling frames exist.} 
We similarly obtain a full-video letter sequence from the predicted
segments. We suppress detections with score $f_i$ below $\delta_f$ and apply local
non-maximum suppression, resulting in a set of non-overlaping
segments $\widehat{x}_1,\ldots,\widehat{x}_n$. Each of these is fed to
the recognizer, producing
$\widehat{\letseq}_i=Rec(I_{\widehat{s}_i:\widehat{t}_i})$. Concatenating
these in the same way as in~\eqref{eq:lstar} gives us the full-video
predicted letter sequence $\widehat{\mathbf{L}}(\delta_f)$. We can now treat
$\mathbf{L}^\ast$ and $\widehat{\mathbf{L}}$ as two letter sequences,
and compute the transcription accuracy. Maximum sequence accuracy
(MSA) is defined as

\begin{equation}
  \label{eq:msa}
    MSA=\displaystyle\max_{\delta_f}Acc(\mathbf{L}^\ast,\widehat{\mathbf{L}}(\delta_f)).
\end{equation}
Like AP@Acc, MSA depends on both the detector and the given recognizer
$Rec$.  By comparing the MSA for a given detector and for an ``oracle detector'' that produces the
ground-truth segments, we can obtain an indication of how far the detector output is from the ground-truth.

\section{Models for Fingerspelling Detection}

\subsection{Baseline Models}

\textbf{Baseline 1: Frame-based detector}
This model classifies every frame as positive
(fingerspelling) or negative (non-fingerspelling), and is
trained using the per-frame cross-entropy loss, weighted to control
for class imbalance.  Each frame is passed through a convolutional network to extract visual features, which are then passed to a multi-layer bidirectional LSTM to take temporal context into account, followed by a linear/sigmoid layer producing per-frame class posterior probabilities.

\begin{figure}[btp]
  \centering
  \includegraphics[width=0.5\linewidth]{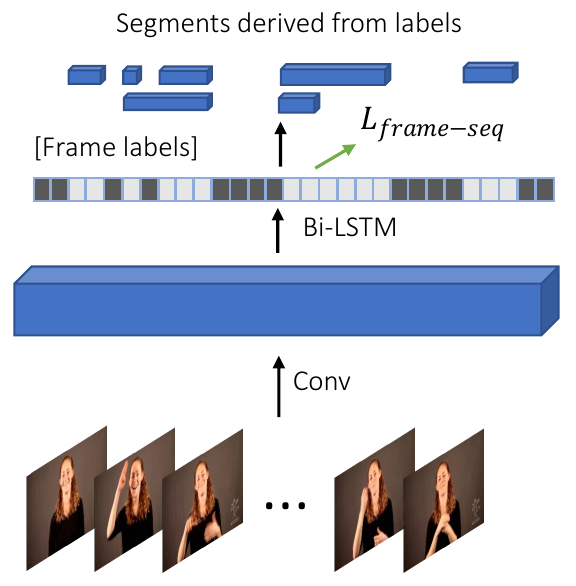}
  \caption{\label{fig:model_base12} Illustration of baseline 1 and 2}
\end{figure}

To convert the frame-level outputs to predicted segments, we first assign
hard binary frame labels by thresholding the posteriors by $\bar{p}$. Contiguous sequences of positive labels are taken to be the
predicted segments, with a segment score $f$ 
computed as the average posterior of the fingerspelling class for its
constituent frames.
We repeat this process for decreasing values of $\bar{p}$, producing a
pool of (possibly overlapping) segments. Finally, these are culled using 
thresholding on $f$ and
non-maximum suppression limiting overlap, producing a set of
$\widehat{x}$s as the final output.

\textbf{Baseline 2: Recognition-based detector}
Instead of classifying each frame into two classes, this baseline addresses the task as a sequence prediction problem, where the target is the
concatenation of
true letter sequences separated by $\emptyset$ indicating
non-fingerspelling segments, as
in~\eqref{eq:lstar}.

The frame-to-letter alignment in fingerspelling spans is usually
unknown during training, so we base the model on connectionist temporal classification (CTC)~\cite{Graves2006ConnectionistTC}: We generate frame-level label softmax posteriors over possible letters,
augmented by $\emptyset$ and a special $blank$ symbol, and train by
maximizing the marginal log probability of label sequences that
produce the true sequence under CTC's ``label collapsing function''
that removes duplicate frame labels and then $blank$s (we refer to
this as the CTC loss).

In this case we have a partial alignment of sequences and frame labels, since we know the
boundaries between $\emptyset$ and fingerspelling segments, and we 
use this information to explicitly compute a frame-level log-loss in
non-fingerspelling regions. This modification stabilizes training and improves performance.
At test time, we use the model's per-frame posterior
probability of fingerspelling, $1-p(\emptyset)$, and follow the same
process as in baseline 1 (frame-based) to convert the per-frame
probabilities to span predictions.

\textbf{Baseline 3: Region-based detector} 
This model directly predicts variable-length temporal segments that potentially contain fingerspelling, and is adapted from R-C3D~\citep{xu2017rc3d}, a 3D version of the Faster-RCNN~\citep{ren2015faster}.
The model first applies a 2D ConvNet on each frame; additional 3D convolutional layers are applied on the whole feature tensor to capture temporal information. Unlike~\cite{xu2017rc3d}, we did not directly apply an off-the-shelf 3D ConvNet such as C3D~\cite{tran2015learning}, since its large stride may harm our ability to capture the delicate movements and often short sequences in fingerspelling.
\begin{figure}[btp]
  \centering
  \includegraphics[width=0.5\linewidth]{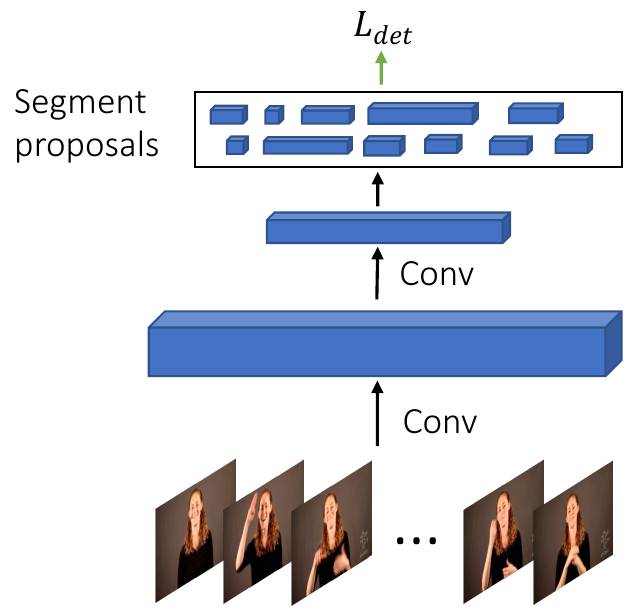}
  \caption{\label{fig:model_base3} Illustration of baseline 3}
\end{figure}

A {region proposal network}  is applied {to the} feature tensor,
predicting shifts of potential fingerspelling segments with
respect to (temporal, 1D) anchors, and a binary label indicating
whether the predicted proposal contains fingerspelling. The
  detector is trained with a loss composed of two terms for (binary)
  classification and (positional) regression, $L_{det} = L_{cls} +
  L_{reg}$.

The anchors are $K$ pre-defined windows of different fixed lengths centered at every timestep of the feature tensor, where $T_{feat}$ is the length of the feature tensor and  $K$ is a hyperparameter. Specifically, a proposal network is composed of convolutional layers that pool a 5D feature tensor ($\in\mathbb{R}^{T\times C\times h\times w}$) into a 2D feature tensor ($\in\mathbb{R}^{T/n\times C^\prime}$). The feature vector in each temporal location is used to predict a relative shift $\{\delta c_i, \delta l_i\}$ to the center location and the length of each anchor as well as a binary label indicating the proposal being fingerspelling or not. A proposal is labeled positive if the IoU with any ground-truth segment is higher than a certain threshold and negative if it is lower than a threshold. We apply the same coordinate transformation in~\citep{xu2017rc3d} as follows:
\begin{equation}
  \label{eq:coord_transform}
  \begin{cases}
    & \delta c_i=(c_i^\star - c_i)/l_i\\
    & \delta l_i=\log(l_i^\star/l_i)\\
  \end{cases}
\end{equation}
where $c_i$ and $l_i$ are center location and length of anchor while $c_i^\star$ and $l_i^\star$ is the center location and length of the ground-truth segment with highest IoU with that anchor.

At test time, we use greedy non-maximum suppression (NMS) on fingerspelling proposals to eliminate highly overlapping and low-confidence intervals.
Since fingerspelling detection is a binary classification task, we do not use a second stage classification subnet as in the original RC3D~\citep{xu2017rc3d}.

\subsection{Multi-Task Fingerspelling Detector}

Our model
is based on the region-based detector, with the key difference being that fingerspelling recognition and
{pose estimation} are incorporated into training the model.

\textbf{Recognition loss} is computed by passing 
{the fingerspelling segments} to a recognizer. 

\begin{figure}[btp]
  \centering
  \includegraphics[width=0.8\linewidth]{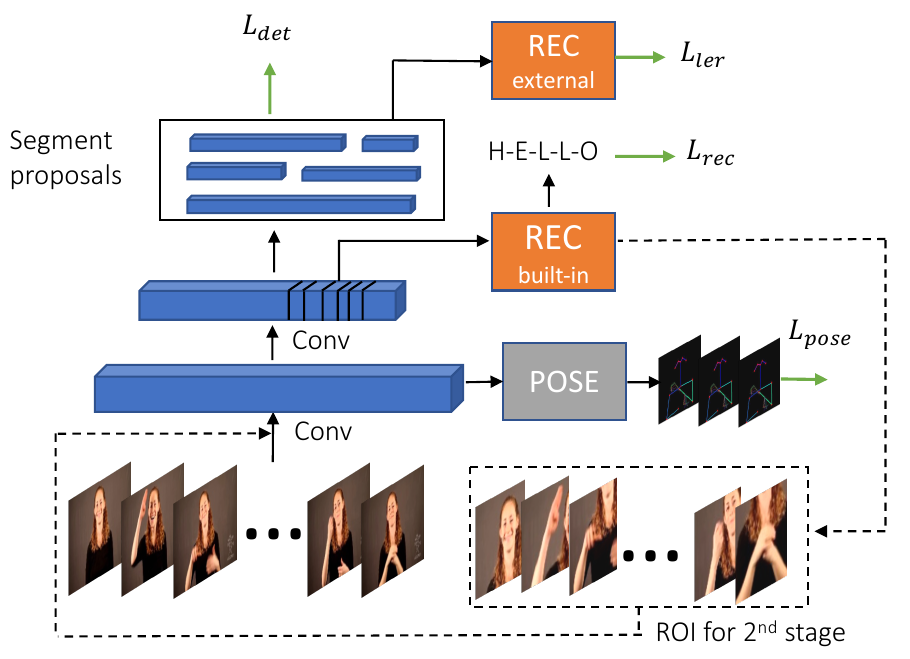}
  \caption{\label{fig:model_multitask} Illustration of multi-task fingerspelling detector}
\end{figure}

Our intuition is that including the recognition task may help
the model learn richer features for fingerspelling, which may improve its ability to distinguish between fingerspelling and non-fingerspelling.
The recognizer here plays a role similar to the classification
subnet in RC3D. But since we don't assume that frame-letter alignment is
available at training time, we directly build a sub-network for
letter sequence prediction (the {orange ``REC''} in Figure~\ref{fig:model_multitask}).

The recognition sub-network follows the attention-based model proposed for
fingerspelling recognition in chapter~\ref{ch:e2efsr}, using only the ground-truth segment for the recognition loss.
The recognition loss $L_{rec}$ is computed as the CTC loss summed over
the proposed regions predicted by the detector:

\begin{equation}
 \label{eq:model_rec_ctc_loss}
 L_{rec} = \displaystyle\sum_{i=1}^nL_{ctc}(Rec(I_{s_i^\star:t_i^\star}), \letseq^\ast_i)
\end{equation}

where $n$ is the number of true fingerspelling segments.

\textbf{Letter error rate loss} Though $L_{rec}$ should help
learn an improved image feature space, the performance of the detector does not impact the
recognizer directly since $L_{rec}$ uses only the ground-truth segments.
%
$L_{det}$ encourages the
model to output segments that are spatially close to the ground truth.
What's missing is the objective of making the detector work well
\emph{for the downstream recognition}. To this end we add a loss measuring the letter
error rate of a recognizer applied to proposals from the detector:
\begin{equation}
  \label{eq:model_rec_ler_loss}
  L_{ler}=-\sum_{i=1}^m p(\widehat{x}_i)Acc(\letseq^\ast_{k(i)},Rec(I_{\widehat{s}_i:\widehat{t}_i})),
\end{equation}
where as before, $k(i)$ is the index of the ground-truth segment matched (based
on IoU) to the
$i^\text{th}$ proposal $\widehat{x}_i=(s_i, t_i)$, $m$ is the number of proposals output by the detector, and $Acc$ is the {recognizer accuracy.}
The loss can be interpreted as the expectation of the negative
letter accuracy of the recognizer on the proposals given to it by
  the detector.
  Since $Acc$, which depends on
the edit distance, is non-differentiable, we approximate the gradient
of~\eqref{eq:model_rec_ler_loss}
{as in} 
REINFORCE~\citep{Williams2004SimpleSG}:
\begin{equation}
  \label{eq:model_rec_ler_loss_grad}
  \nabla L_{ler}\approx
  -\sum_{i=1}^Mp(\widehat{x}_i)Acc(\letseq^\ast_{k(i)},Rec(I_{\widehat{s}_i:\widehat{t}_i})\nabla\log p(\widehat{x}_i),
\end{equation}
where the sum is over the $M$ highest scoring proposals, and
$p(\widehat{x}_i)$ is the normalized score $f_i/\sum_{i=1}^Mf_i$.

\textbf{Pose estimation loss} Since sign language is
to a large extent based on body articulation,
it is natural to consider
incorporating pose
into the model.
However, we can not assume access to ground-truth \kledit{pose}
even in the training data.
Instead, we can rely on general-purpose human pose
estimation models, such as OpenPose~\cite{cao2019openpose}, trained on large sets of annotated images.

Prior work on sign language
has used pose estimates
extracted from an external model as input to models for sign
language-related tasks~\cite{Joze2019MSASLAL,Moryossef2020sld,Parelli2020slpose,li2020word,Charles2013AutomaticAE}. On the controlled studio data used in that work,
the quality of extracted keypoints is reliable. On more challenging
real-world data like ChicagoFSWild/ChicagoFSWild+, we find the
detected keypoints from OpenPose to be much less reliable, 
presumably in part due to
the widespread motion blur in the data (see Figure~\ref{fig:pose_samples} for examples).  {Indeed, as we show in our experiments, using automatically extracted poses as input to the model does not significantly improve performance.}

Instead of relying on estimated pose at test time, we treat
the estimated pose as a source of additional supervision for our model
at training time. We use keypoints from OpenPose as pseudo-labels to
help distill knowledge from the pre-trained pose model into
our detection model. As pose is {used only} as an auxiliary task, the
quality of those pseudo-labels has less impact on detection
performance than does using pose as input at test time.

The pose estimation sub-network takes the feature maps extracted by the model (shared with
the detector and recognizer) and applies several transposed
convolutional layers to increase spatial resolution, producing for
frame $I_t$ a set of heat
maps $\mathbf{b}_{t,1},\ldots,\mathbf{b}_{t,P}$ for $P$ keypoints in the OpenPose
model. We also use OpenPose to extract keypoints from $I_t$; each
estimated keypoint $p$ is accompanied by a confidence score
$\sigma_{t,p}$. We convert these estimates to pseudo-ground truth heatmaps
$\mathbf{b}^\ast_{t,1},\ldots,\mathbf{b}^\ast_{t,P}$.  The pose loss is the
per-pixel Euclidean distance between the predicted and pseudo-ground truth maps:
\begin{equation}
  \label{eq:model_pose_loss}
  L_{pose}=\displaystyle\sum_{t=1}^{T}\displaystyle\sum_{p=1}^{P}\|\v
  b_{t,p} -\v b^\ast_{t,p}\|^2\cdot 1_{\sigma_{t,p}>\tau},
\end{equation}
where the threshold on the OpenPose confidence is used to ignore
low-confidence pseudo-labels.

Putting it all together, the loss for training our model is
\begin{equation}
  \label{eq:model_total_loss}
  L=L_{det} + \lambda_{rec} L_{rec} + \lambda_{ler} L_{ler} + \lambda_{pose} L_{pose},
\end{equation}
with tuned weights $\lambda_{ler}$, $\lambda_{rec}$,
$\lambda_{pose}$ controlling the relative importance of different loss components. 

\textbf{Second stage refinement}
%
Since we expect reasoning about
fine-grained motion and handshape differences to be helpful,
we would like to use high-resolution input
images. However, since the input video clip covers hundreds of
frames, the images need to be downsampled to fit into the memory
of a typical GPU.

To mitigate the issue of low resolution in local regions, we
{``zoom in'' on} the original image frames using the attention map produced by the reocognizer sub-network.  This idea is based on the iterative attention approach presented in section~\ref{sec:end2end-method}.  As large attention values indicate higher importance {of} the corresponding region for fingerspelling, cropping the ROI surrounding {a location with high attention values} helps increase the resolution of that part while avoiding other irrelevant areas.  To achieve this, we first complete training of the whole model described above. Suppose the image sequence of original resolution is $I^o_{1:N}$, the lower-resolution frame sequence used in the first round of training is $I^g_{1:N}$, and the trained model is $\mathcal{H}$. We run inference on $I^g_{1:N}$ with the recognition sub-network $\mathcal{H}_{rec}$ to produce a sequence of attention maps $\mathbf{A}_{1:N}$. We use  $\mathbf{A}_{1:N}$ to crop  $I^o_{1:N}$ into a sequence of local ROIs  $I^l_{1:N}$. Specifically, at timestep $n$, we put a bounding box $b_n$ of size $R|I_n|$ centered on the peak of attention map $\v A_n$, where $R$ is the zooming factor. We average {the} bounding boxes of {the} $2a+1$ frames centered on the {$n^\textrm{th}$} frame to produce a smoother cropping ``tube'' $b_{1:N}^s$:
\begin{equation}
  \label{eq:model_second_bbox}
  b^s_n=\frac{1}{2a+1}\displaystyle\sum_{i=-a}^{a}(b_{n+i})
\end{equation}
The local ROIs $I^{l}_{1:N}$ are cropped from $I^o_{1:N}$ with $b_{1:N}^s$.

Finally, we perform a second stage of training with both
$I^{g}_{1:N}$ and $I^{l}_{1:N}$
as input. At each timestep $n$, the backbone conv layers are applied
on $I^{g}_n$ and $I^{l}_n$ to obtain global and
local feature maps, respectively. The two feature maps
are concatenated for
detection and recognition. The pseudo-ground truth pose maps are estimated on $ I^{(g)}_t$ and $I^{(l)}_t$ separately. The overall loss for the second-stage training is
\begin{equation}
  \label{eq:model_second_loss}
    L^{\text{final}}\,=\,L_{det} + \lambda_{rec} L_{rec} + \lambda_{ler} L_{ler} +\,\lambda_{pose} (L^{(g)}_{pose} + L^{(l)}_{pose}) 
\end{equation}
The key difference between our approach and the iterative attention in section~\ref{sec:end2end-method} is that we do not drop the input images, but rather use the newly generated ROIs as extra input. In fingerspelling detection, both global context (e.g., upper body position) and local details (e.g., handshape) are important.

\section{Experiments}
\label{sec:detection-exp}
\subsection{Setup}

We conduct experiments on ChicagoFSWild and
ChicagoFSWild+,
two large-scale ASL fingerspelling datasets collected in the wild.
Though the datasets were introduced for fingerspelling recognition (with the boundaries given), the URLs of the raw ASL videos and the fingerspelling start/end timestamps are provided.
We split each video clip into 300-frame chunks \kledit{($\sim$12s)} with a 75-frame overlap between chunks.
The longest fingerspelling sequence in the data is 290 frames long.
We use the same training/dev/test data split as in the
original datasets \kledit{(see additional data statistics in the supplementary material)}.
The image frames are
center-cropped and resized to $108\times 108$.

We
take the convolutional layers from VGG-19~\cite{Simonyan2015very}
pre-trained on ImageNet~\cite{deng2009imagenet} as our backbone
network, and fine-tune the weights during training. For baselines 1 and 2, an average pooling layer is applied on
the feature map, giving a 512-dimensional vector for each frame, which
is fed into a one-layer Bi-LSTM with 512 hidden units. In baseline 3
and our model, the feature map is further passed through a 3D conv +
maxpooling layer (with temporal stride 8). In \kledit{the region proposal network}, \kledit{the lengths of the anchors are fixed at 12 values ranging from 8 to 320,}
which are chosen according to the typical lengths of
fingerspelling sequences in the data. The IoU thresholds for
positive/negative anchors are respectively 0.7/0.3. The predicted
segments are refined with NMS at a threshold of 0.7. The magnitude of
optical flow is used as a prior attention map \kledit{as in~\cite{shi2019fingerspelling}}.
\bsedit{We use the same recognizer ($Rec$) for~\eqref{eq:model_rec_ctc_loss} and ~\eqref{eq:model_rec_ler_loss}.}
The pose sub-net is composed of one transposed convolutional layer with
stride 2. We use OpenPose~\cite{cao2019openpose} to extract 15 body
keypoints and $2\times 21$ hand keypoints (both hands) as pseudo pose
labels. Keypoints with confidence below $\tau=0.5$ are dropped. For
\kledit{second-stage refinement,} the moving averaging of bounding boxes
in~\eqref{eq:model_second_bbox} uses 11 frames ($a=5$) \bsedit{and the frame sequence is downsampled by a factor of 2 to save memory.} \kledit{The loss weights ($\lambda$s) are tuned on the dev set.}

To evaluate with AP@Acc and MSA, we train a reference \kledit{recognizer}
with the same architecture as the recognition-based detector on
the fingerspelling training data.
\bsedit{The recognizer achieves an accuracy (\%) of 44.0/62.2 on
  ground-truth fingerspelling segments on
  ChicagoFSWild/ChicagoFSWild+.} \kledit{The accuracy is
  slightly worse than that of~\cite{shi2019fingerspelling} because we used a
  simpler recognizer.} \bsedit{In particular, we skipped the iterative
  training in~\cite{shi2019fingerspelling},
  used lower-resolution input, and did not use a language model.}
We consider $\delta_{IoU}\in\{0.1, 0.3, 0.5\}$ and $\delta_{acc}\in\{0,0.2,0.4\}$; $\delta_{IoU}$ is fixed at 0 for AP@Acc.

\subsection{Main Results}

\begin{table}[htb]
\centering
  \caption{\label{tab:det-main-results}Model comparison on the ChicagoFSWild and
    ChicagoFSWild+ \bsedit{test sets}.  BMN refers to boundary matching network~\cite{lin2019bmn}.  The right column (GT) shows results when the
    ``detections'' are given by ground-truth fingerspelling segments.} 
  \begin{tabular}{cc|cccccc|c}\hline
    \multicolumn{2}{c|}{ChicagoFSWild} & Base 1 & Base 2 & Base 3 & MS-TCN & BMN & Ours & GT \\ \hline
   \multirow{3}{*}{\shortstack{AP@\\IoU}} & AP@0.1 & .121 & .310 & .447 &.282 & .442 & \textbf{.495}  & 1.00  \\
   & AP@0.3 & .028 & .178 & .406 & .177 & .396 & \textbf{.453} & 1.00 \\
   & AP@0.5 & .012 & .087 & .318 & .095 & .284 & \textbf{.344} & 1.00 \\ \hline
   \multirow{3}{*}{\shortstack{AP@\\Acc}} & AP@0.0 & .062 & .158 & .216 & .141 & .209 & \textbf{.249}  & .452  \\
   & AP@0.2 & .028 & .106 & .161 & .093 & .157 & \textbf{.181} & .349 \\
   & AP@0.4 & .006 & .034 & .069 & .036 & .070 & \textbf{.081} & .191 \\ \hline
   \multicolumn{2}{c|}{MSA} & .231 & .256 & .320 & .307 & .319 & \textbf{.341} & .452 \\ \hline\hline
       \multicolumn{2}{c|}{(b). ChicagoFSWild+} & Base 1 & Base 2 & Base 3 & BMN & MS-TCN & Ours & GT \\ \hline
   \multirow{3}{*}{\shortstack{AP@\\IoU}} & AP@0.1 & .278 & .443 & .560 & .429 & .580 & \textbf{.590}  & 1.00  \\
   & AP@0.3 & .082 & .366 & .525 & .345  & .549 & \textbf{.562} & 1.00 \\
   & AP@0.5 & .025 & .247 & .420 & .179 & .437 & \textbf{.448} & 1.00 \\ \hline
   \multirow{3}{*}{\shortstack{AP@\\Acc}} & AP@0.0 & .211 & .323 & .426 & .350 & .433 & \textbf{.450}  & .744  \\
   & AP@0.2 & .093 & .265 & .396 & .299 & .401 & \textbf{.415} & .700 \\
   & AP@0.4 & .029 & .152 & .264 & .147 & .260 & \textbf{.277} & .505 \\ \hline
   \multicolumn{2}{c|}{MSA} & .267 & .390 & .477 & .414 & .470 & \textbf{.503} & .630 \\ \hline
  \end{tabular}
\end{table}

\begin{figure}[btp]
 \centering
 \caption{\label{fig:pr_curve_frame}Precision-recall \kledit{curves for frame classification of three baselines and} our approach.}
 \includegraphics[width=0.8\linewidth]{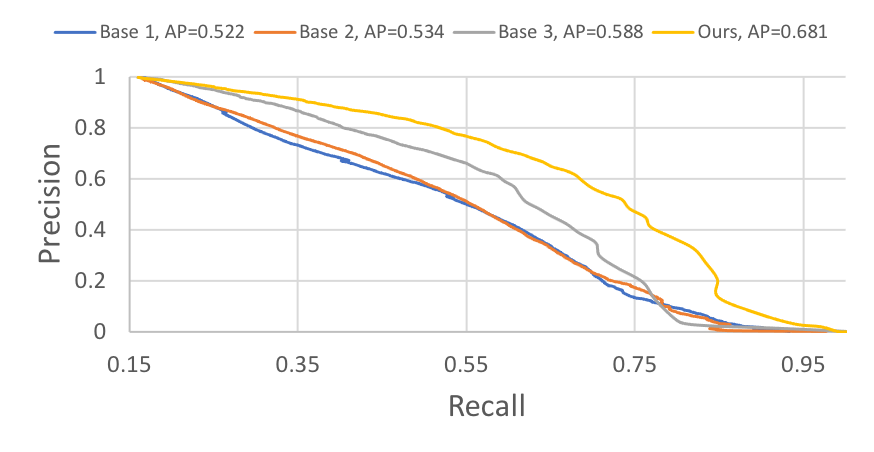}
\end{figure}

Table~\ref{tab:det-main-results} compares models using the three proposed metrics. The values of AP@Acc and MSA depend on the reference recognizer. For ease of comparison, we also show the oracle results when the same recognizer is given the ground-truth fingerspelling segments.
Overall, the relative model performance is consistent across metrics.  Methods that combine detection and recognition outperform those that do purely detection (baseline 2 vs.~1, our model vs.~baseline 3).
In addition, region-based methods (baseline 3 and our model) outperform frame-based methods (baseline 1 \& 2), whose segment predictions lack smoothness.  We can see this also by examining the frame-level performance of the three baselines.\footnote{For the region-based model (baseline 3) we take the maximum probability of all proposals containing a given frame as that frame's probability.} Figure~\ref{fig:pr_curve_frame} shows the precision-recall curves for frame classification of the three baseline models and our proposed model. On the one hand, our approach dominates the others in terms of these frame-level metrics as well. In addition, the differences in frame-level performance among the three baselines are much smaller than the differences in sequence-level performance reported in the main text. 
The frame-level average precisions of three baselines are 0.522 (baseline 1), 0.534
(baseline 2), and 0.588 (baseline 3), which are much closer than the
segment-based metrics, \kledit{showing how frame-level metrics can obscure important differences between models.}
\kledit{The trends in results are similar on the two datasets.}

\kledit{In addition to the baseline models, we compare against two additional approaches---boundary matching network (BMN)~\citep{lin2019bmn} and multi-stage temporal convolutional network (MS-TCN)~\citep{farha2019mstcn}---on our task of fingerspelling detection. Those two methods are state-of-the-art on temporal action proposal generation in ActivityNet1.3~\citep{Heilbron2015ActivityNetAL} and sign language segmentation~\citep{Renz2021signsegmentation}. The implementations are based on~\cite{bmn-code,mstcn-code}. For fair comparison, we use the same backbone network as in the other methods. We use the same network architecture for the individual submodules of the two models and tune hyperparameters on our datasets. As MS-TCN does frame classification in principle, we follow the same post-processing steps as in baseline 1 and 2 to convert frame probabilities into sequence predictions for evaluation.}

\kledit{As is shown in Table~\ref{tab:det-main-results}, these two approaches do not outperform our approach. Comparing BMN and our baseline 3, we notice that the size of the training set has a large impact. The more complex modeling choices in BMN, which searches over a wider range of proposals, leads to better performance mostly when using the larger training set of ChicagoFSWild+. The discrepancy in performance of these two models as measured by different metrics (e.g., AP@IoU vs.~AP@Acc) also shows that a model with lower localization error does not always enable more accurate downstream recognition. The MS-TCN model is generally better than other frame-based approaches (baseline 1, 2) but remains inferior to region-based approaches including baseline 3 and ours. Our post-processing steps lead to inconsistency between the training objective and evaluation. Similarly in~\cite{Renz2021signsegmentation}, it is noted that the model sometimes over-segments fingerspelled words.}

For the remaining analysis we report results on the ChicagoFSWild dev set.

\subsection{Analysis of Evaluation Metrics} 

\begin{figure}[btp]
  \centering
  \caption{\label{fig:ap_heatmap}AP@Acc with different IoU thresholds \bsedit{on ChicagoFSWild dev set}. Left: baseline 3. Right: our model.}
  \includegraphics[width=\linewidth]{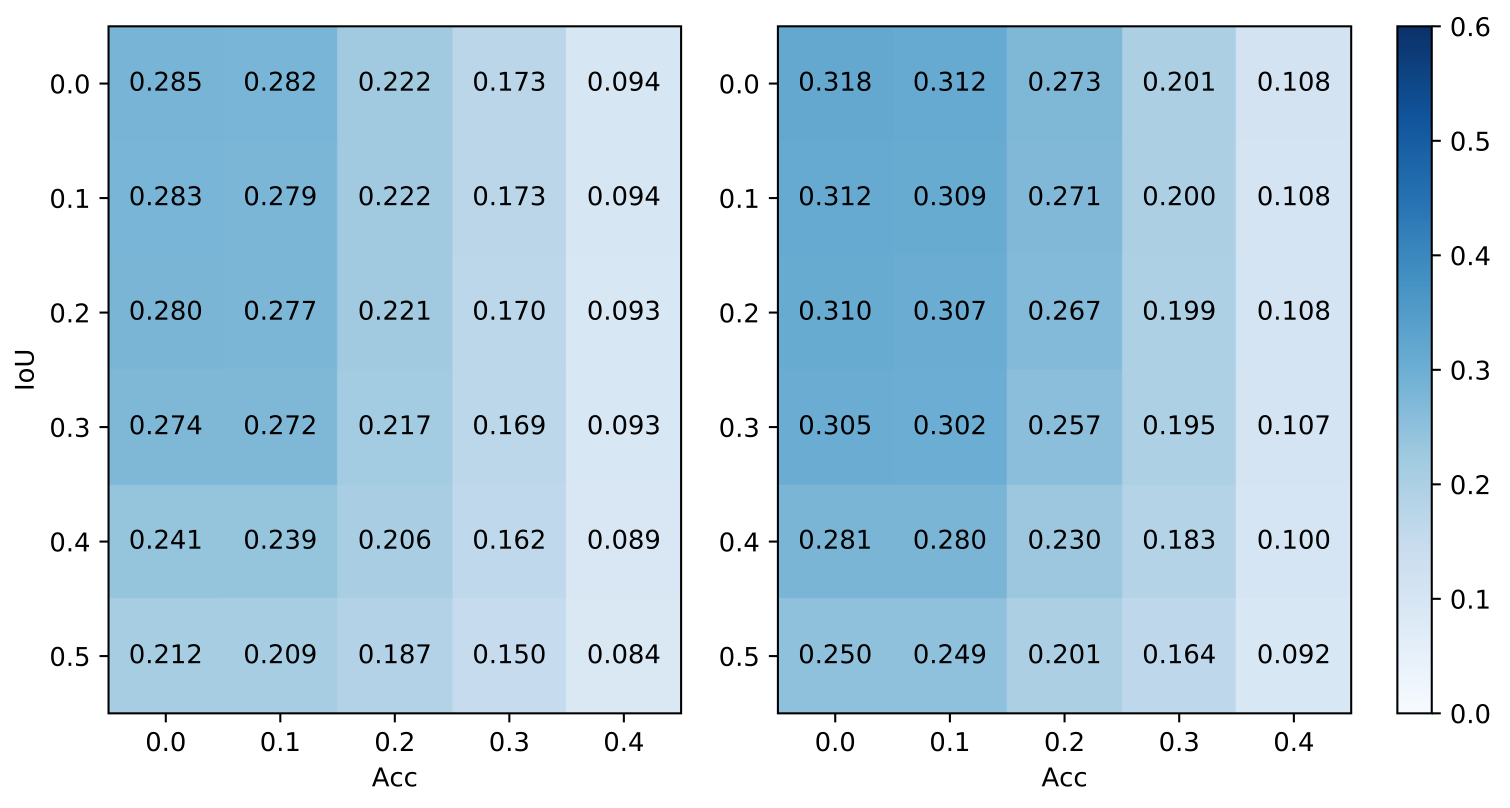}
\end{figure}

Figure~\ref{fig:ap_heatmap} shows how varying $\delta_{IoU}$ and $\delta_{acc}$ impacts the value of \kledit{AP@Acc}.
The accuracy threshold $\delta_{acc}$ has a much larger impact on AP than does $\delta_{IoU}$. This is primarily because a \kledit{large} overlap between predicted and ground-truth segments is often necessary in order to achieve high accuracy. Therefore, we set the default value of $\delta_{IoU}$ to 0.

\begin{figure}[htb]
\centering
  \includegraphics[width=0.5\linewidth]{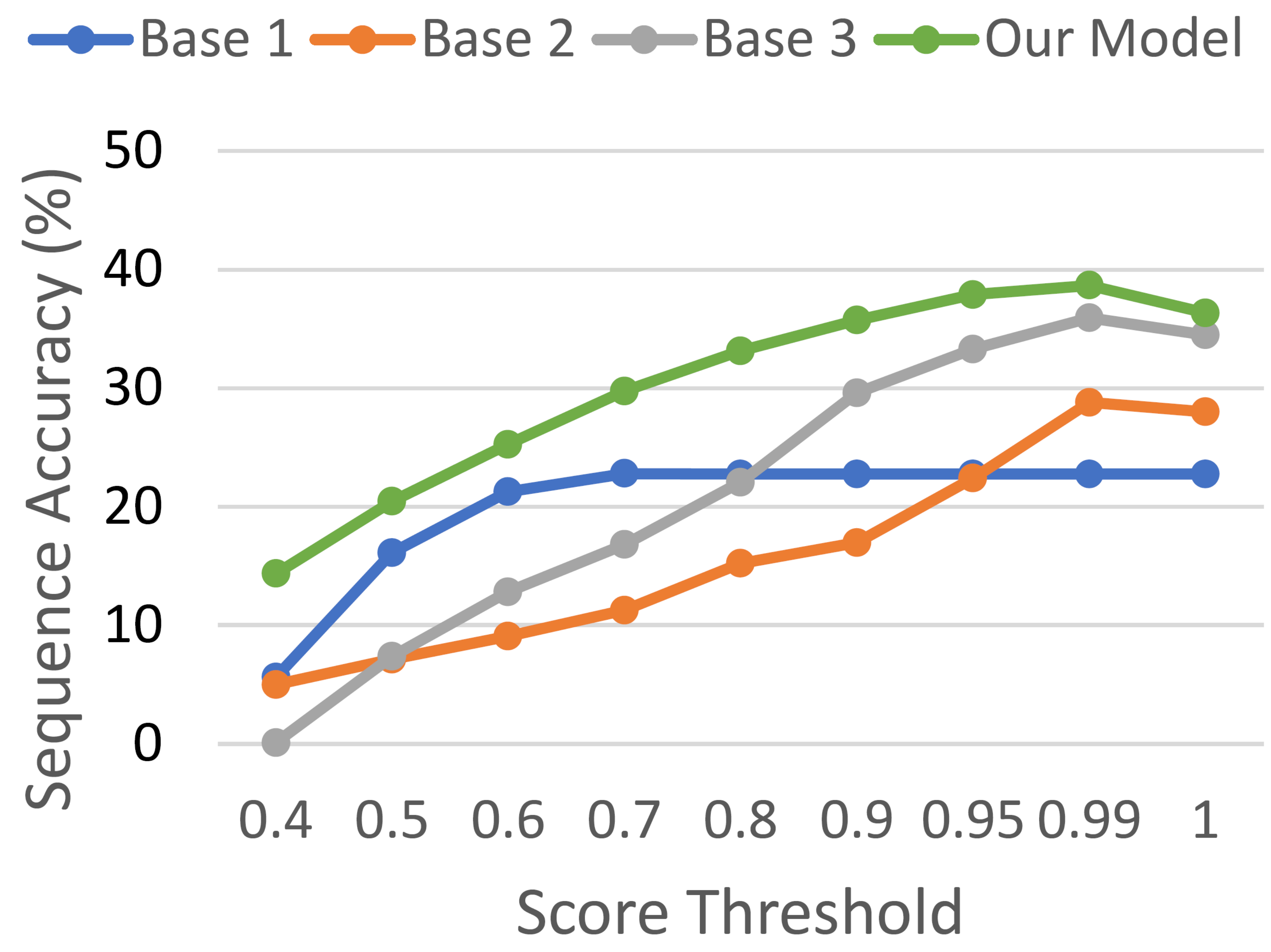}    
  \caption{\label{fig:sa_thr}Dependence of sequence accuracy on score
    threshold $\delta_f$. \kl{}}  
\end{figure}

 As the AP@Acc results
are largely invariant to $\delta_{IoU}$, we report results mainly for $\delta_{IoU}=0$ (i.e., matching ground truth segment to detections
based on accuracy, subject to non-zero overlap). 
We also examine the relationship between sequence accuracy and the
score threshold of each model (Figure~\ref{fig:sa_thr}). Our model
achieves higher sequence accuracy across all thresholds. The threshold
producing the best accuracy varies for each model.

\begin{figure}[btp]
 \centering
 \includegraphics[width=0.8\linewidth]{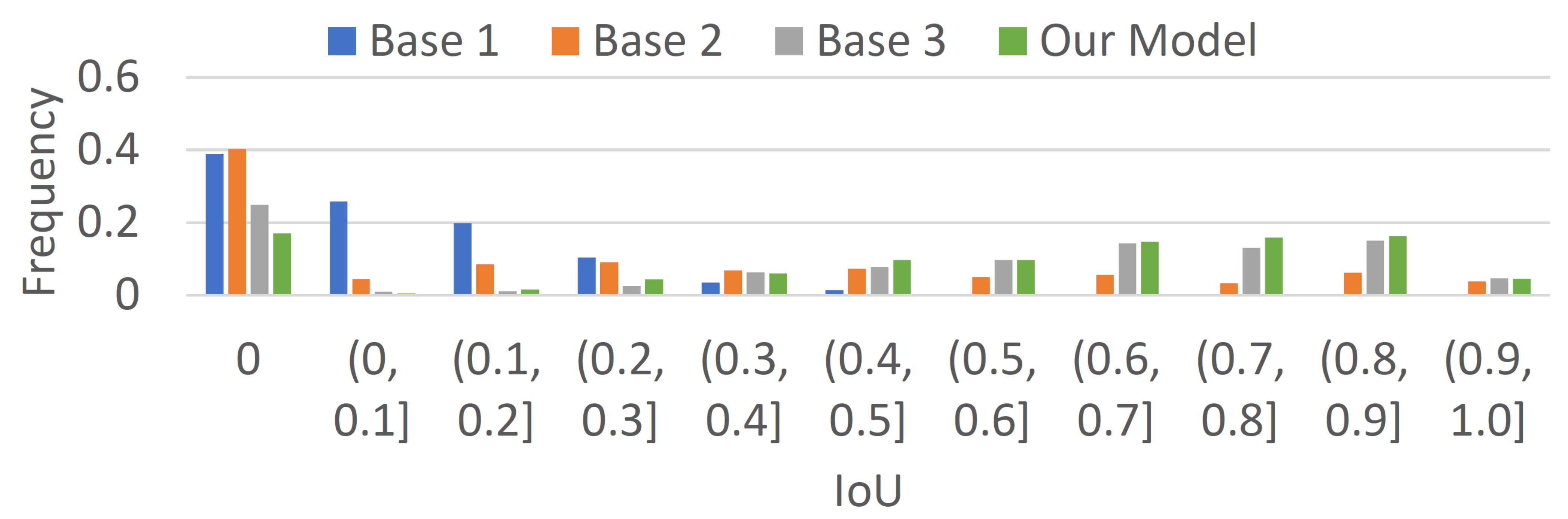}
 \caption{Histogram of IoU at peak thresholds.}\label{fig:iou_histo}
\end{figure}

\kledit{Figure~\ref{fig:iou_histo} shows histograms of IoU of predicted segments with respect to the ground truth at peak thresholds used in the MSA computation.
Our model has overall higher IoU than the three baselines.
The average IoUs of the three baselines and our model for the optimal (peak) threshold
$\delta_f$ are 0.096, 0.270, 0.485, and 0.524 respectively.}
\bsedit{The average IoUs of baseline 3 and our model} suggest that for AP@IoU, AP@0.5 is more meaningful to compare in terms of recognition performance for \bsedit{those two models}.

\begin{figure}[btp]
\begin{tabular}{@{}c@{}c@{}}
    (a) & \raisebox{-.5\height}{\includegraphics[width=0.9\linewidth]{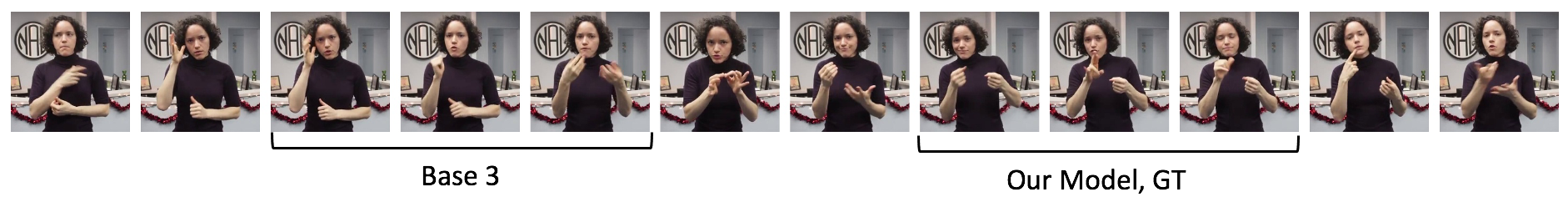}} \\
    (b) & \raisebox{-.5\height}{\includegraphics[width=0.9\linewidth]{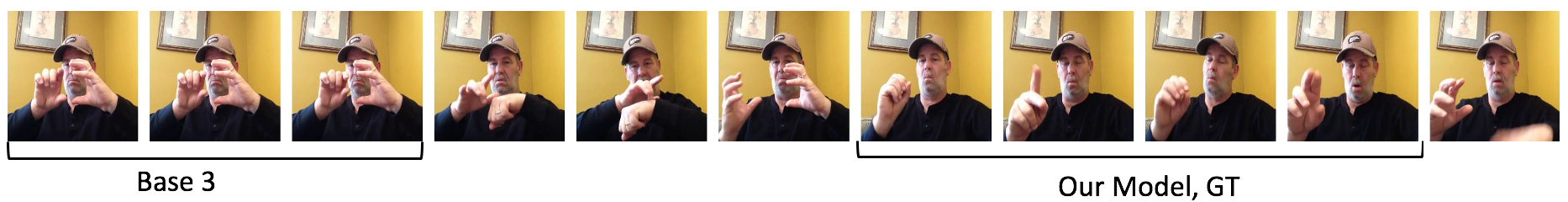}} \\
    (c) & \raisebox{-.5\height}{\includegraphics[width=0.9\linewidth]{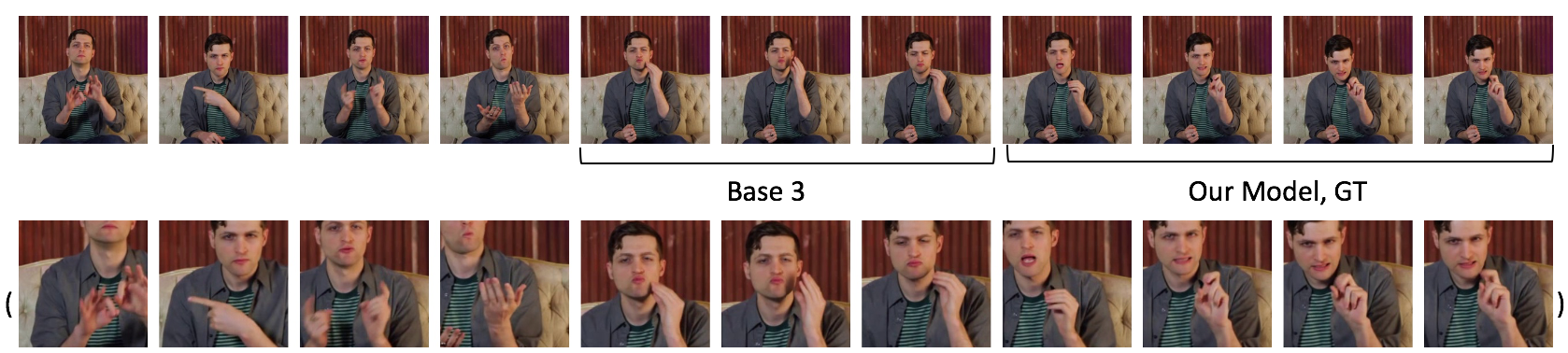}} \\
\end{tabular}
  \caption{\label{fig:quals}Example segments detected by different
    models. Bottom row: ROIs used in \kledit{second-stage refinement}. GT: Ground-truth segment. \bsedit{The sequence is downsampled.}}
\end{figure}

\subsection{Ablation Study} 

Our model reduces to baseline 3 when all loss terms except the detection loss are removed.  Table~\ref{tab:analysis_loss} shows how model performance improves as more tasks are added. The gain due to the recognition loss alone is smaller than for the frame-based models (base 1 vs. base 2). The recognition sub-network contributes more through the LER loss, which communicates the recognition error to the detector directly. The second-stage refinement does not always boost AP@IoU; the gain is more consistent in the accuracy-based metrics (AP@Acc, MSA).
The zoom-in effect of the second-stage refinement increases the resolution of the hand region and improves recognition,
though the IoU
may remain similar.
The downsampling in the second-stage refinement also leads to some positioning error, reflected in the slight drop in AP at IoU=0.5, though a minor temporal shift does not always hurt recognition.

\begin{table}[btp]
\centering

\begin{tabular}{cc|ccccc}\toprule
    & & Base 3 & + rec & +LER & +pose & +2nd stage \\ \midrule
    \multirow{3}{*}{\shortstack{AP@\\IoU}} & AP@0.1 & .496 & .505 & .531 & .539  & .551  \\
    & AP@0.3 & .466 & .479 & .498 & .500 & .505 \\
    & AP@0.5 & .354 & .362 & .392 & .399 & .393 \\ \midrule
    \multirow{3}{*}{\shortstack{AP@\\Acc}} & AP@0.0 & .285 & .290  & .299  & .302  & .313  \\
    & AP@0.2 & .222 & .231 & .245  & .255 & .267 \\
    & AP@0.4 & .094 & .097 & .103  & .107 & .112 \\ \midrule
    \multicolumn{2}{c|}{MSA} & .359 & .365  & .378 & .383 & .386 \\ \bottomrule
  \end{tabular}
    \caption{\label{tab:analysis_loss}Impact of adding loss components
    to our model training.  Results are on \bsedit{the ChicagoFSWild dev set}. \kl{}}
\end{table}

{\bf Examples} Figure~\ref{fig:quals} shows examples in which our
model correctly detects fingerspelling segments while baseline 3
fails. Fingerspelling can include handshapes that are visually similar
to non-fingerspelled signs (Figure~\ref{fig:quals}a). Fingerspelling
recognition, as an auxiliary task, may improve detection by helping
the model distinguish among fine-grained handshapes. \kledit{The signer's pose}
may provide additional information for detecting fingerspelling (Figure~\ref{fig:quals}b). Figure~\ref{fig:quals}c shows an example where the signing hand is a small portion of the whole image; baseline 3 likely fails due to the low resolution of the signing hand. The second-stage refinement enables the model to access a higher-resolution ROI, leading to a correct detection.

\begin{figure}[btp]
  \centering
  \includegraphics[width=0.8\linewidth]{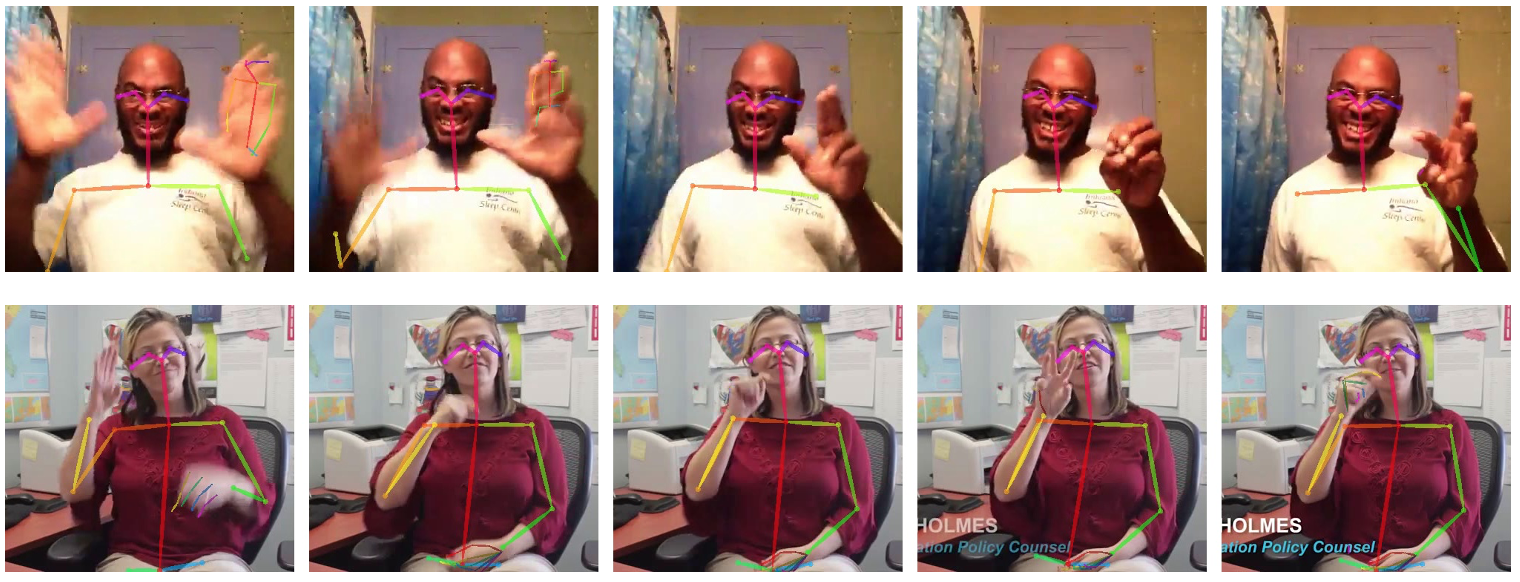}
    \caption{\label{fig:pose_samples} \kledit{Pose estimation failure cases.}}
\end{figure}

\subsection{Error Analysis} 
\begin{figure}[btp]
  \begin{minipage}{0.04\linewidth}
    (a),
  \end{minipage}
  \begin{minipage}{0.96\linewidth}
    \includegraphics[width=\linewidth]{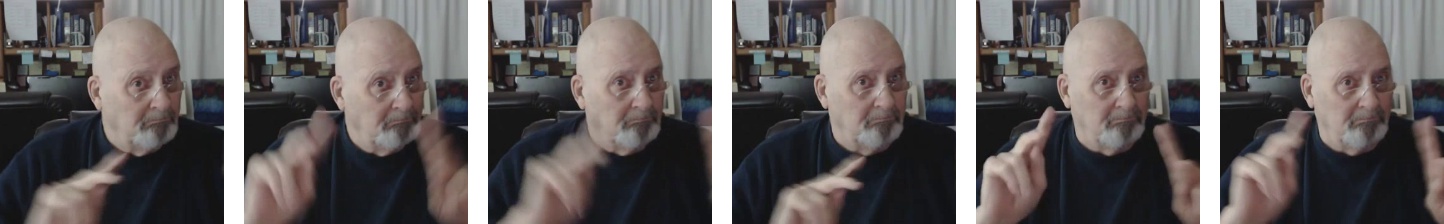}
  \end{minipage}\\ 
  \begin{minipage}{0.04\linewidth}
    (b),
  \end{minipage}
  \begin{minipage}{0.96\linewidth}
    \includegraphics[width=\linewidth]{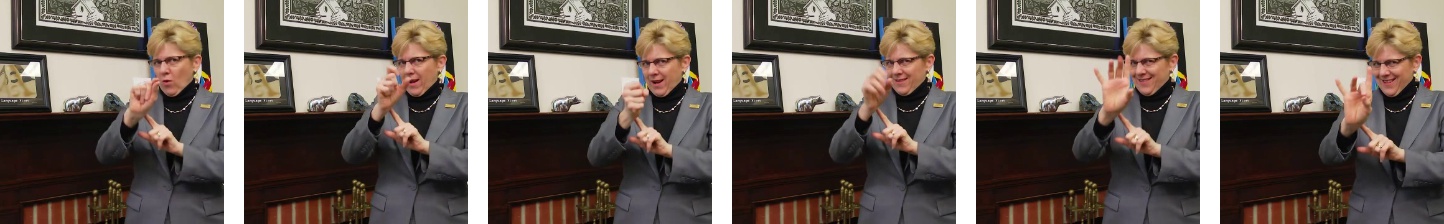}
  \end{minipage}\\ 
  \begin{minipage}{0.04\linewidth}
    (c),
  \end{minipage}
  \begin{minipage}{0.96\linewidth}
    \includegraphics[width=\linewidth]{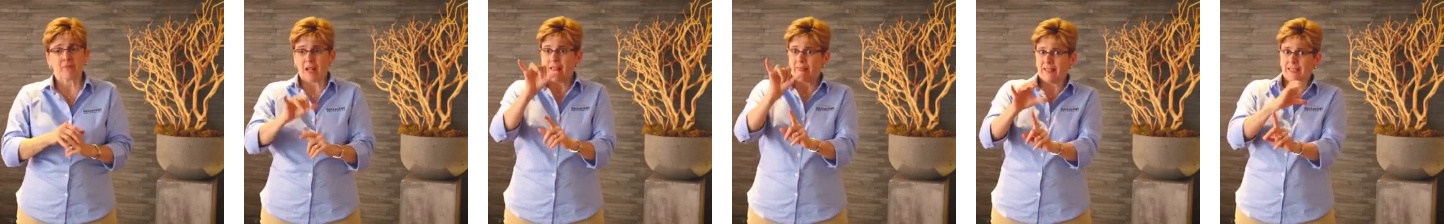}
  \end{minipage}
  \caption{\label{fig:fps}False positive detections. The glosses corresponding to the rows (a) to (c) are respectively ``PEOPLE'', ``2018'', ``THAT-[C..\}''. \kl{}}
\end{figure}

Qualitatively, we notice three common sources of false positives: (a) near-face sign, (b) numbers, and (c) fingerspelling handshape used in regular signing \kledit{(see Figure~\ref{fig:fps})}. Such errors can potentially be reduced by incorporating linguistic knowledge in the detector, which is left as future work. 

\begin{figure}[btp]
  \centering
  \includegraphics[width=0.8\linewidth]{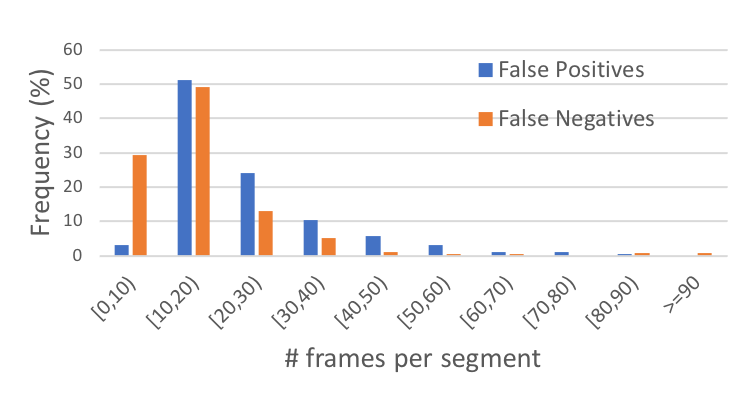}
   \caption{\label{fig:fp_fn_distribution}Distribution of lengths of false positives and false negatives.}
\end{figure}

\kledit{We further separate raw video clips into three categories based on the duration of the fingerspellng segments: short ($<$20 frames), medium (20-80 frames), an
d long ($\geq$80 frames). This division is based on the statistics of the dataset. The performance of our model for the three categories is shown in Table~\ref{tab:perf_dur}, and can be compared to the overall performance in \kledit{Table~\ref{tab:det-main-results}} of the paper.\kl{} Shorter fingerspelling segments are harder to spot with
in regular signing. The typical fingerspelling pattern (relatively static arm and fast finger motion) is less obvious in short segments. In addition, Figure~\ref{fig:fp_fn_distribution} shows the length distribution of false positive and false negative detections from our model.} The length distribution of false po
sitives roughly matches that of ground-truth segments in the dataset. 

 \begin{table}[htp]
    \caption{Performance on segments of different durations.}
    \label{tab:perf_dur}
    \centering
    \begin{tabular}{c|ccc|ccc|c}\toprule
    \multirow{2}{*}{} &
         \multicolumn{3}{c|}{AP@IoU} & \multicolumn{3}{c|}{AP@Acc} & \multirow{2}{*}{MSA}\\ 
         & AP@0.1 & AP@0.3 & AP@0.5 & AP@0.0 & AP@0.2 & AP@0.4 & \\ \midrule
         Short & .411 & .346 & .235 & .149 & .140 & .051 & .357 \\ 
Medium & .675 & .671 & .623& .476 & .361 & .156 & .435 \\ 
Long & .781 & .703 & .420 & .704 & .362& .130 & .393 \\ \bottomrule
    \end{tabular}
\end{table}

\subsection{Other Analysis}

\textbf{Other input modalities}
Our model is trained on RGB images. Motion and pose are two common modalities used in sign language tasks~\cite{Joze2019MSASLAL,Moryossef2020sld,yanovich2016detection}, so it is natural to ask whether they would be helpful instead of or in addition to RGB input.
We use magnitude of optical flow~\cite{Farnebck2003TwoFrameME} as the motion image and the pose heatmap from OpenPose as the pose image.
\kledit{Raw RGB frames as input outperform the other two modalities, although each of them can be slightly helpful when concatenated with the RGB input (see Table~\ref{tab:modalities}).}
The pose image is less consistently helpful, likely because pose estimation is very challenging on our data and OpenPose often fails to detect the keypoints especially in the signing hand (see Figure~\ref{fig:pose_samples}).  However, treating pose estimation as a secondary task, as is done in our model, successfully ``distills'' the \kledit{pose model's} knowledge
and outperforms the use of additional modalities as input. \kledit{We note that using all three modalities can boost performance further. Using both RGB and motion images as input while jointly estimating pose, the detector achieves .523/.495/.367 for AP@IoU(0.1/0.3/0.5), improving over the best model in Table~\ref{tab:modalities}.  However, optimizing performance with multiple modalities is not our main focus, and we leave further study of this direction to future work.}

\begin{table}[btp]
    \caption{\label{tab:modalities}Comparison among modalities for the region-based detector (baseline 3) \bsedit{on the ChicagoFSWild dev set}. RGB+Pose(in): both RGB and pose image used as input. RGB+Opt(in): both RGB and optical flow used as input. RGB+Pose(out): detector trained jointly with pose estimation \kledit{(our model)}.} 
  \centering
  \begin{tabular}{c|cccccc}\toprule
    AP@IoU & RGB & Pose & Opt & \shortstack{RGB+Opt\\ (in)} & \shortstack{RGB+Pose\\ (in)} & \shortstack{RGB+Pose\\ (out)} \\ \midrule
    AP@0.1 & .496 & .368 & .476 & .503 & .501 & \textbf{.505} \\
    AP@0.3 & .466 & .332 & .438 & .472 & .470 & \textbf{.478} \\
    AP@0.5 & .354 & .237 & .315 & .357 & .346 & \textbf{.366} \\ \bottomrule
  \end{tabular}
\end{table}

\begin{figure*}[t]
  \caption{\label{fig:examples}Detection examples. Red: ground-truth segment, green: predicted segment. The \kledit{sequences are} downsampled.}
  \begin{minipage}{\textwidth}
    \includegraphics[width=\linewidth]{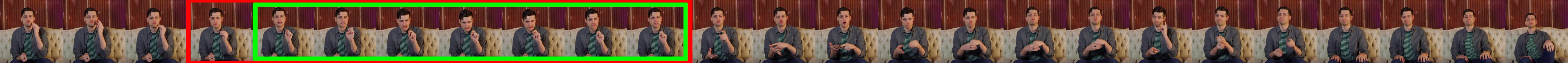}
  \end{minipage}
  \vspace{0.1in}

  \begin{minipage}{\textwidth}
    \includegraphics[width=\linewidth]{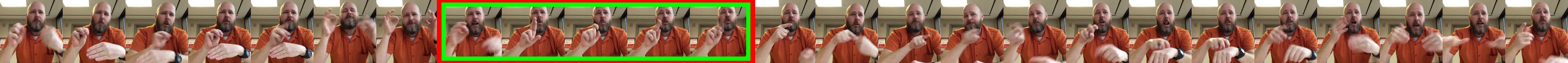}
  \end{minipage}
  \vspace{0.1in}

  \begin{minipage}{\textwidth}
    \includegraphics[width=\linewidth]{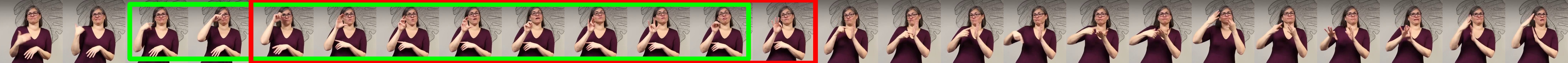}
  \end{minipage}
  \vspace{0.1in}

  \begin{minipage}{\textwidth}
    \includegraphics[width=\linewidth]{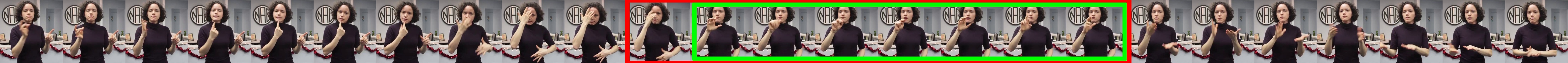}
  \end{minipage}
  \vspace{0.1in}

  \begin{minipage}{\textwidth}
    \includegraphics[width=\linewidth]{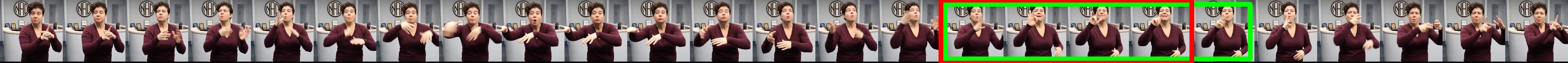}
  \end{minipage}
  \vspace{0.1in}

  \begin{minipage}{\textwidth}
    \includegraphics[width=\linewidth]{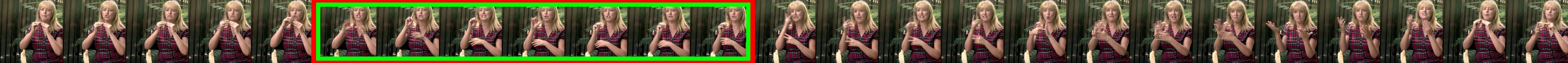}
  \end{minipage}
  \vspace{0.1in}

\begin{minipage}{\textwidth}
    \includegraphics[width=\linewidth]{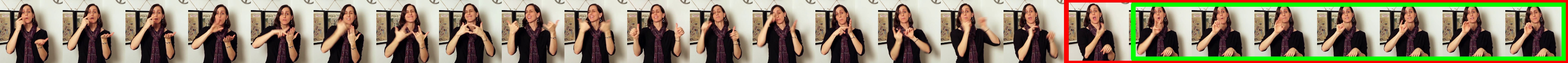}
  \end{minipage}
  \vspace{0.1in}

  \begin{minipage}{\textwidth}
    \includegraphics[width=\linewidth]{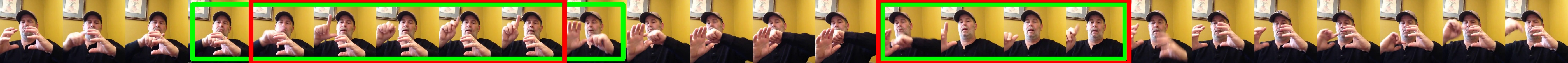}
  \end{minipage}
  \vspace{0.1in}

  \begin{minipage}{\textwidth}
    \includegraphics[width=\linewidth]{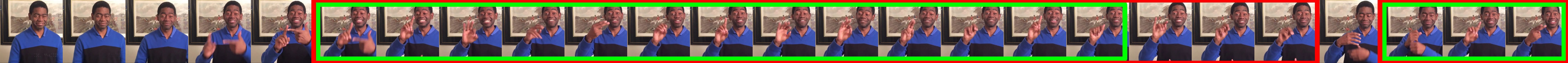}
  \end{minipage}
  \vspace{0.1in}

  \begin{minipage}{\textwidth}
    \includegraphics[width=\linewidth]{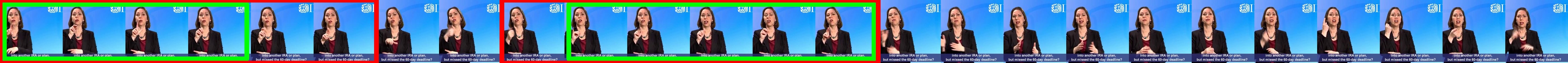}
  \end{minipage}
  \vspace{0.1in}

  \begin{minipage}{\textwidth}
    \includegraphics[width=\linewidth]{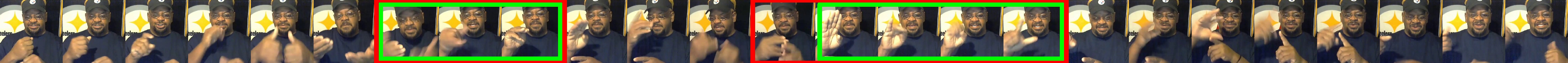}
  \end{minipage}
  \vspace{0.1in}

  \begin{minipage}{\textwidth}
    \includegraphics[width=\linewidth]{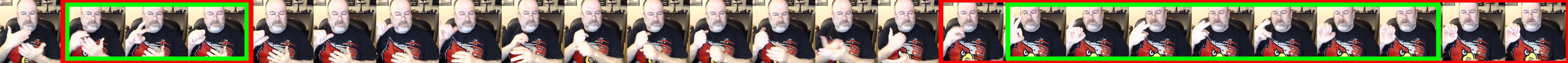}
  \end{minipage}
  \vspace{0.1in}

  \begin{minipage}{\textwidth}
    \includegraphics[width=\linewidth]{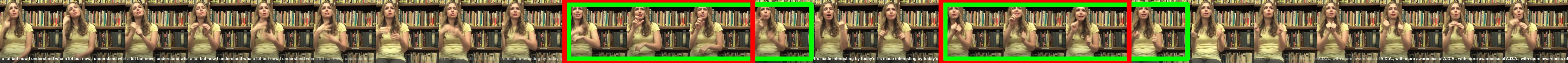}
  \end{minipage}
  \vspace{0.1in}

  \begin{minipage}{\textwidth}
    \includegraphics[width=\linewidth]{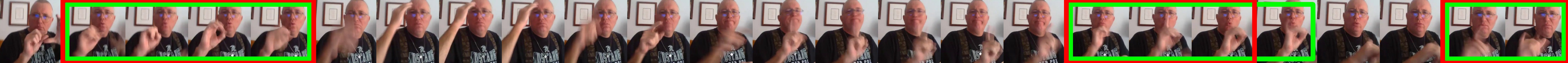}
  \end{minipage}
  \vspace{0.1in}

\end{figure*}

\textbf{Detection examples}
\label{sec:det-examples}
Figure~\ref{fig:examples} shows various detection examples from the ChicagoFSWild dev set. 

\textbf{Speed Test}
\label{sec:det-speed-test}
The inference time per video clip is shown in Table~\ref{tab:inference_time}. The speed test is conducted on one Titan X GPU.  Inference times for all models are under 1 second. Baselines 1 and 2 are faster as the model architecture is simpler. Our model takes roughly twice the time of baseline 3, which is mainly due to the second-stage refinement. 

\begin{table}[htp]
\centering
  \caption{\label{tab:inference_time} Inference time per 300-frame video clip}
  \begin{tabular}{c|cccc}\toprule
    & Base 1 & Base 2 & Base 3 & Ours \\ \midrule
    Inference time (ms) & 10.9 & 11.6 & 284.5 & 511.1 \\ \bottomrule
  \end{tabular}
\end{table}

\section{Summary}

We study the task of sign language fingerspelling detection, motivated
by the goal of developing it as an upstream component in an automatic
recognition system. We propose a benchmark, based on extending a
previously released fingerspelling recognition dataset, and establish a
suite of metrics to evaluate detection models both on their own merits
and on their contribution to downstream recognition. We investigate
approaches adapted from frame classification, fingerspelling
recognition and action detection, and demonstrate that a novel,
multi-task model achieves
the best results across metrics. Beyond
standard detection loss, this model incorporates losses derived from
recognition and pose estimation; we show that each of these
contributes to the superior performance of the model.

Our results provide, for the first time, a practical recipe for fully
automatic detection and recognition of fingerspelling in real-world
ASL media. 
Beyond this novelty, our contributions are as follows. First, we propose an evaluation
framework for fingerspelling detection that incorporates the
downstream recognition task into the metrics, and introduce
a benchmark based on extending a publicly available data
set. Second, we investigate a number of approaches for fingerspelling detection, adapted from fingerspelling recognition and action detection, and develop a novel multi-task
learning approach that incorporates pose estimation to boost detector training.
While our focus has been on fingerspelling, we expect that
the techniques, both in training and in evaluation of the methods,
will be helpful in the more general sign language detection and
recognition domain.

%% file: search.tex
\chapter{Fingerspelling Search and Retrieval}
\label{ch:fs-search}

Chapter~\ref{ch:pipeline}-\ref{ch:closing-gap} \kledit{have} focused on recognition, \kledit{that is, transcription of a fingerspelling} 
video clip into text. However, automatic recognition may not be the end goal in real-world use cases.
In addition, complete transcription may not be necessary to extract the needed information.

In this chapter, we study the problem of searching for fingerspelled content in raw ASL videos.
Fingerspelling search is a crucial aspect of general sign language video search.
This chapter is based on~\citep{shi2022searching}.

\section{Related Work}
\label{sec:search-related-work}

In existing \kledit{work on} sign language \kledit{video} processing, search and retrieval tasks \kledit{have been} studied much less than sign language recognition \kledit{(mapping from sign language video to gloss labels)}~\citep{koller2017resign,forster-etal-2014-extensions} and translation \kledit{(mapping from sign language video to text in another language)}~\citep{Yin2020BetterSL,Camgoz2018neural}.
\kledit{Work thus far on sign language search has been framed mainly} as the retrieval of \bsedit{lexical signs \kledit{rather than} fingerspelling}.  %
Pfister et al.~\citep{Pfister2013LargescaleLO} and Albanie et al.~\citep{albanie2020bsl1k} employ mouthing to detect keywords in sign-interpreted TV programs with coarsely aligned subtitles. Tamer et al.~\citep{Tamer2020CrossLingualKS,Tamer2020KeywordSF} utilize whole-body pose estimation to search \kledit{for} sign language keywords (gloss or translated word) \kledit{in a} German Sign Language translation dataset PHOENIX-2014T~\citep{Camgoz2018neural}. %
All prior work \kledit{on} keyword search for sign language \kledit{has been done in} a closed-vocabulary setting, which assumes that only words \bsedit{from a pre-determined set} will be queried. %
\kledit{Searching in an open-vocabulary setting, including proper nouns, typically requires searching for fingerspelling.}
\kledit{Some related tasks in the speech processing literature are spoken term detection (STD) and query-by-example search, which are the tasks of automatically retrieving speech segments 
from a 
database that match a given text or audio query~\cite{knill2013investigation,mamou2007vocabulary,Chen2015QuerybyexampleKS}}. 
In terms of methodology, our model \kledit{also shares some aspects with prior work on moment retrieval}~\citep{gao2017tall,Xu2019MultilevelLA,zhang2020span}, \kledit{which also combines candidate generation and matching components}.
\kledit{However, we incorporate additional}
\bsedit{task-specific}
\kledit{elements 
that consistently improve performance}. 

\section{Tasks}
\label{sec:task}

\begin{figure}[htp]
    \centering
    \includegraphics[width=0.7\linewidth]{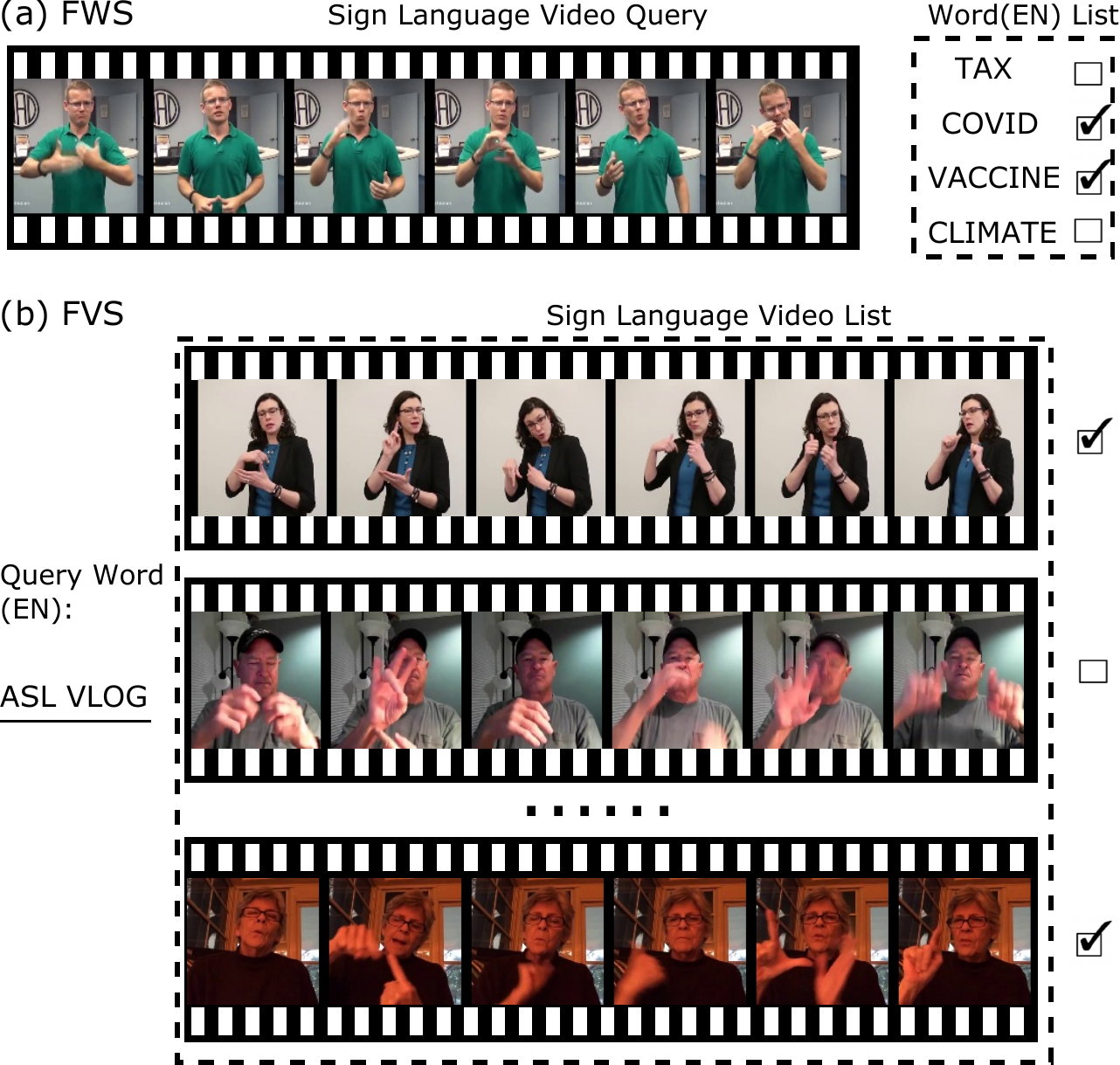}
       \caption{\label{fig:search-task}
       Our two search tasks:  (a) \emph{fingerspelled word search} (FWS) for determining which words are fingerspelled in a sign language video \kledit{clip},
       and (b) \emph{fingerspelling video search} (FVS) for searching for sign language videos that include a fingerspelled query word/phrase. The sign language \kledit{videos are untrimmed, i.e.~they include} regular signs in addition to fingerspelling\kledit{, and
       are downsampled here for visualization.}
       }
\end{figure}
We consider two \kledit{tasks:}
Fingerspelled Word Search (FWS) and Fingerspelling-based Video Search (FVS) (see Figure~\ref{fig:search-task}). FWS and FVS respectively \kledit{consist} of detecting %
fingerspelled \kledit{words} within a given raw ASL video stream and detecting \kledit{video clips} of interest containing a given fingerspelled word.\footnote{\kledit{We use "word" to refer to a fingerspelling sequence, which could be a single word or a phrase.}} %
Given a query video clip $v$ and a list of $n$ words $w_{1:n}$, \kledit{FWS is the task of} finding 
\kledit{which (if any) of $w_{1:n}$ are present in $v$.} 
Conversely, \kledit{in FVS the input} is a query word $w$ and $n$ video clips $v_{1:n}$, and the task consists of finding 
all videos \kledit{containing the fingerspelled word $w$}. We consider \kledit{an} open-vocabulary setting where the word $w$ is \kledit{not constrained} to a pre-determined set.
\kledit{The two tasks correspond to} two directions of search (video$\longrightarrow$text and text$\longrightarrow$video), 
\kledit{as is standard practice in other retrieval work such as} video-text search~\citep{zhang2018crossmodal,krishna2017dense,ging2020coot}.

\section{Model}
\label{sec:models}

\begin{figure}[htb]
  \centering
  \includegraphics[width=1.0\linewidth]{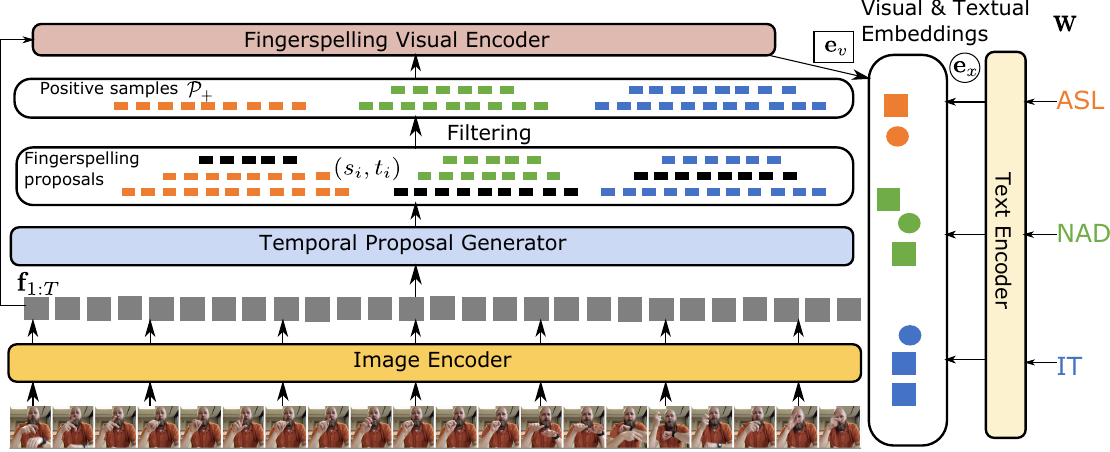}
    \caption{\label{fig:fss-net-model}  FSS-Net: The proposed model for fingerspelling search and \kledit{retrieval.} The model maps candidate fingerspelling segments and text into a shared embedding space. The circle and square symbols refer to textual and visual embeddings, respectively. The colors \kledit{correspond to  different input fingerspelling sequences}. As pictured, this is the training time model, where the pairing between text and video segments is known.  At test time, the labels (colors) of the visual embeddings are unknown \bsedit{and we do not filter the proposals}. }
\end{figure}

We propose a single \kledit{model, FSS-Net (for "FingerSpelling Search Network"),} summarized in \kledit{Figure}~\ref{fig:fss-net-model}, to address the two aforementioned search tasks. FSS-Net receives a pair of \kledit{inputs---a raw ASL video clip, and a written text sequence---and produces a score indicating the degree of match between the video clip and the text}. The text is encoded into \bsedit{an embedding} vector via a learned encoder. The visual branch of \kledit{FSS-Net} generates a number of fingerspelling segment proposals and each proposed visual segment is  encoded into a feature space shared with the text embeddings. Paired embeddings from both modalities are drawn towards each other in the shared embedding space during training.

\textbf{Image encoding} The input image frames are encoded into a sequence of feature vectors via \bsedit{an image} encoder, which consists of \kledit{the} VGG-19~\citep{Simonyan2015very} convolutional layers followed by \kledit{a} Bi-LSTM.\footnote{Transformers~\citep{Vaswani17transformer} can also be used, \kledit{but in our initial experiments, they were outperformed by BiLSTMs on our tasks and data}.} 
We use raw RGB images as \kledit{input,} instead of signer pose as used in \kledit{some prior work}~\citep{Tamer2020KeywordSF,Tamer2020CrossLingualKS} on sign language search, as estimating pose for hands is particularly hard for signing videos in the wild (see \kledit{Section}~\ref{sec:exp} for details).

\textbf{Temporal proposal generation} 
Suppose the visual feature sequence is $\v f_{1:T}$, where $T$ is the number of frames \kledit{in} the video clip.
The purpose of temporal proposal generation is to produce a number of candidate fingerspelling segments \bsedit{$\mathcal{H}(\v I_{1:T})=\{(s_i, t_i)\}_{1\leq i\leq |\mathcal{H}(\v I_{1:T})|}$ %
from $\v f_{1:T}$, where $s_i$,$t_i$ \kledit{are} the start and end frame indices of \kledit{the $i^{\textrm{th}}$ proposed} segment. Below we use $\mathcal{H}$ as a shorthand for $\mathcal{H}(\v I_{1:T})$.} Here we adopt the \kledit{strategy} in~\citep{xu2017rc3d}, \bsedit{which is commonly used to generate proposals} for action detection. \bsedit{Briefly, the model assigns a probability $p_{det}$ of each proposal being fingerspelling.} \bsedit{See~\citep{xu2017rc3d} for more details. We denote the detection loss as $L_{det}$.} %



Note \kledit{that} the training requires known ground-truth fingerspelling \kledit{boundaries}. \kledit{In the fingerspelling datasets we use here~\citep{shi2018american,shi2019fingerspelling}, the fingerspelling boundaries are already annotated, so no further annotation is needed.}

\textbf{Filtering} 
\kledit{A} visual embedding is produced 
\kledit{for each} segment.
\kledit{We} denote \bsedit{a labeled fingerspelling segment (shortened as fingerspelling segment below)} as a tuple $(s, t, w)$, where $s$, $t$ and $w$ represent \kledit{the} start frame index, \kledit{the} end frame index, and the written text it represents. 
\kledit{A} naive approach 
\kledit{would be to use only the} ground-truth fingerspelling segments $\mathcal{P}_g=\{(s_i, t_i, w_i)\}_{1\leq i\leq |\mathcal{P}_g|}$ \kledit{for training}.
However, \kledit{this} approach does not take \kledit{into account the potential shifts (errors) that may exist at test time between the ground-truth and generated segment proposals}. 
The \kledit{embeddings} produced by the \kledit{fingerspelling encoder} %
should be robust to such shifts. To this end, we incorporate proposals in forming positive pairs at training time. Formally, 
\kledit{let} the set of time intervals from \kledit{the} temporal proposal generator \kledit{be}  $\mathcal{H}=\{(s_i, t_i)\}_{1\leq i\leq |\mathcal{H}|}$. 
We sample $K$ intervals from $\mathcal{P}_t$ to form the set of generated fingerspelling segments: 
\begin{equation}
\label{eq-pos-pairs}
\begin{split}
\mathcal{P}_k = &\{(s_k, t_k, w_g)|IoU((s_k, t_k), (s_g, t_g))>\delta_{IoU}, \\
& IS((s_t, t_k), (s_g, t_g))>\delta_{IS}, \\
& (s_k, t_k)\in \mathcal{H}, (s_g, t_g, w_g)\in\mathcal{P}_g\}\\
\end{split}
\end{equation}
\noindent \kledit{where} 
$\text{IS}(x, y)=\frac{\text{Intersection}(x,y)}{\text{Length}(y)}$ and $\text{IoU}(x, y)=\frac{\text{Intersection}(x,y)}{\text{Union}(x,y)}$. We use $\delta_{IoU}$ and $\delta_{IS}$ to control the degree to which the proposals can deviate from the ground-truth.  In addition to \kledit{the intersection over union (IoU)}, we use \kledit{the normalized intersection IS} to eliminate proposals with many missing frames.
We take the union of the two \kledit{sets, $\mathcal{P}_{+}=\mathcal{P}_g \cup\mathcal{P}_k$, as the filtered proposal set to be encoded.} 

 \textbf{Fingerspelling visual encoding \kledit{(FS-encoding)}}
\bsedit{The visual encoding of each segment $(s, t,  w)\in\mathcal{P}_+$ is $\v e_{v}^{(s,t)}=\text{Bi-LSTM}(\v f_{s: t})$.}\footnote{We compared the Bi-LSTM encoder with average/max pooling of $f_{s:t}$, and found the former to perform better.}

\textbf{Text encoding} A written word (or phrase) $\v w$ is mapped to an embedding vector $\v e_x^{w}$ via a text encoder. To handle words \kledit{not seen at training time (and better handle rarely seen words)}, we first decompose $\v w$ into a sequence of characters $c_{1:|w|}$ (e.g. `ASL'=`A'-`S'-`L') and feed the character sequence $c_{1:|w|}$ into a text encoder (\kledit{here, a} Bi-LSTM\footnote{\kledit{Again, transformers can also be used, but in our experiments Bi-LSTMs show} better performance.}). 

\textbf{Visual-text matching} With the above pairs of visual and textual embeddings, we use a training objective function consisting of two triplet loss terms: 
\begin{equation}
\centering
\label{eq:fss-net-triplet}
\begin{split}
& L_{tri}(\v I_{1:T}, \mathcal{P}_+) =\\ &\displaystyle\sum_{(s,t, w)\in\mathcal{P}_{+}}\max\{m+d(\v e^{(s,t)}_v, \v e^{w}_x)\\ 
& -\frac{1}{|\mathcal{N}_w|}\displaystyle\sum_{ w^\prime\in\mathcal{N}_w}d(\v e^{(s,t)}_v, \v e_x^{w^\prime}), 0\} \\
& + \max\{m+d(\v e^{(s,t)}_v, \v e^w_x)\\
& -\frac{1}{|\mathcal{N}_v|}\displaystyle\sum_{(s^\prime,t^\prime)\in\mathcal{N}_v}d(\v e^{(s^\prime,t^\prime)}_v, \v e^w_x), 0\} \\
\end{split}
\end{equation}
\noindent where \kledit{$d$} denotes cosine distance $d(\v a,\v b)=1-\frac{\v a\cdot\v b}{\|\v a\|\|\v b\|}$\kledit{, $m$ is a margin, and}
$\mathcal{N}_v$ and $\mathcal{N}_{w}$ are \kledit{sets} of negative samples of proposals and words. To form negative pairs we 
\kledit{use} semi-hard negative \kledit{sampling~\citep{Schroff2015FaceNetAU}:}
\begin{equation}
\label{eq:semi-hard-neg}
\centering
\begin{split}
\mathcal{N}_v &= \{(s^\prime, t^\prime)|d(\v e^{(s^\prime, t^\prime)}_v, \v e^w_x) > d(\v e^{(s,t)}_v, \v e^w_x)\} \\
\mathcal{N}_w &= \{w^\prime|d(\v e^{(s,t)}_{v}, \v e^{w^\prime}_x) > d(\v e^{(s,t)}_v, \v e^w_x)\} \\
\end{split}
\end{equation}
\noindent For efficiency, negative samples are selected 
\kledit{from the corresponding} mini-batch. 

\textbf{Overall loss} The model is trained 
with \kledit{a combination of the} detection loss and triplet loss: %
\begin{equation}
\label{eq:fss-overall-loss}
 L_{tot}(\v I_{1:T}, \mathcal{P}_g) = \lambda_{det} L_{det}(\v I_{1:T}, \mathcal{P}_g) + L_{tri}(\v I_{1:T}, \mathcal{P}_+)
\end{equation}
 with \kledit{the} tuned weight  $\lambda_{det}$ controlling the relative importance of detection versus visual-textual matching.%

\textbf{Inference} At test time, the model assigns a score $sc(\v I_{1:T}, w)$ \kledit{to a given video clip $\v I_{1:T}$ and word $w$.} The word is encoded into the word embedding $\v e^{w}_x$.
\bsedit{Suppose the set of fingerspelling proposals generated by the temporal proposal generator is $\mathcal{H}(\v I_{1:T})$.}
\kledit{We define a} scoring function for the proposal $h\in\mathcal{H}(\v I_{1:T})$ and word $w$ 
\begin{equation}
\label{eq:fss-search-score}
{sc}_{word}(h_m, w)=p_{det}^\beta(1-d(\v e_v^{h_m}, \v e_x^w))
\end{equation}
\noindent where $p_{det}$ is the probability  \bsedit{given by the temporal proposal generator}
and $\beta$ controls the relative weight between detection and matching.  \kledit{In other words, in order for a segment and word to receive a high score, the segment should be likely to be fingerspelling (according to $p_{det}$) and its embedding should match the text.} 

 Finally, the \kledit{score for the} video clip $\v I_{1:T}$ and the word $w$ 
 \kledit{is defined as} the highest score among \bsedit{the set of proposals $\mathcal{H}(\v I_{1:T})$:}
\begin{equation}
\label{eq:fss-final-score}
{sc}(\v I_{1:T},w)=\displaystyle\max_{\begin{subarray}{l}{h\in\mathcal{H}(\v I_{1:T})}\end{subarray}}{sc}_{word}(h, w)
\end{equation}

\section{Experimental Setup}
\label{sec:search-exp-setup}

We conduct experiments on ChicagoFSWild~\citep{shi2018american} and ChicagoFSWild+~\citep{shi2019fingerspelling}. We follow the setup of Chapter~\ref{ch:fsdet} \kledit{and} split \kledit{the} raw ASL videos into 300-frame clips with a 75-frame overlap between neighboring chunks and remove clips without fingerspelling.

\subsection{Implementation}
\textbf{Data Pre-processing} 
The raw images in ChicagoFSWild and ChicagoFSWild+ datasets contain diverse visual scenes which can involve multiple persons. We adapt the heuristic approach used in~\citep{shi2019fingerspelling} to select the target signer. Specifically, we use an off-the-shelf face detector to detect all the faces in the image. We extend each face bounding box by 1.5 times size of the bounding box in 4 directions and select the largest one with highest average magnitude of optical flow~\citep{Farnebck2003TwoFrameME}. We further use the bounding box averaged over the whole sequence to crop the ROI area, which roughly denotes the signing region of a signer. Each image is resized to $160\times160$ before feeding into the model.

\textbf{Implementation Detail}
The backbone convolutional layers are taken from VGG-19~\citep{Simonyan2015very}. We pre-train the convolutional layers with a fingerspelling recognition task using the video-text pairs from the corresponding dataset. In pre-training, the VGG-19 layers are first pre-trained on ImageNet~\citep{deng2009imagenet} and the image features further go through a 1-layer Bi-LSTM with 512 hidden units per direction. The model is trained with CTC loss~\citep{Graves2006ConnectionistTC}. The output labels include the English alphabet plus the few special symbols, $<$space$>$, ', \&, ., @, as well as the blank symbol for CTC. The model is trained with SGD with batch size 1 at the initial learning rate of 0.01. The model is trained for 30 epochs and the learning rate is decayed to 0.001 after 20 epochs. 
The VGG-19 convolutional layers are frozen in FSS-Net training.

In FSS-Net, the visual features output from convolutional layers are passed through a 1-layer Bi-LSTM with 256 hidden units per direction to capture temporal information. To generate proposals, we first transform the feature sequence via a 1D-CNN with the following architecture: conv layer (512 output dimension, kernel width 8), max pooling (kernel width 8, stride 4), conv layer (256 output dimension, kernel width 3) and conv layer (256 output dimension, kernel width 3). The scale of anchors is chosen from the range: $\{1, 2, 3, 4, 5, 6, 7, 8, 9, 10, 12, 14,$
$16, 18, 20, 24, 32, 40, 60, 75\}$,  according to the typical fingerspelling lengths in the two datasets. The positve/negative threshold of the anchors are 0.6/0.3 respectively. $\delta_{IoU}/\delta_{IS}$ are 1.0/0.8 (chosen from \{0.4, 0.6, 0.8, 1.0\}). The FS-encoder and text encoder are 3-layer/1-layer BiLSTM with 256 hidden units respectively. The margin $m$, number of negative samples in $\mathcal{N}_v$ and $\mathcal{N}_w$ are tuned to be 0.45, 5 and 5. The model is trained for 25 epochs with Adam~\citep{Kingma2015AdamAM} at initial learning rate of 0.001 and batch size of 32. The learning rate is halved if the mean average precision on the dev set does not improve for 3 epochs. $\lambda_{det}$ in equation~\ref{eq:fss-overall-loss} is 0.1 (chosen from \{0.1, 0.5, 1.0\}). At test time, we generate $M=50$ proposals after NMS with IoU threshold of 0.7. $\beta$ is tuned to 1 (chosen from \{0.5, 1, 2, 3\}).

\subsection{Baselines}

\kledit{We compare the proposed model, FSS-Net,} %
to the following baselines adapted from common approaches for search and retrieval in related fields.  %
To facilitate comparison, the network architecture for \kledit{the} visual and text encoding in \kledit{all baselines is the same as in} FSS-Net. 

\textbf{Recognizer} In this approach, we train a %
\kledit{recognizer that transcribes the video clip into text.}
\kledit{Specifically, we train a recognizer to output a sequence of symbols consisting of either fingerspelled letters or a special non-fingerspelling symbol $<$x$>$.} 
\kledit{We train the recognizer with a connectionist temporal classification (CTC) loss~\citep{Graves2006ConnectionistTC}, which is commonly used for speech recognition}. At test time, we use beam search to generate a list of hypotheses $\hat{\v w}_{1:M}$ for the target video clip $\v I_{1:T}$. Each hypothesis $\hat{w}_m$ is split into a list of words $\{\hat{w}^n_m\}_{1\leq n\leq N}$ \kledit{separated} by $<$x$>$. The matching score between video $\v I_{1:T}$ and \kledit{$\v w$} is defined as:
\begin{equation}
    \label{eq:word-list-score}
    {sc}(\v I_{1:T}, w)=1-\displaystyle\min_{1\leq m\leq M}\displaystyle\min_{1\leq n\leq N}\text{LER}(\hat{w}^n_m, w)
\end{equation}
\noindent where \kledit{the letter error rate} $\text{LER}$ \kledit{is the Levenshtein edit distance.}
This approach is adapted from~\citep{Saralar2004LatticeBasedSF} for spoken utterance retrieval. \bsedit{Fingerspelling boundary information is not used in \kledit{training} this baseline \kledit{model}.}

\textbf{Whole-clip} \kledit{The} whole-clip baseline encodes the whole video clip $\v I_{1:T}$ into a visual embedding  $\v e_v^{I}$, which \kledit{is} matched to the textual embedding $\v e_{x}^w$ of \kledit{the query} $\v w$. 
The model is trained with contrastive loss as in equation~\ref{eq:fss-net-triplet}. At test time, the 
\kledit{score for video clip $\v I_{1:T}$ and word $\v w$ is}:
\begin{equation}
    \label{eq:word-list-score}
    {sc}(\v I_{1:T},\v w)=1-d(\v e_{v}^{I}, \v e_{x}^{w})
\end{equation}
\noindent where $d$ is the cosine distance \kledit{as} in FSS-Net. \bsedit{Fingerspelling boundary information is \kledit{again} not used in this baseline.}



\textbf{External detector (Ext-Det)} \kledit{This baseline uses} 
the off-the-shelf fingerspelling detectors of ~\citep{Shi2021FingerspellingDI} to generate fingerspelling proposals\kledit{, instead of our proposal generator, and is otherwise identical to FSS-Net.} \kledit{For each dataset (ChicagoFSWild, ChicagoFSWild+), we use the detector trained on the training subset of that dataset.} This baseline uses ground-truth fingerspelling boundaries for the detector training.

\textbf{Attention-based \kledit{keyword search} (Attn-KWS)} %
\kledit{This model is} 
adapted from~\citep{Tamer2020KeywordSF}\kledit{'s approach for} keyword search in sign language. The model employs \kledit{an} attention mechanism to match \kledit{a} text query with \kledit{a} video clip, where each frame is weighted based on the query embedding.  The attention mechanism enables the model to implicitly localize frames relevant to the text. \kledit{The model of~\citep{Tamer2020KeywordSF} is designed for lexical signs rather than fingerspelling.} To adapt the model \kledit{to our} open-vocabulary fingerspelling setting, we use the same text encoder as \kledit{in} FSS-Net to map words into embeddings instead of using \kledit{a} word embedding matrix as in~\citep{Tamer2020KeywordSF}. \kledit{Fingerspelling boundary information is again not used in training this model, which arguably puts it at a disadvantage compared to FSS-Net.} 

The model assigns a score to video clip $\v I_{1:T}$ and word $w$ via equation~\ref{eq:attn-kws}, where $\v e_v^{1:T}$ is the visual feature sequence of $\v I_{1:T}$ and $\v e_x^w$ is the text feature of $w$, $\v W$ and $\v b$ are learnable parameters. The model is trained with cross-entropy loss.

\begin{equation}
\label{eq:attn-kws}
    \begin{split}
        & s(\v e_v^{t}, \v e_x^w)=\alpha(\frac{\v e_v^t\cdot\v e_x^w}{||\v e_v^t||\cdot||\v e_x^w||})^2+\theta\\
        & a(t)=\frac{\exp(s(\v e_v^{t}, \v e_x^w))}{\sum_t\exp(s(\v e_v^{t}, \v e_x^w))}\\
        & sc(\v I_{1:T}, \v e_x^w)=\sigma(\v W\displaystyle\sum_{t=1}^{T}a(t)\v e_v^{t} +\v b) \\
    \end{split}
\end{equation}

\subsection{Evaluation}
 For FWS, we use all words in the test set as the test vocabulary \kledit{$w_{1:n}$}. For FVS, all video clips in the test are used as candidates \kledit{and the text queries are again the entire test vocabulary}. 
 \bsedit{We report the results in terms of standard metrics from the video-text retrieval literature~\citep{Momeni2020SeeingWW,Tamer2020CrossLingualKS}: mean Average Precision (mAP)} \kledit{and mean F1 score (mF1), where the averages are over words for FVS and over videos for FWS.}
 Hyperparameters are chosen to maximize the mAP on the dev set, independently for the two tasks (though ultimately, the best hyperparameter values in our search are identical for both tasks). 
 
 \section{Results and Analysis}
 \label{sec:search-results-analysis}
 
 \subsection{Main Results}
 \label{sec:search-main-results}
 
  Table~\ref{tab:fss-perf-fswild} shows the performance of 
 \kledit{the} above approaches on \kledit{the} two datasets.  First, we \kledit{notice} that embedding-based approaches consistently outperform the recognizer baseline in the larger data setting (ChicagoFSWild+) \kledit{but not the smaller data setting (ChicagoFSWild), which 
 suggests that} embedding-based models generally require more training data. \kledit{The inferior performance of recognizer} also shows that \kledit{explicit} fingerspelling recognition
 is not necessary for the \kledit{search tasks.} 
 In addition, explicit \kledit{fingerspelling detection (Ext-Det, FSS-Net)
 improves performance over} implicit fingerspelling detection (Attn-KWS) and detection-free \kledit{search} (Whole-clip).  \kledit{Explicit fingerspelling detection requires boundary information during training.  Of the models that don't use such supervision, Attn-KWS is the best performer given enough data, but is still far behind FSS-Net.} %
 Our model outperforms all \kledit{of the alternatives}. %
 The \kledit{relative performance of} different models remains consistent across \kledit{the} various metrics \kledit{and the} two search tasks. 

\input{table-tex/search/main-short-fws-fvs}

\bsedit{For completeness, we also measure the performance of different models in \kledit{terms of} ranking-based metrics (e.g., Precision@N, Recall@N), as in prior \kledit{work on} video-text retrieval~\citep{ging2020coot,krishna2017dense}. 
 For top-N retrieved X, we compute the percentage of correct X among N retrieved X as precision@N and the percentage of correct X among all correct X as recall@N, where X is text for FWS and video for FVS. Note the maximum value of R@1 and P@10 can be less than 1 as there are clips with multiple fingerspelling sequences and clips with fewer than 10 fingerspelling sequences. The performance of different models measured by all the above metrics is shown in table~\ref{tab:fss-perf-fswild-long}.
The \kledit{relative} performance of different models \kledit{remains} consistent \kledit{on} these metrics.}
\input{table-tex/search/main-fws-fvs}

The analysis below is done on ChicagoFSWild for simplicity. The conclusions also hold for ChicagoFSWild+.

\subsection{Model Analysis}
\textbf{Does better localization lead to better \kledit{search}?}
\kledit{In the previous section we have seen that models that explicitly detect and localize fingerspelling outperform ones that do not.  Next we look more closely at how well several models---Ext-Det, Attn-KWS and FSS-Net---perform
on the task of localizing fingerspelling, which is a byproduct of these models' output.
We measure performance via AP@IoU, a commonly used evaluation metric for action detection~\citep{Heilbron2015ActivityNetAL} that has also been used for fingerspelling detection~\citep{Shi2021FingerspellingDI}.  AP@IoU measures the average precision of a detector under the constraint that the overlap of its predicted segments with the ground truth is above some threshold Intersection-over-Union (IoU) value.
For Attn-KWS, the model outputs an attention vector, which we convert to segments as in~\citep{Shi2021FingerspellingDI}. }

\input{table-tex/search/table-ap-iou}

In general, \kledit{the models with} more accurate localization 
\kledit{also have} higher search and retrieval performance\kledit{, as seen by comparing Table~\ref{tab:ap_iou} with Table~\ref{tab:fss-perf-fswild}}. 
\kledit{However, differences in AP@IoU do not directly translate to differences in search performance.}
For example, \kledit{the} AP@IoU of Ext-Det (0.344) is \kledit{an order of} magnitude higher than \kledit{that of} Attn-KWS (0.035) while their FVS mAP \kledit{results are much closer (0.593 vs.~0.573)}. %

\textbf{\kledit{Raw images vs.~estimated pose as input}} %
Prior \kledit{work on sign language search~\citep{Tamer2020CrossLingualKS,Tamer2020KeywordSF} has used estimated pose keypoints as input, rather than raw images as we do here.} 
For comparison, we extract body and hand keypoints with \kledit{OpenPose}~\citep{cao2019openpose} and train a model with \kledit{the} pose \bsedit{skeleton} \kledit{as input}. 

\input{table-tex/search/pose-vs-rgb-short}

As is shown in \kledit{Table}~\ref{tab:pose-vs-rgb}, \kledit{the} pose-based model 
\kledit{has much poorer search performance 
than the RGB image-based} models. \kledit{We believe this is largely because, while pose estimation works well for large motions and clean visual conditions, in our dataset much of the handshape information is lost
in the estimated pose (see Figure~\ref{fig:pose_samples}).}

\subsection{Ablation Study} 

\input{table-tex/search/ablation-short-table}

 Within our model, the proposal generator 
\kledit{produces a subset of all possible fingerspelling proposals, intended to represent the most likely fingerspelling segments.}
\kledit{To measure whether this component is important to the performance of the model, we compare our full model with the proposal generator to one where the proposal generator is removed (see Table~\ref{tab:full-ablation-study}).}
When the proposal generator is not used, the model is \kledit{trained only} with ground-truth fingerspelling segments ($\mathcal{P}_g$) and \kledit{considers all possible proposals within a set of sliding windows.}  
%
%
\kledit{Such a "sliding-window" approach
is commonly used in previous} sign language keyword search~\citep{albanie2020bsl1k,Pfister2013LargescaleLO} \kledit{and} spoken \kledit{keyword spotting}~\citep{Chen2015QuerybyexampleKS}. As can be seen from \kledit{Table}~\ref{tab:full-ablation-study} \kledit{(Full model vs. row (1))}, the proposal generator greatly improves \kledit{search}
performance.  \kledit{This is not surprising, since the proposal generator greatly reduces the number of non-fingerspelling segments, thus lowering the chance of a mismatch between the text and video, and also refines the segment boundaries through regression, which should improve the quality of the visual segment encoding.}

\kledit{The fingerspelling detection component of our model has two aspects that may affect performance: imposing an additional loss during training,} and rescoring during inference. We disentangle \kledit{these} two factors and show their respective benefits \kledit{for} our model in \kledit{Table}~\ref{tab:full-ablation-study} \kledit{(row (2) and (3))}. The auxiliary detection task, which includes classification between fingerspelling and non-fingerspelling proposals, helps encode 
more comprehensive visual information 
into the visual embedding. 
In addition, the proposal probability output by the detector contains extra information and merging it into the matching score further improves the \kledit{search} performance.

Table~\ref{tab:full-ablation-study} \kledit{(row (4))} shows the effect of sampling additional proposals ($\mathcal{P}_k$) in fingerspelling detection. Additional positive samples make the visual embedding more robust to \kledit{temporal shifts} in the generated proposals, thus improving \kledit{search} performance. 

\subsection{Result Analysis} 
The performance of our model is \kledit{worse} for short fingerspelled \bsedit{sequences} than for long \bsedit{sequences} (see \kledit{Figure}~\ref{fig:performance-length}). 
This 
\kledit{may be} because shorter words are harder to spot, as is shown from the trend in 
fingerspelling detection in the same figure.
\input{table-tex/search/localization}

\input{table-tex/search/perf-len-plot}

The datasets we use are collected from multiple sources, and the video quality varies between them.  To quantify the effect of visual quality on search/retrieval performance, we categorize the ASL videos into three categories according to \kledit{their source:}
YouTube, DeafVIDEO, 
\kledit{and other miscellaneous sources (misc)}. YouTube videos are mostly ASL lectures with high resolution. DeafVIDEO videos are 
\kledit{vlogs} from deaf users of the social media \kledit{site} \texttt{deafvideo.tv}, where the style, \kledit{camera angle},
and image quality \kledit{vary greatly.}
The visual quality of \kledit{videos in the miscellaneous category tends to fall between the other} 
two categories. Typical image examples from the three categories can be found in Figure~\ref{fig:source-example}.

\input{table-tex/search/source-examples}

\input{table-tex/search/pr-curve}

\bsedit{The \kledit{FWS performance of our model on videos} in YouTube, DeafVIDEO, and misc are 0.684, 0.584, 0.629 (mAP) respectively.}
The results are overall consistent with the \kledit{perceived relative visual qualities of these categories}.

\kledit{As a qualitative analysis, we examine example words and videos on which our model is more or less successful.  Table~\ref{tab:map-examples} shows the query words/phrases with the highest/lowest FVS performance.  The best-performing queries tend to be long and drawn from the highest-quality video source.}

\input{table-tex/search/fvs-examples}

We also visualize the top \kledit{FWS} predictions made by our model in 
\kledit{several} video clips \kledit{(see Figure~\ref{fig:kws-examples})}.
\kledit{Another common source of error is confusion between letters with similar handshapes (e.g., "i" vs. "j").  A final failure type is} 
fingerspelling detection failure. \kledit{As our model includes a fingerspelling detector, detection errors can harm search performance.} 

\input{table-tex/search/kws-examples}

\section{Summary}

This chapter takes one step toward better addressing the need for language technologies for sign languages, by defining fingerspelling search tasks and developing a model tailored for these tasks.  These tasks are complementary to existing work on keyword search for lexical signs, in that it addresses the need to search for a variety of important content that tends to be fingerspelled, like named entities.  Fingerspelling search is also more challenging in that it requires the ability to handle an open vocabulary and arbitrary-length queries.  Our results demonstrate that a model tailored for the task in fact improves over baseline models based on related work on signed keyword search, fingerspelling detection, and speech recognition.  \kledit{However, there is room for improvement between our results and the maximum possible performance.  
One important aspect of our approach is the use of explicit fingerspelling detection within the model.  An interesting avenue for future work is to address the case where the training data does not include segment boundaries for detector training.  Finally, a complete sign language search system should consider both fingerspelling and lexical sign search.} 

%% file: table-tex/search/main-short-fws-fvs.tex
\begin{table}[htb]
    \caption{\label{tab:fss-perf-fswild}FWS/FVS 
    \kledit{performance} on \kledit{the} ChicagoFSWild and ChicagoFSWild+ \kledit{test sets}. \bsedit{The range of mAP and mF1 is [0, 1].}
    $\bigstar$: methods that use fingerspelling boundaries in training.}
    \centering
    \begin{tabular}{l c c c c }
        \toprule
        \multicolumn{1}{c}{} &
        \multicolumn{2}{c}{FWS (Video$\implies$Text)} &
        \multicolumn{2}{c}{FVS (Text$\implies$Video)}   \\
        \cmidrule(r){2-3}
        \cmidrule(r){4-5}
         \multicolumn{5}{c}{\textbf{ChicagoFSWild}} \\ 
        \cmidrule(r){2-3}
        \cmidrule(r){4-5}
        Method & mAP & mF1 & mAP & mF1 \\
        \cmidrule(r){2-3}
        \cmidrule(r){4-5}
        Whole-clip & 0.175 & 0.154 & 0.142 & 0.119  \\
        Attn-KWS & 0.204 & 0.181 & 0.246 & 0.229 \\
        Recognizer & 0.318 & 0.315 & 0.331 & 0.305  \\ 
        Ext-Det$\bigstar$ & 0.383 & 0.385 & 0.332 & 0.312 \\
        FSS-Net$\bigstar$ &\textbf{0.434}&\textbf{0.439} & \textbf{0.394} & \textbf{0.370}   \\
        \cmidrule(r){2-3}
        \cmidrule(r){4-5}
         \multicolumn{5}{c}{\textbf{ChicagoFSWild+}} \\ 
        \cmidrule(r){2-3}
        \cmidrule(r){4-5}
        Method & mAP & mF1 & mAP & mF1  \\
        \cmidrule(r){2-3}\cmidrule(r){4-5}
        Whole-clip & 0.466 & 0.457
        & 0.548 & 0.526 \\
        Attn-KWS & 0.545 & 0.530 &
        0.573 & 0.547 \\
        Recognizer & 0.465 & 0.462 & 0.398 & 0.405  \\
        Ext-Det$\bigstar$ &
        0.633 & 0.641 &
        0.593 & 0.577 \\
        FSS-Net$\bigstar$ & \textbf{0.674}
 & \textbf{0.677} & \textbf{0.638} &
\textbf{0.631} \\ \bottomrule
    \end{tabular}
\end{table}

%% file: table-tex/search/main-fws-fvs.tex
\begin{table*}[htb]
    \caption{\label{tab:fss-perf-fswild-long}FWS/FVS 
    \kledit{performance} on \kledit{the} ChicagoFSWild and ChicagoFSWild+ \kledit{test sets}. The maximum value of each metric is given in the parentheses (below each metric). \bsedit{The minimum value of each metric is 0.}
     }
    \centering
    \resizebox{\columnwidth}{!}{
    \begin{tabular}{l c c c c c c  c c c c c c }
        \toprule
        \multicolumn{1}{c}{} &
        \multicolumn{6}{c}{FWS (Video$\implies$Text)} &
        \multicolumn{6}{c}{FVS (Text$\implies$Video)}   \\
        \cmidrule(r){2-7}
        \cmidrule(r){8-13}
         \multicolumn{13}{c}{\textbf{ChicagoFSWild}} \\ 
        \cmidrule(r){2-7}
        \cmidrule(r){8-13}
        Method & mAP & mF1 & P$@$1 & P$@$10 & R$@$1 & R$@$10 & mAP & mF1 & P$@$1 &  P$@$10 & R$@$1 & R$@$10 \\
         & (1) & (1) & (1) & (0.16) & (0.75) & (1) & (1)  & (1)  & (1) &  (0.17) & (0.86) & (1)  \\
         \cmidrule(r){2-7}
         \cmidrule(r){8-13}
        Whole-clip & 0.175 & 0.154 & 0.116 & 0.043 & 0.092 & 0.293 & 0.142 & 0.119 & 0.106 & 0.039 & 0.070 & 0.251 \\
        Attn-KWS & 0.204 & 0.181 & 0.158 & 0.059 & 0.108 & 0.358 & 0.246 & 0.229 & 0.238 & 0.061 & 0.179  & 0.411\\
        Recognizer & 0.318 & 0.315 & 0.352 & 0.072 &0.284 & 0.465 & 0.331 & 0.305 & 0.323 & 0.071 & 0.220 & 0.474  \\ 
        Ext-detector & 0.383 & 0.385 & 0.334  & 0.085 & 0.268 & 0.529 & 0.332 & 0.312 & 0.296 & 0.079 & 0.205 & 0.510 \\
        FSS-Net &\textbf{0.434}&\textbf{0.439}&\textbf{0.384} & \textbf{0.093} & \textbf{0.300} & \textbf{0.591} & \textbf{0.394} & \textbf{0.370} & \textbf{0.370} & \textbf{0.091} & \textbf{0.255} & \textbf{0.580}  \\ 
        \cmidrule(r){2-7}
        \cmidrule(r){8-13}
         \multicolumn{13}{c}{\textbf{ChicagoFSWild+}} \\ 
         \cmidrule(r){2-7}
        \cmidrule(r){8-13}
        Method & mAP & mF1 & P$@$1 & P$@$10 & R$@$1 & R$@$10 & mAP & mF1 & P$@$1 &  P$@$10 & R$@$1 & R$@$10 \\
        & (1) & (1) & (1) & (0.16) & (0.76) & (1) & (1)  & (1)  & (1) &  (0.18) & (0.84) & (1)  \\
        \cmidrule(r){2-7}\cmidrule(r){8-13}
        Whole-clip & 0.466 & 0.457&0.416&0.100&0.326&0.626
        & 0.548 & 0.526 & 0.546 & 0.101 & 0.421 & 0.711\\
        Attn-KWS & 0.545 & 0.530 & 0.485 & 0.112 & 0.392 &  0.727 &
        0.573 & 0.547 & 0.541 & 0.111 & 0.408 & 0.748 \\
        Recognizer & 0.465 & 0.462 & 0.470 & 0.094 & 0.390 & 0.620 & 0.398 & 0.405 & 0.394 & 0.090 & 0.292 & 0.617 \\
        Ext-detector &
        0.633 & 0.641 & 0.589 & 0.118 & 0.491 & 0.769 &
        0.593 & 0.577 & 0.568 & 0.114 & 0.419 & 0.786\\
        FSS-Net & \textbf{0.674}
 & \textbf{0.677} & \textbf{0.637} &
\textbf{0.123} & \textbf{0.530} & \textbf{0.796} & \textbf{0.638} &
\textbf{0.631} & \textbf{0.596} &
\textbf{0.123} & \textbf{0.442} & \textbf{0.825}\\ \bottomrule
    \end{tabular}
    }
\end{table*}

%% file: table-tex/search/table-ap-iou.tex
\begin{table}[htp]
    \centering
        \caption{\label{tab:ap_iou}\kledit{Fingerspelling localization performance for detection-based models.} The minimum and maximum values of each metric are 0 and 1.}
    \begin{tabular}{cccc}
    \toprule
         & AP@0.1 & AP@0.3 & AP@0.5 \\ 
         \midrule
        Attn-KWS & 0.268 & 0.104 & 0.035 \\
        Ext-Det & 0.495 & 0.453 & 0.344 \\
        Ours & \textbf{0.568} & \textbf{0.519} & \textbf{0.414} \\
        \bottomrule
    \end{tabular}
\end{table}

%% file: table-tex/search/pose-vs-rgb-short.tex
\begin{table}[hbt]
    \caption{\label{tab:pose-vs-rgb}Impact of input \kledit{type (pose vs.~raw RGB images) on search} performance.}
    \centering
    \begin{tabular}{l c c c c }
        \toprule
        \multicolumn{1}{c}{} &
        \multicolumn{2}{c}{FWS (Video$\implies$Text)} &
        \multicolumn{2}{c}{FVS (Text$\implies$Video)}   \\
        \cmidrule(r){2-3}
        \cmidrule(r){4-5}
        Input & mAP & mF1 & mAP & mF1 \\ \cmidrule(r){2-3}\cmidrule(r){4-5}
        Pose & 0.142 & 0.147 & 0.127 & 0.121 \\
        RGB & \textbf{0.434}&\textbf{0.439} & \textbf{0.394} & \textbf{0.370}  \\
        \bottomrule
    \end{tabular}
\end{table}

%% file: table-tex/search/ablation-short-table.tex
\begin{table}[hbt]
    \caption{\label{tab:full-ablation-study}Effect of \kledit{various} components of FSS-Net on FWS and \kledit{FVS.} 
    }
    \centering
    \begin{tabular}{l c c c c }
        \toprule
        \multicolumn{1}{c}{} &
        \multicolumn{2}{c}{FWS} &
        \multicolumn{2}{c}{FVS}   \\
        \cmidrule(r){2-3}
        \cmidrule(r){4-5}
         & mAP & mF1 & mAP & mF1 \\
        \cmidrule(r){2-3}
        \cmidrule(r){4-5}
        Full model & \textbf{0.434}&\textbf{0.439} & \textbf{0.394} & \textbf{0.370} \\
         \cmidrule(r){1-3}
         \cmidrule(r){4-5}
        (1) w/o generator & 0.186 & 0.180 & 0.259 & 0.270 \\
        (2) $\lambda_{det}=0$, $\beta=0$ & 0.411 & 0.420 & 0.373 & 0.350 \\
        (3) $\lambda_{det}=0.1$, $\beta=0$ & 
        0.418 & 0.432 & 0.360 & 0.348  \\
        (4) w/o $\mathcal{P}_k$ & 
        0.411 & 0.420 & 0.386 & 0.366 \\
        \bottomrule
    \end{tabular}
\end{table}

%% file: table-tex/search/localization.tex
\begin{figure}[btp]
\centering
\caption{\label{fig:localization}
\kledit{Examples of} fingerspelling localization produced by different methods. Upper: \textcolor{red}{Ground-truth}, Bottom: Attention \kledit{weight curve} and \textcolor{blue}{proposals} generated by our model. }
\begin{tabular}{cc}
    \includegraphics[width=0.45\linewidth]{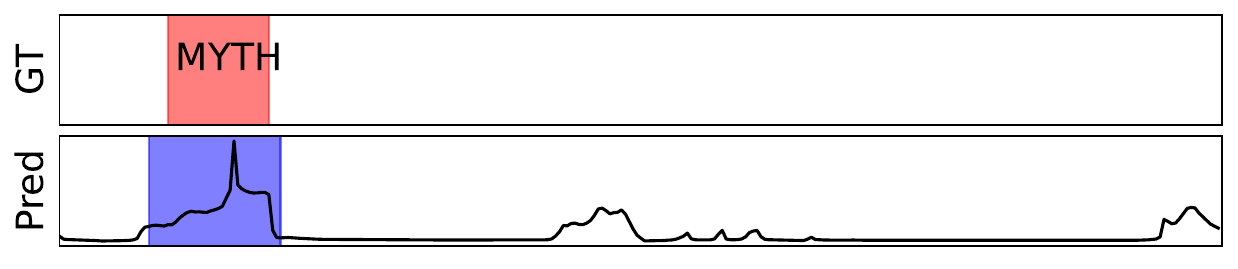} & \includegraphics[width=0.45\linewidth]{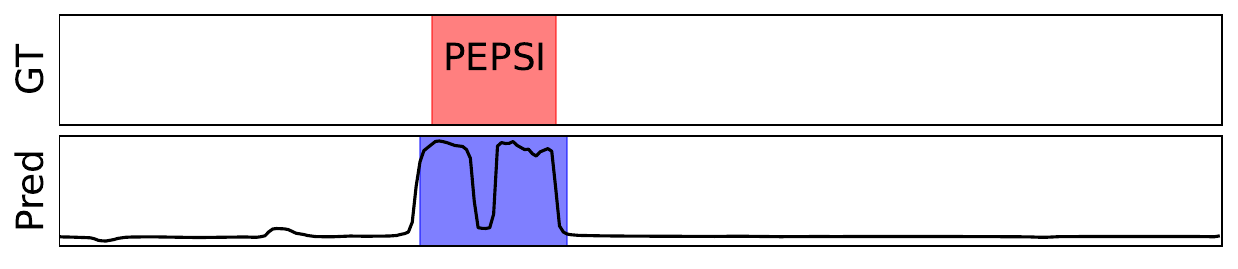}\\
    \includegraphics[width=0.45\linewidth]{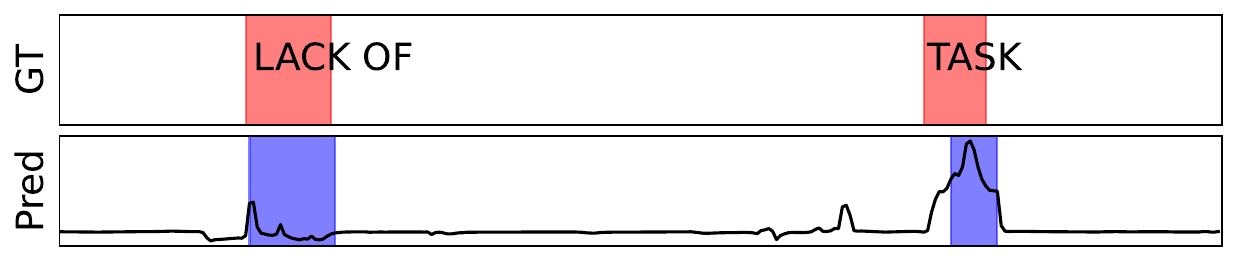} & \includegraphics[width=0.45\linewidth]{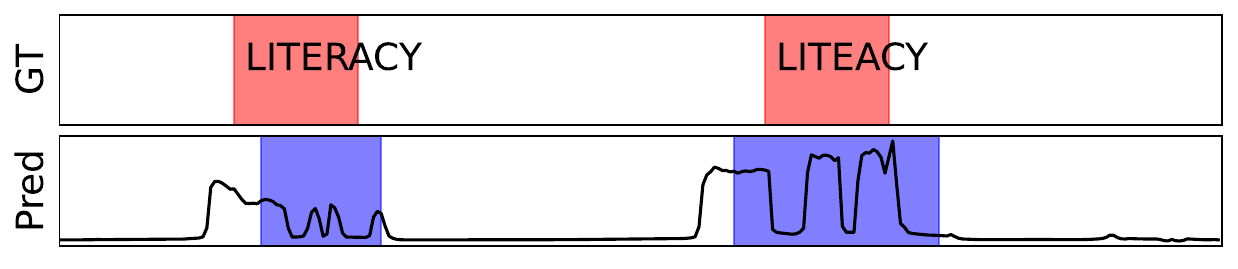} \\
    \includegraphics[width=0.45\linewidth]{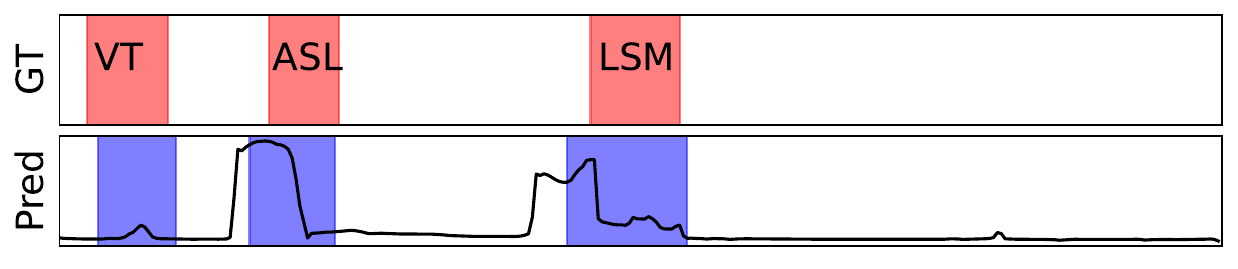} & \includegraphics[width=0.45\linewidth]{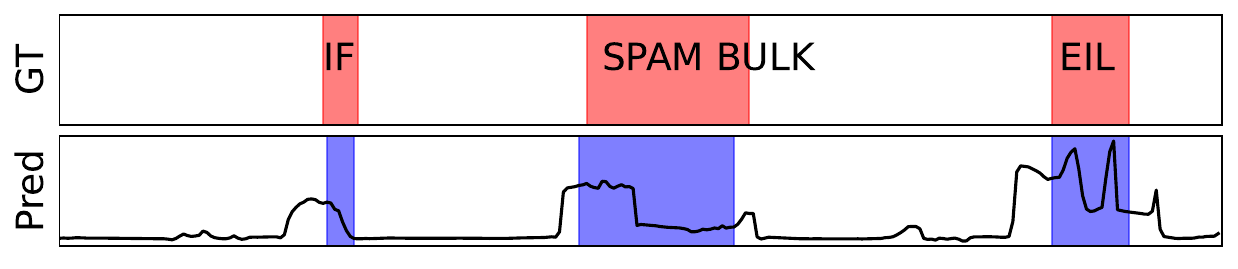}\\
    \includegraphics[width=0.45\linewidth]{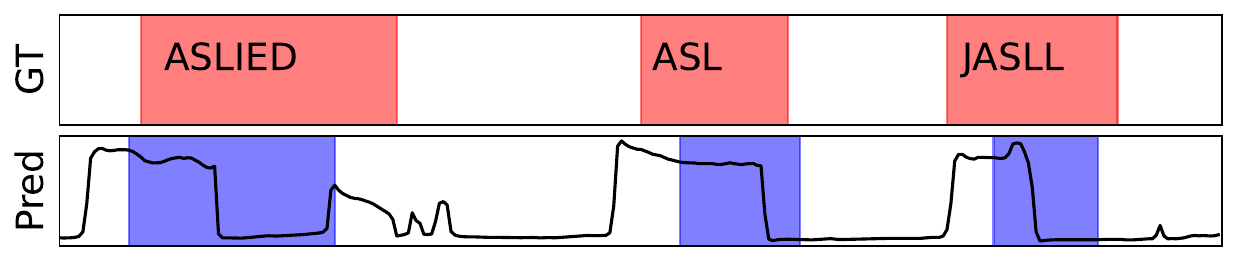} & \includegraphics[width=0.45\linewidth]{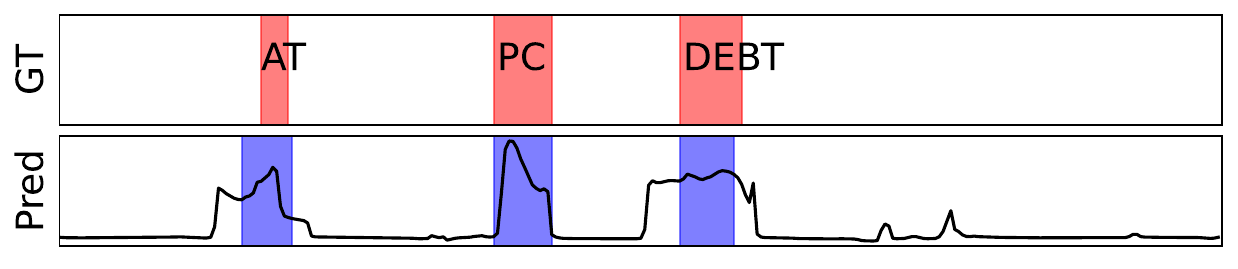} \\
    \end{tabular}
\end{figure}

%% file: table-tex/search/perf-len-plot.tex
\begin{figure}[btp]
\centering
    \caption{\label{fig:performance-length}Performance as a function of fingerspelled word length. Red: FVS mAP, Blue: detection AP@IoU. }
    \includegraphics[width=0.6\linewidth]{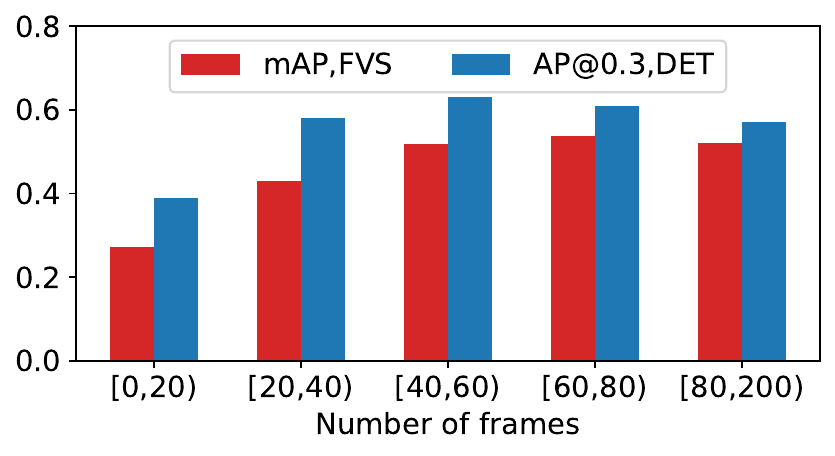}
\end{figure}

%% file: table-tex/search/source-examples.tex
\begin{figure*}[btp]
\caption{\label{fig:source-example}Example image frames from different sources in ChicagoFSWild and ChicagoFSWild+.}
\begin{tabular}{ll}
\toprule
    YouTube & \raisebox{-.3\height}{\includegraphics[width=0.8\linewidth]{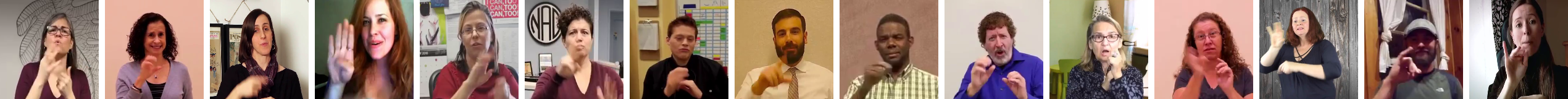}} \\
\midrule
    DeafVIDEO & \raisebox{-.3\height}{\includegraphics[width=0.8\linewidth]{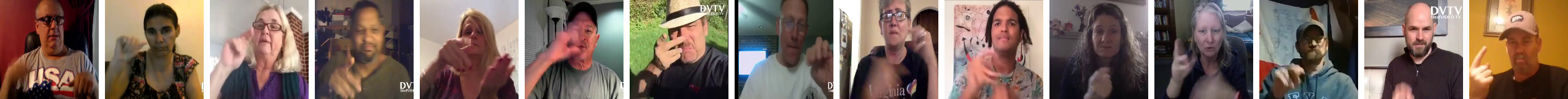}}\\
\midrule    
    misc & \raisebox{-.3\height}{\includegraphics[width=0.8\linewidth]{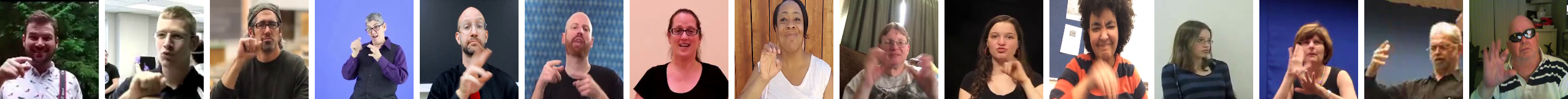}} \\
\bottomrule
\end{tabular}
\end{figure*}

%% file: table-tex/search/pr-curve.tex
\begin{figure*}[btp]
\centering
\caption{\label{fig:pr-curves}FVS precision-recall curve of common words in ChicagoFSWild+ test set. Inside (): mAP}
\begin{tabular}{c@{}c@{}c}
    \includegraphics[width=0.32\linewidth]{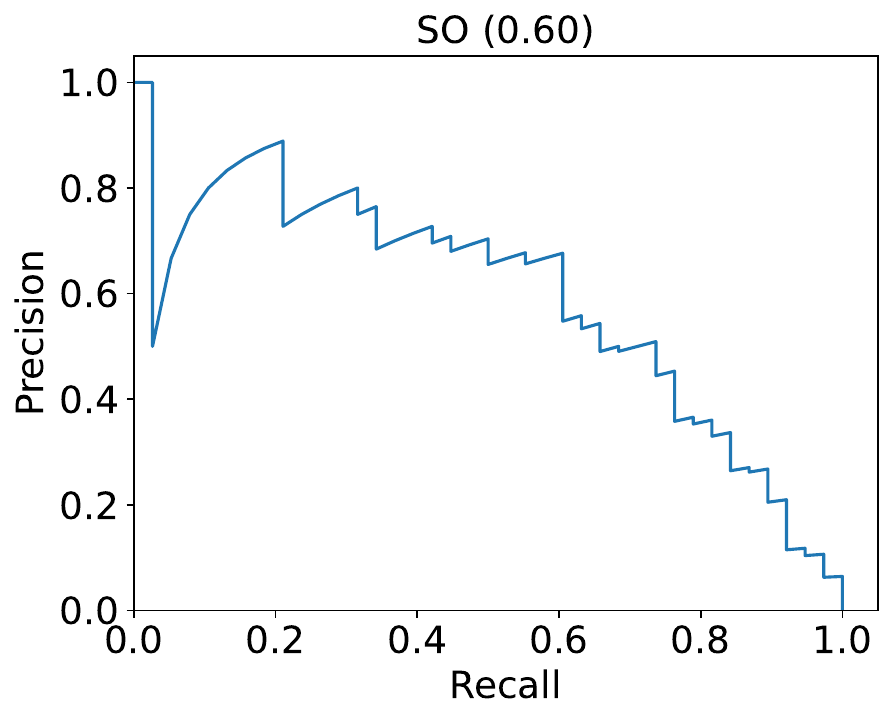} & \includegraphics[width=0.32\linewidth]{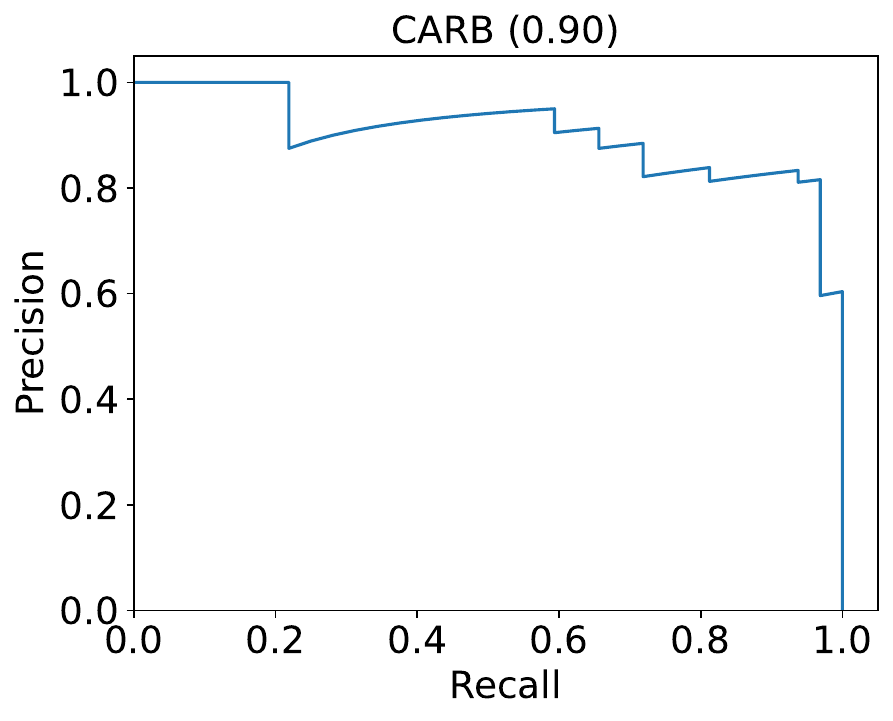} & \includegraphics[width=0.32\linewidth]{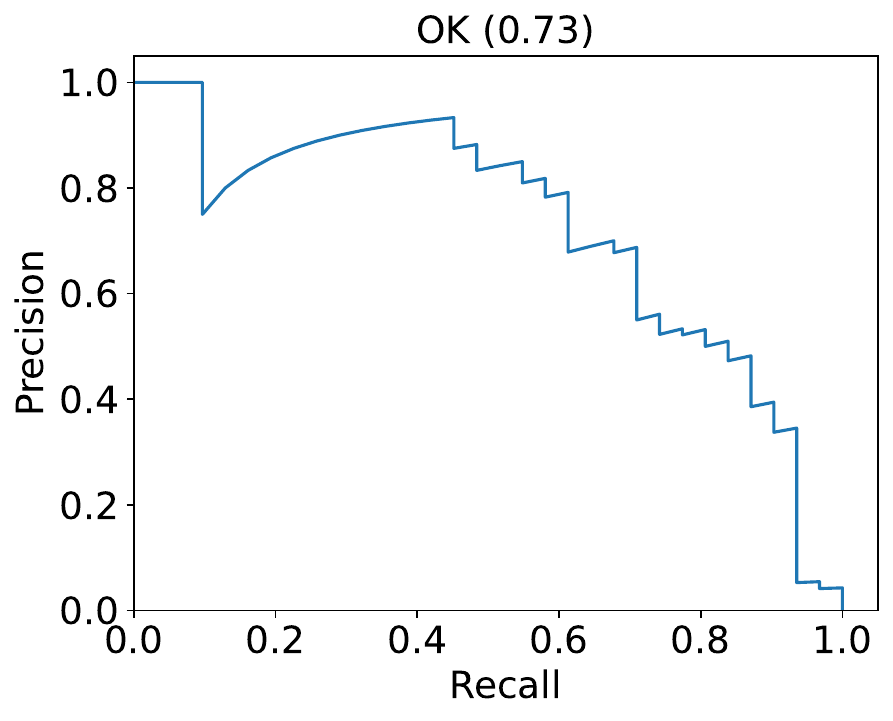} \\
    \includegraphics[width=0.32\linewidth]{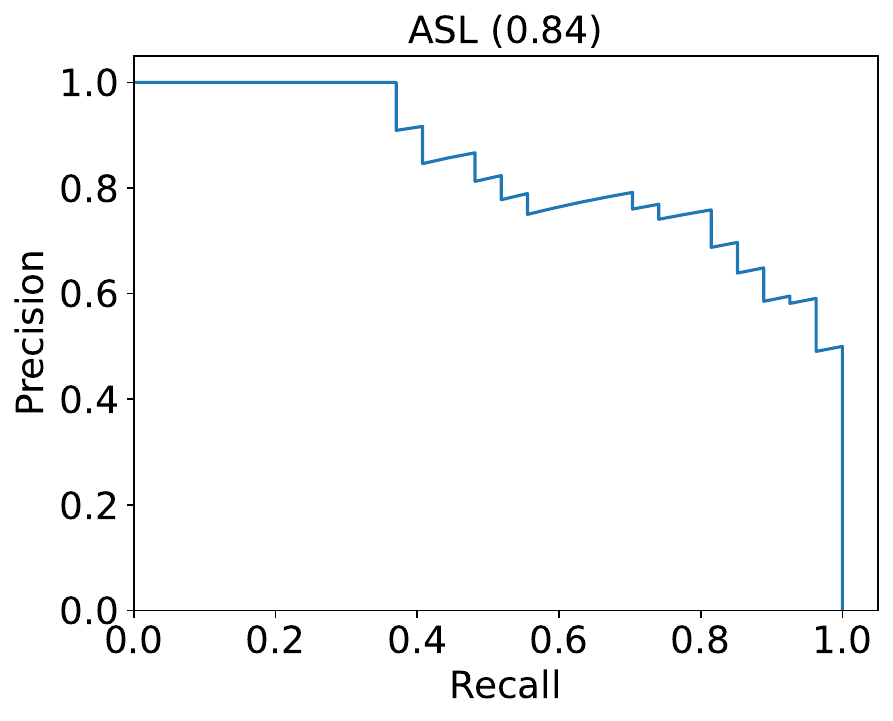} & \includegraphics[width=0.32\linewidth]{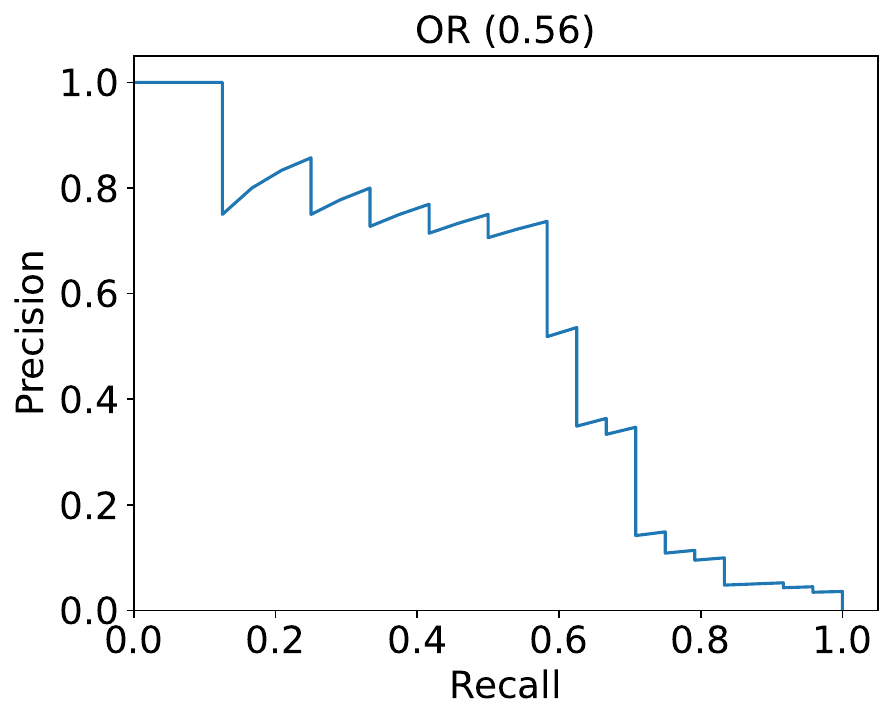} & \includegraphics[width=0.32\linewidth]{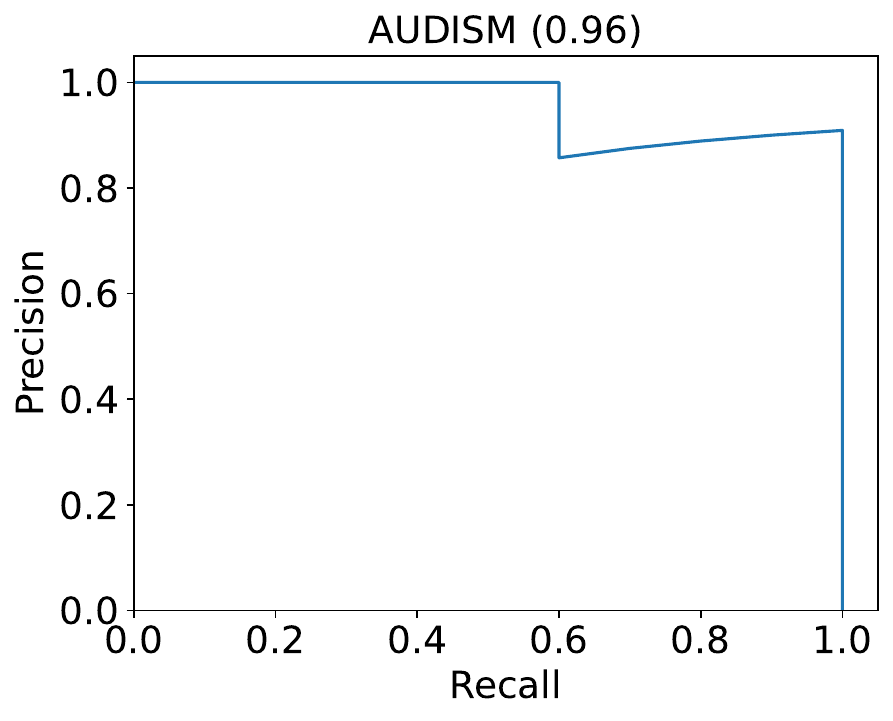} \\
    \includegraphics[width=0.32\linewidth]{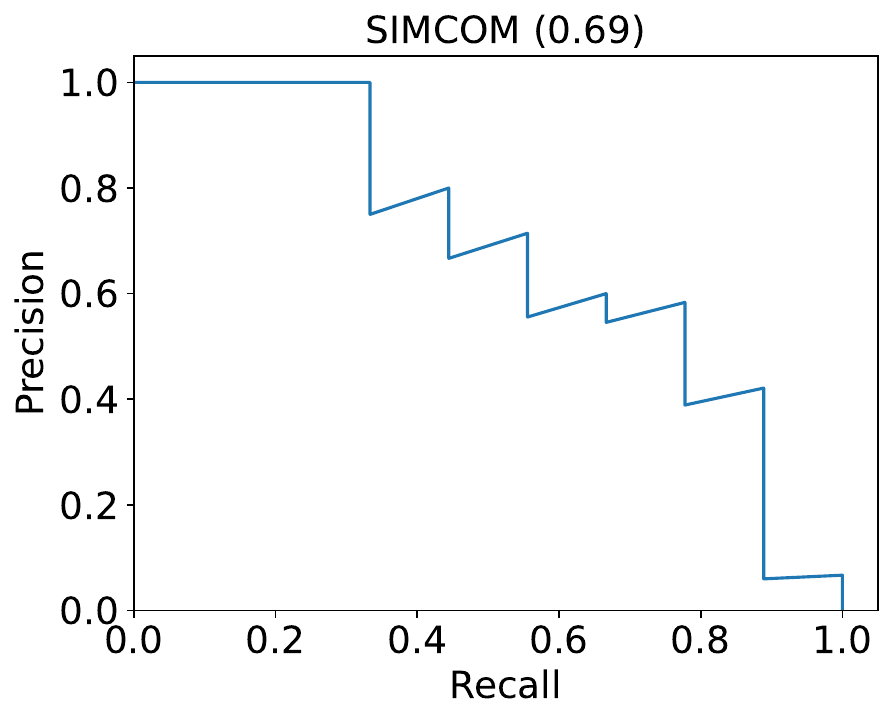} & \includegraphics[width=0.32\linewidth]{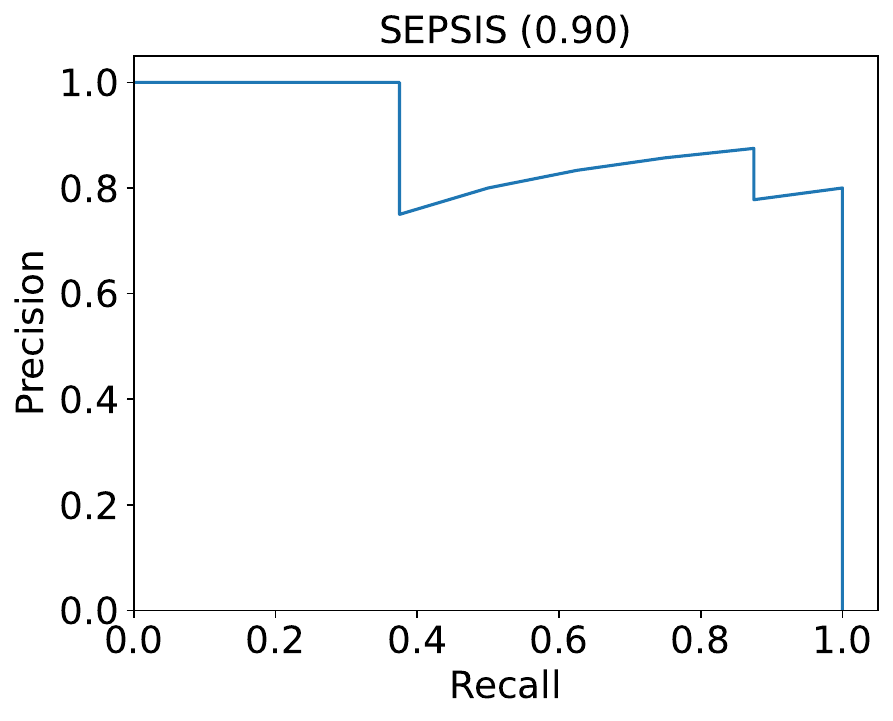} & \includegraphics[width=0.32\linewidth]{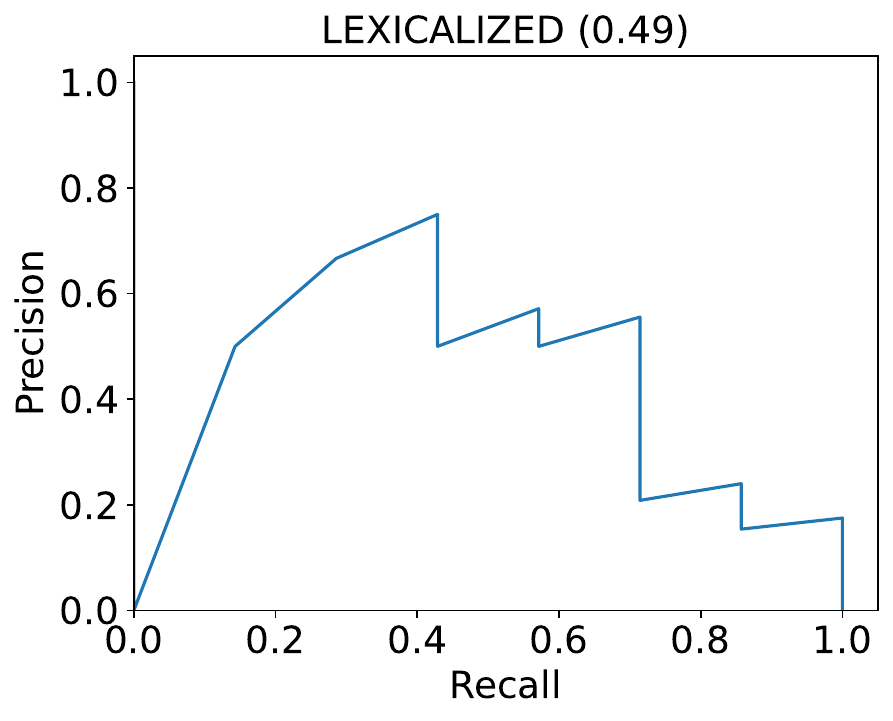} \\
    \end{tabular}
\end{figure*}

%% file: table-tex/search/fvs-examples.tex
\begin{table}[htp]
    \centering
    \caption{\label{tab:map-examples}Example words with low/high mAP in FVS. The source of the corresponding video is given in parentheses}
    \begin{tabular}{c|c}
    \toprule
         Low & High \\ 
         \midrule
         \begin{tabular}{@{}c@{}}
script (YouTube)\\
agent (misc)\\
kc (YouTube)\\
pati (DeafVIDEO)\\
mexer (DeafVIDEO)\\
flow (YouTube)\\
yr (DeafVIDEO)\\
exalted (misc)\\
poem (YouTube)\\
         \end{tabular}
 &          \begin{tabular}{@{}c@{}}
cabol erting (YouTube)\\
vp ron stern (YouTube)\\
co chairs (YouTube)\\
dr kristin mulrooney (YouTube)\\
myles (YouTube)\\
camaspace (YouTube)\\
electronics (YouTube)\\
brain (YouTube)\\
land (DeafVIDEO)\\
         \end{tabular} \\
        \bottomrule
    \end{tabular}
\end{table}

%% file: table-tex/search/kws-examples.tex
\newcommand\widthratio{0.99}
\begin{figure*}[btp]
\caption{\label{fig:kws-examples} Examples of FWS \kledit{predictions}. \kledit{For each example video, the ground truth (GT) is shown along with the top 5} predicted \kledit{fingerspelling sequences}. Top red line: ground-truth fingerspelling segment. Bottom blue line: \kledit{highest-scoring} predicted fingerspelling segment. \kledit{Segment locations are shown here for qualitative analysis, but they are not part of the task evaluation.}  \kledit{Note that many fingerspelling sequences (both ground-truth and predictions) are abbreviations, and some are misspelled; we include all fingerspelling sequences that appear in the \bsedit{test set} in the query vocabulary.} 
}
\begin{tabular}{l}
\toprule
\multicolumn{1}{c}{Successful retrieval}\\
\midrule
\midrule
GT: \textcolor{red}{LITERACYS TO} $\leftrightarrow
$ Pred: \textcolor{blue}{LITERACYS TO},  LITERACY, DISTRACT, ILOW, LIST  \\
\includegraphics[width=\widthratio\linewidth]{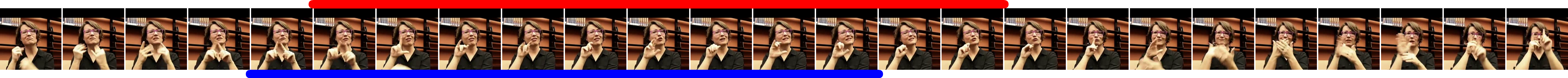}\\
\midrule
GT: \textcolor{red}{ASL} $\leftrightarrow
$ Pred: \textcolor{blue}{ASL}, ALL, ASLIED, ALLAH, HOME  \\
\includegraphics[width=\widthratio\linewidth]{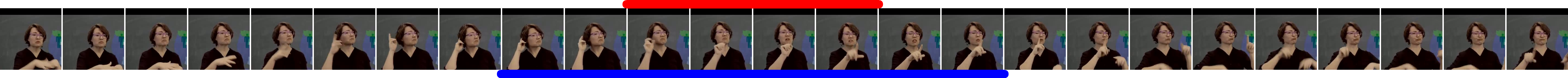}\\
\midrule
GT: \textcolor{red}{US} $\leftrightarrow
$ 
Pred: \textcolor{blue}{US}, USA, CAMUS, LS, SUCH AS  \\
\includegraphics[width=\widthratio\linewidth]{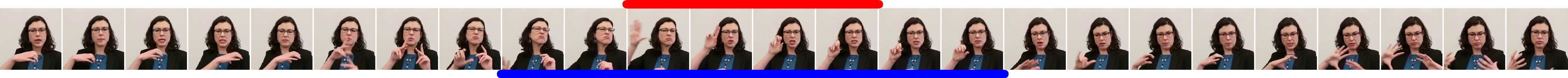}\\
\midrule
\multicolumn{1}{c}{Failure cases}\\
\midrule
\midrule
GT: \textcolor{red}{BACK} $\leftrightarrow
$ 
Pred: BA, AEBSP, BAK, AS, AT BTH BEACH  \\
\includegraphics[width=\widthratio\linewidth]{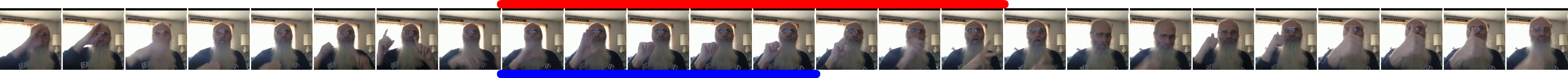}\\
\midrule
GT: \textcolor{red}{JETS} $\leftrightarrow
$ 
Pred: IT, OF, OFF, IE, IX  \\
\includegraphics[width=\widthratio\linewidth]{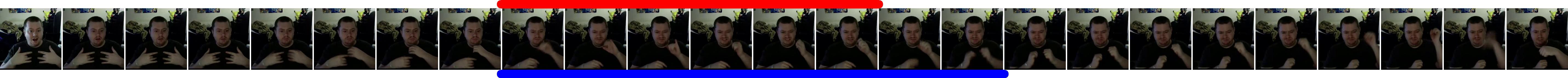}\\
\midrule
GT: \textcolor{red}{TXPU} $\leftrightarrow
$ 
Pred: FISH, F EST, RG, GER, TOSS  \\
\includegraphics[width=\widthratio\linewidth]{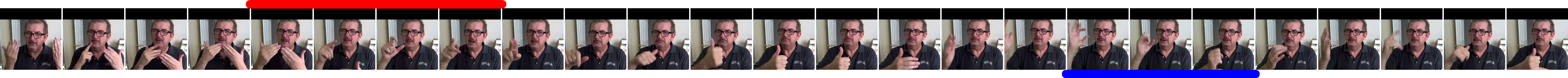}\\
\end{tabular}
\end{figure*}

%% file: translation.tex
\chapter{Sign Language Translation in the Wild}
\label{ch:slt-wild}

Prior chapters have studied various fingerspelling-related tasks including recognition, detection and search. 
In this chapter, we study sign language translation (SLT),\footnote{Note that "SLT" is often used as an abbreviation for "spoken language technology".  Here we use it exclusively for "sign language translation."} the task 
\kledit{of} translating continuous signing video into written language sentences. 
Prior work in SLT is mostly restricted to specific domains (e.g., weather \kledit{forecasts}~\cite{Camgoz2018neural}, emergency situations~\cite{ko2019neural}), characterized by small vocabulary size and lack of visual variability. 
%
%

To tackle the challenges \kledit{of} sign language translation \kledit{in realistic settings and without glosses}, 
we propose a \kledit{set} of techniques including sign search as a \kledit{pretext} task for pre-training and fusion of mouthing and handshape features. 
\kledit{The proposed techniques 
produce consistent and large improvements in translation quality, over baseline models based on prior work}.
This chapter is based on~\citep{shi2022open}.

\section{Related Work}
\label{sec:translation-related-work}

Direct translation from videos of continuous signing is practically appealing and has received 
\kl{} growing interest recently. \kledit{Ko et al.~\citep{ko2019neural}} \kledit{study translation of} common Korean sign language sentences (in video) \kledit{that may be} used in an emergency scenario. 
In this specific domain with restricted vocabulary size (419 words), the model can achieve BLEU-4 score higher than 60. 
In a \bsedit{larger-vocabulary} setting,
\kledit{Camgoz et al.~\citep{Camgoz2018neural}} \kledit{study translation} of German sign language \kledit{weather forecast} videos 
under various labeling setups. In particular, one of \kledit{their} main findings is the drastic improvement achieved \kledit{when} using gloss \kledit{labels} in training an SLT model. It is hypothesized in~\cite{camgoz2020sign} that glosses, as an intermediate representation of sign language, can provide more direct guidance in learning sign language video representation. Therefore, most followup work~\cite{camgoz2020sign,Chen2022ASM,Zhou2021ImprovingSL,Yin2020BetterSL} largely relies on gloss \kledit{sequences} in training.

Given the high cost of gloss labeling, conducting gloss-free SLT is practically appealing \kledit{but} introduces modeling challenges. Glosses, which are monotonically aligned to the video, provide stronger supervision than text in written language \bsedit{translation} %
and facilitate learning of a more effective video representation.  \kledit{On the} Phoenix-2014T benchmark, \kledit{a} model trained without glosses~\cite{Camgoz2018neural} falls behind its counterpart with glosses by over 10.0 (absolute) BLEU-4 score~\cite{camgoz2020sign}. %
Improving translation in real-world sign language video without gloss labels is the modeling focus of this paper. 
\kledit{There is little prior work addressing} SLT without glosses. \bsedit{In a gloss-free setting, }\kledit{Li et al.~\citep{Li2020TSPNetHF} study} the use of segmental structure in translation to boost translation performance. \kledit{Alptekin et al. ~\citep{alptekin2020neural} incorporate handshape features} into the translation model.  
\bsedit{In this paper, we consider sign spotting pre-training and fusion of multiple local features for gloss-free translation.}
\kl{}\bs{} \kl{}\bs{} \kl{}\bs{}

A typical SLT model is composed of a visual encoder and a sequence model. The visual encoder maps input video into intermediate visual features. In~\cite{Camgoz2018neural}, a sign recognizer CNN-LSTM-HMM trained with gloss labels was used to extract image features. The continuous sign recognizer was replaced by a CTC-based model in~\cite{camgoz2020sign}. In addition to RGB-based images, pose is also used~\cite{ko2019neural,Gan2021SkeletonAwareNS} as a complementary input modality, which is commonly encoded by graph convolutional neural networks. The sequence models in SLT are usually based on standard sequence-to-sequence models in machine translation with either recurrent neural networks~\cite{Camgoz2018neural} or transformers~\cite{camgoz2020sign,Yin2020BetterSL,Chen2022ASM} as the backbone.

\kledit{Two key components of our proposed approach are searching for} spotted signs from video-sentence pairs and fusing multiple local visual features.
There has been \kledit{a substantial} \bsedit{amount of} {prior} 
work~\cite{Buehler2009learning,albanie2020bsl1k,Varol2021ReadAA,momeni2020watch,shi2022searching} devoted to spotting signs in real-world sign language videos. In contrast to \kledit{this prior work} where sign search is the end goal, here we treat sign spotting as a \kledit{pretext task} in the context of SLT. 

\kledit{The use of} multi-channel visual features has \kledit{also} been previously explored \kledit{for} multiple tasks, including sign spotting~\cite{albanie2020bsl1k} and continuous sign language recognition~\cite{koller2020weakly}. Specifically for SLT, \kledit{Camgoz et al.~\citep{camgoz2020multi}} \kledit{learn} a multi-channel translation model by including mouthing and handshape features. However, \kledit{these} local modules are trained with in-domain data 
\bsedit{whose labels are inferred from glosses, which makes it  inapplicable for gloss-free translation.}
\kledit{In contrast, we utilize models pre-trained on out-of-domain data to extract local features and study the effect of feature transfer to translation.}

\section{Model and Pre-Training for Gloss-Free Translation}
\label{sec:translation-method}

A translation model 
\kledit{maps a sequence of $T$ image frames $\v I_{1:T}$ to a sequence of $n$ words $w_{1:n}$.}
\kledit{
In the most recent state-of-the-art approaches~\citep{camgoz2020sign,Li2020TSPNetHF} for SLT, a visual encoder ${M}_g^v$ first maps
$\v I_{1:T}$ 
to a visual feature sequence $\v f_{1:T^\prime}$,
and a transformer-based sequence-to-sequence model decodes $\v f_{1:T_g}$ into $w_{1:n}$.
Our approach is based on the same overall} \bsedit{ architecture (see Figure~\ref{fig:translation-model}). We further incorporate several techniques for pre-training and local feature modeling, described next.
}

\subsection{Sign Spotting Pre-Training}
\kledit{For} $M^v_g$, we use \kledit{an inflated 3D convnet (I3D) developed for action recognition}~\cite{Carreira2017QuoVA}.
Ideally, the visual encoder 
\kledit{should capture} signing-related visual cues (\kledit{arm movement, handshape, and so on}).  
However, the translated sentence in the target language may not provide \kledit{sufficiently} direct guidance \kledit{for} learning \kledit{the} visual representation\bsedit{, as is observed in prior work~\cite{Camgoz2018neural}.}

To alleviate this issue, we 
\kledit{pre-train} the I3D network on %
\kledit{relevant tasks that provide} more direct supervision \kledit{for} the convolutional layers than {full} translation. Specifically,
we pre-train I3D \kledit{for the task of isolated sign recognition on} 
WL-ASL~\cite{li2020word}, a large-scale isolated \kledit{ASL sign dataset.}
Empirically, we observe considerable gains from isolated sign recognition pre-training (see \kledit{Section}~\ref{sec:app-effect-pretrain-dataset}).

\begin{figure}[btp]
    \centering
    \includegraphics[width=0.8\linewidth]{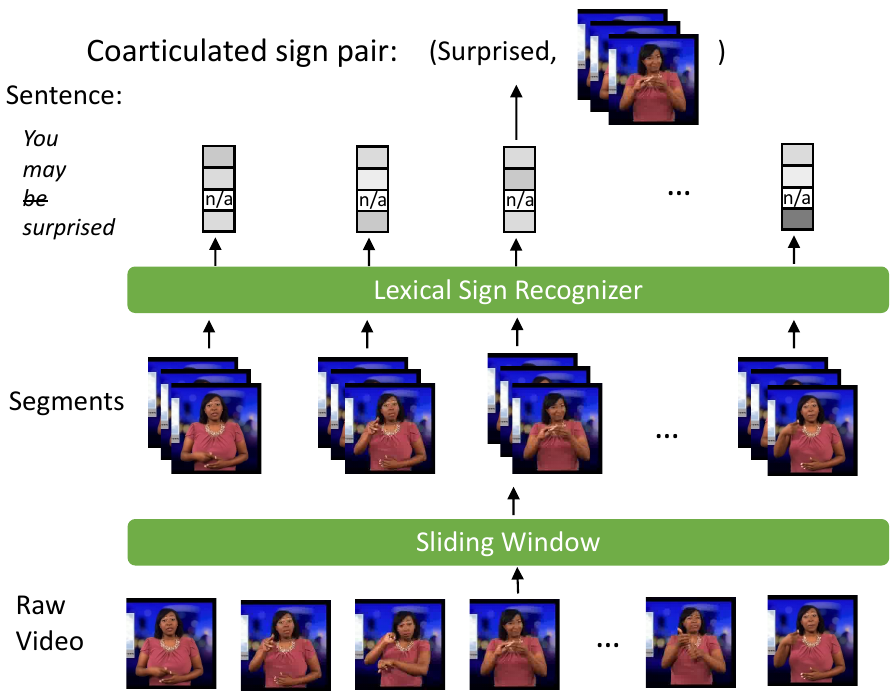}
    \caption{\label{fig:sign-search} Sign spotting. For illustration \kledit{purposes,} only lexical sign search is shown. Fingerspelling sign search works similarly.}
\end{figure}

Despite the aforementioned benefits, the isolated sign recognition \kledit{pre-training causes} 
two potential problems \kledit{for} the translation model. First, there 
\kledit{is} substantial domain mismatch between isolated signs and \kledit{the continuous} signing data used in translation. The coarticulation in a continuous signing stream is not reflected in isolated sign datasets. 
\kledit{In addition,} the {isolated sign} videos are collected from sources such as online \kledit{lexica, which}
usually \kledit{have simpler visual backgrounds 
and less motion blur than}
real-world signing \kledit{video}. 
\kledit{Finally,} existing isolated sign datasets mainly consist of lexical signs and have few instances of fingerspelling. Fingerspelling is used frequently in day-to-day signing and many important content words are commonly fingerspelled. 
Features related to fingerspelling \kledit{may not be encoded well}
due to the lack of fingerspelling-specific pre-training data. 

\input{table-tex/translation/algo-search}

To mitigate the above issues, we propose to search \kledit{for} 
signs from the signing video (see Figure~\ref{fig:sign-search}). \bsedit{The searched signs} are used to pre-train \kledit{the} visual backbone for translation. 
The \kledit{search} relies on a lexical sign recognizer ${M}^{l}$ and a fingerspelling recognizer ${M}^{f}$, which \kledit{map} a video segment into \kledit{a} word probability vector $\v p\in[0,1]^V$ ($V$: vocabulary size) and \kledit{a} letter sequence $\v c_{1:|\v c|}$.
Given a translation video-sentence pair ($\v I_{1:T}$, $w_{1:n}$), the 
\kledit{task} 
is to spot 
\textbf{lexical} and \textbf{fingerspelled} signs $\mathcal{P}=\{(\v I_{s_i:t_i}, w_i)\}_{1\leq i\leq |\mathcal{P}|}$\bsedit{, where the $w_i$ are selected from $w_{1:n}$.} \kl{not clear how $\mathcal{P}$ is related to $w_{1:S}$}  The search process is described briefly below (see Algorithm~\ref{alg:sign-search} for details). 
%

%
%
\kledit{We} generate a list of candidate time intervals for lexical signs and fingerspelling signs respectively with a sliding window approach and a fingerspelling detector $M^d$. For each interval,
we infer its word probability $\v p$ for lexical signs or word hypothesis (i.e., a sequence of characters) $\hat{w}_f$ for fingerspelling. We assign a word from the translated sentence to the target interval if the word probability $p_w$ is high or its edit distance with \kledit{the} fingerspelling hypothesis is low.

\kledit{Unlike} the isolated sign dataset, the spotted signs are sampled from \bsedit{the same data used for translation \kledit{training}.}
Additionally, the detected fingerspelling signs \kledit{should} also enhance the model's ability
to transcribe signs that are fingerspelled.

\subsection{\bsedit{Hand and Mouth} ROI Encoding}
In sign language, 
meaning is usually conveyed via a combination of multiple elements \kledit{including motion of the arms, fingers, mouth, and eyebrows}. 
\kledit{The corresponding local regions in the %
image frame play} an important role in %
\kledit{distinguishing} signs. For instance, SENATE and COMMITTEE have the same place of articulation and \kledit{movement;
the difference 
lies only} in the handshape. Furthermore, mouthing (i.e., mouth movement) is commonly used \kledit{for} adjectives or adverbs to add descriptive meaning
~\cite{Nadolske2008Occurence}.  \kl{I removed "associated with an ASL word" as it seemed redundant, but maybe I'm missing something}
\kledit{Our model's} visual backbone does not explicitly employ 
\kledit{local visual cues.}
In \kledit{principle, learned global features can include sufficient information about the important local cues, but this may require a very large amount of training data.}
\kledit{However, it may be helpful to guide the translation model more explicitly by learning local discriminative features using external tasks.}

Here we focus on learning features for two local visual modalities: handshape and mouthing. To extract handshape \kledit{features}, we train a fingerspelling recognizer\footnote{The implementation is based on~\cite{shi2019fingerspelling}.} \kledit{on} two large-scale fingerspelling datasets~\cite{shi2019fingerspelling} and 
\kledit{use} it to extract features for \kledit{the hand region of interest (ROI)}.  
\kledit{ASL} fingerspelling 
\kledit{includes many handshapres that}
are also used in lexical signs. Recognizing fingerspelling requires distinguishing 
\kledit{quick hand motions} and nuance in finger positions. The features are extracted for both hands in each frame and are concatenated before feeding into the translation model. We denote \kledit{the} hand feature sequence as $\v f_{1:T}^{(h)}$, where $T$ is the video length in frames.

{For} mouthing, we 
\kledit{use}
an external English lip-reading model\footnote{We use the publicly available model of~\cite{shi2022avhubert} without any additional training.}~\cite{shi2022avhubert} \kledit{to}
extract 
\kledit{features} $\v f^{(m)}_{1:T}$ \kledit{from the} lip regions of the signer. 
 \kledit{Although mouthing in ASL is not used to directly "say" words, we assume there is sufficient shared lip motion between speaking and ASL mouthing.} 

\begin{figure}[btp]
    \centering
    \includegraphics[width=0.6\linewidth]{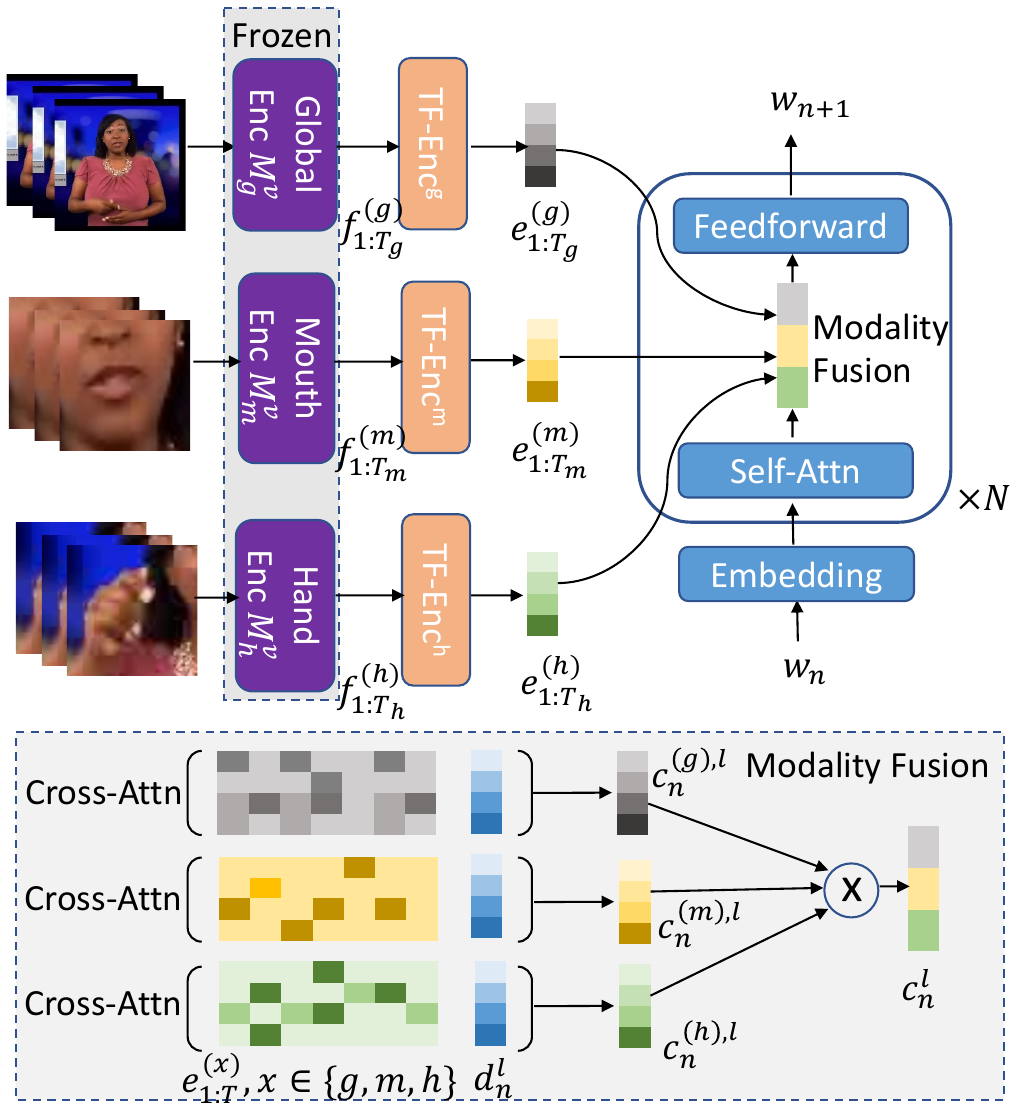}
    \caption{\label{fig:translation-model}Multi-stream sign language translation.}
\end{figure}

\subsection{Fusion and Sequence Modeling}

Given the global/handshape/mouthing feature sequences $\v f^{(g)}_{1:T^\prime}$/$\v f^{(m)}_{1:T}$/$\v f^{(h)}_{1:T}$, the %
\kledit{sequence model 
maps them to} text \kledit{$w_{1:n}$, as
illustrated in Figure~\ref{fig:translation-model}.}
%
\kledit{Since we have multiple feature sequences each with its own sequential properties},  we adopt three independent transformer~\cite{Vaswani17transformer} encoders \kledit{for the three types of features}:
\begin{align*}
\label{eq:enc-transformer}
\begin{split}
    \v e^{(x)}_{1:T_x} &= \text{TF-Enc}^{(x)}(\v f^{(x)}_{1:T_x}), x\in\{g,m,h\} \\
\end{split}
\end{align*}
where $\text{TF-Enc}^{(g)}$, $\text{TF-Enc}^{(m)}$, $\text{TF-Enc}^{(h)}$ denote the transformer encoders for global, mouthing and hand feature sequences respectively.  \kl{shortened}
\kledit{For decoding, we use a single transformer decoder that takes all three encoder representations as input.} 
\bsedit{At decoding timestep $n$, we compute modality-specific context vectors:}
\begin{equation}
\label{eq:cross-attn}
\begin{split}
    \v c^{(x),l}_{n} &= \text{Cross-Attn}^{(x)}(\v d^l_n, \v e^{(x)}_{1:T_x}), x\in\{g,m,h\}  \\
\end{split}
\end{equation}
where $\text{Cross-Attn}^{(g)}$, $\text{Cross-Attn}^{(m)}$ and $\text{Cross-Attn}^{(h)}$ are cross-attention layers~\cite{Vaswani17transformer} for global / mouthing / hand features. 
We concatenate \kledit{the context vectors from the three modalities to form the decoder context vector $\v {c}_n^l=[\v c^{(g),l}_{n},\v c^{(m),l}_{n},\v c^{(h),l}_{n}]$, which is passed to a feedforward layer and then the next layer.  The final layer output is then passed to a linear projection, followed by a final softmax to produce a probability vector over words in the vocabulary.} 

\section{\kledit{Experiments}}
\label{sec:exp}

\input{table-tex/translation/main_result}

\subsection{Setup}
\label{sec:exp-setup}

\textbf{Evaluation} For evaluation, we report BLEU-\{1,2,3,4\}~\cite{Papineni2002BleuAM} and ROUGE~\cite{Lin2004ROUGEAP} \kledit{scores, 
as in prior work on SLT~\cite{Camgoz2018neural,ko2019neural,camgoz2021content}.}
As there is only one English sentence as reference for evaluation, we 
\kledit{also report BLEURT~\cite{sellam2020bleurt} score, a metric that provides a measure of semantic similarity between the prediction and ground truth.}

\textbf{Preprocessing}
For training, we use the time boundaries in the associated video caption to segment raw videos into short clips. 
We extend the time boundaries of a video clip by 0.5 second in both the beginning and end to reduce the proportion of potential missing frames caused by misalignment between subtitle and signing video. 
Each video clip is cropped to include only the signing region of the target signer. Specifically, we employ DLIB face detector~\cite{King2009DlibmlAM} to detect the face of the target signer and crop an ROI centered on the face which is 4 times of the original bounding box. In case of multiple face detected, we employ a simple heuristic to determine the target face track (tracks with the highest optical flow~\cite{Farnebck2003TwoFrameME} magnitude). The selected ROI is resized to $224\times 224$. 

\textbf{Visual Backbone}
For global visual feature extraction, we adopt I3D network~\cite{Carreira2017QuoVA} as our backbone. 
The I3D, pre-trained on Kinetics-400~\cite{Carreira2017QuoVA} is further fine-tuned on WLASL~\cite{li2020word}, an isolated ASL sign dataset with 14,289 isolated training videos from 2000 distinct ASL signs.
The isolated sign recognizer achieves 42.6\% accuracy on WLASL test set.

For hand feature extraction, we train a fingerspelling recognizer on ChicagoFSWild~\cite{shi2018american} and ChicagoFSWild+~\cite{shi2019fingerspelling} datasets, which include 61,536 ASL fingerspelling sequences.
The recognizer is based on Conv-LSTM architecture~\cite{Shi2021FingerspellingDI}
 consisting of first 11 conv layers of VGG-19 followed by a one-layer Bi-LSTM with 512 hidden units per direction.  The model is trained with CTC loss~\cite{Graves2006ConnectionistTC} and achieves 64.5\% letter accuracy on ChicagoFSWild test set. In order to extract hand features on our data, we use HR-Net whole-body pose estimator~\cite{Sun2019DeepHR} to detect hands of the signer and extract features in the hand ROI. Features for left and right hand are concatenated before feeding into the translation model. 

To obtain mouthing feature, we employ AV-HuBERT~\cite{shi2022avhubert}, a state-of-the-art lip reading model for English. The mouth ROI, cropped and resized to $96\times 96$ based on the facial landmarks detected with DLIB facial keypoint detector~\cite{King2009DlibmlAM}, are fed into the lip reading model for feature extraction.

\textbf{Sign Search}
To search lexical signs, we run inference with the aforementioned I3D isolated sign recognizer on 32-frame windows. The window is swept across the whole video clip at a stride of 8 frames. To search fingerspelling, we use the off-the-shelf fingerspelling detector~\cite{Shi2021FingerspellingDI} trained on raw ASL videos of ChicagoFSWild+, which has achieved 0.448 AP@0.5. The aforementioned fingerspelling recognizer is used for searching fingerspelling signs. We keep proposals with confidence score higher than 0.5. The thresholds $\delta_l$/$\delta_f$ are tuned to be 0.6/0.2 respectively. The total number of coarticulated signs detected from our translation data is 32,602. 
We combine WLASL and the spotted signs for pre-training I3D (see section~\ref{sec:effect-sign-pretrain}). 
The model is trained with SGD for 50 epochs at batch size of 8. The learning rate and momentum of SGD  are 0.01 and 0.9 respectively. The learning rate is reduced to half at epoch 20 and 40. 

\textbf{Sequence Model}
The visual backbones are frozen in training translation model. Both transformer encoder and decoder have 2 layers with 512 hidden dimension and 2048 feedforward dimension. The model is trained with Adam~\cite{Kingma2015AdamAM} for 14K iterations at batch size of 64. The learning rate is linearly increased to 0.001 for the first 2K iterations and then decayed to 0 in the later iterations. At test time, we use beam search for decoding. The beam width and length penalty are tuned on the validation set. 

\textbf{Real-time performance} \kledit{Although real-time performance is not a goal of this work, we note that the} whole proposed system (including all pre-processing such as mouth/hand ROI estimation) processes $\sim$25 frames per second on average for a sign language video from scratch on one RTX A4000 GPU.

\subsection{Main Results}

The performance of our proposed approach is shown in Table~\ref{tab:main-result}.
We compare it to two baseline approaches adapted from prior \kledit{work}. Conv-GRU, which uses ImageNet-pretrained AlexNet as \kledit{a} visual backbone, is an RNN-based sequence-to-sequence model proposed \kledit{by} Camgoz et al.~\citep{Camgoz2018neural} for sign language translation without glosses. I3D-transformer is a similar model architecture to \kledit{ours, but it uses only}
global visual \kledit{features} and the CNN backbone is pre-trained \kledit{only on the WLASL isolated sign recognition task}. See \kledit{Section}~\ref{sec:app-rwth2014t-perf} for the performance of \kledit{these} two baseline methods on the popular DGS-German benchmark Phoenix-2014T.  

\kledit{From the results in Table~\ref{tab:main-result}, we observe:}
(1) Conv-GRU 
\kledit{has} the worst performance among the three models. One key difference 
lies in the data \kledit{used} to pre-train the visual \kledit{encoder:  Conv-GRU}
is pre-trained on ImageNet while the latter two are pre-trained on \kledit{sign language-specific data}. 
\kledit{There are, of course, also differences in the model architecture and training pipeline.  To isolate the effect of sign language-specific pre-training, we 
compare I3D-transformer pre-trained with different types of data, and find that isolated sign pre-training leads to consistent gains.}
See \kledit{Section}~\ref{sec:app-effect-pretrain-dataset} for details.
(2) Our proposed approach achieves the best performance. \kledit{On average,}
the relative gain over I3D transformer is $\sim15\%$ in ROUGE and BLEU scores.
This \kledit{demonstrates}
the effect of including spotted signs in visual backbone pre-training and \kledit{of incorporating the multiple} local visual features.
(3) The performance measured by BLEU, ROUGE and BLEURT scores are consistent \kledit{for} different models.

Despite \kledit{the} improvement over baseline approaches, our \kledit{model's performance is still quite poor.}  
We show some qualitative examples of translated sentence in \kledit{Section}~\ref{sec:app-translation-example}. 
\bsedit{The low performance of current translation models \kledit{has also been} observed in prior work {on other sign languages}~\citep{Albanie2021bobsl,camgoz2021content}, highlighting the challenging nature of sign language translation.}  \kl{}

In the next sections,
we analyze the \kledit{effects of the} main components in our model. For \kledit{the purpose of these analyses}, we report BLEU and ROUGE scores on the validation set.

\subsection{Ablation Study}
\label{sec:effect-sign-pretrain}

\textbf{Effect of sign spotting pre-training} In Table~\ref{tab:effect-sign-search}, we compare the performance of models with different pre-training data: WLASL only, WLASL + spotted lexical signs. \kledit{For both models, the visual backbone} is I3D and we do not incorporate local visual features.  
\begin{table}[htp]

\centering
\begin{tabular}{lrrrrr}
\toprule
Model & {\small ROUGE} & {\small BLEU-1} & {\small BLEU-2} & {\small BLEU-3} & {\small BLEU-4} \\
\midrule  
iso only & 18.88 & 18.26 & 10.26 & 7.17 & 5.60 \\ 
+spotted & \textbf{19.65} & \textbf{19.72} & \textbf{11.18} & \textbf{8.56} & \textbf{6.51} \\ \bottomrule
\end{tabular}

\caption{\label{tab:effect-sign-search} Effect of sign spotting pre-training (iso: isolated sign, spot: spotted signs) on the development set.
}
\end{table}

\kledit{The results show that} sign search 
\kledit{consistently} improves performance. Compared to training with WLASL only, including \kledit{lexical sign and fingerspelling spotting produces} $\sim10\%$ relative \kledit{improvements, averaged across 
metrics.} We attribute \kledit{these gains} to the adaptation of I3D to our translation \kledit{data, which includes coarticulation and visual challenges that the isolated sign data lacks}. \kl{}

\kledit{An alternative strategy could be to fine-tune the visual backbone on our translation data.}
However, 
\kledit{this} strategy downgrades translation performance by a large margin (see \kledit{Section}~\ref{sec:app-finetune-freeze} for details). 
\bsedit{Qualitatively, the spotted sign pairs are high-quality in general (see \kledit{Section}~\ref{sec:search-sign-example}). 
%
}
%
%

%
%

%

%

%

%

\textbf{Effect of local feature incorporation} Table~\ref{tab:effect-local-feature} compares models without local visual \kledit{features}, \kledit{and} with both mouthing and handshape \kledit{features}. All models are pre-trained with spotted signs. Overall the incorporation of local features 
\kledit{produces} 5\% gains in different metrics. \bsedit{The gain is relatively larger in BLEU scores of lower orders (e.g., BLEU-1).} \kl{}\bs{}
See \kledit{Section}~\ref{sec:search-sign-example} for qualitative examples of improved translation \kledit{when} using mouthing \kledit{features}.


\begin{table}[htp]
\centering
\begin{tabular}{l|rrrrr}
\toprule
Model & {\small ROUGE} & {\small BLEU-1} & {\small BLEU-2} & {\small BLEU-3} & {\small BLEU-4} \\
\midrule  
global & 19.65 & 19.72 & 11.08 & 8.06 & 6.30 \\ 
+ local & \textbf{20.43} &	\textbf{20.10} &	\textbf{11.81} &	\textbf{8.43} &	\textbf{6.57} \\ \bottomrule
\end{tabular}

\caption{\label{tab:effect-local-feature} Effect of incorporating local visual features. \kl{}\bs{.}}

\end{table}

\subsection{Analysis}

For \kledit{a more} detailed analysis \kledit{of our model}, we \kledit{measure its} performance 
on different evaluation subsets, divided by several criteria.

\textbf{Duplicate vs. non-duplicate} Certain sentences appear frequently in our dataset, which leads to duplicated sentences appearing in both training and evaluation. \kledit{The} duplicate sentences account for 10.9\% (105 out of 967) of the dev set and 10.6\% (103 out of 976) of the test set. Most \kledit{of these are sentences that} are used frequently in the \kledit{news,} such as "Hello", "Thank you", \kledit{and "See you tomorrow".}

\begin{figure}[btp]
    \centering
    \includegraphics[width=0.75\linewidth]{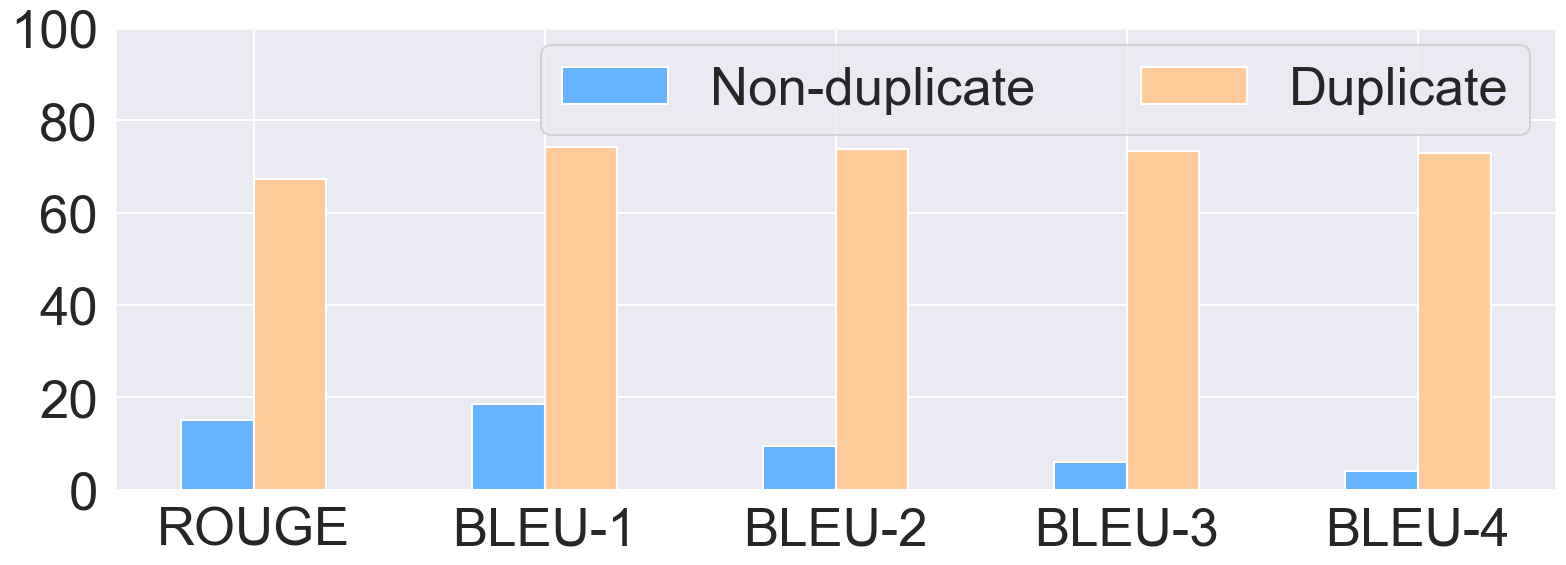}
    \caption{\label{fig:dup-vs-nondup}Comparison of translation performance on duplicate and non-duplicate sentences. Duplicate sentences are \kledit{ones that appear in the training set}.}
\end{figure}

%
\kledit{Our model translates} videos associated with duplicate sentences with high accuracy (see Figure~\ref{fig:dup-vs-nondup}). On \kledit{the duplicate} subset, the BLEU-4 score is \bsedit{72.91}. 
\kledit{Duplicates}
tend to be short clips, which 
\kledit{are} easy for the model to 
memorize.
%
\kledit{In contrast,} the BLEU-4 \kledit{score on the non-duplicate subset} is only 4.09. \kl{}

\begin{figure}[btp]
    \centering
    \includegraphics[width=0.75\linewidth]{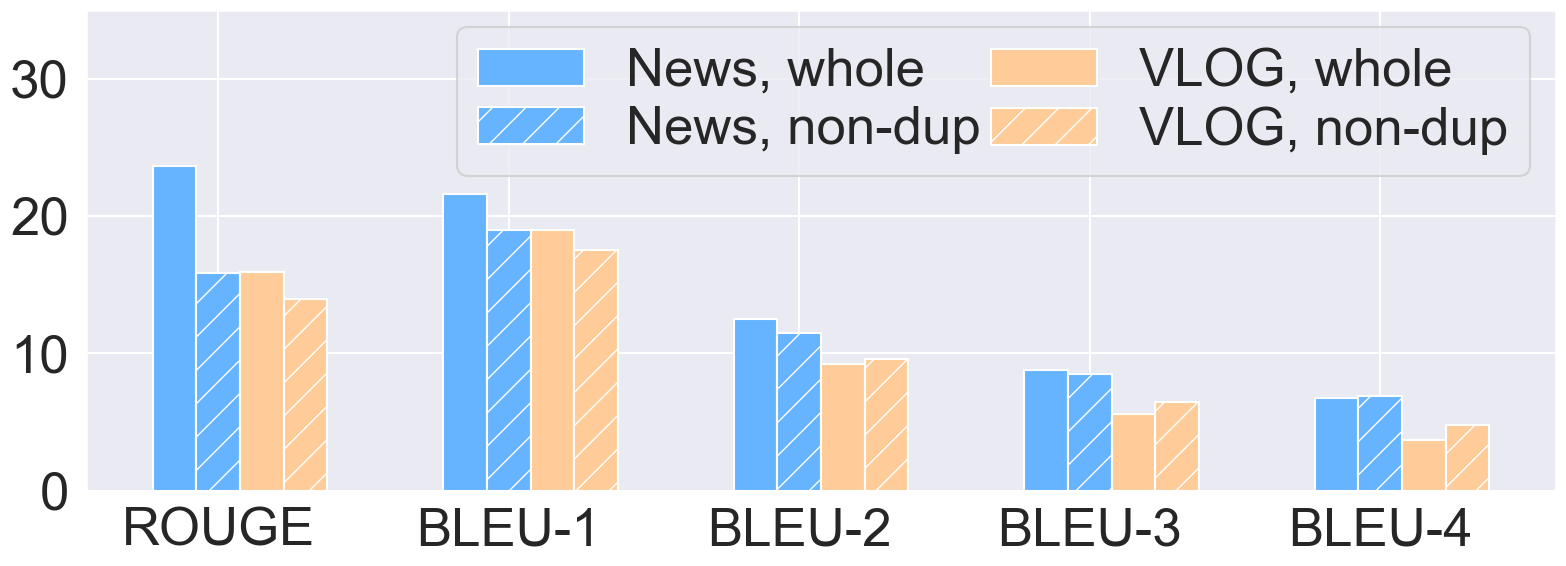}
    \caption{\label{fig:news-vs-vlog}Translation performance for ASL news and VLOGs.}

\end{figure}

\textbf{News vs. VLOGs} Our data are collected from online sign language resources from two categories: news (\texttt{Sign1News} and \texttt{TheDailyMoth}) and VLOGs (\texttt{NAD}). The two sources differ in multiple \kledit{aspects, including} 
visual conditions and linguistic \kledit{content.}
\kledit{In} the dev set, videos from \kledit{these two categories} account for 63.6\%/36.4\% \kledit{of sentences} respectively. We break the performance down according to the source (see Figure~\ref{fig:news-vs-vlog}).  
To avoid the impact of duplicate sentences, we also 
\kledit{perform this comparison 
separately on} non-duplicate sentences. 

Our model 
\kledit{performs better on scripted news} videos regardless of whether the duplicates are included or not, which \kledit{may be} attributed to multiple factors.
On \kledit{the} one hand, \kledit{the} data from 
\kledit{NAD} VLOGs contain a larger set of signers than the news videos. The 
\kledit{variability} in signing among different signers increases the difficulty \kledit{of} translation. \kledit{NAD} VLOG videos also have higher visual variance in terms of image resolution \kledit{and} background diversity. 
\kledit{It is also possible that} the news videos are more likely to be scripted beforehand while the VLOG videos are \kledit{more likely to be} spontaneous. 
Spontaneous ASL videos are \kledit{expected to be} more challenging to translate than scripted videos.

\begin{figure}[btp]
    \centering
    \includegraphics[width=0.75\linewidth]{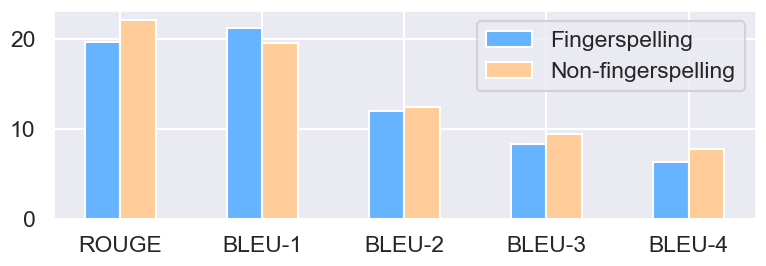}
    \caption{\label{fig:fs-vs-nonfs}Comparison of translation performance between subsets with and without fingerspelling.}
\end{figure}

\textbf{Fingerspelling vs. non-fingerspelling} 
Fingerspelling is an important component in real-world ASL videos. 
In our dev set, $54.7\%$ 
\kledit{of the clips} have at least one fingerspell\kledit{ed word}. 
We measure the performance on the subsets with and without fingerspelling respectively. According to Figure~\ref{fig:fs-vs-nonfs}, the translation quality in non-fingerspelling subsets is consistently higher than the other part. Typical fingerspelled words which our model fails to translate are either proper nouns with low frequency in training (e.g., \uppercase{schmidt}, \uppercase{whaley}), or long words (e.g., \uppercase{massachusetts}, \uppercase{salt lake city}).
Though the visual backbone of our translation model is pre-trained with fingerspelling sequences, transcribing the fingerspelling segment(s) is still problematic. As our model is based on whole word, it is incapable of translating words unseen during training. Thus proper nouns, typically fingerspelled in ASL, are difficult to translate by our model.
In practice, we observed many fingerspelled words are simply replaced with $<$UNK$>$. 

\subsection{Other Analysis}

\textbf{Baseline performance on Phoenix-14T}\label{sec:app-rwth2014t-perf}
Table~\ref{tab:ph14t-perf} shows the performance of the two baseline approaches on Phoenix-14T. In contrast to results on OpenASL, I3D-transformer does not generally outperforms Conv-GRU, which is probably due to the linguistic discrepancy between the isolated sign data used to pre-train I3D (WLASL: ASL) and the translation data (Phoenix-14T: DGS).

\input{table-tex/translation/ph14t-perf}

\textbf{Which pre-training data to use?}
\label{sec:app-effect-pretrain-dataset}
To show the effect of isolated sign pre-training, we compare I3D pre-trained with Kinetics-400 and WLASL in Table~\ref{tab:pretrain-data}. Kinetics-400~\cite{Carreira2017QuoVA} is a large-scale action recognition dataset including over 306,245 video clips from 400 action categories, while WLASL contains 14,289 clips from 2,000 ASL signs.
Though the size of WLASL is one order of magnitude smaller, using WLASL for pre-training 
outperforms pre-training with Kinetics only \kledit{by a large margin}. Utilizing isolated sign data, despite its amount being scarce, greatly boosts the visual representation that further benefits translation.

\input{table-tex/translation/pt-datasets}

\textbf{Does fine-tuning the visual encoder help?}\label{sec:app-finetune-freeze}
To show the effect of isolated sign pre-training, we compare I3D pre-trained with Kinetics-400 and WLASL in Table~\ref{tab:pretrain-data}. Kinetics-400~\cite{Carreira2017QuoVA} is a large-scale action recognition dataset including over 306,245 video clips from 400 action categories, while WLASL contains 14,289 clips from 2,000 ASL signs.
Though the size of WLASL is one order of magnitude smaller, using WLASL for pre-training 
outperforms pre-training with Kinetics only \kledit{by a large margin}. Utilizing isolated sign data, despite its amount being scarce, greatly boosts the visual representation that further benefits translation.

%

\input{table-tex/translation/finetune-vs-freeze}

\textbf{Impact of output units}
By default we use word as output unit in our experiments, thus it is unable to predict fingerspelling words which do not appear in the training set. A simple way to mitigate this issue is incorporating characters. We experimented this option and compared the following three types of output units: (a) word, (b) character, (c) a mixture of words and characters.
In option c, we only decompose named entities into characters for training, which are more likely to be fingerspelled.

\begin{table}[htp]
\centering
\begin{tabular}{l|rrrrr}

\toprule
Output Unit & ROUGE & BLEU-1 & BLEU-2 & BLEU-3 & BLEU-4 \\
\midrule  
 Char & 16.38 & 17.05 & 8.65 & 5.02 & 3.95  \\
 Char + Word & 18.75 & 17.83 & 9.10 & 6.58 & 4.86 \\
 Word & \textbf{20.43} & \textbf{20.10} & \textbf{11.81} & \textbf{8.43} & \textbf{6.57}  \\
 \bottomrule
\end{tabular}

\caption{\label{tab:impact-output-label} Impact of output units on translation performance}
\end{table}

The comparison between these three methods is shown in Table~\ref{tab:impact-output-label}. We do not see any improvement when using characters as output units. This is probably because many proper nouns appear frequently in the data. For example, the frequency of fingerspelled words "JOHN" and "JOHNSON" is 219 and 179, respectively. Using words as output units can help avoid spelling mistakes that could potentially be made by the character-based model.

\textbf{Does having a fingerspelling-specific module help?} Fingerspelling is a common occurrence in our data, but our model doesn't give it any special treatment. To address this, we experimented with a fingerspelling module to improve the translation quality. We transcribed the fingerspelling portion of an ASL video using a state-of-the-art fingerspelling recognizer (see Chapter~\ref{ch:closing-gap}) and detector (see Chapter~\ref{ch:fsdet}), and fed the word hypotheses as an additional input stream to the translation model.

Table~\ref{tab:effect-fingerspelling-module} presents the performance of the fingerspelling-enhanced translation model in comparison to the original translation model. The results were mixed, with slight improvements seen in ROUGE and BLEU-1 metrics, while falling short in the other metrics. This outcome could be due to several factors, including errors introduced by the fingerspelling recognizer and detector that may have negatively impacted the downstream translation performance. Additionally, the overall low translation quality with model predictions unrelated to the original video made it difficult for the fingerspelling module to demonstrate a significant effect. Further exploration is needed to determine effective approaches for handling fingerspelling in sign language translation.

\begin{table}[htp]
\centering
\begin{tabular}{c|rrrrr}
\toprule
With FS-module? & ROUGE & BLEU-1 & BLEU-2 & BLEU-3 & BLEU-4 \\
\midrule  
 \xmark & {20.43} & {20.10} & \textbf{11.81} & \textbf{8.43} & \textbf{6.57}  \\
\cmark & \textbf{21.00} & \textbf{20.68} & {11.08} & {7.95} & {6.31}  \\
\bottomrule
\end{tabular}
\caption{\label{tab:effect-fingerspelling-module} Effect of adding a fingerspelling-specific module on the translation performance}
\end{table}

\subsection{Qualitative Examples}

\textbf{Translation Examples}
\label{sec:app-translation-example}
We randomly select 15 examples from dev set and compare the model prediction against the reference (see Table~\ref{tab:translation-examples}). The exactly correct translations are mostly short and commonly used sentences in daily communication (e.g., thank you). For longer and more complex sentences, the model frequently fails to capture their general meaning though some keywords can be predicted correctly.

\textbf{Mouthing Examples}
\label{sec:app-mouthing-example}
Table~\ref{fig:mouthing-examples} shows some examples
where the model with mouthing features better translates the sentence. We highlight words mouthed by the signer, which are also correctly predicted by the model after the incorporation of mouthing feature.

\textbf{Signs spotted by our model}
\label{sec:search-sign-example}
The spotted lexical signs and fingerspelling sequences
are shown in figure~\ref{fig:lexical-sign-examples} and~\ref{fig:fs-sign-examples} respectively. Note that those examples are randomly selected. The spotted signs are mostly accurate. Below are our main observations.
First, in lexical sign spotting, the target clip often includes a partial (or whole) segment from adjacent signs. For instance, the third clip of UNIVERSITY has an extra sign of GALLAUDET. This is due to the fixed window size we use for lexical sign search. 2. False positives occur especially when two signs are of similar appearance. The second clip of BEFORE, which has a similar body posture to BEFORE, is a pointing sign indicating that one thing is happening prior to something else. 3. Using a sophisticated fingerspelling detector enables us to spot fingerspelling sequences more precisely compared to lexical signs.

\section{Conclusion}
\label{sec:conclusion}

Our work advances sign language translation "in the wild" (i.e., directly translating real-world sign language videos into written language) both (1) by introducing a new large-scale ASL-English translation dataset, OpenASL, and (2) by developing methods for improved translation in the absence of glosses and in the presence of visually challenging data.  OpenASL is the largest publicly available ASL translation dataset to date.  By using online captioned ASL videos, we have been able to collect a large amount of high-quality and well-aligned translation pairs (as verified by professional translators) that represent a wide range of signers, domains, and visual conditions.  Our translation approach, which combines pre-training via sign spotting and multiple types of local features, outperforms 
alternative methods from prior \kledit{work} by a large margin.  Nevertheless, the overall translation quality for sign language videos, in both our work and prior work, is significantly lower than that of machine translation for written languages.  There is therefore much room for future improvement, and we hope that OpenASL will enable additional progress on this task.

\input{table-tex/translation/translation-example-table}

\input{table-tex/translation/mouthing-examples}

\input{table-tex/translation/lexical-sign-examples}

\input{table-tex/translation/fs-sign-examples}

%% file: table-tex/translation/algo-search.tex
\algrenewcommand\algorithmicprocedure{\textbf{Procedure}}

\begin{algorithm}[btp]
\caption{Coarticulated Sign Search}\label{alg:sign-search}
\centering
\begin{algorithmic}[1]
\State \textbf{Data}: {Translation dataset $\mathcal{D}^t$}

\State \textbf{Model}: {isolated sign recognizer $M^{l}$,fingerspelling recognizer $M^f$, fingerspelling detector $M^d$}
\State \textbf{Hyperparameters}: {Lexical/FS threshold $\delta_l$/$\delta_f$}
\State \textbf{Output}: {Coarticulated lexical and fingerspelling sign dataset $\mathcal{D}^s$}
\Procedure{SearchSign}{$\mathcal{D}^t$, $M^{\{l,d,f\}}$, $\delta_{\{l,f\}}$}
\State $\mathcal{D}^s\gets \emptyset$
\For{$(\v I_{1:T}, w_{1:L})\in\mathcal{D}^t$}
\State Sliding windows $\Omega_s=\{(s_i, e_i)\}_{1:|\Omega_s|}$
\For{$(s, e)\in \Omega_s$}
\State Let $\v p\gets M^l(\v I_{s:e})$ be probability vector
\State Let $\tilde{w}_{1:L^\prime}\gets$ the subset of $w_{1:L}$ within the vocabulary of $M^l$
\State Let $\v q\gets (p_{\tilde{w}_1}, p_{\tilde{w}_2},...,p_{\tilde{w}_{L^\prime}})$
\State $k\gets \text{argmax}\{\v q\}$
\If{$q_k>\delta_l$}
\State $\mathcal{D}^s\gets \mathcal{D}^s\cup\{(\v I_{s:e}, \tilde{w}_k)\}$
\EndIf
\EndFor
\State Fingerspelling proposals $\Omega_f=\{(s_i, e_i)\}_{1:|\Omega_f|}=M^d(\v I_{1:T})$

\For{$(s, e)\in \Omega_f$}
\State Word hypothesis $\hat{w}_f=M^f(\v I_{s:e})$
\State Accuracies $\v y=(A(\hat{w}_f, w_1),...,A(\hat{w}_f, w_L))$, $A$: letter accuracy function
\State Let $k\gets \text{argmax}\{\v y\}$
\If{$y_k>\delta_l$}
\State $\mathcal{D}^s\gets \mathcal{D}^s\cup\{(\v I_{s:e}, w_k)\}$
\EndIf
\EndFor
\EndFor
\State \textbf{return} $\mathcal{D}^s$
\EndProcedure

\end{algorithmic}
\end{algorithm}

%% file: table-tex/translation/main_result.tex
\begin{table*}[btp]

\centering
\setlength{\tabcolsep}{3pt}
\resizebox{1\linewidth}{!}{
\begin{tabular}{lrrrrrr|rrrrrr}
\toprule
   & \multicolumn{6}{c|}{DEV} &\multicolumn{6}{c}{TEST} \\ \midrule
Models & {\small ROUGE} & {\small BLEU-1} & {\small BLEU-2} & {\small BLEU-3} & {\small BLEU-4} & {\small BLEURT} & {\small ROUGE} & {\small BLEU-1} & {\small BLEU-2} & {\small BLEU-3} & {\small BLEU-4} & {\small BLEURT} \\
\midrule  
\multirow{2}{*}{\begin{tabular}{@{}l@{}}Conv-GRU \\ {\small \cite{Camgoz2018neural}}$\dagger$\end{tabular}} &
\multirow{2}{*}{16.25} & \multirow{2}{*}{16.72} & \multirow{2}{*}{8.95} & \multirow{2}{*}{6.31} & \multirow{2}{*}{4.82} &
\multirow{2}{*}{25.36} &
\multirow{2}{*}{16.10} & \multirow{2}{*}{16.11} & \multirow{2}{*}{8.85} & \multirow{2}{*}{6.18} & \multirow{2}{*}{4.58} &
\multirow{2}{*}{25.65} \\ 
& & &  &  &  & & &  &  &  \\
I3D-transformer & 18.88	& 18.26 &	10.26 &	7.17 &	5.60 & 29.17 & 18.64 &	18.31 &	10.15 &	7.19 &	5.66 &	28.82  \\ \midrule\midrule
 Ours & \textbf{20.43} &	\textbf{20.10} &	\textbf{11.81} &	\textbf{8.43} &	\textbf{6.57} & \textbf{31.22} &	\textbf{21.02} &	\textbf{20.92} &	\textbf{12.08} &	\textbf{8.59} &	\textbf{6.72}	& \textbf{31.09}  \\ \bottomrule
\end{tabular}
}
\caption{\label{tab:main-result} Translation performance of baseline models and our proposed approach. $\dagger$: based on the public code released by the authors.}
\end{table*}

%


%
%
%

%
%
%
%
%
%
%

%
%
%

%% file: table-tex/translation/ph14t-perf.tex
\begin{table}[htp]
\centering
\resizebox{0.7\linewidth}{!}{
\begin{tabular}{c|rrrrr}

\toprule
Model & ROUGE & BLEU-1 & BLEU-2 & BLEU-3 & BLEU-4 \\
\midrule  
Conv-GRU~\cite{Camgoz2018neural} & \textbf{31.80} & \textbf{32.24} & \textbf{19.03} & 12.83 & 9.58 \\ 
I3D-transformer & 27.92	& 26.88	& 18.18 &	\textbf{13.42} &	\textbf{10.66} \\ \bottomrule
\end{tabular}
}
\caption{\label{tab:ph14t-perf} Performance of baseline approaches on Phoenix-14T test set.}
\end{table}

%% file: table-tex/translation/pt-datasets.tex
\begin{table}[htp]
\centering
\begin{tabular}{l|rrrrr}

\toprule
Data & ROUGE & BLEU-1 & BLEU-2 & BLEU-3 & BLEU-4 \\
\midrule  
 K & 13.63 & 12.25 & 5.07 & 3.14 & 2.32  \\
 K$\rightarrow$W & \textbf{18.88} & \textbf{18.26} & \textbf{10.26} & \textbf{7.17} & \textbf{5.60} \\  \bottomrule
\end{tabular}

\caption{\label{tab:pretrain-data} Effect of pre-training data (K: Kinetics-400~\cite{Carreira2017QuoVA}),W: WLASL~\cite{li2020word}).}
\end{table}

%% file: table-tex/translation/finetune-vs-freeze.tex

%
%

\begin{table}[htp]
\centering
\begin{tabular}{c|rrrrr}
\toprule
Fine-tuning? & ROUGE & BLEU-1 & BLEU-2 & BLEU-3 & BLEU-4 \\
\midrule  
\xmark & \textbf{18.88} & \textbf{18.26} & \textbf{10.26} & \textbf{7.17} & \textbf{5.60} \\ 
\cmark & 18.91 & 16.95 & 9.12 & 5.87 & 4.38 \\ \bottomrule
\end{tabular}

\caption{\label{tab:finetune-vs-freeze} Comparison between fine-tuning and freezing visual backbone.}
\end{table}

%% file: table-tex/translation/translation-example-table.tex
\begin{table*}[htp]
\centering
\resizebox{\linewidth}{!}{
    \begin{tabular}{|rl|}
    \hline
\#1 & \\ 
 Ref: & thank you\\ 
 Hyp: & thank you \\ \hline\hline 
\#2 & \\ 
 Ref: & come on\\ 
 Hyp: & come on \\ \hline\hline 
\#3 & \\ 
 Ref: & now i’ve come this far and it ’s a different team\\ 
 Hyp: & how do you feel about it \\ \hline\hline 
\#4 & \\ 
 Ref: & i was there from the beginning to the end and time went by fast\\ 
 Hyp: & the students were thrilled by this \\ \hline\hline 
\#5 & \\ 
 Ref: & i'm here at nad's 50th wow\\ 
 Hyp: & the nad has been $<$unk$>$ for many years \\ \hline\hline 
\#6 & \\ 
 Ref: & i entered the yap 2018 competition and won\\ 
 Hyp: & the competition was started with ideas \\ \hline\hline 
\#7 & \\ 
 Ref: & you can check out their kickstarter in the link below\\ 
 Hyp: & you can watch the conversation at lake county \\ \hline\hline 
\#8 & \\ 
 Ref: & that is one thing i found interesting and wanted to share with you today\\ 
 Hyp: & i also am the president of the jr. nad conference here \\ \hline\hline 
\#9 & \\ 
 Ref: & those are the different types of bills\\ 
 Hyp: & schools have switched to teaching students \\ \hline\hline 
\#10 & \\ 
 Ref: & dry january has picked up in popularity since it began in 2012\\ 
 Hyp: & krispy kreme is bringing back its original playstation in 2016 \\ \hline\hline 
\#11 & \\ 
 Ref: & we will be happy to respond give you support and listen to your concerns\\ 
 Hyp: & please review and submit your time passion and support this important issue \\ \hline\hline 
\#12 & \\ 
 Ref: & there were videos posted on the internet that showed a person walking on the grass completely engulfed in flames\\ 
 Hyp: & a video shows the officer walking up to his shoulder and before he was shot \\ \hline\hline 
\#13 & \\ 
 Ref: & and people would become carpenters laborers mechanics plowers and farmers\\ 
 Hyp: & the next year 1880 the nad was established in the first operation 30 of the house in 2015 \\ \hline\hline 
\#14 & \\ 
 Ref: & for nad youth programs related information please contact us via facebook at the nad youth programs or email us through\\ 
 Hyp: & you can contact us through our website where you can check our facebook page online at $<$unk$>$ \\ \hline\hline 
\#15 & \\ 
 Ref: & last week suspects gregory mcmichael and his son travis were arrested and charged with felony murder and aggravated assault\\ 
 Hyp: & last week a black man named $<$unk$>$ was arrested and charged with felony murder and aggravated assault \\ \hline\hline 

    \end{tabular}
}
\caption{\label{tab:translation-examples}Qualitative translation examples.  (Ref: reference, Hyp: prediction from our SLT model). Note the examples are randomly chosen without cherrypicking.}
\end{table*}

%
%
%

%% file: table-tex/translation/mouthing-examples.tex
\begin{figure*}[btp]
\centering
\resizebox{\linewidth}{!}{
\begin{tabular}{l}
\toprule
\# 1 \\
Ref: \textbf{what} are {you} \textbf{trying to} \textbf{do}\\
Hyp w/o M: that was a tough par \\
Hyp w/ M: so {what} should {you do} \\
\includegraphics[width=\linewidth]{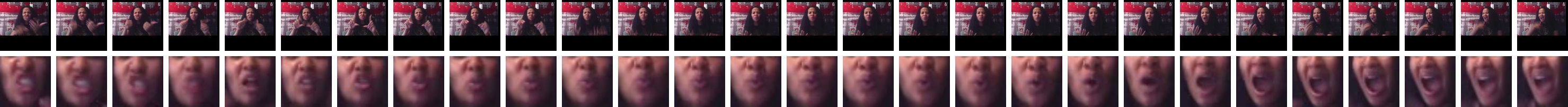}\\
\midrule  
\# 2 \\
Ref: \textbf{i have one for you}\\
Hyp w/o M: we have it \\
Hyp w/ M: {i have one of you} \\
\includegraphics[width=\linewidth]{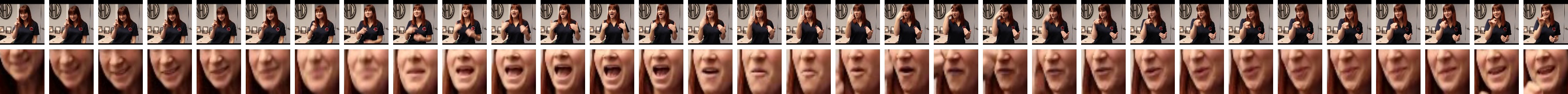}\\
\midrule  
\# 3 \\
Ref: so that's a \textbf{big concern}\\
Hyp w/o M: i don't want to be a part of the process \\
Hyp w/ M: it is {a big} challenge \\
\includegraphics[width=\linewidth]{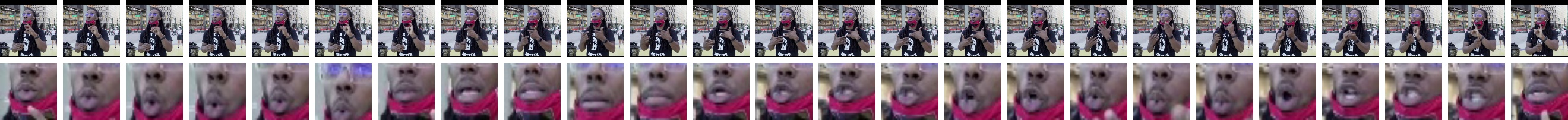}\\
\midrule  
\# 4 \\
Ref: i am from here \textbf{san francisco} born and raised\\
Hyp w/o M: i'm $<$unk$>$ this is our sign name \\
Hyp w/ M: i am from {san francisco} \\
\includegraphics[width=\linewidth]{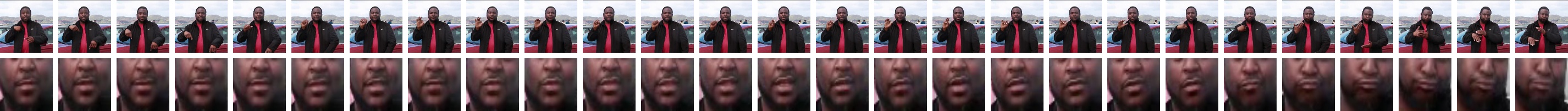}\\
 \bottomrule

\end{tabular}
}
\caption{\label{fig:mouthing-examples} Comparison between predictions of the original model and model using mouthing feature. M: mouthing feature. The words that are mouthed are boldfaced in the ground-truth sentence.}
\end{figure*}

%% file: table-tex/translation/lexical-sign-examples.tex
\begin{figure*}[htp]
\centering

\resizebox{\linewidth}{!}{
    \begin{tabular}{c}
\hline
president \\ 
 \includegraphics[width=\linewidth]{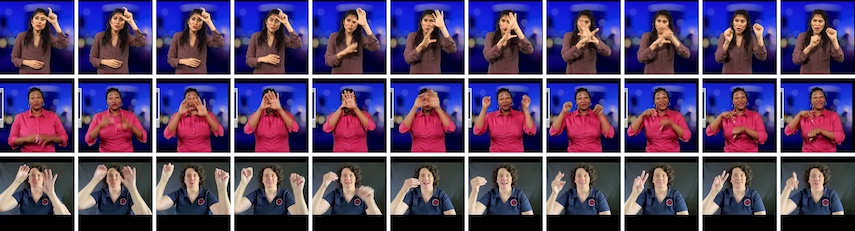}\\ \hline
\hline
family \\ 
 \includegraphics[width=\linewidth]{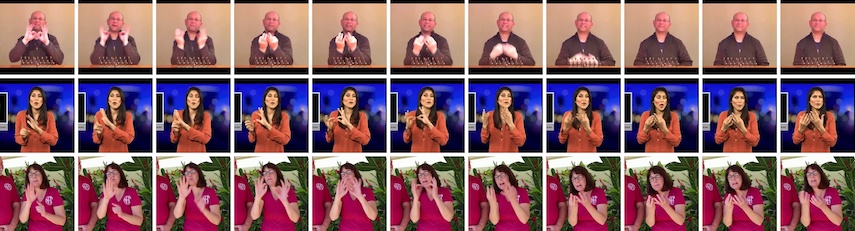}\\ \hline
\hline
people \\ 
 \includegraphics[width=\linewidth]{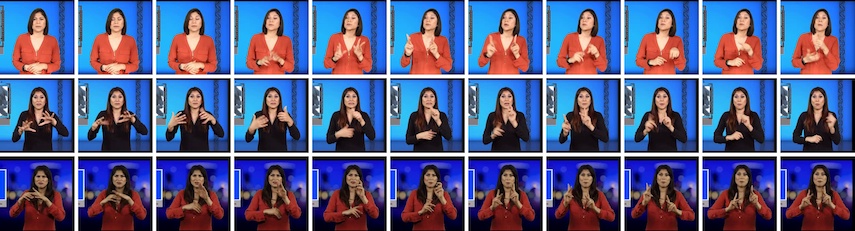}\\ \hline
\hline
more \\ 
 \includegraphics[width=\linewidth]{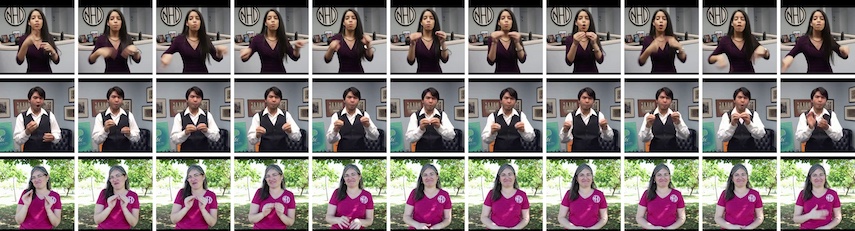}\\ \hline
    \end{tabular}
}
\caption{\label{fig:lexical-sign-examples}Qualitative examples of lexical signs searched by our model.}
\end{figure*}

%% file: table-tex/translation/fs-sign-examples.tex
\begin{figure*}[htp]
\centering
\resizebox{\linewidth}{!}{
    \begin{tabular}{c}
\hline
CNN \\ 
 \includegraphics[width=\linewidth]{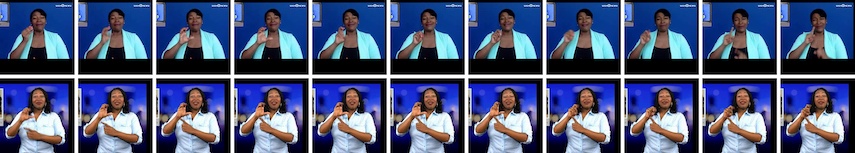}\\ \hline
\hline
MODERNA \\ 
 \includegraphics[width=\linewidth]{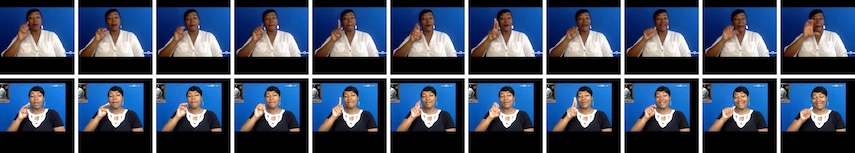}\\ \hline
\hline
ROBERT \\ 
 \includegraphics[width=\linewidth]{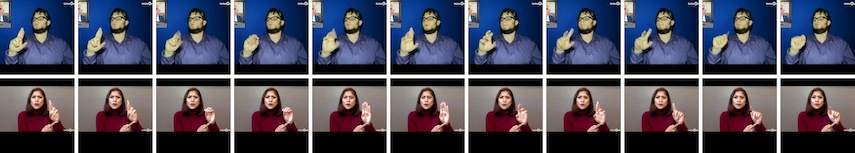}\\ \hline
\hline
US \\ 
 \includegraphics[width=\linewidth]{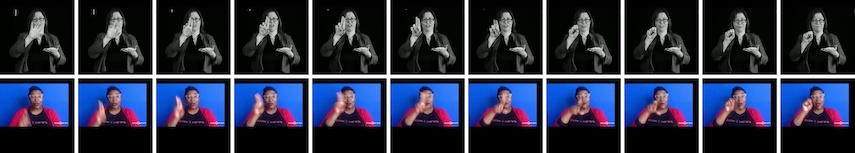}\\ \hline
\hline
IHOB \\ 
 \includegraphics[width=\linewidth]{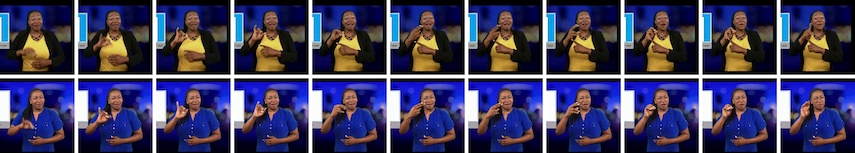}\\ \hline
\hline
RIT \\ 
 \includegraphics[width=\linewidth]{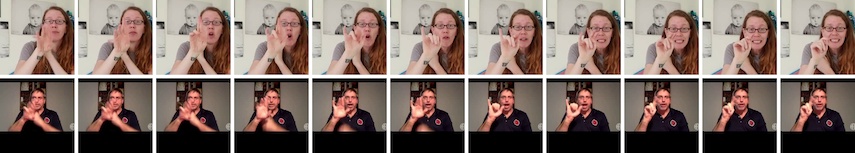}\\ \hline
\hline

    \end{tabular}
}
\caption{\label{fig:fs-sign-examples}Qualitative examples of fingerspelling signs searched by our model.}
\end{figure*}

%% file: conclusion.tex
\chapter{ Conclusion}
\label{ch:conclusion}

This thesis presents my efforts, in collaboration with my co-authors, in building sign language processing techniques for real-life applications, particularly on fingerspelling processing in real-world ASL media. 
To address the lack of realistic sign language data, we collected two fingerspelling datasets (ChicagoFSWild, ChicagoFSWild+) and one open-domain sign language translation dataset (OpenASL) using online sign language videos (Chapter~\ref{ch:data}). 
Those datasets, amongst the largest public sign language video corpora to date, resemble natural daily signing and pose more significant modeling challenges (e.g., more considerable visual variability, broader domain) than lab-based sign language datasets.
Chapter~\ref{ch:pipeline} to~\ref{ch:fs-search} focuses on fingerspelling, with a goal of enabling machines to understand fingerspelled content from raw ASL videos "in the wild".
Specifically, in Chapter~\ref{ch:pipeline}-\ref{ch:closing-gap} we study fingerspelling recognition, which aims at transcribing pure fingerspelling sequences. Chapter~\ref{ch:pipeline} establishes baselines for fingerspelling recognition under visually challenging conditions and identifies the importance of localizing the signing hand to a recognizer. 
To relieve the burden of hand annotation, we propose an end-to-end fingerspelling recognizer that implicitly detects the signing hand through iterative attention and substantially outperforms prior work using a hand detector (Chapter~\ref{ch:e2efsr}).
Chapter~\ref{ch:closing-gap} describes an approach jointly using mouthing and hand sequences for fingerspelling recognition, which achieves a new SOTA on ChicagoFSWild/ChicagoFSWild+ and dramatically reduces the gap between machine and human performance.
To enable fully automatic detection and recognition of fingerspelling in real-world "untrimmed" ASL media, in Chapter~\ref{ch:fsdet} we establish a new task --- fingerspelling detection . Motivated by the goal of using it as an upstream component in an automatic recognition system, we establish a suite of metrics to evaluate detection models both on their own merits and on their contribution to downstream recognition, and demonstrate a model incorporating losses derived from recognition and pose estimation achieves the best results across metrics.
In Chapter~\ref{ch:fs-search}, we address
the problem of searching for fingerspelled keywords or key phrases in raw sign language
videos, and develop a model tailored for those tasks.
Our model makes use of explicit fingerspelling detection to handle arbitrary-length queries and improves over baseline models based on related
work on signed keyword search, fingerspelling detection, and speech recognition.
Finally, based on OpenASL we advance sign language translation "in the wild" by developing methods for improved translation in the absence of glosses (Chapter~\ref{ch:slt-wild}). The proposed techniques, including pre-training with sign search and incorporating local features, produce consistent and large improvements in translation quality.

This thesis has focused on ASL fingerspelling as a case study to investigate sign language in the wild. The extensive collection of large-scale real-world fingerspelling data and integration of the latest network architectures has allowed us to achieve close-to-human level performance in fingerspelling recognition. However, the tools we developed and the findings we discovered have broader applicability beyond ASL and sign language video research.

Since fingerspelling is a common feature in many sign languages, the techniques we established for detecting and searching it in raw sign language videos, as demonstrated in Chapters~\ref{ch:fsdet} and \ref{ch:fs-search}, are similarly useful for different sign languages. Additionally, pose estimation plays a crucial role in sign language research nowadays~\cite{Moryossef2020sld,saunders2020progressive,Saunders2020EverybodySN}. However, in our ``in-the-wild" sign language datasets, off-the-shelf pose estimation toolkits like OpenPose~\cite{cao2019openpose} often fail to detect keypoints in the signing region, potentially harming downstream tasks if directly used as the sign language representation. To address this, we proposed a pose estimation approach as an auxiliary task during model training (in Chapter~\ref{ch:fsdet}) to provide a practical recipe for using pose in general sign language video processing.

Finally, the iterative attention approach that we proposed for end-to-end fingerspelling recognition is a universal mechanism for obtaining high-resolution regions of interest (ROI) from videos without explicit object detection. Therefore, it can be applied to other fine-grained gesture or action recognition tasks. Overall, the techniques, methodologies, and models developed in this thesis provide contributions to the broader field of 
computer vision research, particularly in the areas of fine-grained gesture and action recognition.

The work presented in this thesis is just the first step toward developing real-world sign language processing technologies. While our focus has been on fingerspelling processing and sign language translation, many other practical tasks (e.g., SLP, SLS) have not been explored in this work. It is clear that there is still a long way to go before an AI system that can understand sign language becomes ubiquitous in Deaf people's lives. Below we describe some future directions for sign language processing.



\textbf{Effective handling of fingerspelling in sign language translation} Despite the high accuracy of our standalone fingerspelling recognizer, we have not been successful in integrating it into a sign language translation system. In Chapter~\ref{ch:slt-wild}, we introduced a translation model that did not explicitly handle fingerspelling. However, given its frequent use in ASL, effective handling of fingerspelling is critical for developing a full-fledged ASL-English translation system. Several factors, such as the overall low quality of the translation model, error accumulation in cascading a fingerspelling recognizer with the downstream translation model, may contribute to the unsuccessful attempts at integration. How to handle fingerspelling effectively within a sign language translation system remains an open question and requires further research.
 
\textbf{Collecting larger real-world datasets}
Modern machine learning models require extensive amounts of data to achieve optimal performance, particularly in complex scenarios. Sign language processing is no exception, and large-scale sign language datasets are essential for advancing the development of sign language processing techniques. However, the lack of sufficient labeled data, particularly sign language video-text pairs, remains a major obstacle to improving the performance of current SLT approaches, including the one discussed in Chapter~\ref{ch:slt-wild}, as well as in prior research~\cite{camgoz2021content,Albanie2021bobsl}.
Data scarcity is a pervasive issue in current research on sign language processing, and one way to address it is by collecting larger and more diverse datasets. In this thesis, we utilized web videos and crowdsourcing annotation to collect datasets that were larger than previous studio-based sign language corpora. However, these datasets are still several orders of magnitude smaller than many spoken language corpora. For instance, the largest SLT video corpus we are aware of, BOBSL, contains 1,400 hours of British Sign Language (BSL) videos from 40 signers, covering a vocabulary size of 2K signs. In contrast, large-scale public speech recognition datasets, such as MLS~\cite{Pratap2020mls} and People's Speech~\cite{Galvez2021ThePS}, typically contain 10K to 50K hours of transcribed speech from over 1K speakers, covering several million unique words. In-house datasets can be even larger~\citep{Soltau2016neural}. Furthermore, many speech datasets, such as those used in~\cite{Gales2014SpeechRA,Ardila2020CommonVA,Pratap2020mls}, are multilingual, while existing sign language datasets are limited to a few languages, such as ASL, BSL, and DGS.
Scaling up sign language datasets presents a few challenges compared to spoken languages. First, annotating sign language requires more expertise and is therefore likely more expensive than annotating speech datasets of a similar scale. Although crowdsourcing offers potential cost savings, quality control issues arise, particularly when annotating a translation dataset that requires extensive knowledge. Second, creating public sign language video corpora raises potential legal concerns. For example, online sign language videos may be subject to copyright protections. Furthermore, signing footage typically includes a signer's face, which raises privacy concerns. Prior work has attempted to address these issues by masking faces~\cite{Duarte2021how,camgoz2021content}, but this approach may lead to the loss of signing information from non-manual articulators. These issues are a few of the hurdles that must be overcome to scale up naturalistic sign language datasets.

\textbf{Developing Unsupervised Models} Most sign language processing research, including the work presented in this thesis, has focused on a fully supervised setting. However, developing unsupervised methods for sign language can reduce the reliance on labeled sign language videos, potentially improving sign language processing techniques at a lower cost. In speech and language processing, unsupervised representation learning has made significant progress in recent years. For example, the self-supervised speech representation learning framework, wav2vec-2.0~\cite{Baevski2020wav2vec2A}, has outperformed former state-of-the-art speech recognition systems using 100 times fewer labeled data. Similarly, large language models like GPT~\cite{Brown2020LanguageMA} have achieved competitive performance in machine translation for written languages without using translation-specific training data. However, these methods are not directly applicable to video-based sign language processing due to differences in input type (video vs. audio or text), despite the shared language nature of signed, spoken, and written languages. While unsupervised representation learning for sign language has received limited attention, a few studies~\cite{Martinez2022Unsupervised,Hu2021SignBERTPO,Shi2017MultitaskTW,Gebre2014UnsupervisedFL} have been conducted under heavily restrictive settings (e.g., small data regimes~\cite{Martinez2022Unsupervised,Shi2017MultitaskTW}, isolated sign recognition only~\cite{Hu2021SignBERTPO}), and have not generally shown considerable gains. The lack of self-supervised representation learning for sign language may arise from the small scale of existing sign language corpora and the lack of a proper benchmark for evaluating the models. We hope that increased efforts in data collection, standardization of sign language processing tasks, and advancements in modeling tools will eventually lead to successful unsupervised representation learning approaches for sign language.


%
%

%
%

%

%